Fall 2019

# Analysis and Control of Fiber-Reinforced Elastomeric Enclosures (FREEs)

Soheil Habibian
*Bucknell University*, sh051@bucknell.edu

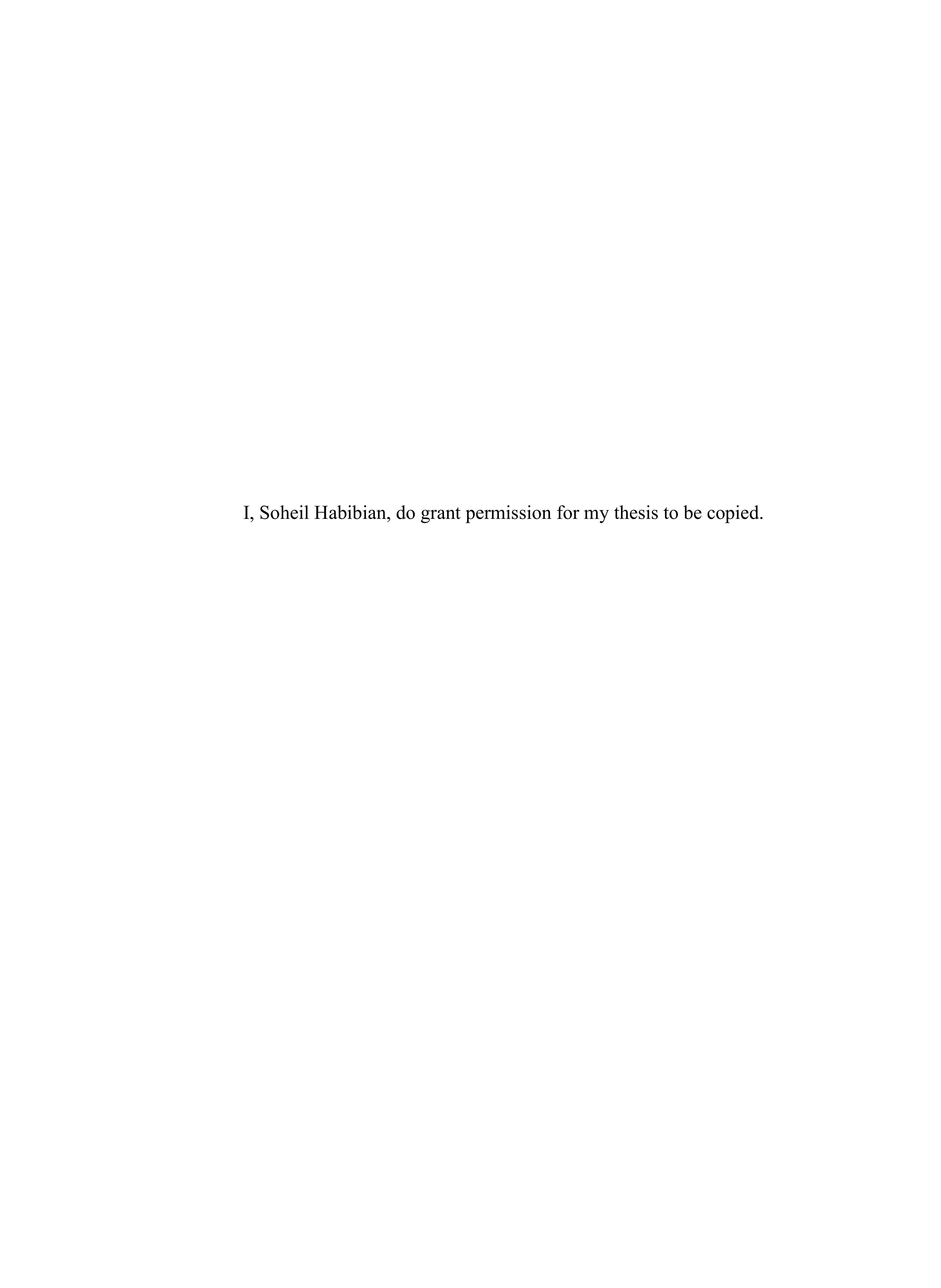

ANALYSIS AND CONTROL

OF

FIBER-REINFORCED ELASTOMERIC ENCLOSURES (FREEs)

By

SOHEIL HABIBIAN

A Thesis

Presented to the Faculty of
Bucknell University
In Partial Fulfillment of the Requirements for the Degree of
Master of Science in Mechanical Engineering

Approved:

Keith W. Buffinton, Adviser

Constance W. Ziemian, Dept. Chair

Charles J. Kim, Committee Member

Benjamin B. Wheatley, Committee Member

Date: December 2019

# Acknowledgement

I would first like to express my sincere gratitude to my thesis advisor Professor Keith W. Buffinton. He opened a gateway of the learning experience for me as well as provided professional and patient mentoring throughout my Master's program. Professor Buffinton insightfully guided me in this journey and facilitated my transition to graduate school. His consistent support, encouragement, and precise evaluations of my work are greatly appreciated. I would also like to acknowledge Prof. Charles J. Kim and Prof. Benjamin B. Wheatley as the second readers of this thesis, and I am gratefully indebted to their very valuable comments on my research. My thanks to the Office of Graduate Studies and the Department of Mechanical Engineering at Bucknell University for providing funding and equipment for this project. Special thanks to Daniel Bruder and Audrey Sedal, the Ph.D. students of Prof. C. David Remy and Prof. Brent Gillespie at the University of Michigan that offered their support and collaboration on this work. Finally, I must express my very profound gratitude to my parents for providing me with unfailing support and continuous encouragement throughout my years of study and through the process of starting higher education in another country. This accomplishment would not have been possible without them.

# List of Parameters

$R$ – Initial radius
$\Gamma$ – Initial fiber winding angle
$L$ – Initial length
$r$ – Final radius
$\gamma$ – Final fiber winding angle
$l$ – Final length
$\varphi$ – Rotation angle of the free end
$s$ – Displacement of the free end
$P$ – Pneumatic pressure
$F_l$ – External load
$F_e$ – Elastomer force
$T_{net}$ – Net force generated by fibers
$F_P$ – Force exerted by the pressure at the end cap
$m_l$ – Mass of the end cap
$M_l$ – External moment
$M_e$ – Elastomer moment
$I_l$ – Mass moment of inertia of the end cap
$\sigma_\theta$ – Circumferential stress
$F$ – Force exerted due to hoop stress
$t_c$ – Thickness of the cylinder
$A$ – Area of the wall of the cylinder
$F_y$ – Net force generated by pressure at fibers
$k_e$ – Elastomer linear stiffness
$c_e$ – Elastomer axial damping coefficient
$k_t$ – Elastomer torsional stiffness
$c_t$ – Elastomer torsional damping coefficient
$F_{gravity}$ – Force exerted by the end cap
$K_p$ – Proportional gain
$K_i$ – Integral gain
$K_d$ – Derivative gain
$\varphi_d$ – Desired rotation angle
$t$ – Dynamic response time
$t_f$ – Trajectory following time
$C(s)$ – System output
$G(s)$ – Plant transfer function
$H(s)$ – Feedback transfer function

$R(s)$ − Reference input
$X(s)$ − Laplace transform of extension $s$
$Y(s)$ − Laplace transform of rotation $\varphi$
$Y_d$ − Reference output of the closed-loop control system
$K$ − Gain "rlocus" MATLAB function
$C$ − Constant coefficient of the general solution of a differential equation
$a$ − Constant coefficient of the general solution of a differential equation
$a_0$ − Constant coefficient of the cubic polynomial function
$a_1$ − Constant coefficient of the cubic polynomial function
$a_2$ − Constant coefficient of the cubic polynomial function
$a_3$ − Constant coefficient of the cubic polynomial function
$\varphi_f$ − Goal angle of rotation in trajectory planning
$\varphi_0$ − Initial angle of rotation in trajectory planning
$b/B$ − Normalized radial expansion of FREE (FEA)
$\lambda_z$ − Normalized axial extension of FREE (FEA)
$\tau$ − Twist angle per length (FEA)
$\Psi$ − Strain energy function for a first-order Ogden model
$\bar{\lambda}_i$ − Deviatoric principal stretches
$J$ − Volume ratio
$\mu$ − Material parameter (Ogden model)
$\alpha$ − Material parameter (Ogden model)
$D$ − Material parameter (Ogden model)
$\delta_r$ − Radial expansion (thin-walled cylindrical pressure vessel)
$p$ − Internal pressure (thin-walled cylindrical pressure vessel)
$b$ − Wall thickness (thin-walled cylindrical pressure vessel)
$E$ − Elastic modulus
$\nu$ − Poisson's ratio
$e$ − Normalized root-mean-square deviation (RMSD)

# Abstract

While rigid robots are extensively used in various applications, they are limited in the tasks they can perform and can be unsafe in close human-robot interactions. Soft robots on the other hand surpass the capabilities of rigid robots in several ways, such as compatibility with the work environments, degrees of freedom, manufacturing costs, and safe interactions with the environment. This thesis studies the behavior of Fiber Reinforced Elastomeric Enclosures (FREEs) as a particular type of soft pneumatic actuator that can be used in soft manipulators. A dynamic lumped-parameter model is created to simulate the motion of a single FREE under various operating conditions and to inform the design of a controller. The proposed PID controller determines the response of the FREE to a defined step input or a trajectory following polynomial function, using rotation angle to control the orientation of the end-effector. Additionally, Finite Element Analysis method is employed, incorporating the inherently nonlinear material properties of FREEs, to precisely evaluate various parameters and configurations of FREEs. This tool is also used to determine the workspace of multiple FREEs in a module, which is essentially a building block of a soft robotic arm.

Both of these models provided a great understanding of a FREE's behavior in various working conditions. This understanding led to employing a group of FREEs in a module to explore new applications, Although, the finite element model was not able to fully and accurately predict the system response in all cases. It did however provide a basis of understanding for the trends in FREEs' behavior in single and module configurations, and demonstrated the necessity of improving the fabrication process of FREEs. Results of the two models point to the importance of the manufacturing process in minimizing variations in FREE behavior. Overall, the developed models in this project efficiently predict the behavior of FREEs and they can potentially be used for future studies of FREEs and similar soft actuators.

# 1
# Introduction

## 1.1 Motivation

In human-robot interactions (HRI), safety is one of the top priorities for robotics engineers (Vasic & Billard, May 2013). Conventional robots use rigid materials to create an accurate and controllable robotic system. While rigid links and discrete joints ensure performance of a robotic system, they can cause accidents in the working environment (Jiang & Gainer, 1987). Therefore, increasing attention has been focused in recent years on the development and analysis of "soft" robots that can perform a variety of simple tasks in and around humans with minimal risk of injury to the humans as well as to the work environments. In addition to safety concerns, the inherent compliant structure of soft robots has the potential to exceed the capabilities of traditional robots. Nature presents many examples of soft biological structures in animals and plants. Examples such as octopus' arms and elephant trunks, known as muscular hydrostats, have inspired engineers to design soft robots to operate in unstructured environments. These compliant structures offer the advantage of an infinite number of degrees of freedoms in robotic systems and allow them to perform complex tasks which traditional robots confront with difficulty (Trivedi, Rahn, Kier, & Walker, 2008). Recent developments in this field suggest that traditional methods for the design, fabrication, and modeling of robots are not appropriate for soft robots (Rus & Tolley, 2015). Current research now strives to create techniques that deliver the full potential of a soft machine (Sedal, 2019). This thesis seeks to address these opportunities and challenges through in-depth modeling and control studies of a particular type of soft actuator known as Fiber Reinforced Elastomeric Enclosures (FREEs) for use in a soft robotic arm.

## 1.2 Fiber Reinforced Elastomeric Enclosure (FREE)

Most of soft robots are driven by soft actuators, which are often driven by fluids. The fiber reinforced elastomeric enclosure (Bishop-Moser, Krishnan, Kim, & Kota, October 2012) is a special pneumatic-driven type of actuator in this class. A Fiber Reinforced Elastomeric Enclosure (FREE) (Figure 1.1) consists primarily of two components, an elastomer and a fiber, and thus represent a composite material. The elastomer has the role of the matrix of material supporting the fibers, which provides additional resistance to loads. FREEs can be used as pneumatic actuators in mechanical systems by applying pneumatic pressure to their internal surface.

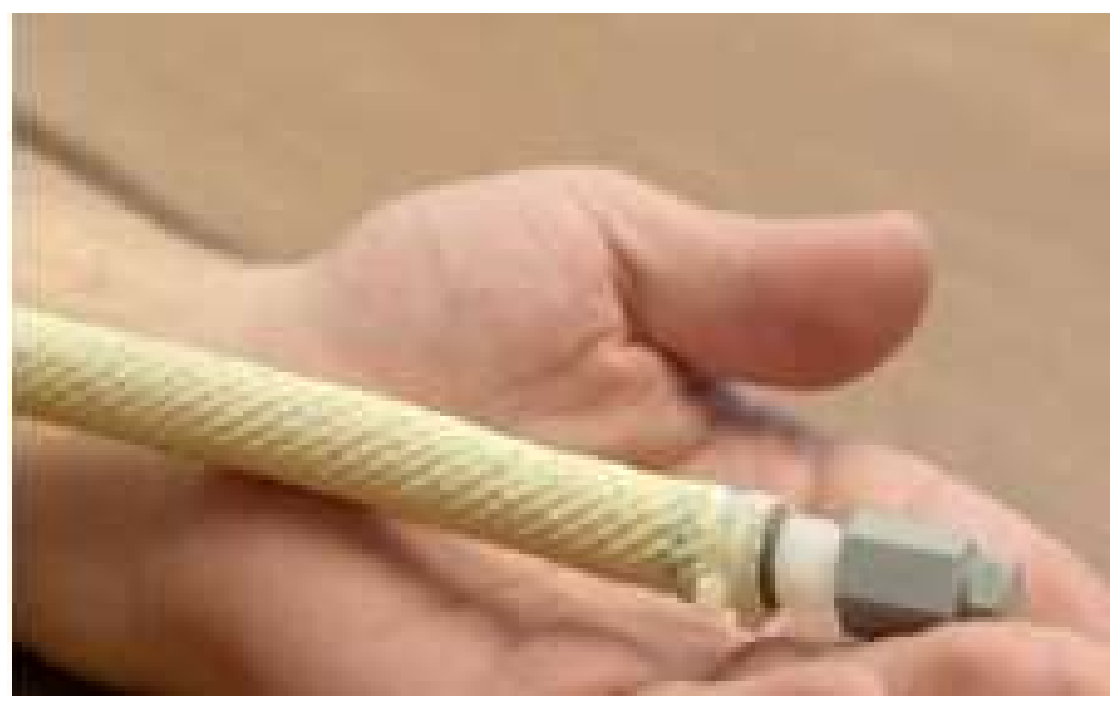

*Figure 1.1. Fiber Reinforced Elastomeric Enclosure (FREE)*

Pressurized FREEs generate sophisticated motions, including axial rotation, extension, and compression, and serve as building blocks for soft robotic manipulators. There is a variety of parameters that affect such an actuator's response, including both geometric and material properties. The choice of these parameters in designing FREEs is determined by the desired application and overall response characteristics.

## 1.3 Prior Work

In recent years, the relatively high stiffness of traditional robots has led roboticists to become increasingly interested in the use of soft robots (Rus & Tolley, 2015). Traditional rigid robots with discrete joints create predictable systems. However, soft robots (Figure 1.2) often mimic properties found in nature, such as plants and especially animals (Kier, 1985). For example, soft tissues in cephalopods or compliant bones in human vertebrates make them appropriate to perform sophisticated motions. Soft robots made from thin wires or compliant materials (Majidi, 2019), and operated by an electrical, thermal, or pneumatic actuation, are employed in a variety of applications. Hawkes, Blumenschein, Greer, & Okamura (2017) have designed a soft pneumatic robot to perform active-controlled navigation by growth in constrained environments. This fluid-driven soft robot is inspired by fungal hyphae that navigate in their surroundings through growth. The design of "Pneu-Net" helped soft robots to step into the world of autonomous mobile robots by using lightweight and resilient silicon rubber (Tolley *et al.*, 2014). Marchese, Onal, & Rus (2014) have introduced another example of bioinspired autonomous mobile robots in the form of a soft-bodied fish, which performs an escape response analogous to biological fish in terms of kinematics and controllability. The role of soft robots becomes more significant in environments in which robots interact closely with the environment or humans. Having structurally soft components is essential to the protection of humans, robots, and their environments. Galloway, Kevin C, *et al.* (2016) presented a soft robotic gripper mounted on a remotely operated vehicle for sampling fragile species under the sea, which profoundly decreased the destructive interaction between machines and nature for "*in situ*" testing and field collections. Using the "Pneu-Net" soft actuators (Polygerinos, Wang, Galloway, Wood, & Walsh, 2015) to create an open-palm glove for rehabilitation purposes demonstrated that robots can interact with humans closely. In addition to these applications, design and fabrication of soft robots are often cheaper and more accessible by rapid prototyping techniques (Marchese, Katzschmann, & Rus, 2015), whereas rigid parts of traditional robots often require sophisticated manufacturing procedures.

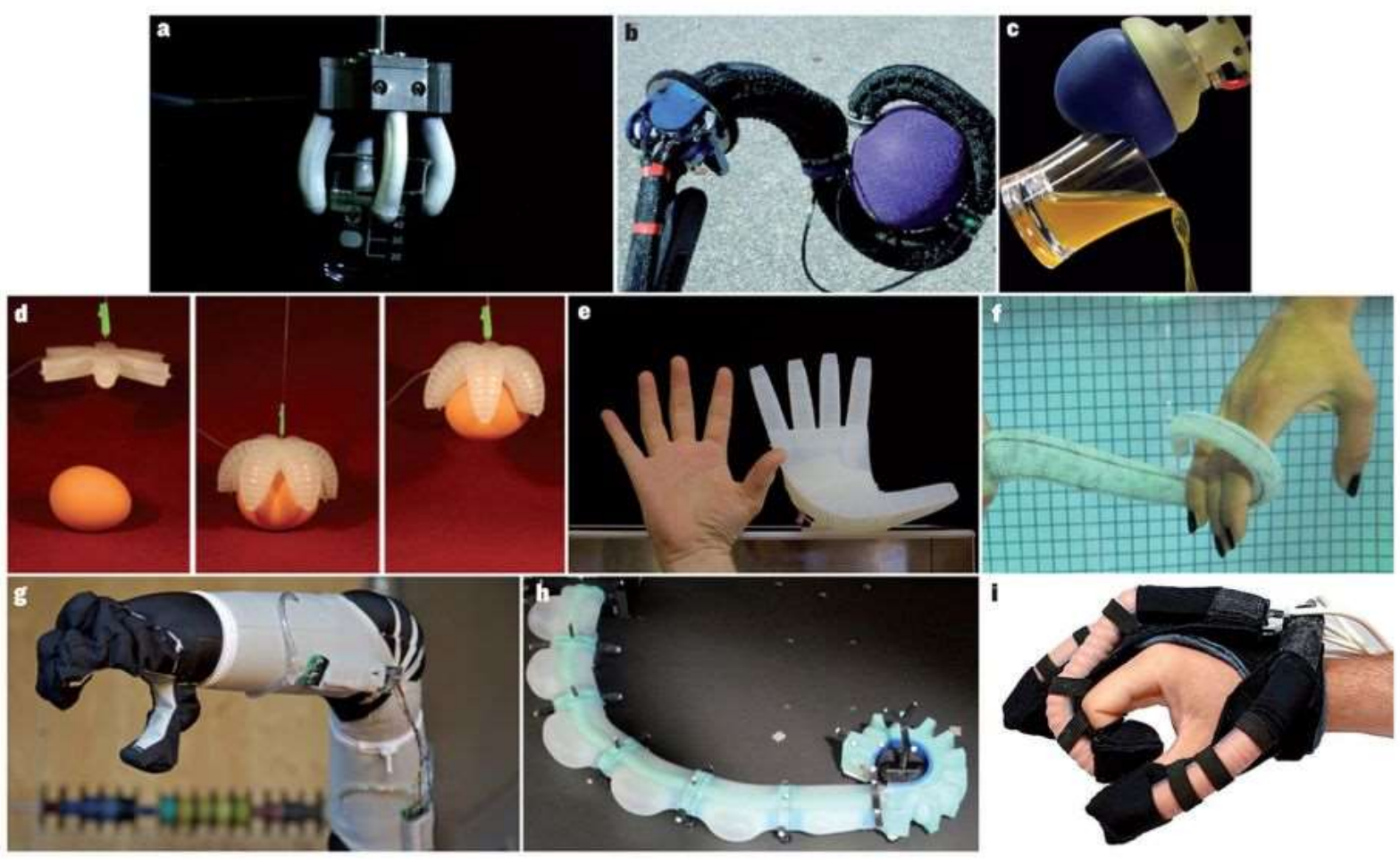

*Figure 1.2. Soft robots inspired by nature demonstrating superior performance (Rus & Tolley, 2015)*

Most soft robots are created with soft pneumatic actuators. For instance, OctArm is a soft robotic manipulator that achieved its adaptability by using air muscle extensors (Grissom *et al.*, 2006). One type of well-known pneumatic actuators is the McKibben artificial muscle, which is made of a tube wrapped with braided cords (Tondu, 2012). McKibben actuators use pneumatic pressure to generate circumferential stress on the tube and transmit contraction forces to the whole mechanism. To analyze this actuator a linearized model has been developed to do static/dynamic length measurements of the inflated muscle. The FORA is an improved McKibben-type actuator, which has double the typical range of motion, an enhanced force profile, and lower actuation pressure (Yi, Chen, Song, & Wang, 2018). To avoid dynamic uncertainties, a quasi-static analytical model was developed to characterize the performance of the FORA. Polygerinos *et al*. (2015) have combined a quasi-static analytical and a finite element model to characterize the motion and force generation of a bending actuator. The analytical model using a feedback control system was sufficient in estimating the bending angle in real-time for their application. In some cases, the motion of soft actuators is constrained by using a conformal cover to make the modeling more convenient (Galloway, Kevin, Polygerinos, Walsh, & Wood, 2013). The large number of degrees of freedom (DOF) provided by soft actuators has made them popular among roboticists; however, modeling and controlling them, usually because of the nonlinearity of materials, is still a challenge.

Fiber-Reinforced Elastomeric Enclosures (FREEs) is a subset of pneumatic actuators and have been modeled with several techniques. Bishop-Moser, Joshua, Krishnan, Kim, & Kota (2012) used the geometric relationship between fibers and fluid forces to develop a kinematic model of a FREE. In continuance of this study, (Krishnan, Bishop-Moser, Kim, & Kota, 2015) found that fiber alignment around the circumference of the tube influences the pitch of the motion and may cause a condition in which the FREE does not deform with inflation (known as a locked manifold). A quasi-static model was also created (Bruder, Sedal, Bishop-Moser, Kota, & Vasudevan, 2017) to perform open-loop control of rotation angle. A continuum model developed by Sedal, Bruder, Bishop-Moser, Vasudevan, & Kota (2018) has focused on the FREE from a different perspective. This computational method accounts for the nonlinear characteristics of the FREE and describes the relationship between pressure and output forces and deformations. Research on parallel combinations of FREEs (Bruder, Sedal, Vasudevan, & Remy, 2018; J. Bishop-Moser, G. Krishnan, C. Kim, & S. Kota, 2012) opened a path to studying parallel manipulation tasks and force generation by using multiple FREEs in a module.

Each of these studies has contributed to a better understanding of the FREEs' behavior by addressing a different modeling technique. However, controlling the position and rotation of FREEs has remained challenging, particularly when multiple FREEs are coupled. The research in this thesis specifically concentrates on dynamic modeling by including mass, inertia, and damping coefficients in the equations of motion to determine a controlled time-dependent response. Additionally, finite element analysis is used to broadly explore the FREEs' behavior, particularly when other modeling techniques become complicated and laborious.

## 1.4 Thesis Statement

The goal of this thesis is to characterize and control FREEs' behavior based on both a lumped-parameter dynamic lumped-parameter model and a finite element material model. The outcomes of this work are (1) a dynamic simulation that visualizes FREEs' behavior with respect to pressure, (2) a model-driven PID controller for following a desired rotation of a single FREE, (3) a finite element model as an additional design and verification tool to supplement the dynamic simulation, (4) the determination of effective parameters for use in the design of FREEs, and (5) a simulations of the workspace of multiple FREEs in a module based on finite element analysis. The overreaching goal is to demonstrate the capabilities of FREEs as an actuator for soft robots performing typical daily tasks, such as cooking on a stovetop, by exploiting the created tools and controller.

## 1.5 Organization

The remainder of this thesis is organized as follows. Chapter 2 presents the lumped-parameter model used to derive differential equations of motion for a FREE. Chapter 3 addresses the dynamic simulation of the FREE with respect to pressure and the PID control of a FREE to achieve a desired rotation of the end-effector. Chapter 4 describes the finite element model of FREEs in single and module configurations, the parametric study of FREE behavior, and the workspace of multiple FREEs in a module. Chapter 5 presents the experimental apparatus and software interface utilized in this work. Chapter 6 discusses the experimental test data corresponding to the models developed in Chapters 3 and 4. Finally, the overall behavior and capabilities of FREEs as an actuator for use in soft robotic manipulators are reviewed in Chapter 7.

# 2

# Dynamic Modeling

In the majority of cases, scientists and engineers engage in modeling to analyze and predict the behavior of a system quantitatively. To study the characteristics of a FREE and broaden understanding of its capabilities as a soft robotic actuator, a simple mathematical model is considered. In this chapter, the geometry of a FREE and its relation to internal pneumatic pressure is analyzed to create the governing differential equation of motions.

## 2.1 Lumped Parameter Model

This section introduces the geometric and dynamic parameters of a FREE and relates them to the variables used to develop dynamical equations of motion.

### Geometrical Relationship

As depicted in Figure 2.1, the notation used for the parameters and variables of the model of the FREE are defined in two different configurations: pressurized (final) and unpressurized (initial). The length and rotation of a FREE change with pressure. To address these changes, the following variables and parameters are used:

$R$ – initial radius - constant parameter
$\Gamma$ – initial fiber winding angle - constant parameter
$L$ – initial length - constant parameter
$r$ – final radius - variable
$\gamma$ – final fiber winding angle - variable
$l$ – final length - variable
$\varphi$ – rotation angle of the free end - variable
$s$ – displacement of the free end $(l - L)$ - variable
$P$ – pneumatic pressure - variable

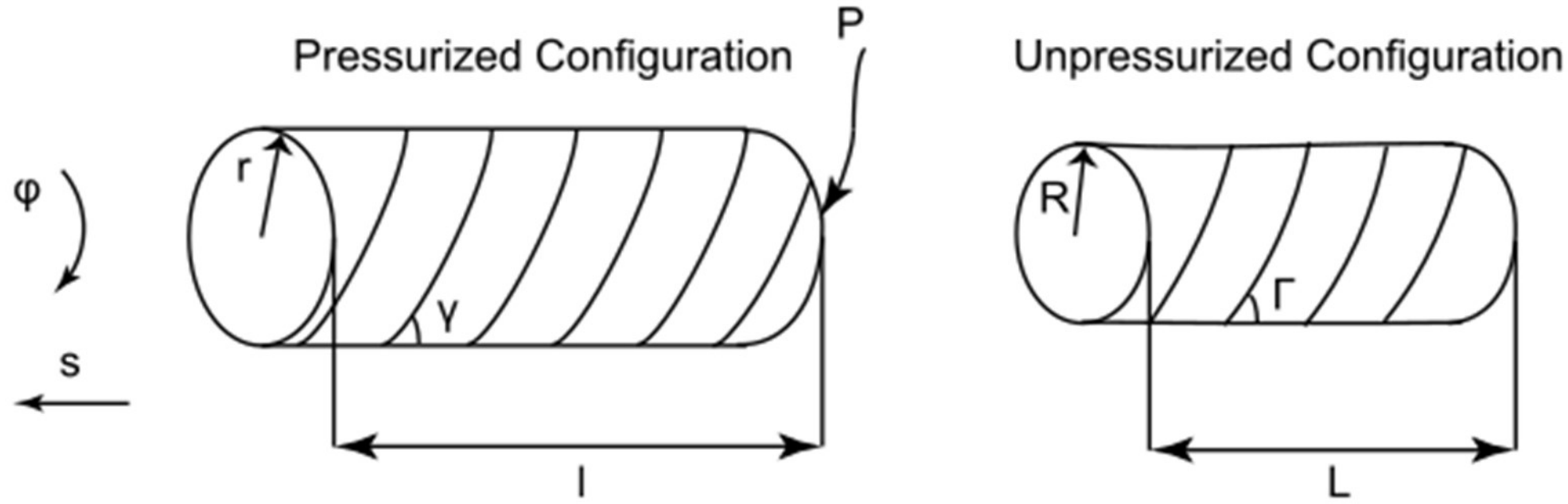

*Figure 2.1. Geometric parameters of a FREE in pressurized and unpressurized configurations*

As depicted in Figure 2.1, each fiber is wrapped in the form of a helix around the tube. Since the fibers are modeled as inextensible, the following geometric relationships in Eqs. (2.1) and (2.2) for fiber angle γ and radius *r* of a FREE at a pressure *P* are taken from the literature published by Bruder *et al.* (2017):

$$\gamma = cos^{-1}\left(\frac{l\cos\Gamma}{L}\right), \tag{2.1}$$

$$r = \frac{L\tan\gamma}{\frac{L\tan\Gamma}{R} + \varphi}. \tag{2.2}$$

**Free Body Diagram**

Figure 2.2 shows the FREE fixed at one end and closed at the other end by a cap. The motion of the end cap describes the behavior of the FREE and the force and moment equilibrium of the end cap is used to develop dynamical equations of motion. Assume that the end cap with a mass of $m_l$ and a control moment of inertia about on axis along the FREE $I_l$ is the only significant mass of the FREE. There are multiple forces applied to the cap by each component of the FREE. The pressure applies a force in the outward direction. The elastomer and the fiber also apply forces to the cap. Additionally, an external load may apply force and torque to the end cap.

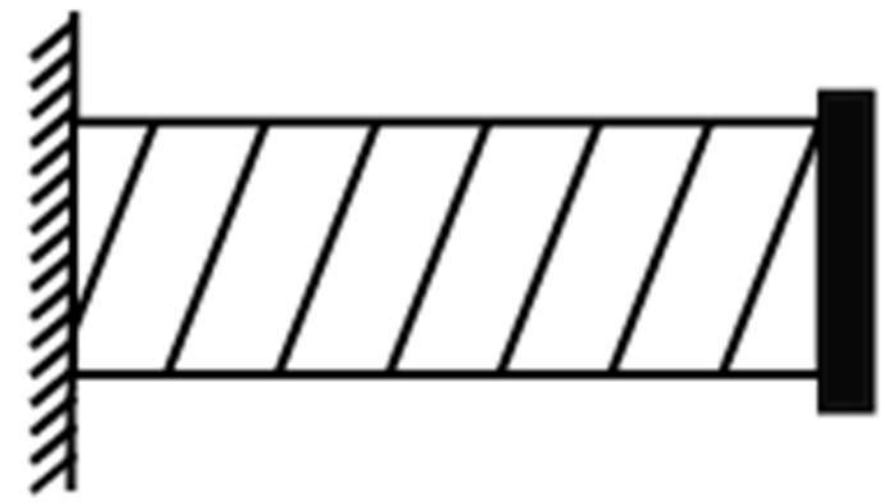

*Figure 2.2. Boundary conditions of a FREE*

Figure 2.3 presents a free body diagram of the end cap: $F_l$ is the external load, $M_l$ is the external moment, $F_p$ is the exerted force by the pressure, $F_e$ is the force applied by the elastomer, $M_e$ is the moment applied by the elastomer, and $T_{net}$ is the net force due to the fibers. For simplicity of the analysis, the FREE is modeled only with one family of fibers (wound with a group of parallel fibers of the same material).

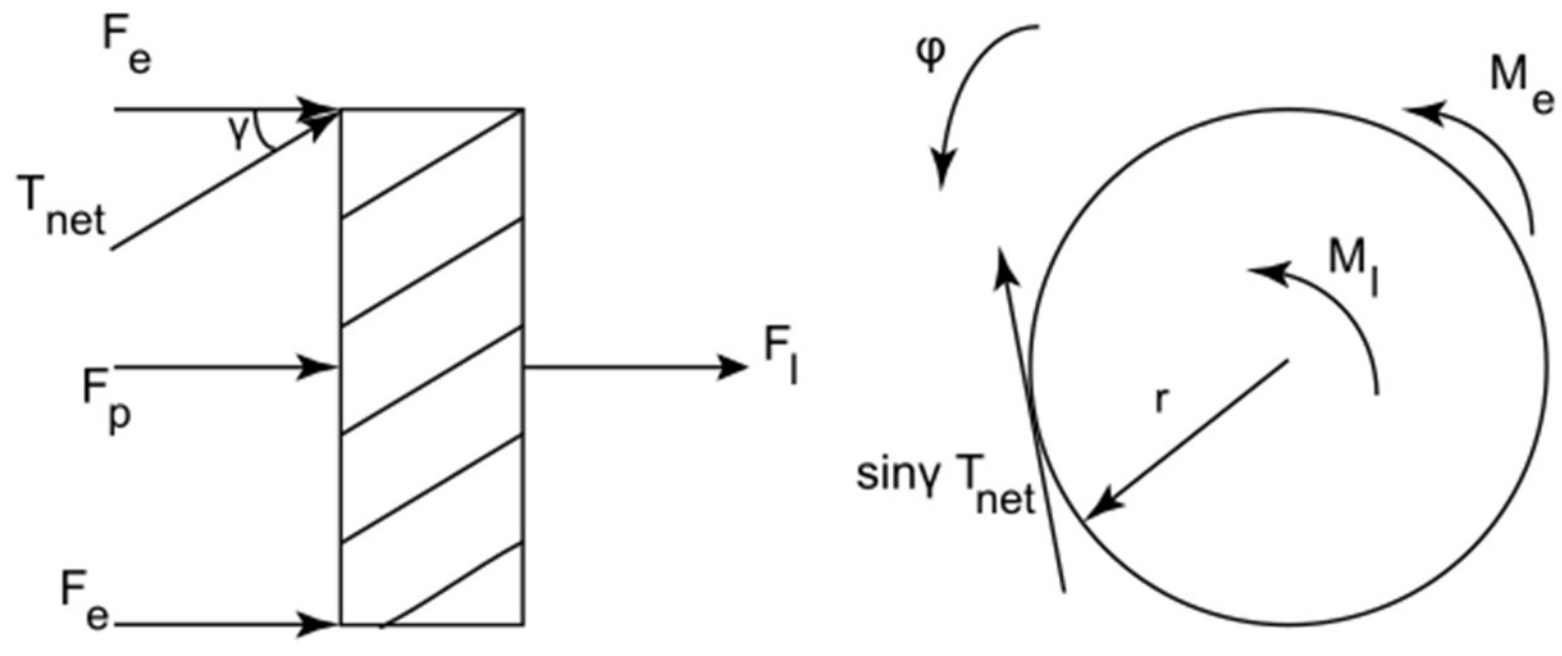

*Figure 2.3. Free body diagram of the end cap*

Writing the force and moment balance according to the Newton's second law:

$$F_l + F_e + T_{net}\cos\gamma + F_P = m_l \frac{d^2 s}{dt^2}\ , \tag{2.3}$$

$$M_l + M_e - rT_{net}\sin\gamma \ = \ I_l \frac{d^2 \varphi}{dt^2}\ , \tag{2.4}$$

where $F_P$ is defined:

$$F_P = P\pi r^2\ . \tag{2.5}$$

## 2.2 Differential Equations of Motion

The FREE is modeled as a thin-walled tube with uniform thickness and diameter along its length, wound with inextensible fibers with a constant cross-section and perfectly adhered to the outer surface of the tube. These assumptions are employed to derive fiber tension and elastomer force and moment.

### Fiber Tension Derivation

The pressure within the tube creates circumferential (hoop) stresses in the tube's wall (Figure 2.4). The stiffness of the fiber is relatively high in comparison to the elastomer and thus the fiber is modeled as inextensible. Hence, nearly all of the circumferential force is carried by the fiber and the contribution of the elastomer can be neglected. Treating the tube as a thin-walled cylinder (no more than one-tenth of its radius) and assuming the fluid pressure is distributed evenly throughout the internal surface of the tube (Beer, 2009), the relationship between the pressure *P* and circumferential stress $\sigma_\theta$ can be described as:

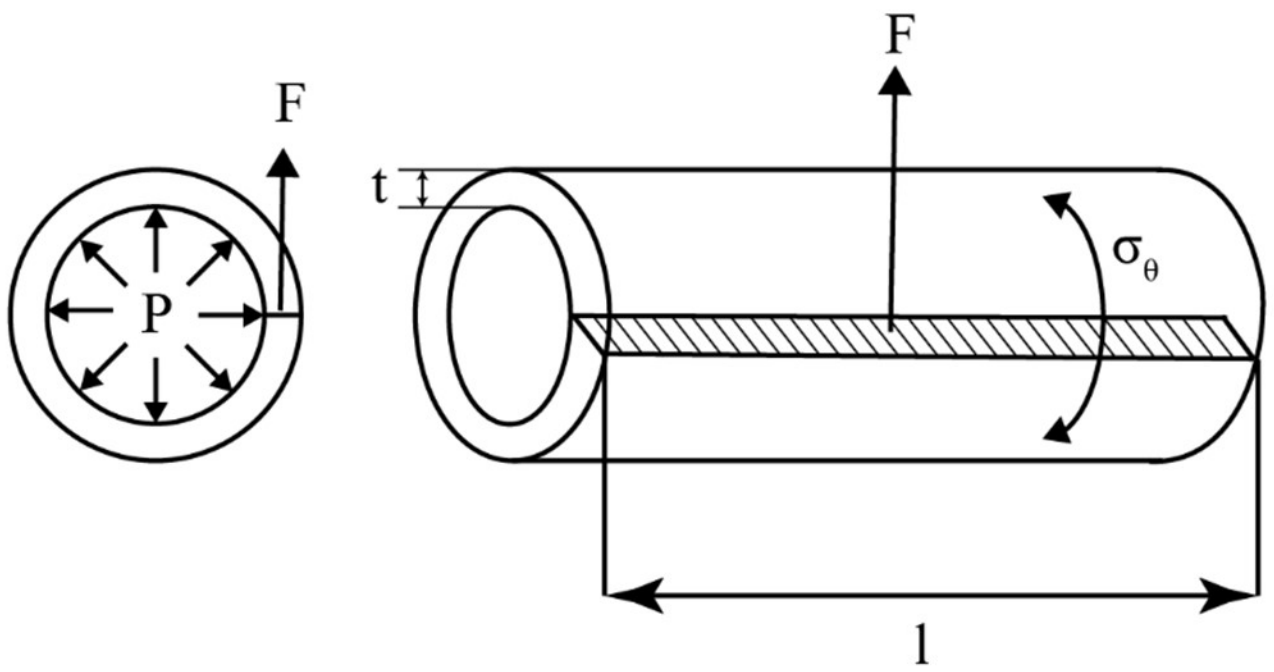

*Figure 2.4. Circumferential hoop stress due to the internal pressure*

$$\sigma_\theta = \frac{Pr}{t} \; , \tag{2.6}$$

where *r* and *t* are the (mean) radius and thickness of the inflated cylinder. The area of the wall with the length *l* is:

$$A = lt \; . \tag{2.7}$$

Hence, the force exerted on the area is:

$$F = Prl \; . \tag{2.8}$$

Figure 2.5 shows a FREE cut in half along the longitudinal direction. By considering the force equilibrium between the fiber tension T and the net force generated (upward) by the pressure on half of the wall of the cylinder $F_y$:

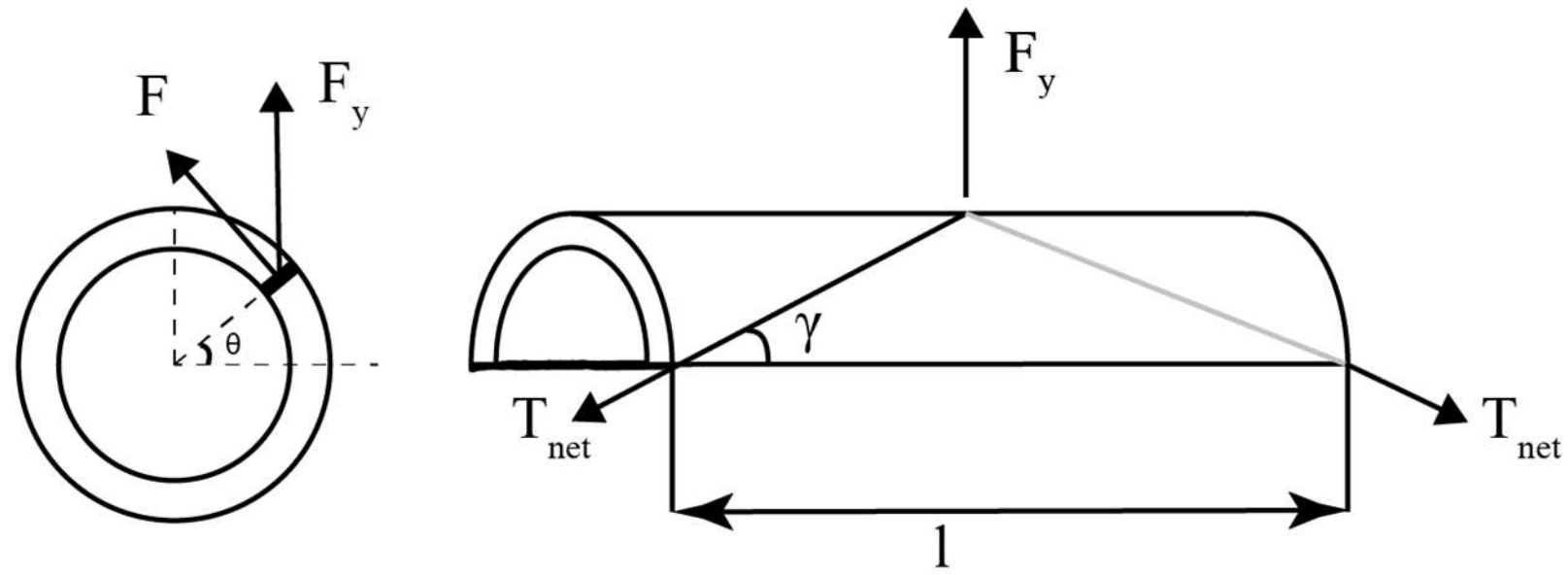

*Figure 2.5. Fiber tension generated by pressure*

$$F_y = 2T_{net} \sin\gamma \,, \tag{2.9}$$

By using Eq. (2.9) integrating $F_y$ over a half of the cylinder:

$$\tan\gamma = \frac{\pi r}{l} \,, \tag{2.10}$$

$$F_y = 2\int_0^{\pi} Prl \sin\theta \, d\theta = 2Prl = 4\pi r^2 P \cot\gamma \,, \tag{2.11}$$

$$T_{net} = 2\pi r^2 P \frac{\cot\gamma}{\sin\gamma} \,. \tag{2.12}$$

Substituting Eqs. (2.5) and (2.12) into Eqs. (2.3) and (2.4) gives:

$$F_l + F_e + \pi r^2 P(1 - 2\cot^2\gamma) = m_l \frac{d^2 s}{dt^2} \,, \tag{2.13}$$

$$M_l + M_e - 2\pi r^3 P \cot\gamma = I_l \frac{d^2\varphi}{dt^2} \,. \tag{2.14}$$

The elastomeric part of a FREE is made of latex which is considered as a nonlinear hyper-elastic material. This non-linearity of latex arises at large strains; for small strains it is roughly linear. To simplify the dynamic model, the force $F_e$ and moment $M_e$ created by the elastomer at low pressures can be modeled as a linear function of the free end extension $s$ and rotation $\varphi$ and their derivatives:

$$F_e = -k_e s - c_e \frac{ds}{dt} \,, \tag{2.15}$$

$$M_e = -k_t \varphi - c_t \frac{d\varphi}{dt} \,. \tag{2.16}$$

where $k_e$ and $k_t$ are the linear and torsional stiffnesses of the FREE, and $c_e$ and $c_t$ are the linear and torsional damping constants of the FREE. The final equations of motion can be obtained by substituting Eqs. (2.15) and (2.16) into Eqs. (2.13) and (2.14):

$$F_l - k_e s - c_e \frac{ds}{dt} + \pi r^2 P(1 - 2\cot^2\gamma) = m_l \frac{d^2 s}{dt^2} \,, \tag{2.17}$$

$$M_l - k_t \varphi - c_t \frac{d\varphi}{dt} - 2\pi r^3 P \cot\gamma = I_l \frac{d^2\varphi}{dt^2} \,. \tag{2.18}$$

## Numerical Solution

The mathematical model of a FREE represented by Eqs. (2.17) and (2.18) can be used to analyze the response of the system numerically. In Eqs. (2.17) and (2.18), $s$ and $\varphi$ are the generalized coordinates that describe the overall motion of the FREE. In order to test the correctness and functionality of the model, a set of simple tests with a constant pressure are developed (see Appendix A for the MATLAB script), and results for rotation and elongation of a FREE ($\Gamma$= -40°, $L = 11$ cm, and $R = 0.7$ cm) are illustrated in Figure 2.6 to Figure 2.13. To solve the equations of motion the *"ode45"* solver in MATLAB was used. All of the assumptions and constraints have been defined as simply as possible in this analysis.

System remains at initial conditions with no internal pressure, stiffness, and damping:

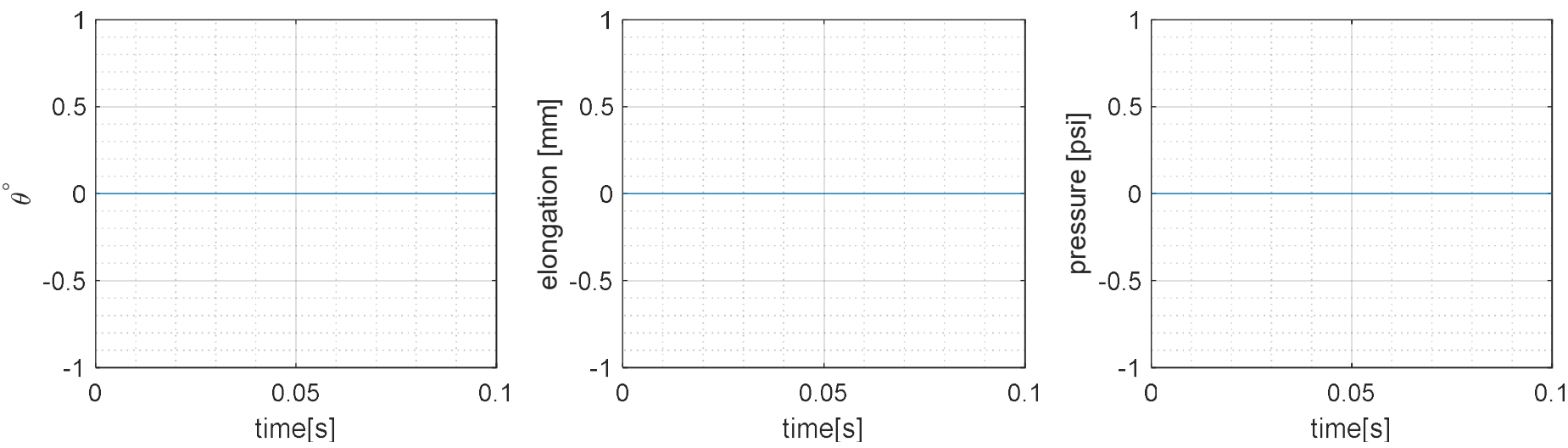

*Figure 2.6. Simulated response with* $P = 0;\ F_l = 0;\ M_l = 0;\ c_e, c_t = 0;\ k_e, k_t = 0$

System has no internal pressure, stiffness, and damping. The external load $F_{gravity}$ (weight of the end cap) makes the elongation monotonically increase:

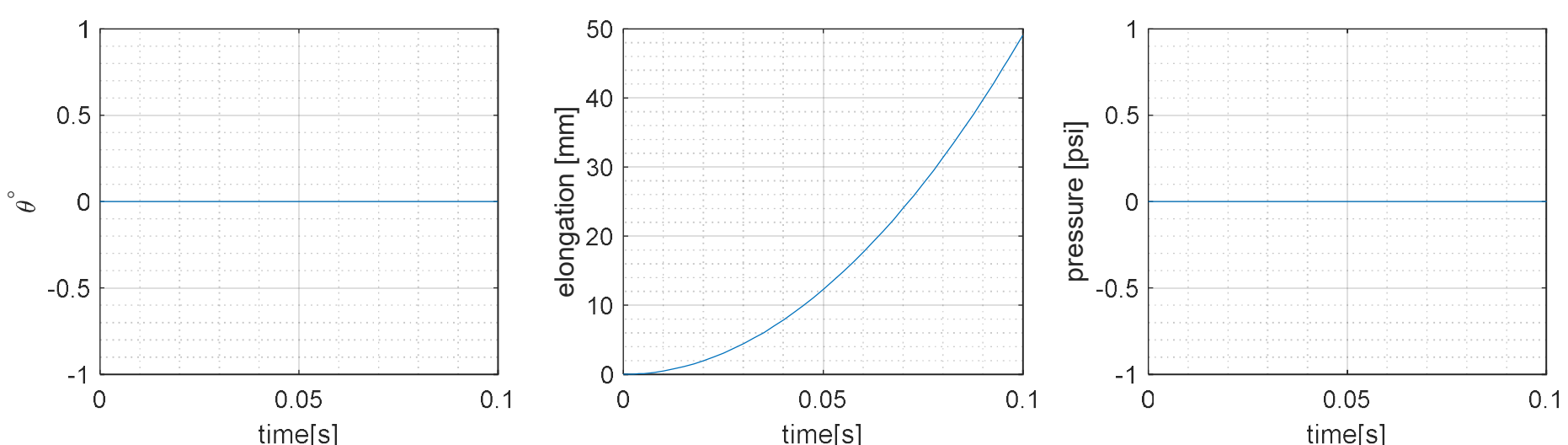

*Figure 2.7. Simulated response with* $P = 0;\ F_l = F_{gravity};\ M_l = 0;\ c_e, c_t = 0;\ k_e, k_t = 0$

System has no internal pressure and damping. The external load $F_{gravity}$ makes the elongation oscillate, starting from zero position (I.C.):

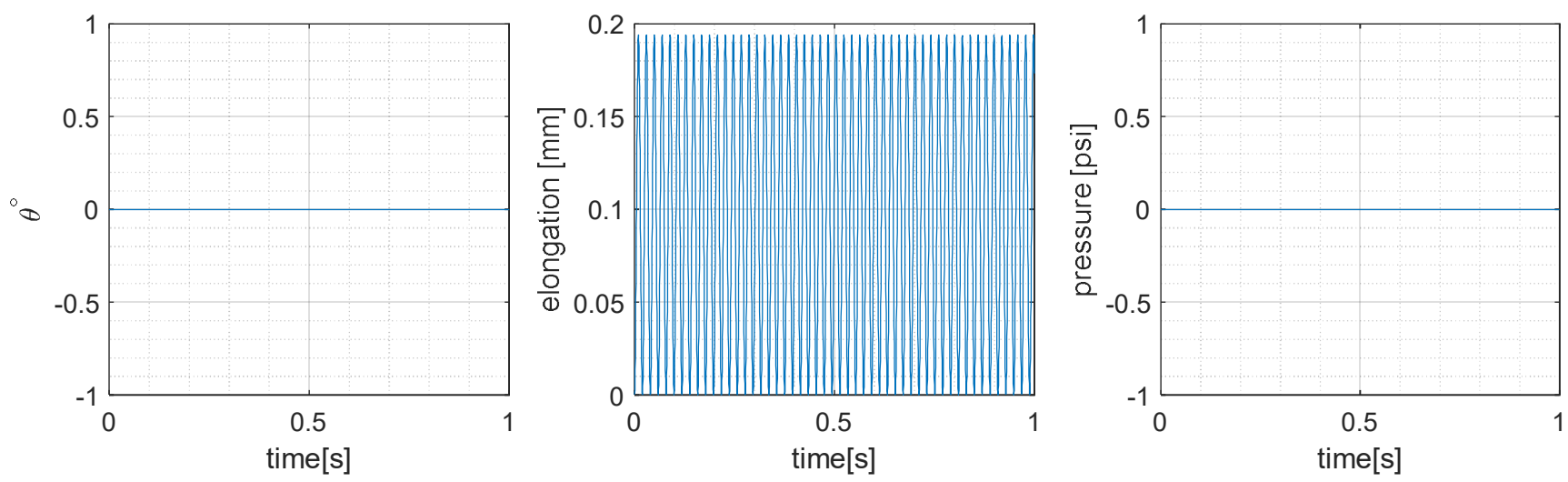

*Figure 2.8. Simulated response with* $P = 0;\ F_l = F_{gravity};\ M_l = 0;\ c_e, c_t = 0;\ k_e = 10110\ \frac{N}{m};\ k_t = 0.18\ \frac{Nm}{rad}$; *zero I.C.*

System with no internal pressure and damping maintains the equilibrium position (displacement caused by the weight of the end cap):

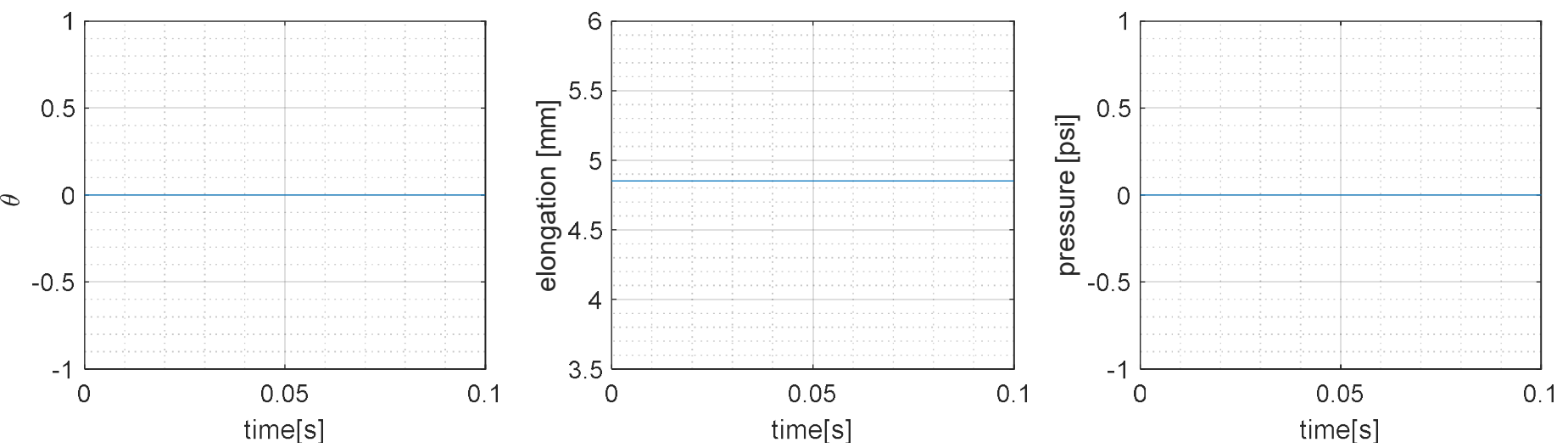

*Figure 2.9. Simulated response with* $P = 0;\ F_l = F_{gravity};\ M_l = 0;\ c_e, c_t = 0;\ k_e = 10110\ \frac{N}{m};\ k_t = 0.18\ \frac{Nm}{rad}; equilibrium\ I.C.$

System continuously deforms with no stiffness and no damping at a constant pressure:

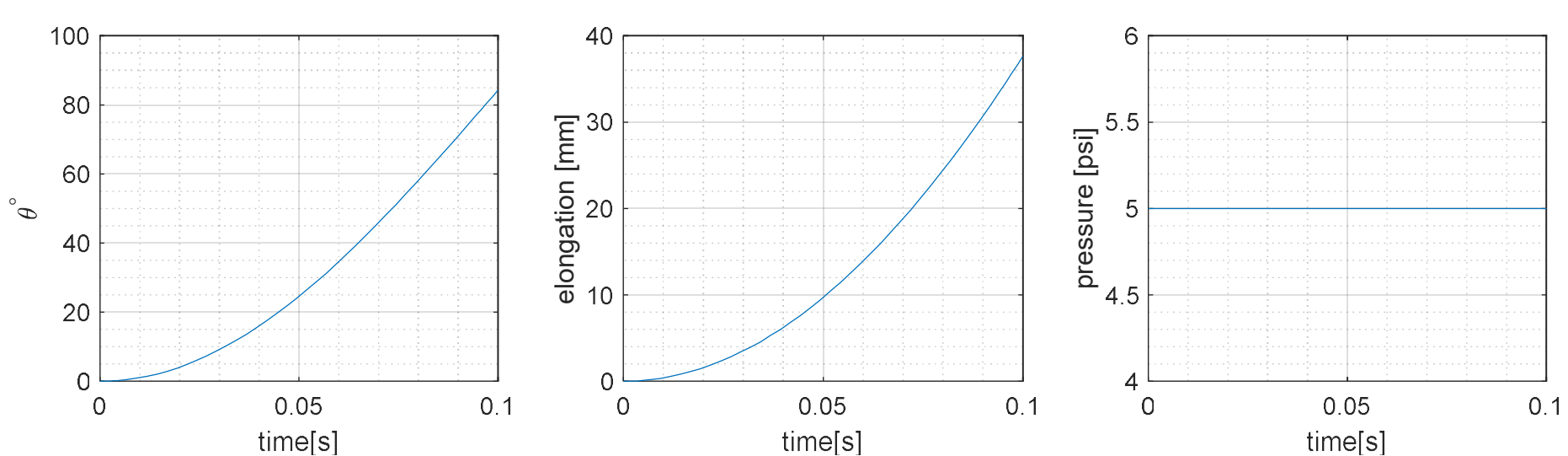

*Figure 2.10. Simulated response with* $P = 5\,psi;\ F_l = 0;\ M_l = 0;\ c_e, c_t = 0;\ k_e, k_t = 0$

System rotates and elongates monotonically with damping and no stiffness at a constant pressure:

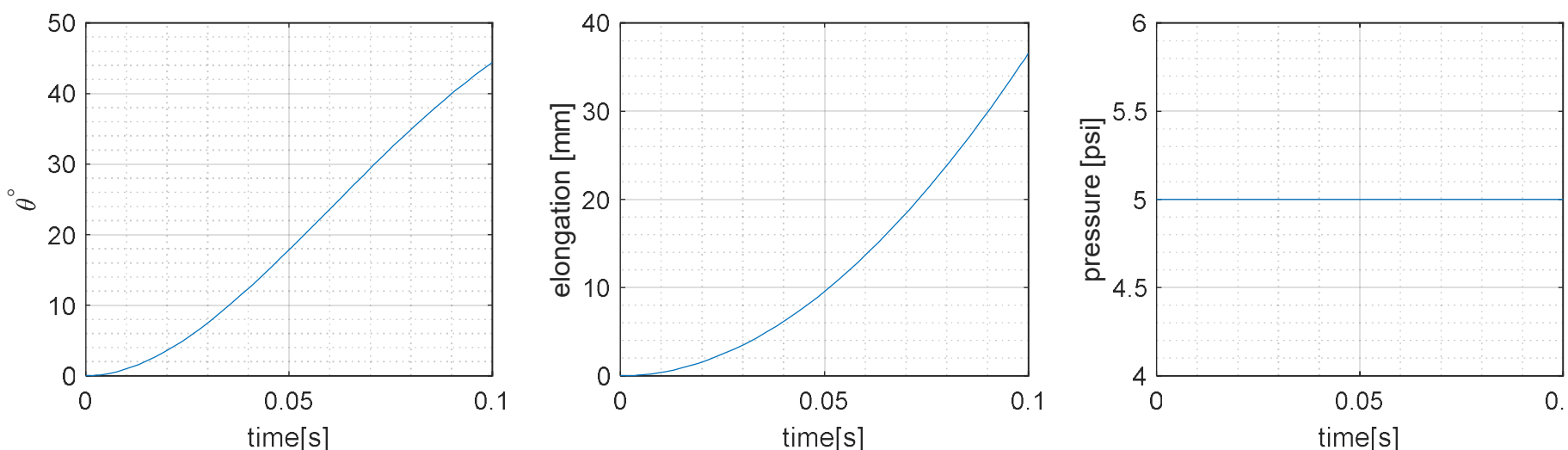

*Figure 2.11. Simulated response with* $P = 5\ psi;\ F_l = 0;\ M_l = 0;\ c_e = 5\ \frac{N.s}{m}, c_t = 0.005\ \frac{Nm.s}{rad};\ k_e, k_t = 0$

System rotates and elongates oscillatory with damping, and no stiffness at a constant pressure:

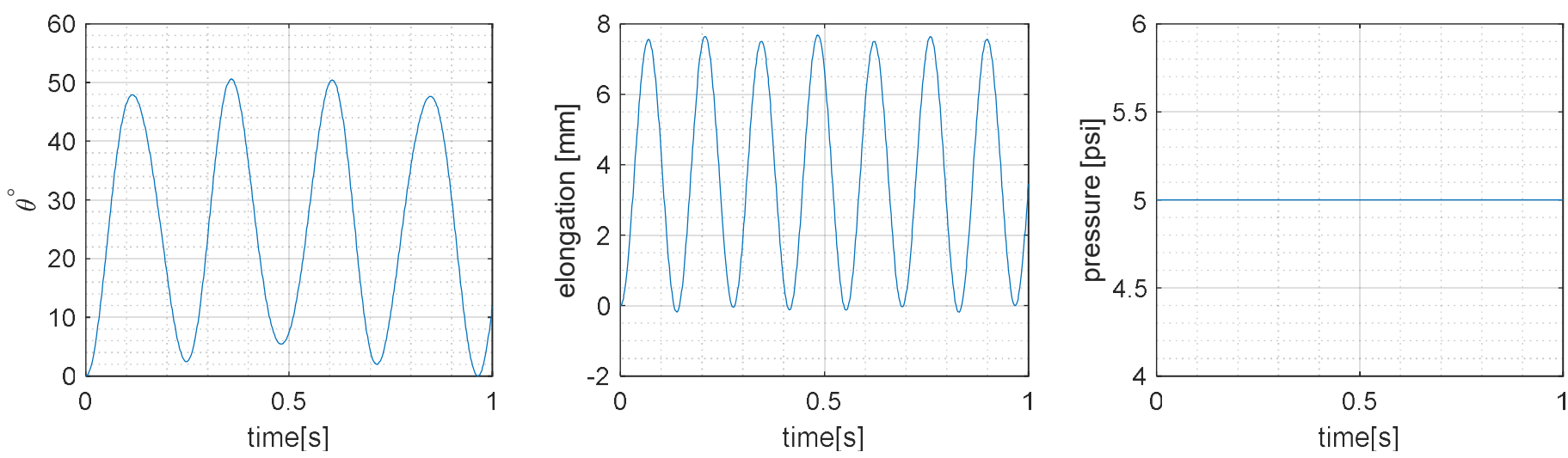

*Figure 2.12. Simulated response with* $P = 5\ psi;\ F_l = 0;\ M_l = 0;\ c_e, c_t = 0;\ k_e = 10110\ \frac{N}{m};\ k_t = 0.18\ \frac{Nm}{rad}$

System reaches a steady-state rotation and elongation at a constant pressure:

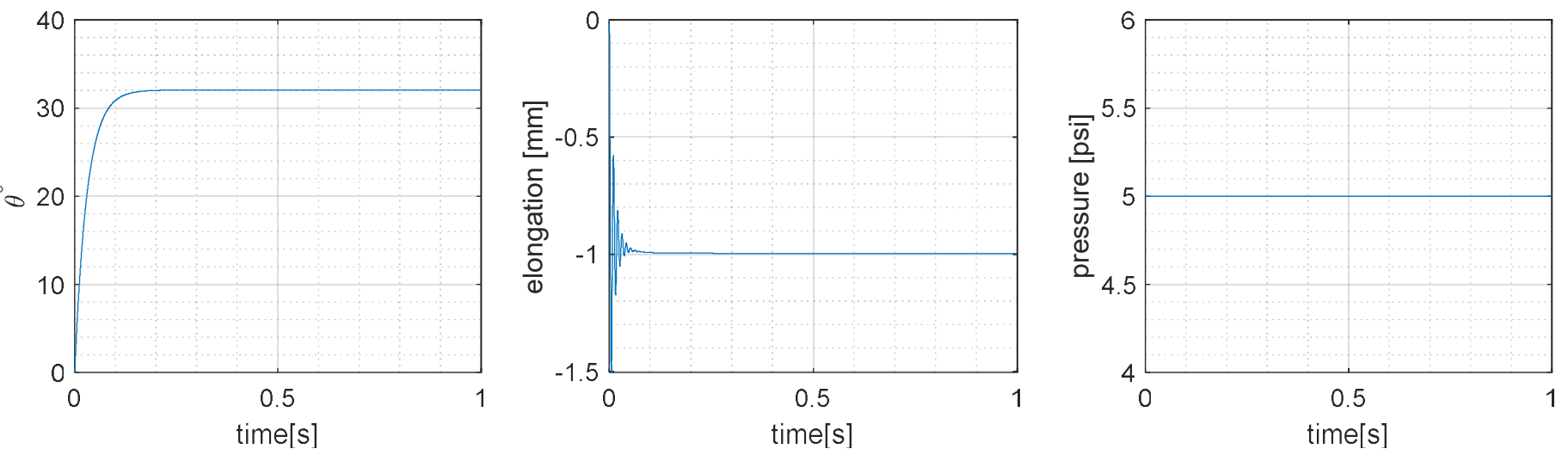

*Figure 2.13. Simulated response with* $P = 5;\ F_l = 0;\ M_l = 0;\ c_e = 5\ \frac{N.s}{m}, c_t = 0.005\ \frac{Nm.s}{rad};\ k_e = 10110\ \frac{N}{m};\ k_t = 0.18\ \frac{Nm}{rad}$

All of the simple tests verify the fidelity of the lumped-parameter model for various conditions. Thus, this model can provide practical insight into influence of each parameter to alter the behavior of the system. One of the dominant parameters defined for the FREE in the lumped-parameter model is torsional stiffness ($k_t$). Figure 2.14 shows the dynamic response of rotation of a specific FREE ($\Gamma = 40°$, $L = 11$ cm, $R = 0.7$ cm, $F_l = F_{gravity}$,

$M_l = 0$, $k_e = 10110\ \frac{\mathrm{N}}{\mathrm{m}}$, $c_e = 5\ \frac{\mathrm{N.s}}{\mathrm{m}}$ , and $c_t = 0.005\ \frac{\mathrm{Nm.s}}{\mathrm{rad}}$) with different torsional stiffnesses $k_t$ where pressurized up to 10 psi. Note that the negative direction of rotation is due to selection of a positive winding angle.

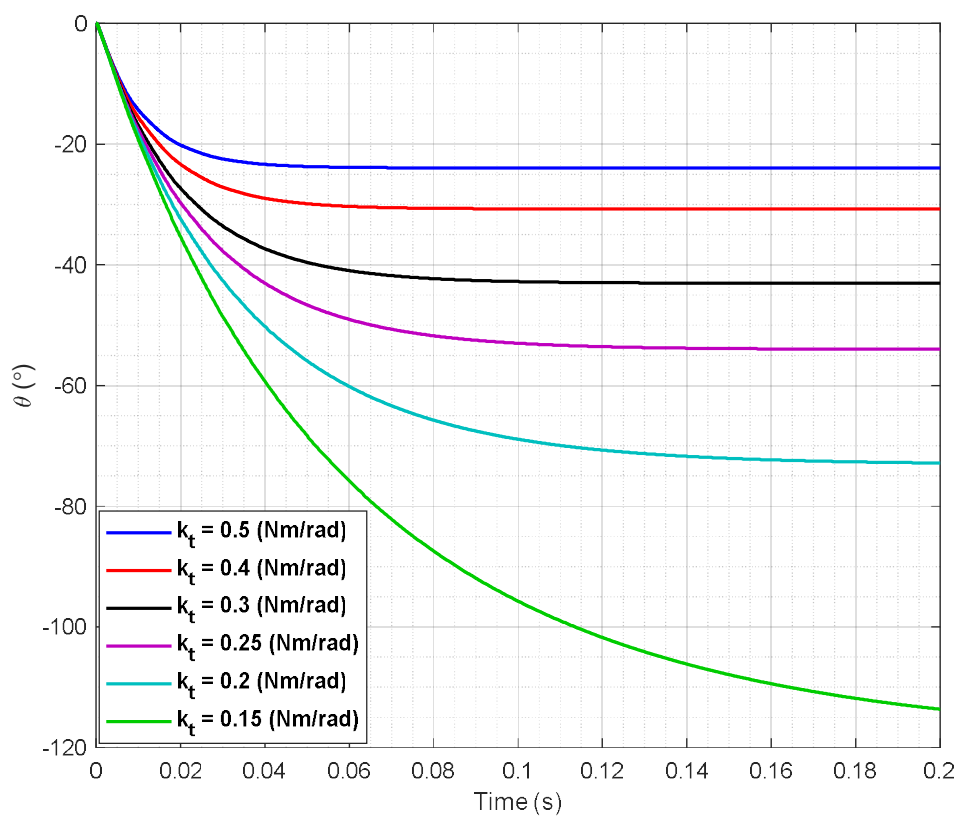

*Figure 2.14. Dynamic response of a 40° FREE using various torsional stiffnesses*

The above simulation illustrates the response of the simple lumped parameter model of the FREE and suggests that as the torsional stiffness decreases, the rotation of the FREE increases (similar to a torsional spring). It also indicates that a difference of 0.35 $\frac{Nm}{rad}$ in torsional stiffness of the 40° FREE changes the rotation angle 90° at 0.2 s shown in the graph. This demonstrates the importance of selecting appropriate system parameters in the design of a FREE. Additional simulation results comparing are presented in Chapters 3 and 5.

## 2.3 Summary

This chapter had presented a lumped-parameter model that simplifies the modeling of a FREE and accurately describes the behavior of the system under certain assumptions. The model is developed based on the relationship between the applied forces, moments, and resulting reactions of a FREE when fixed at one end, and determines a correlation between the internal pressure and displacements (i.e., rotation and elongation) of the FREE. The lumped-parameter model enables the designer to study the dynamic behavior of FREEs with a variety of geometries and loading conditions.

# 3
# Controller Design

In Chapter 2 a simple model was developed to calculate the displacement and rotation of a FREE at the free end by knowing the geometric parameters, stiffness/damping factors, internal pressure, and external loads. Similar to other types of robotic actuators for robotic applications, the motion of the FREE needs to be controlled. This chapter mainly discusses the controllability of the rotation of a single FREE by applying a PID (Proportional-plus-Integral-plus-Derivative) controller. The root locus method is used to tune the control gains and the response of the system is studied for each gain variation. Additionally, a trajectory following controller is explored for a single FREE.

## 3.1 Proportional-plus-Integral-plus-Derivative (PID)

One of the motivations for studying FREEs is the goal of using them as robotic actuators with controlled behaviors including motion, force, and torque. Elongation, rotation, and expansion are the primary motion characteristics of a FREE. Considered in this section is the control of the rotational angle $\varphi$. The controller measures the error between the desired rotation angle and the actual angle and calculates the input pressure $P$ based on Eq. (3.1):

$$P = K_p(\varphi_d - \varphi) - K_d\dot{\varphi} + K_i \int (\varphi_d - \varphi)dt \ , \tag{3.1}$$

where the terms in the equation can be defined as:

$K_p$ – proportional gain (constant)
$K_i$ – integral gain (constant)
$K_d$ – derivative gain (constant)
$\varphi_d$ – desired rotation angle (constant)
$\varphi$ – rotation angle (variable)
$t$ – time (variable)

The controller represented by Eq. (3.1) is known as a PID controller because the control variable $P$ is proportional to the difference between the desired and measured variable $\varphi$ (the error), the integral of the error, and the derivative of the error. To apply the PID controller to the study of the closed-loop behavior of a FREE, Eq. (3.1) must first be substituted into Eqs. (2.17) and (2.18) to give:

$$m_l\ddot{s} = \pi r^2(1-2\cot^2\gamma)\left(K_p(\varphi_d-\varphi) - K_d\dot{\varphi} + K_i\int(\varphi_d-\varphi)dt\right) - k_e s - c_e\dot{s} + F_l\,, \tag{3.2}$$

$$I_l\ddot{\varphi} = (-2\pi r^3\cot\gamma)\left(K_p(\varphi_d-\varphi) - K_d\dot{\varphi} + K_i\int(\varphi_d-\varphi)dt\right) - k_t\varphi - c_t\dot{\varphi} + M_l\,. \tag{3.3}$$

## 3.2 Control System Design and Analysis by the Root-Locus Method

The response of a linear closed-loop system is directly related to the location of the closed-loop poles (the roots of the characteristic equation), which are a function of controller gains. Generally, in the design of a control system, the controller is optimized rather than modifying the system dynamics due to the impracticality of making changes to the physical system. Control engineers thus seek to determine suitable control parameters to reach the desired performance. In this way, the response of a system can be adjusted by simply changing control gains and in the case of controlling the FREE here, the PID control gains. Experimentally determining suitable gains, particularly in a complex system is tedious and sometimes misleading. One well-known method for finding the roots of the characteristic equation corresponding to particular gains was developed by W. R Evans and is called the root-locus method (Evans, 1954). This section discusses using the root-locus method to find the appropriate PID gains to achieve the desired system performance. The closed-loop transfer function of a negative feedback control system is depicted in the block diagram in Figure 3.1.

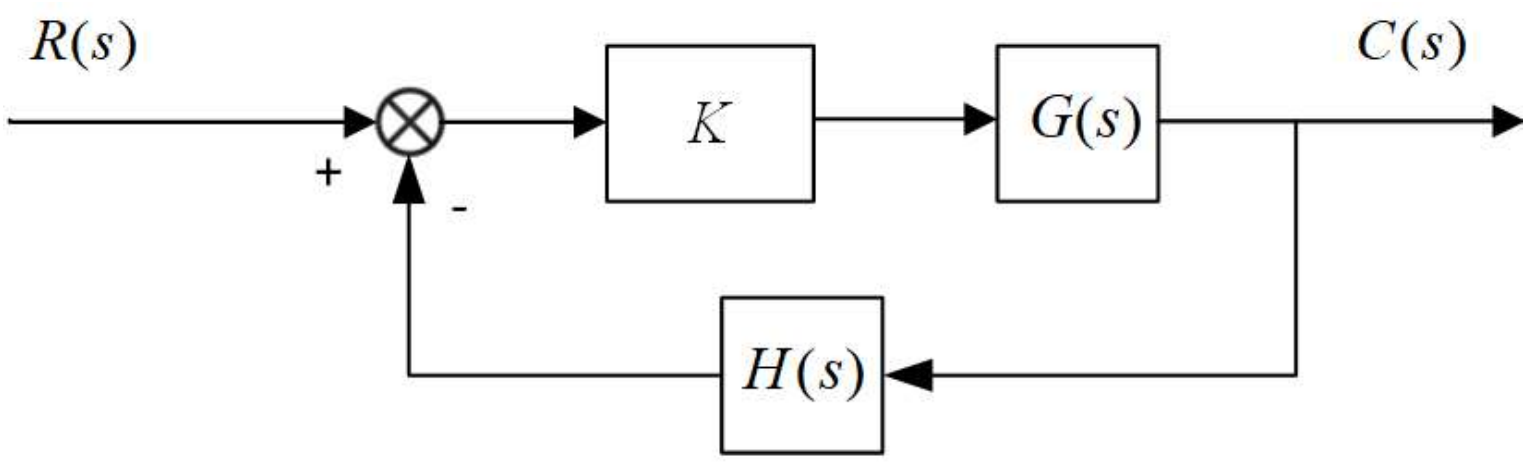

*Figure 3.1. Block diagram of the PID control system*

The relation between the desired reference input and the output of the system is:

$$\frac{C(s)}{R(s)} = \frac{KG(s)}{1 + KG(s)H(s)} \,, \tag{3.4}$$

where $C(s)$, $G(s)$, $H(s)$, and $R(s)$ are the system output, plant transfer function, feedback transfer function, and reference input of the closed-loop control system, respectively, and K is the controller gain. The roots of the characteristic equation are the values of $s$ that cause the denominator of Eq. (3.4) to be equal to zero. In general, the root-locus method enables all the roots to be plotted based on varying a particular gain from 0 to ∞ so as to graphically explore the behavior of the system.

**Linearization of the Nonlinear Mathematical Model**

To use the root locus method a linear representation of Eqs. (3.2) and (3.3) must be created by a Taylor series expansion with respect to $\varphi$ and $s$. All terms of these equations are already linear except the boxed terms:

$$m_l\ddot{s} = \boxed{\pi r^2(1 - 2\cot^2\gamma)\left(-K_p\varphi - K_d\dot{\varphi} - K_i\int \varphi dt\right)} - k_e s - c_e\dot{s} + F_l \,, \tag{3.5}$$

$$I_l\ddot{\varphi} = \boxed{(-2\pi r^3\cot\gamma)\left(-K_p\varphi - K_d\dot{\varphi} - K_i\int \varphi dt\right)} - k_t\varphi - c_t\dot{\varphi} + M_l \,. \tag{3.6}$$

Note that both radius of the FREE $r$ and winding angle $\gamma$ are nonlinear based on Eqs. (2.1) and (2.2). The nonlinear terms are linearized by using the definition of a Taylor series expansion [Eq. (3.7)] about the equilibrium state $s = \varphi = 0$:

$$f(s,\varphi) \approx f(0,0) + s\left.\frac{\partial f}{\partial s}\right|_{\substack{s=0\\ \varphi=0}} + \varphi\left.\frac{\partial f}{\partial \varphi}\right|_{\substack{s=0\\ \varphi=0}} + \cdots \,, \tag{3.7}$$

This gives:

$$\gamma = \cos^{-1}\left[\frac{(L+s)\cos\Gamma}{L}\right] \approx \Gamma - \frac{s}{L\cot\Gamma} + \cdots \,, \tag{3.8}$$

$$r = \frac{L\tan\gamma}{\frac{L\tan\Gamma}{R} + \varphi} \approx R - \frac{sR}{L\sin^2\Gamma} + \frac{\varphi R^2}{L\tan\Gamma} + \cdots \,. \tag{3.9}$$

Similarly, Eqs. (3.10) to (3.13) are linearized expressions for the terms $\cot\gamma$, $\cot^2\gamma$, $r^2$, and $r^3$.

$$cot\,\gamma \approx cot\,\Gamma + \frac{s\,cot\,\Gamma}{L\,sin^2\,\Gamma} + \cdots \ , \tag{3.10}$$

$$cot^2\,\gamma \approx cot\,\Gamma \left(1 + \frac{2s}{L\,sin^2\,\Gamma}\right) + \cdots \ , \tag{3.11}$$

$$r^2 \approx R^2 - \frac{2sR^2}{L\,sin^2\,\Gamma} + \frac{2\varphi R^3}{L\,tan\,\Gamma} + \cdots \ , \tag{3.12}$$

$$r^3 \approx R^3 - \frac{3sR^3}{L\,sin^2\,\Gamma} + \frac{3\varphi R^4}{L\,tan\,\Gamma} + \cdots \ . \tag{3.13}$$

In all of the above equations, only the linear terms are retained in the Taylor series expansion. The linearized version of the nonlinear terms in Eqs. (3.5) and (3.6) can be obtained:

$$\begin{aligned} r^2\,cot^2\,\gamma \approx\ & R^2\,cot^2\,\Gamma \left[1 + s\left(\frac{-8s}{L^2\,sin^2\,\Gamma} + \frac{4\varphi R}{L^2\,sin^2\,\Gamma\,tan\,\Gamma}\right)\Bigg|_{\substack{s=0\\ \varphi=0}} \right. \\ & \left. + \varphi\left(\frac{2R}{L\,tan\,\Gamma} + \frac{4sR}{L^2\,sin^2\,\Gamma\,tan\,\Gamma}\right)\Bigg|_{\substack{s=0\\ \varphi=0}}\right] + \cdots \\ & = R^2\,cot^2\,\Gamma\left(1 + \frac{2\varphi R}{L\,tan\,\Gamma}\right) + \cdots \ , \end{aligned} \tag{3.14}$$

$$\begin{aligned} r^3\,cot\,\gamma \approx\ & R^3 cot\,\Gamma \left[1 - s\left(\frac{-2}{L\,sin^2\,\Gamma} - \frac{6s}{L^2\,sin^4\,\Gamma} + \frac{3\varphi R}{L^2\,sin^2\,\Gamma\,tan\,\Gamma}\right)\Bigg|_{\substack{s=0\\ \varphi=0}} \right. \\ & \left. + \varphi\left(\frac{3R}{L\,tan\,\Gamma} + \frac{3sR}{L^2\,sin^2\,\Gamma\,tan\,\Gamma}\right)\Bigg|_{\substack{s=0\\ \varphi=0}}\right] + \ldots \\ & = R^3\,cot\,\Gamma\left(1 - \frac{2s}{L\,sin^2\,\Gamma} + \frac{3\varphi R}{L\,tan\,\Gamma}\right) + \cdots \ . \end{aligned} \tag{3.15}$$

By substituting Eqs. (3.12) and (3.14) into Eq. (3.5), Eq. (3.15) into Eq. (3.6), and retaining only linear terms, the final linearized equations of motion of the FREE about the equilibrium state $s = \varphi = 0$ are:

$$m_l\ddot{s} = \pi R^2(1 - 2\cot^2\Gamma)\left(-K_p\varphi - K_d\dot{\varphi} - K_i\int \varphi dt\right) - k_e s - c_e\dot{s} + F_l \ , \tag{3.16}$$

$$I_l\ddot{\varphi} = (-2\,\pi R^3 \cot\Gamma)\left(-K_p\varphi - K_d\dot{\varphi} - K_i\int \varphi dt\right) - k_t\varphi - c_t\dot{\varphi} + M_l \ . \tag{3.17}$$

Equations (3.16) and (3.17) represent the motion of a FREE about the equilibrium state $s = \varphi = 0$. These equations would be essentially the same if linearized about another equilibrium state, which allows any desired angle of rotation $\varphi_d$ to be studied.

As depicted in Figure 3.2 the numerical solution for the response of a FREE with $\Gamma = 40°$, $L = 11$ cm, $R = 0.7$ cm, $F_l = F_{gravity}$, $M_l = 0$, $k_e = 10110\ \frac{\text{N}}{\text{m}}$, $c_e = 5\ \frac{\text{N.s}}{\text{m}}$ , and $c_t = 0.005\ \frac{\text{Nm.s}}{\text{rad}}$, controlled with a PD controller ($K_p = 32000\ \frac{\text{Pa}}{\text{rad}}$ and $K_i = 1200000\ \frac{\text{Pa}}{\text{rad.s}}$), shows that the radius and the winding angle do not change more than 2%. The model based on linearized equations of motion reaches a 25° rotation angle (set-point) as fast as the model based on the fully nonlinear equations–the overall response of the system is approximately the same. The only significant difference are the pressures to reach the set-point and 8.5% more elongation with the linear model. This shows that Eqs. (3.16) and (3.17) can determine the overall motion of the FREE to a good approximation as compared with the nonlinear model represented by Eqs. (3.2) and (3.3).

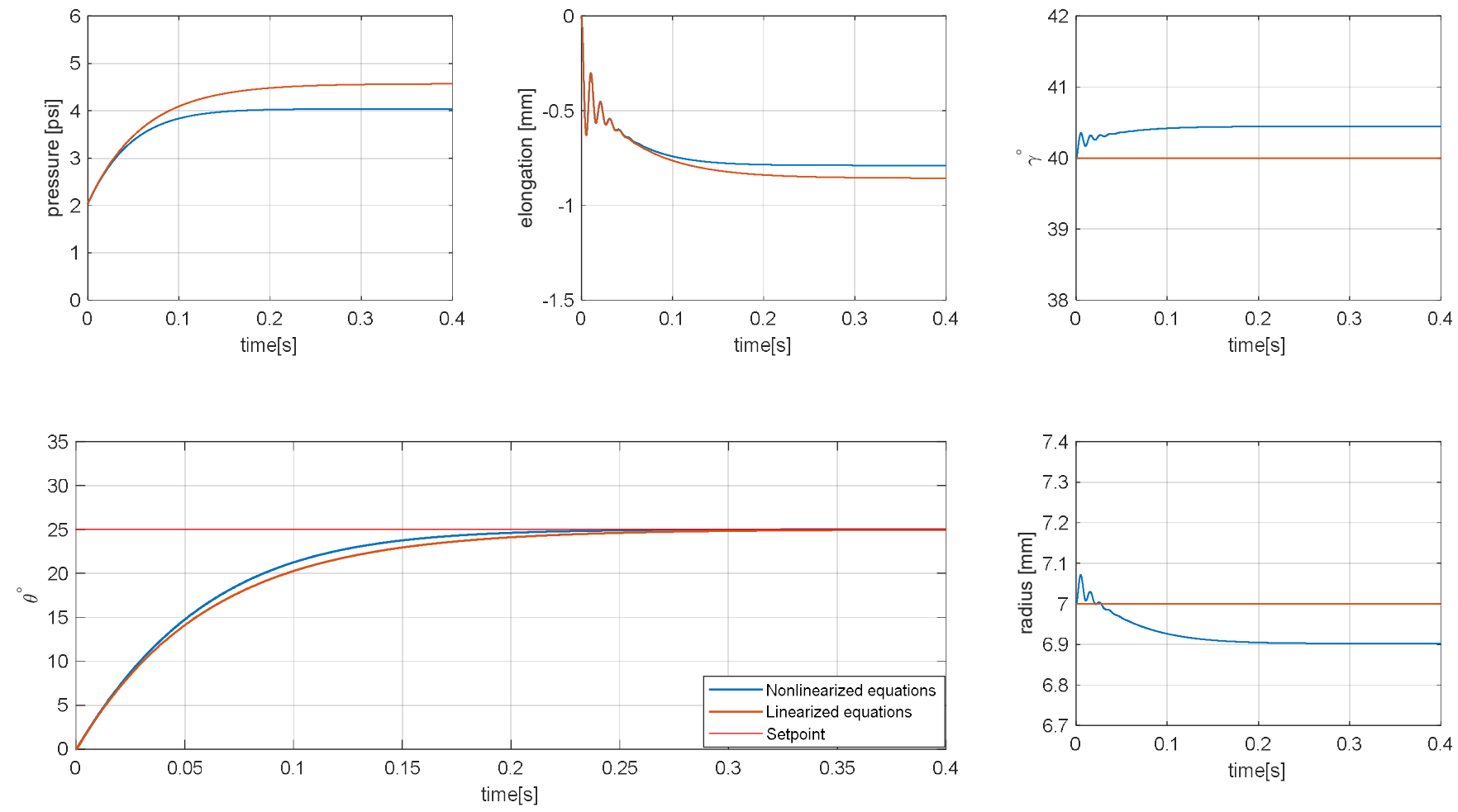

*Figure 3.2. Dynamic response of a FREE with nonlinear Eqs. (3.2) and (3.3); and with linearized Eqs. (3.16) and (3.17)*

Further analysis of the system can be based on the use of Eqs. (3.16) and (3.17), which can be transformed into the Laplace domain after defining the constants:

$$C_1 = \pi R^2(1 - 2\cot^2\Gamma) \ , \tag{3.18}$$

$$C_2 = 2\,\pi R^3 \cot\Gamma \ . \tag{3.19}$$

The dynamical equations of motion can then be expressed in the Laplace domain as follows:

$$m_l s^2 X(s) + k_e X(s) + c_e s X(s) - C_1\left[\left(K_p - K_d s + \frac{K_i}{s}\right)Y(s) - \frac{Y_d\left(K_i + K_p s\right)}{s^2}\right] = 0\,, \tag{3.20}$$

$$I_l s^2 Y(s) + c_t s Y(s) + k_t Y(s) - C_2\left[\left(K_p - K_d s + \frac{K_i}{s}\right)Y(s) - \frac{Y_d\left(K_i + K_p s\right)}{s^2}\right] = 0\,. \tag{3.21}$$

where $X(s)$ and $Y(s)$ are the Laplace transform of the elongation $s$ and rotation $\varphi$, respectively.

Since rotation of the FREE is the variable of primary interest, Eq. (3.21) will be used to generate root loci for various PID gains. Accordingly, $Y_d$ and $Y(s)$ are the reference input and output of the closed-loop control system of the FREE. Considering each of $K_p$, $K_i$, and $K_d$ as the control gains of interest, three relationships for the rotation between $Y(s)$ and $Y_d$ can be produced.

Considering $K_p$ variation:

$$Y(s)/Y_d = \frac{\dfrac{-C_2(K_i + K_p s)}{s(I_l s^3 + c_t s^2 + k_t s - C_2 K_i)}}{1 + K_p \dfrac{-q_2 s}{(I_l s^3 + c_t s^2 + k_t s - C_2 K_i)}}\ , \tag{3.22}$$

Considering $K_i$ variation:

$$Y_s/Y_d = \frac{\dfrac{-C_2(K_i + K_p s)}{s(I_l s^3 + c_t s^2 + (k_t - C_2 K_p)s)}}{1 + K_i \dfrac{-C_2}{(I_l s^3 + c_t s^2 + (k_t - C_2 K_p)s)}}\ , \tag{3.23}$$

Considering $K_d$ variation:

$$ {Y_s}/{Y_d} = \frac{\dfrac{-C_2(K_i + K_p s)}{s\left(I_l s^3 + c_t s^2 + \left(k_t - C_2 K_p\right)s - C_2 K_i\right)}}{1 + K_d \dfrac{-C_2 s}{\left(I_l s^3 + c_t s^2 + \left(k_t - C_2 K_p\right)s - C_2 K_i\right)}} \,. \tag{3.24} $$

**Tuning the PID Controller**

There are various strategies for tuning the PID gains depending on the features of a physical system. The most common method is to vary one gain at a time, determine the gain corresponding to the best response, and then repeat the process for the other gains. In the case of a FREE, the root locus has first been plotted for all values of $K_p$ values using Eq. (3.22) with $K_i = 1200000 \; \frac{\text{Pa}}{\text{rad.s}}$ and $K_d = 0 \; \frac{Pa.s}{rad}$. After selecting the "best" value of $K_p$, similar processes are followed for $K_i$ and $K_d$ using Eqs. (3.23) and (3.24). For generating root loci and finding the desired information from the plots, a very simple MATLAB function "*rlocus(num,dem,K)*" has been used (see Appendix A). The "*rlocus*" function computes and plots the closed-loop poles as a function of the values of the gain *K* with *num* and *dem* obtained by expressing the denominator of Eq. (3.4) in the form:

$$ 1 + K\frac{num}{den} = 0 \,, \tag{3.25} $$

Root loci were generated for a FREE with $\Gamma$= 40°, $L = 11$ cm, $R = 0.7$ cm, $F_l = F_{gravity}$, $M_l = 0$, $k_e = 10110 \; \frac{\text{N}}{\text{m}}$, $c_e = 5 \; \frac{\text{N.s}}{\text{m}}$, and $c_t = 0.005 \; \frac{\text{Nm.s}}{\text{rad}}$, and have been sketched (using MATLAB). Note that a general solution of a second-order, ordinary, linear differential equation with constant coefficients is:

$$ x(t) = Ce^{st}, \tag{3.26} $$

where *C* and *s* are constants. Eq. (3.17) has three roots and thus the response of the system is the summation of three exponential solutions. The values of *s are* the closed-loop poles and determine the response.

Figures 3.3, 3.4, and 3.5 are the root loci for $K_p$, $K_i$, and $K_d$. In each figure, poles on the right side of the imaginary axis cause an unstable response because the response is growing exponentially. Similarly, roots above and below the real axis correspond to poles that creating an oscillatory (underdamped) response. The poles located far to the left of the imaginary axis generally do not strongly influence the response since their contribution to the response dissipates quickly due to the large negative values of the real parts of the closed-loop poles. Generally, gains that produce poles that are located near the real axis

and relatively close to the imaginary axis produce a rapid response with minimal oscillations.

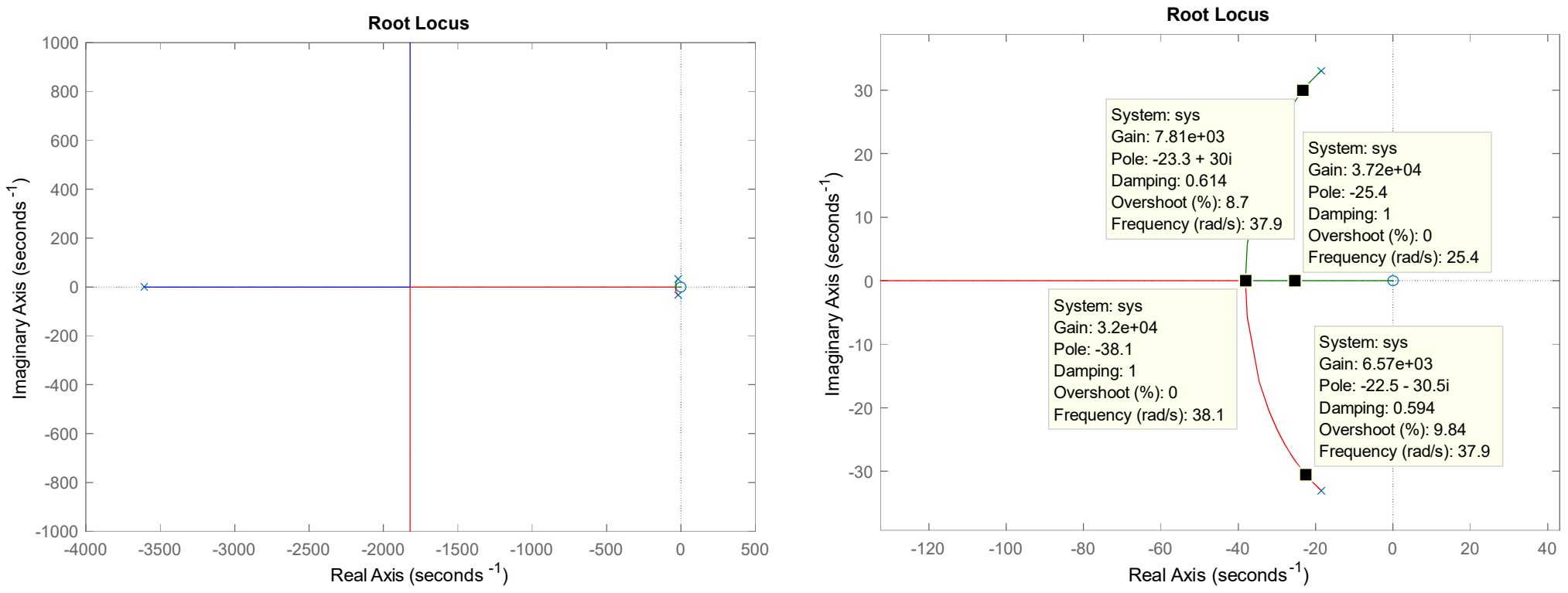

*Figure 3.3. Root-locus plot for* $K_p$ *(*$K_i = 1200000\ \frac{Pa}{rad.s}$*;* $K_d = 0\ \frac{Pa.s}{rad}$*)*

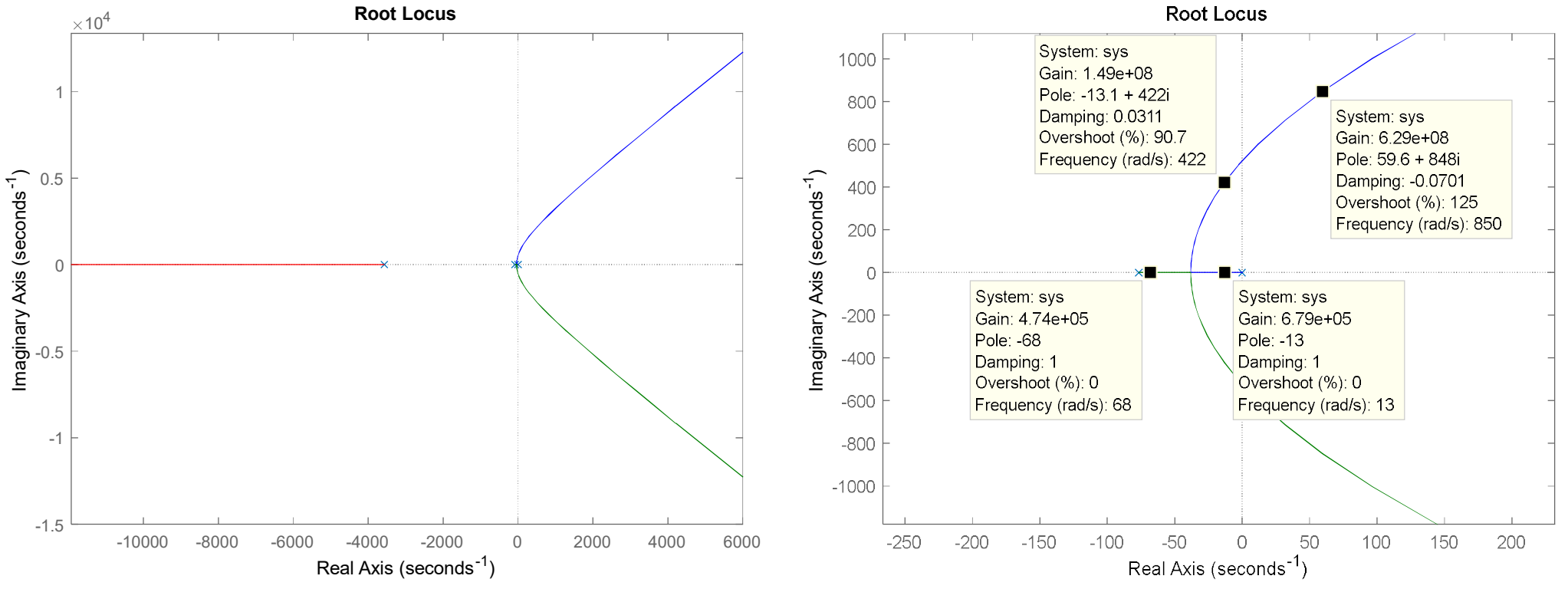

*Figure 3.4. Root-locus plot for* $K_i$ *(*$K_p = 32000\ \frac{Pa}{rad}$*;* $K_d = 0\ \frac{Pa.s}{rad}$*)*

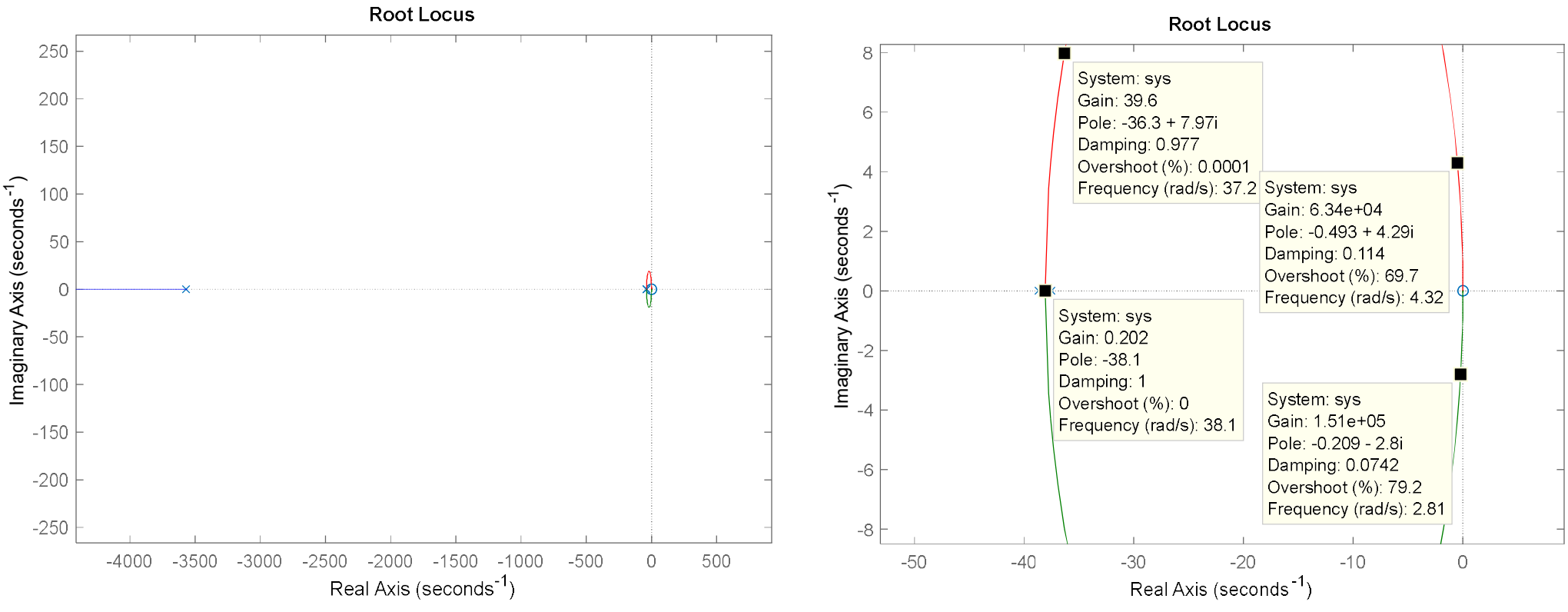

*Figure 3.5. Root-locus plot for* $K_d$ *(*$K_p = 32000\ \frac{Pa}{rad}$*;* $K_i = 1200000\ \frac{Pa}{rad.s}$*)*

In order to employ the insights gleaned from the root locus and verify the PID gain selection, the MATLAB script (see Appendix A) *"freesolve.m"* was used to simulate the behavior of a 40° FREE with nonlinear Eqs. (3.2) and (3.3). For example, the root-locus plots in Figure 3.3 show that $K_p$ = 32000 or 37200 $\frac{Pa}{rad}$ are expected to produce the desired (reaching the setpoint of rotation fast as possible without overshoot or oscillation) response of the system. On the other hand, $K_p$ = 7810 or 6570 $\frac{Pa}{rad}$ should produce an underdamped response. Figure 3.6 illustrates plots of pressure, radius, winding angle, elongation, and rotation of the FREE using these gains and verifies the analysis of the root-loci for $K_p$. Note that $K_p$ = 100000 $\frac{Pa}{rad}$ is a value found by initial try and error without using the root-locus method, and the maximum allowable pressure for the PID controller is 10 psi. Figures 3.7 and 3.8 illustrate the similar verification for gains of $K_i$ and $K_d$.

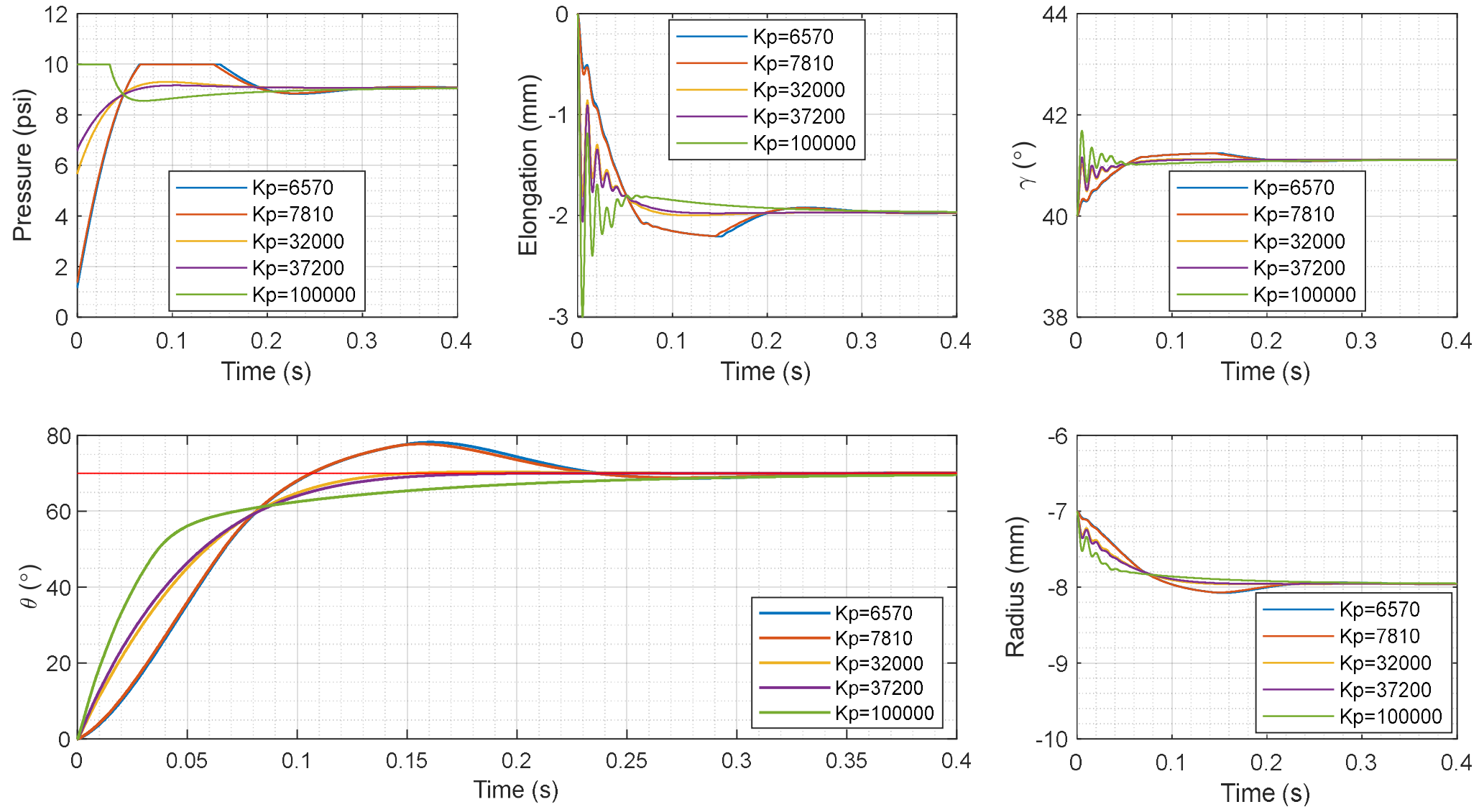

*Figure 3.6. Response of the system to $K_p$ variation ($K_i = 1200000\ \frac{Pa}{rad.s}$; $K_d = 0\ \frac{Pa.s}{rad}$)*

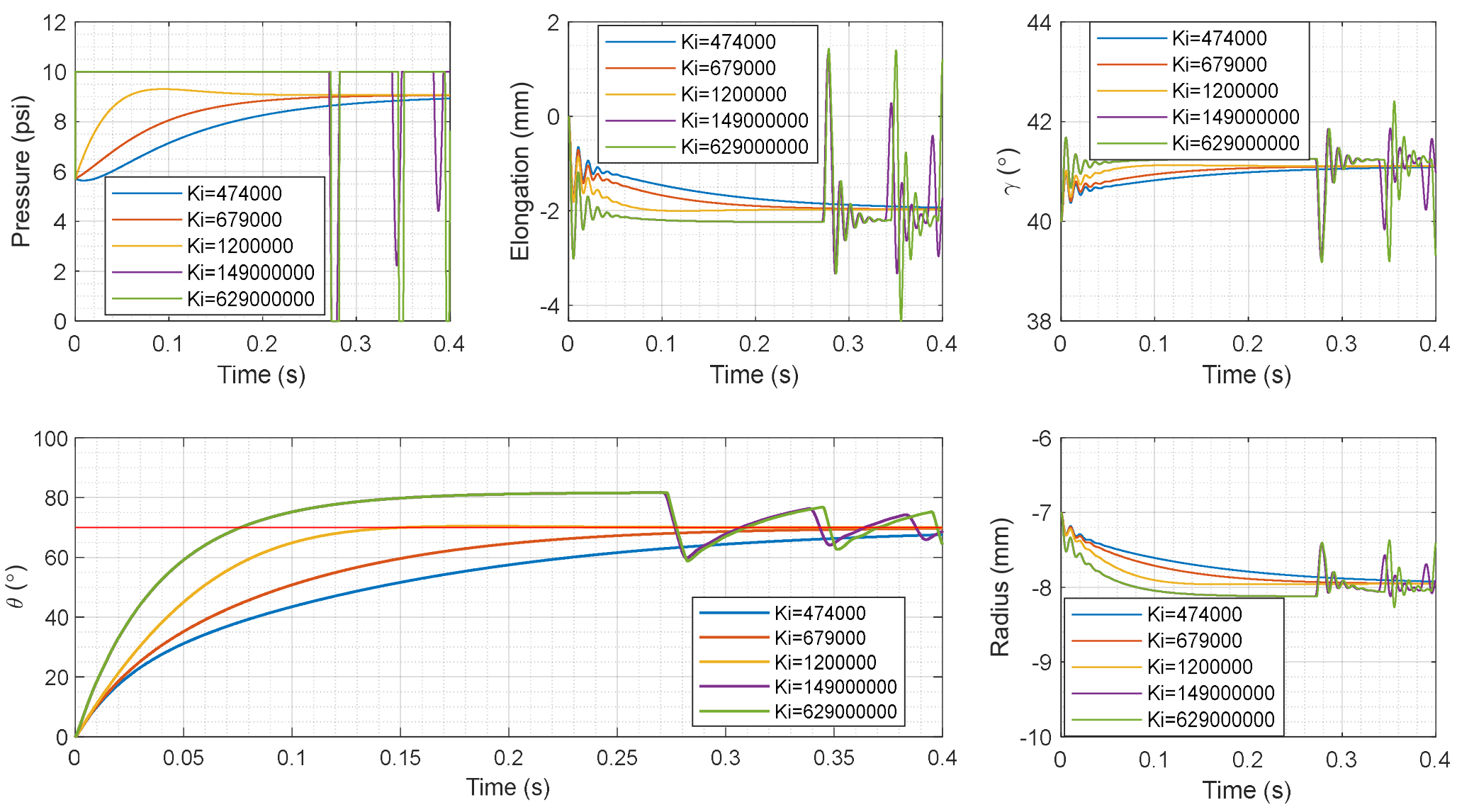

*Figure 3.7. Response of the system to $K_i$ variation ($K_p = 32000 \frac{Pa}{rad}; K_d = 0 \frac{Pa.s}{rad}$)*

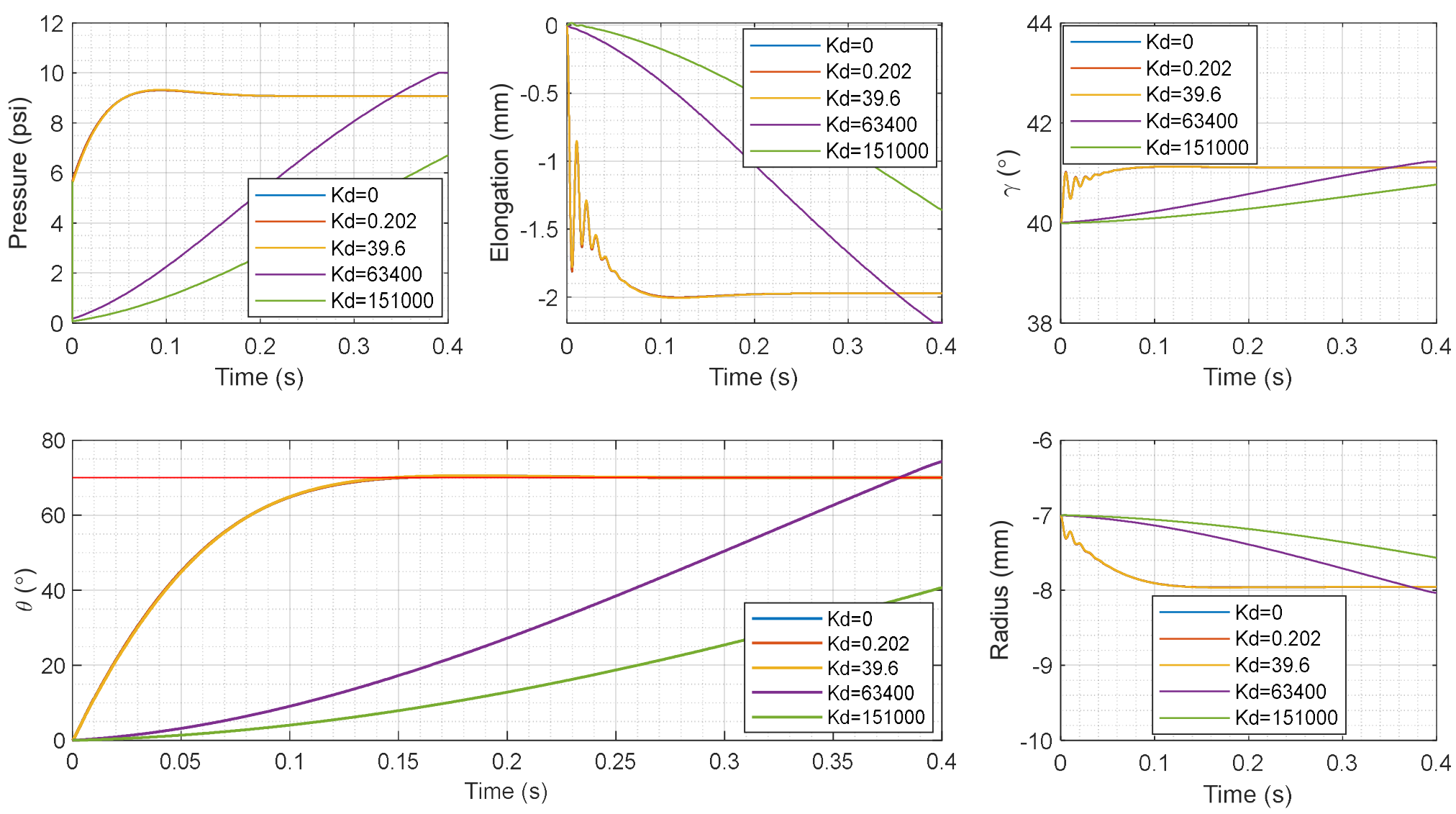

*Figure 3.8. Response of the system to $K_d$ variation ($K_p = 32000 \frac{Pa}{rad}$; $K_i = 1200000 \frac{Pa}{rad.s}$)*

For a robotic system, it is important to eliminate overshoot and oscillatory behavior to avoid collisions with objects. As the dynamic simulation suggests in Figures 3.6, 3.7, and 3.8, gains of $K_p = 32000 \frac{Pa}{rad}$, $K_i = 1200000 \frac{Pa}{rad.s}$, and $K_d = 0 \frac{Pa.s}{rad}$ are suitable to control the rotation of the 40° FREE because it reaches the setpoint in the shortest amount of time without instability and overshoot. Given that the root-locus method is an insightful tool to predict the behavior of a nonlinear system, and thus it assists the designer to efficiently choose the right control parameters for the system.

## 3.3 Trajectory Planning of Rotation

One of the basic maneuvers in controlling robotic arms is moving the end-effector from an initial position to a final position by following a specific trajectory. There is a number of ways for computing the trajectory between two points as shown in Figure 3.9 (Craig, 1986).

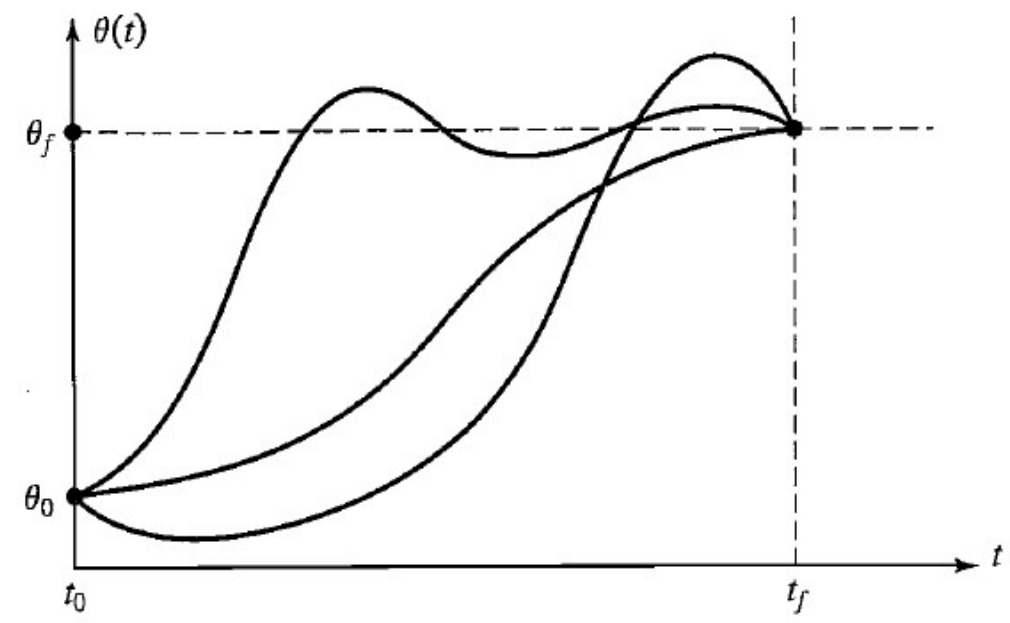

*Figure 3.9. Various chasing trajectories between initial and final points (Craig, 1986)*

Generally, maneuvers in which robotic arms reach the final position smoothly are desired. For the purpose of a soft manipulator made up of FREEs, not only is a smooth motion important, but remaining at the final position is also considered vital. In this section, a trajectory following maneuvers involving only one FREE is simulated based on a cubic polynomial specification of rotation angle. This analysis enables a designer to better understand how a single FREE, as the building block of a module, will physically respond to a planned trajectory. To create a smooth rotational motion when a cubic polynomial trajectory is used:

$$\varphi(t) = a_0 + a_1 t + a_2 t^2 + a_3 t^3, \tag{3.27}$$

where φ is the rotation angle of the free end of the FREE, and $a_i$ $(i = 0, 1, 2, 3)$ are coefficients defined based on satisfying constraints of the motion. Eq. (3.27) is a third degree polynomial, so four constraints are required to uniquely specify all of the $a_i$ $(i = 0, 1, 2, 3)$. The initial and goal position of rotation provide two constraints. Specifying that the angular velocity is zero at the beginning and the end of the trajectory provides two more constraints. Given that $t_f$ and $\varphi_f$ are the desired time of the maneuver and the goal angle of rotation, the four constraints can be written:

$$\begin{aligned} \varphi(0) &= \varphi_0 \ , \\ \varphi(t_f) &= \varphi_f \ , \\ \dot{\varphi}(0) &= 0 \ , \end{aligned} \tag{3.28}$$

$$\dot{\varphi}(t_f) = 0\,.$$

By substituting Eq. (3.28) into Eq. (3.27) and its time derivative, and solving for the $a_i$ $(i = 0, 1, 2, 3)$ the following trajectory equation is obtained:

$$\varphi(t) = \varphi_0 + \frac{3}{{t_f}^2}(\varphi_f - \varphi_0) - \frac{2}{{t_f}^3}(\varphi_f - \varphi_0)\ . \tag{3.29}$$

Using Eq. (3.29) as an expression for $\varphi_d$ in Eqs. (3.2) and (3.3), and running a simulation using the MATLAB m-file in Appendix A, the response of the system can be analyzed. Figure 3.10 shows plots of the response of a FREE ($\Gamma$= 40°, $L = 11$ cm, $R = 0.7$ cm, $F_l = F_{gravity}$, $M_l = 0$, $k_e = 10110\ \frac{\text{N}}{\text{m}}$, $c_e = 5\ \frac{\text{N.s}}{\text{m}}$, and $c_t = 0.005\ \frac{\text{Nm.s}}{\text{rad}}$) following the cubic polynomial trajectory in Eq. (3.29). As the simulation suggests (Figure 3.10) the gains of $K_p = 32000\ \frac{\text{Pa}}{\text{rad}}$, $K_i = 1200000\ \frac{\text{Pa}}{\text{rad.s}}$, and $K_d = 0\ \frac{\text{Pa.s}}{\text{rad}}$ (with the maximum control pressure of 10 psi) are suitable to control the rotation along the specified trajectory.

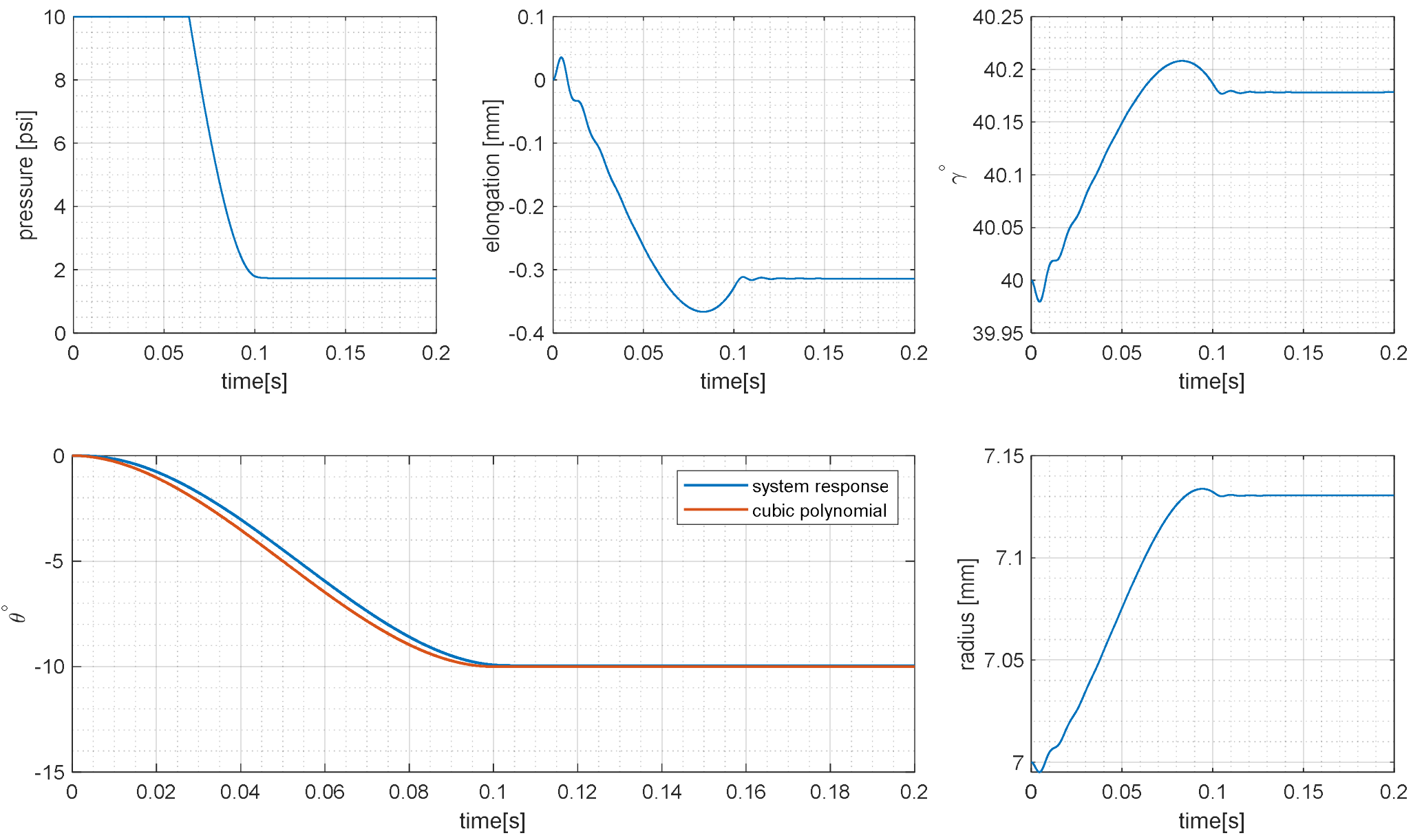

*Figure 3.10. Response of a 40° FREE to a cubic polynomial trajectory following maneuver* ($\varphi_f = -10°, \varphi_0 = 0°, t_f = 0.1\ s$)

From Figure 3.10 it is observed that a single 40° FREE smoothly reaches the setpoint angle of rotation by following a cubic polynomial trajectory. Further experimental validations of the trajectory following control are presented in Section 6.2.

## 3.4 Summary

This chapter has presented the simulated dynamic response of the rotation of a single FREE controlled by a Proportional-plus-Integral-plus-Derivative (PID) controller, which was applied to the lumped-parameter model developed in Chapter 2. The desired control gains were obtained by using the root locus method to predict the response of the system when varying each gain (proportional, derivative, and integral) individually. Additionally, the response of a trajectory following controller based on the same model was simulated for a single FREE. An experimental evaluation of the PID control of FREEs in response to step and trajectory following rotation commands is presented in Section 6.2.

# 4

# Finite Element Analysis

As discussed previously, having a model to predict the behavior of a system assists the designer in avoiding tedious and time-consuming build-and-test processes. The dynamic model developed in Chapter 2 is quite useful for obtaining an understanding the dynamic behavior of a single FREE. However, establishing an idealized lumped-parameter mathematical model for multiple FREEs in a module is difficult. In part, this is because creating a matrix formulation of equations of motion of each FREE in a module is laborious, but more significantly, constructing a relation between moments and forces at the end effector of a module is complicated and fraught with error. As Baumgart (2017) points out, the ability of even a simple model of parallel-actuated FREEs in planar bending to converge to a numerical solution is not guaranteed due to the large reaction forces in particular parameter regimes. In this chapter, attention is focused on developing a finite element model of a FREE in single and module configurations to explore responses, to consider the impact of parameter variations, and to overcome the mathematical modelling difficulties inherent with multiple FREEs. Finite Element Analysis (FEA) is an important tool in the design of components and systems, and this is particularly true for FREEs. The finite element model used here provides just such a modeling tool, allowing a variety of geometries to be studied for capturing displacements: rotation, elongation, and expansion; as well as force and moment generation. Further, a set of parametric studies is presented to evaluate specific behavior criteria of FREEs in single and module configurations.

## 4.1 Model Formulation

FEA is used here to develop a detailed model of an elastomeric tube wound at a specified angle with a thin fiber. The particular FEA software chosen is the commercially available package Abaqus from Daussault Systèmes. Model geometry includes two regions: a three-dimensional elastomeric tube with end caps and multiple fibers wound at the same angle on the exterior of the tube, as shown in Figure 4.1. In this analysis, the elastomer geometry was modeled with 16,830 second-order hybrid tetrahedral elements. Fiber geometry was modeled based on a previously published analysis of soft actuators (Connolly, Polygerinos, Walsh, & Bertoldi, 2015). Fibers were modeled with 1,356 second-order truss elements, which only support axial tensile or compressive loads and not shear or bending. Desirable variables of the FREE were carefully calculated by tracking a set of elastomer surface nodes as a function of pressurization with a custom MATLAB script (see Appendix A).

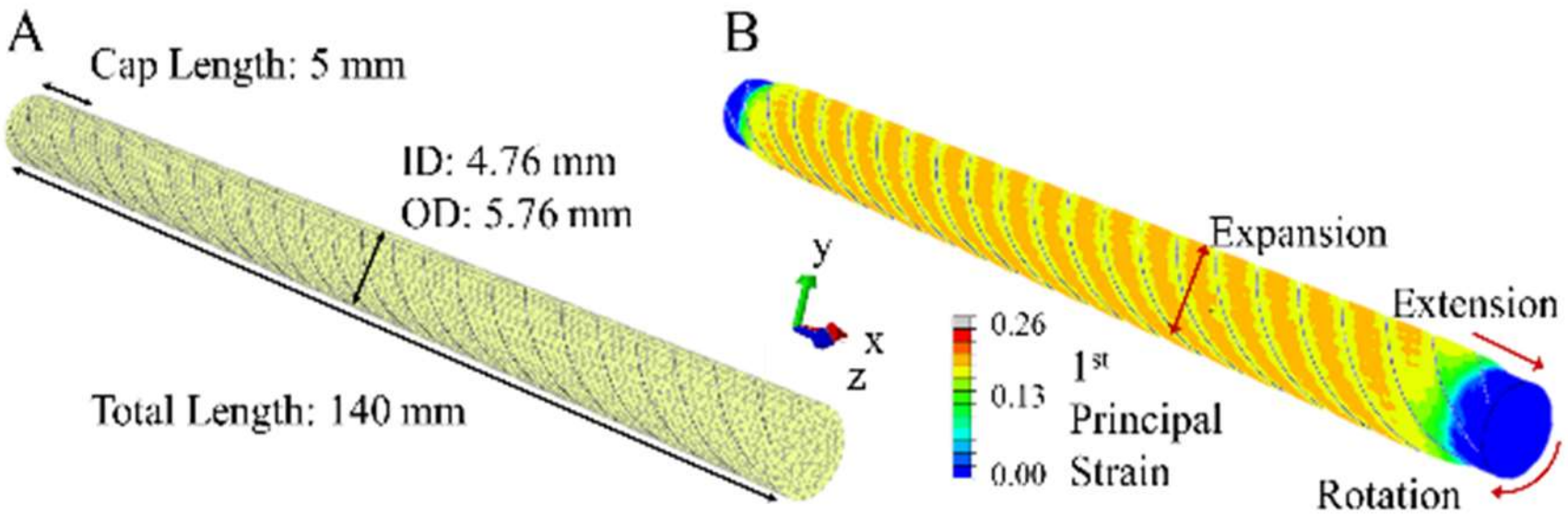

*Figure 4.1. Finite element model of a FREE in Abaqus*

## 4.2 Corroborative Results

Other researchers (Connolly *et al*., 2015; Krishnan *et al*., 2015) have used FEA to study the behavior of fiber-reinforced soft fluidic actuators. Connolly *et al*. (2015) provide particularly relevant comparative data, and these data are used to corroborate results and thus ensure the integrity and fidelity of the FREE finite element model used here. An illustration of the detailed results presented by Connolly *et al.* (2015) is shown in Figure 4.2 (note that Connolly et al. used a silicon elastomer and Kevlar fiber). As can be seen, cases are considered with fiber winding angles ranging from 0° to 90° and pressure increases from 0 to 8.7 psi. As the fiber angle is increased, the radial expansion ratio ($b/B$) increases and the axial extension ratio ($\lambda_z$) decreases until expansion is a maximum and extension is a minimum at a fiber angle of 90°. Note that $b$ is the final and $B$ is the initial radius of the actuator. The axial extension ratio is defined as:

$$\lambda_z = \frac{L + s}{L}, \tag{4.1}$$

where $s$ is the displacement and $L$ is the initial length of the actuator. Noteworthy results are that the extension is non-monotonic (i.e., the length of the FREE decreases for fiber angles in the range of 50° to 90° and that the angle of twist per unit length ($\tau$) reaches a maximum at a fiber angle of approximately 30°. Connolly *et al.* (2015) found that these results were confirmed through physical experiments in which measurements of expansion, extension, and twist per unit length showed "excellent agreement" with FEA results, particularly at low pressures, "with some deviation at higher pressures." They attribute the deviation at higher pressures to "the highly nonlinear response" exhibited by the physical system and suggest that they are "likely due to imperfections in the experiments, and end effects that lead to non-uniform deformations." As noted in Section 4.3, using a neo-Hookean material model is not considered as a good choice for the analysis of large strains, and the possibility that the deviation noted by Connolly *et al*. (2015) at higher pressure is more likely due to the limitations of this model is explored in Sections 4.2 and 4.3. In considering the non-monotonic increase in length described above, note that the theoretical winding angle at which a filament wound pressure vessel reverses direction between elongation and contraction is 54.7° (Roylance, 2001). While this theoretical result is not strictly applicable to a FREE consisting of a soft elastomer wound by a single family of fibers (i.e., all fibers wound at the same angle) and in which the properties of the elastomer play an important role, the non-monotonic change in length determined in the finite element analysis is consistent with expected results.

To verify the finite element model of a FREE and to corroborate the results presented by Connolly *et al.* (2015), this system was analyzed here, again using a silicon elastomer and Kevlar fibers. Figure 4.3 shows the results produced with the same system parameters. As can be seen by comparing Figures 4.2 and 4.3, all of the curves displayed in Figure 4.3 are identical to those in Figure 4.2, including all the noteworthy and particular results demonstrated by Connolly *et al*. (2015) as well as (Sedal *et al.*, 2018), specifically the concave, nonlinear form of the curves. Note that the winding angle convention used by Connolly *et al.* (2015) is the complementary angle of the convention used for FREEs in this thesis (see Chapter 2). Given the very thorough experimental results presented by Connolly *et al.* (2015) and the close agreement of the graphs in Figures 4.2 and 4.3, the model presented here can be confidently used for the studies, described in the following sections.

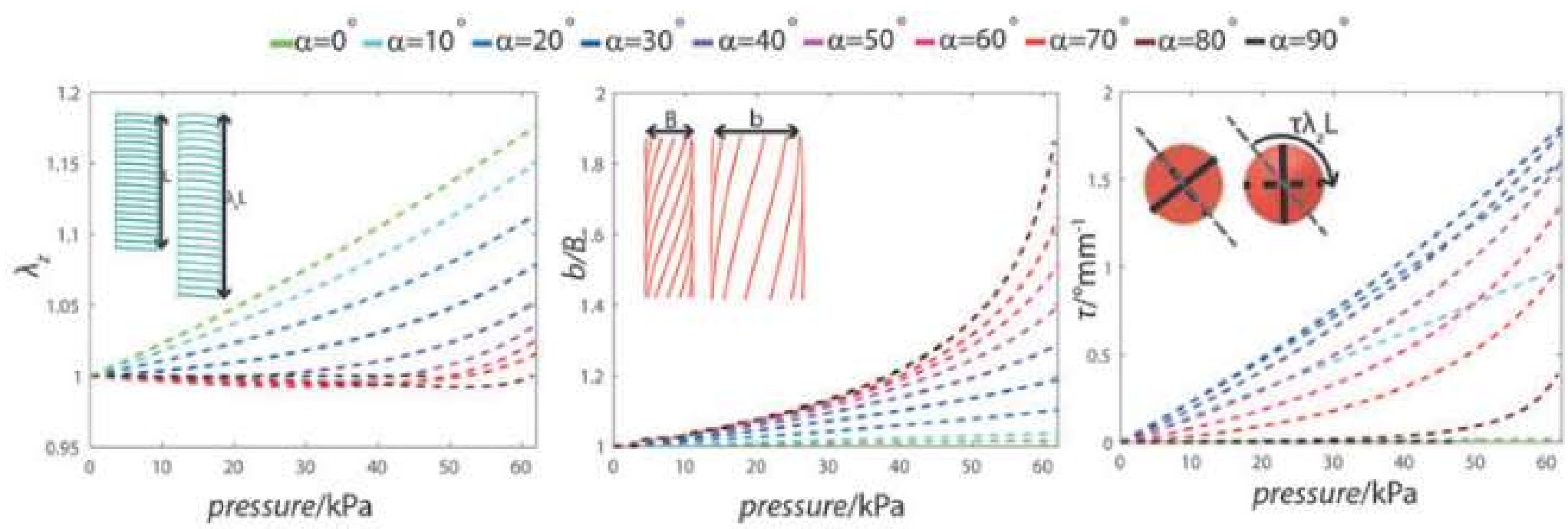

*Figure 4.2. Finite element results showing extension ($\lambda_z$), expansion (b/B), and twist per unit length (τ) as a function of applied pressure for a range of different fiber angles (Connolly et al. 2015)*

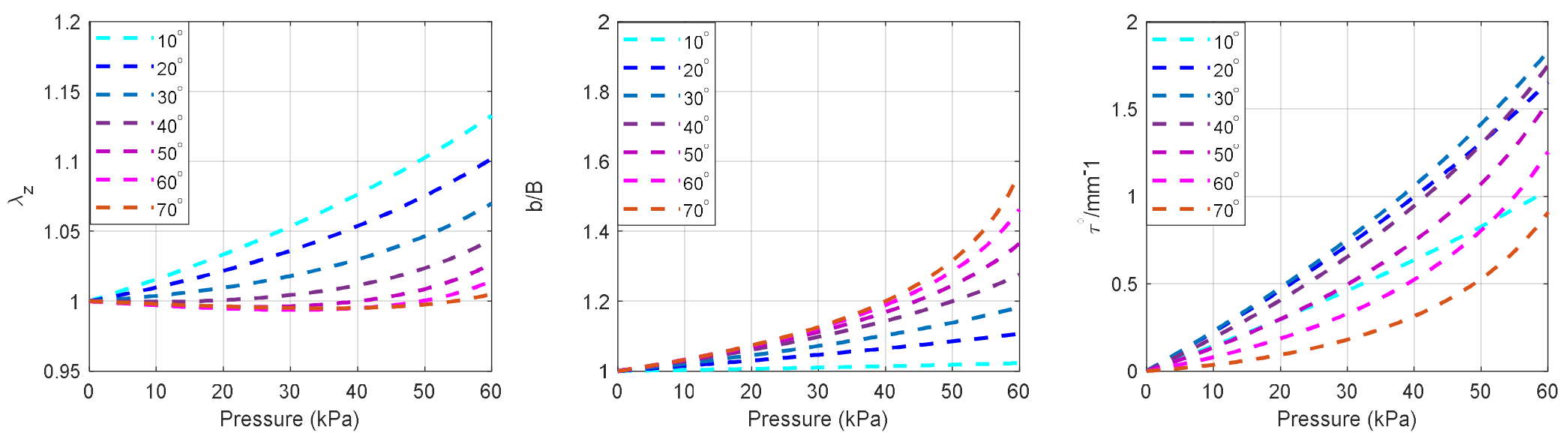

*Figure 4.3. Results of finite element model replicating the results presented by Connolly et al. (2015)*

## 4.3 Numerical Modelling

The extent of finite element analysis of soft pneumatic actuators in the past has been limited and only found to be useful for special cases of materials and geometries (Lipson, 2014). Generally, the research has been narrowly focused on the linear behavior of soft materials at small strains (Deimel & Brock, 2015; Roche *et al.*, 2014). Various constitutive models have been investigated to understand the hyperelastic behavior of soft materials such as the Mooney–Rivlin (Udupa, Sreedharan, Sai Dinesh, & Kim, 2014) and neo-Hookean (Connolly *et al.*, 2015) models. These models are based on linear approximations of the strain invariants and also limited to small strains (Yeoh, 1993). To investigate material properties and the behavior of a FREE, a linear elastic and two different hyperelastic models, neo-Hookean and first-order Ogden were used here. The Ogden model represented by Eq. (4.2) is a polynomial model (Destrade, Murphy, & Saccomandi, 2019; Ogden & Rodney, 1972) that normally yields better results capturing the mechanics of soft materials. Eq. (4.2) presents the strain energy for a first-order Ogden model, where $\bar{\lambda}_i$ $(i = 1, 2, 3)$ represent the deviatoric principal stretches, $J$ is the volume ratio, and μ, α, and $D$ are material parameters. To enforce incompressibility, $D$ was set to zero for both hyperelastic models. For the case of $\alpha = 2$, the first-order Ogden model degenerates to the neo-Hookean model. Fibers were modeled as linearly elastic. Contact and adhesion between elastomer and fibers was modeled with tied conditions.

$$\Psi = \frac{2\mu}{\alpha}\left(\bar{\lambda}_1^{\alpha} + \bar{\lambda}_2^{\alpha} + \bar{\lambda}_3^{\alpha} - 3\right) + \frac{1}{D}(J-1)^2 \qquad (4.2)$$

## 4.4 Material Characterization

Essential to the accurate finite element analysis of any system is an accurate determination of the material characteristics of the system. For the FREEs described here, the needed material characterization involves selection of an appropriate material model (the ones that are considered here are linear, neo-Hookean, and Ogden) as well as the parameter values needed for a particular model (such as elastic modulus, Poisson's ratio, and potentially others). To explore the response of FREEs consisting of a latex elastomer and cotton fibers, the corresponding material properties were determined experimentally from stress-strain relationship for each component (latex and cotton fibers).

**Latex Elastomer Material Properties**

Shown in Figure 4.4 are experimentally determined stress-strain relationships for the latex elastomer. Data were collected using an Instron 5965 universal testing machine as well as by hand. The Instron data were taken using a dog bone-shaped test specimen under uniaxial tension, measuring axial force as strain was slowly incremented to over 50% with a total of over 17,000 data points taken. The hand-measured data (10 data points) were collected by applying a known load to a latex tube made of the same material and having the same cross-sectional dimensions (3/8" inside diameter and 1/32" wall thickness) as that used to create fiber-wound FREEs in later experiments to ensure consistency of data. Both Instron and hand data are plotted as engineering stress (applied load divided by original cross-sectional area) and true stress (applied load divided by actual cross-sectional area, assuming that the actual cross-sectional area varies inversely with longitudinal strain) versus strain. A straight line was fit to the Instron data for both the engineering stress and true stress versus strain plots. Note that both engineering stress and true stress were plotted to illustrate the differences between the two, particular for large strains. When using hyperelastic material models, Abaqus assumes that the material properties are based on true stress. Assuming a linear relationship for the elastic modulus corresponding to the true stress-strain curve fit equals a value of 1.18 MPa. A further check that the data were in a reasonable range was done by calculating the elastic modulus corresponding to typical Shore A hardness numbers for latex [35 +/- 5 (Newtex Latex Tubing, 2018)]. An elastic modulus of 1.18 MPa corresponds to a Shore hardness of 30.8 based on the formula given in Labonte, Lenz, & Oyen (2017), which is within the expected range.

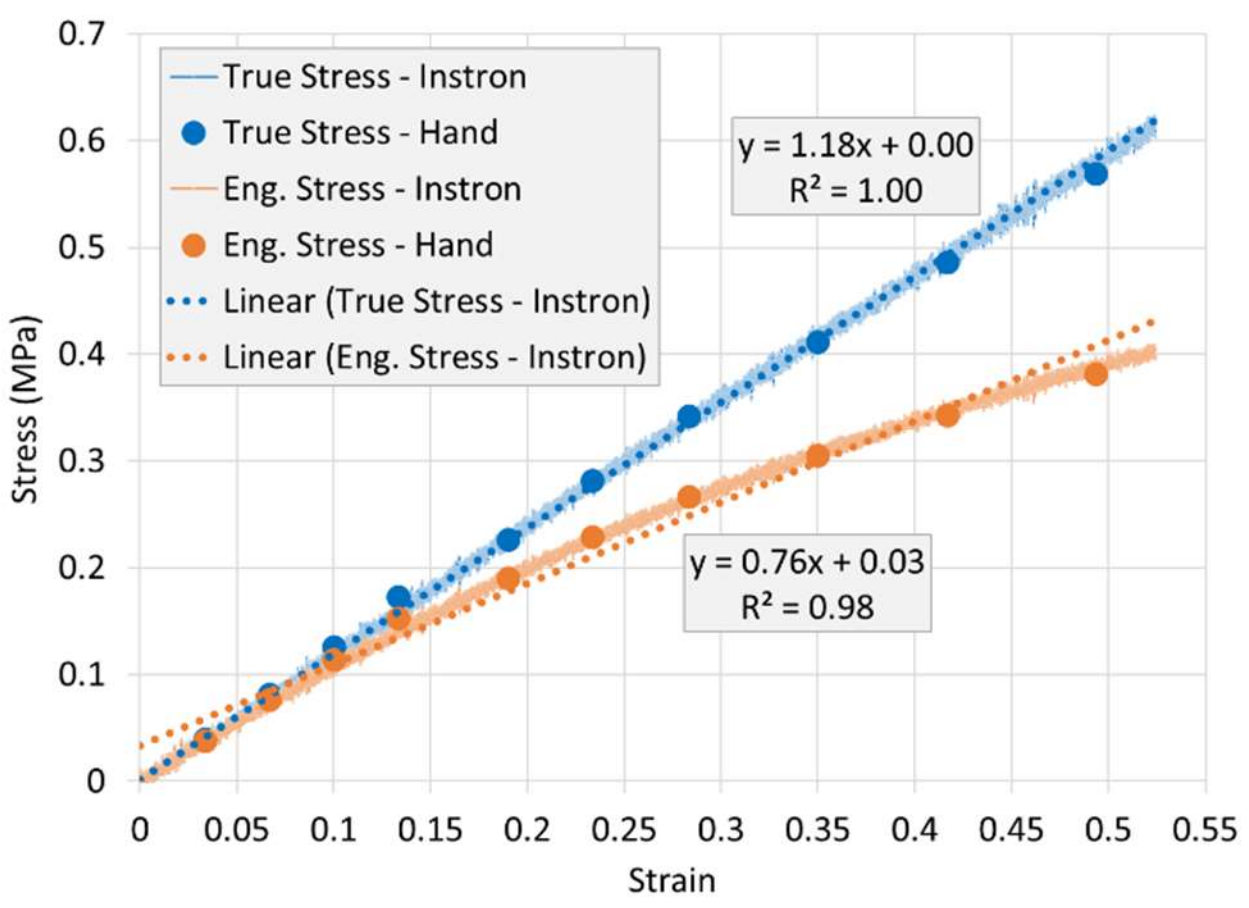

*Figure 4.4. Experimentally determined stress-strain relationships for latex elastomer based on Instron and hand measurements showing both engineering strain and true strain with linear fits to the Instron data*

### Cotton Fiber Material Properties

With the material properties of the latex elastomer well established, similar analyses were performed to determine the properties of the cotton fibers. One challenging aspect of the fibers is that they are made of woven strands and thus a definitive cross-sectional area is difficult to determine. Additionally, as the strands respond to load, they may not be loaded uniformly, particularly for small loads. As a result of these two factors, truss elements with linear material properties were used in Abaqus to model the fibers (as mentioned in Section 4.1). This also eliminates the need to determine the cross-sectional area of the fibers and only requires the experimental determination of applied load versus strain characteristics and the calculation of the product of elastic modulus and cross-sectional area (the stiffness of the fibers) for use in the FEA analysis. Figure 4.5 shows plots of applied load versus strain for two load tests of the fibers, one using the same Instron testing procedure used to measure the properties of the latex (in this case collecting almost 34,000 data points) and another done by hand for a small number of data points (10), with particular attention to the engagement of the individual fiber at small loads. In extracting a value of the fiber stiffness (elastic modulus times cross-sectional area, i.e., EA) from the data, three different approaches were considered as a way to bracket a reasonable value while taking into account the uncertain material response at small loads. As can be seen, a line drawn tangent to the load curve at loads approaching the ultimate load yields a value of EA equal to 644 N. A calculation of EA based solely on the ultimate load and the corresponding strain is 238 N. The value of EA found by fitting a tangent line to the hand measured data is 334 N. A value of EA for FEA purposes is thus constrained within the bounds of approximately 238 and 644 N, with the determination of the most appropriate value within that range dependent on the expected range of strains experienced by the fibers within the model. As will be shown later in this section, the exact value of EA is not critical to the analysis because of the significantly greater stiffness of the cotton fibers relative to the latex elastomer.

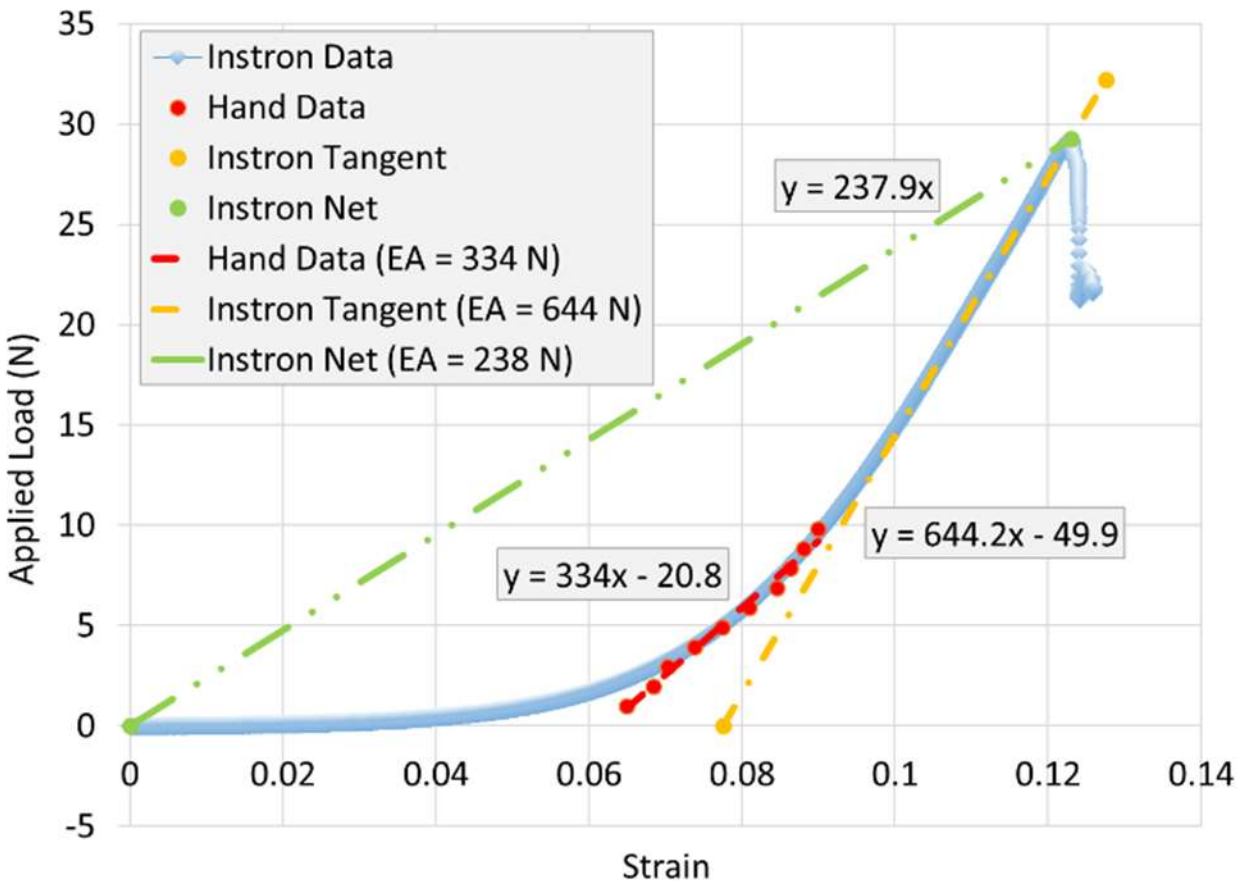

*Figure 4.5. Experimentally determined applied load versus strain relationships for cotton fibers based on Instron and hand measurements Radial Expansion of Latex Elastomer without Fibers*

Further experiments were performed to explore the ability of the FEA model to predict behavior when using the material parameters described above. As a first step in establishing the validity of the model of a FREE consisting of a latex elastomer and cotton fibers, measurements of radial expansion were made using latex tubular segments without fibers. The tubes each had a 9.52 mm (3/8") inside diameter, 0.8 mm (1/32") wall thickness, and approximately 130 mm length. Measurements were taken as pressure within the tubes was increased from 0 to 3.5 psi (corresponding to strains comparable to those experienced in similar fiber-wound FREEs pressurized up to 10 psi) and then decreased from 3.5 psi back to zero. The results are shown in Figure 4.6. Also show in the figure are the radial expansion ratios predicted by 1) simple thin-walled shell theory (Roylance, 2001) as well as FEA analysis using 2) linear material properties allowing only linear deformations, 3) linear material properties allowing nonlinear deformations, and 4) neo-Hookean material properties allowing nonlinear deformations. The equation for radial expansion for a thin-walled cylindrical pressure vessel with closed ends is

$$\delta_r = \frac{pr^2}{bE}\left(1 - \frac{\nu}{2}\right), \tag{4.3}$$

where $\delta_r$ is radial expansion, $p$ is internal pressure, $r$ is nominal radius, $b$ is wall thickness, $E$ is elastic modulus, and $\nu$ is Poisson's ratio. The values used here are $r = 4.76$ mm (3/16"), $b = 0.8$ mm (1/32"), $E = 1.18$ MPa, and $\nu = 0.5$. As can be seen in the figure, simple thin-walled shell theory provides an excellent lower bound on the radial expansion ratio measured experimentally as well as (not surprisingly) showing close agreement with the response determined using the FEA model with linear material properties allowing only linear deformations. Using linear material properties in the FEA model while allowing nonlinear deformations shows good agreement with experimental results as well as demonstrating the importance of nonlinear deformations to the overall response. The response obtained with the FEA model using neo-Hookean material properties is close to the response obtained with linear material properties and nonlinear deformations but shows a small amount of additional material softening over the range of pressures investigated.

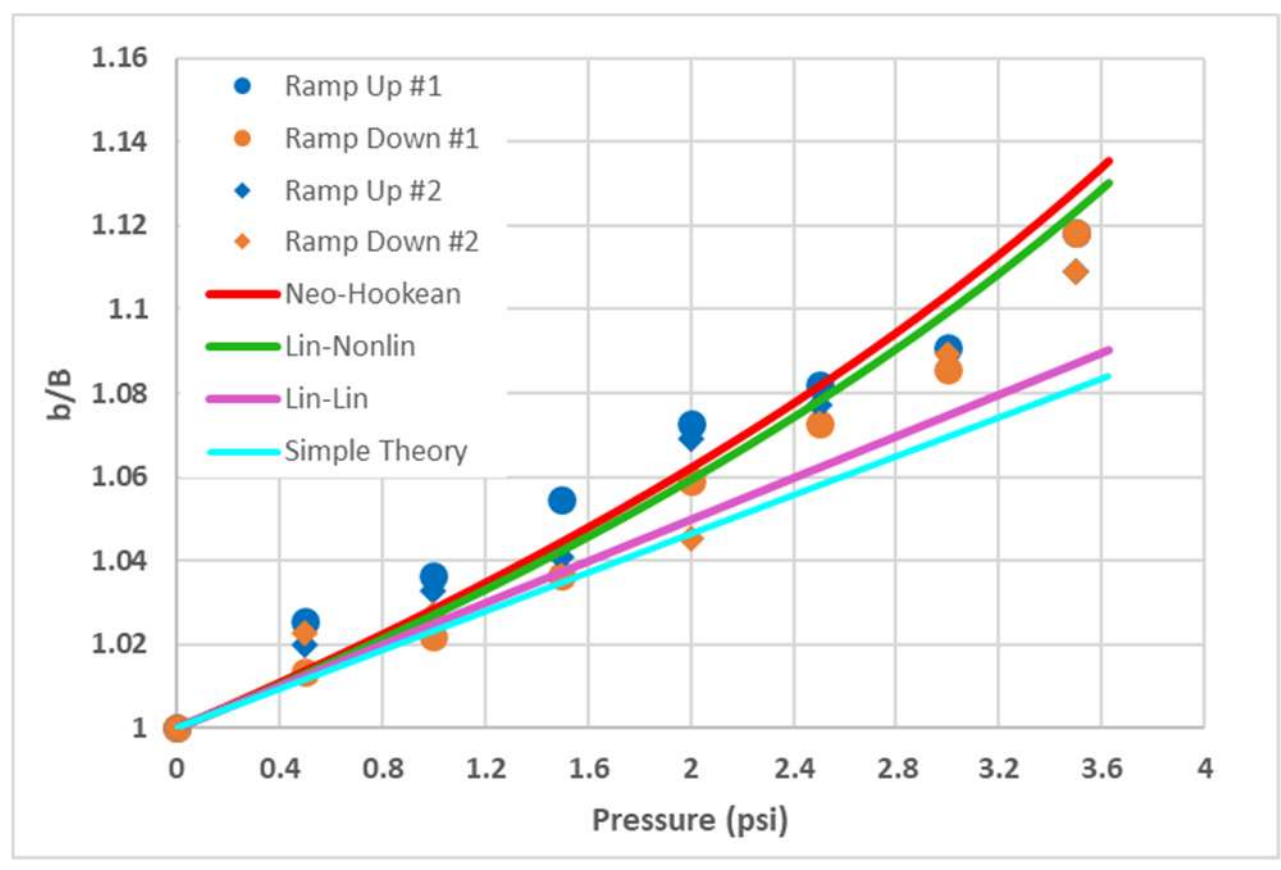

*Figure 4.6. Radial expansion ratio measured with two latex tubes as well as responses predicted using thin-walled shell theory, FEA analysis with linear material properties allowing only linear deformations, linear material properties allowing nonlinear deformations, and neo-Hookean material properties allowing nonlinear deformations*

## 4.5 Determination of Ogden Parameters

After the determination of material properties, experimental results of an actual fiber-wound FREE were considered as well as predications of response generated with the FEA model. Elongation, expansion, and rotation are specifically considered to explore the best combination of Ogden parameters (α and μ). Surprisingly, given the confidence in the model established in Section 4.1, the quality of experimentally determined material parameters, and the consistency seen in the preliminary investigation of responses shown in Figure 4.6, the FEA model did not show good agreement with experimental results, and in the case of rotation differed by over 150%.  It was the realized that even though the material properties were confidently obtained, the fundamental model (neo-Hookean) may not properly represent the behavior of the system as well as seemed to be the case presented by Connolly, *et al*. (2015). Additionally, the significance of components of the FREE were not considered in the FEA model, specifically the adhesive (rubber cement) attaching the cotton fibers to the latex elastomer and a thin latex coating that is brushed over the fibers to seal in the fibers. As a result, additional expansion tests using a simple latex tube without a fiber winding as was done in collecting the data shown in Figure 4.6 were conducted and then compared to the cases in which *only* adhesive and a thin latex coating *without* a fiber winding were applied to gauge their impact on the overall stiffness of the tube. Figure 4.7 shows the experimentally determined expansion ratios measured with the uncoated tube and then measured with the same tube after adhesive and a thin latex coating was applied. As can be seen, the coated tube is in fact significantly stiffer than when uncoated. These results led to the use of material models other than neo-Hookean. The use of a first-order Ogden model was explored as represented by the strain energy function previously presented in Eq. (4.2).

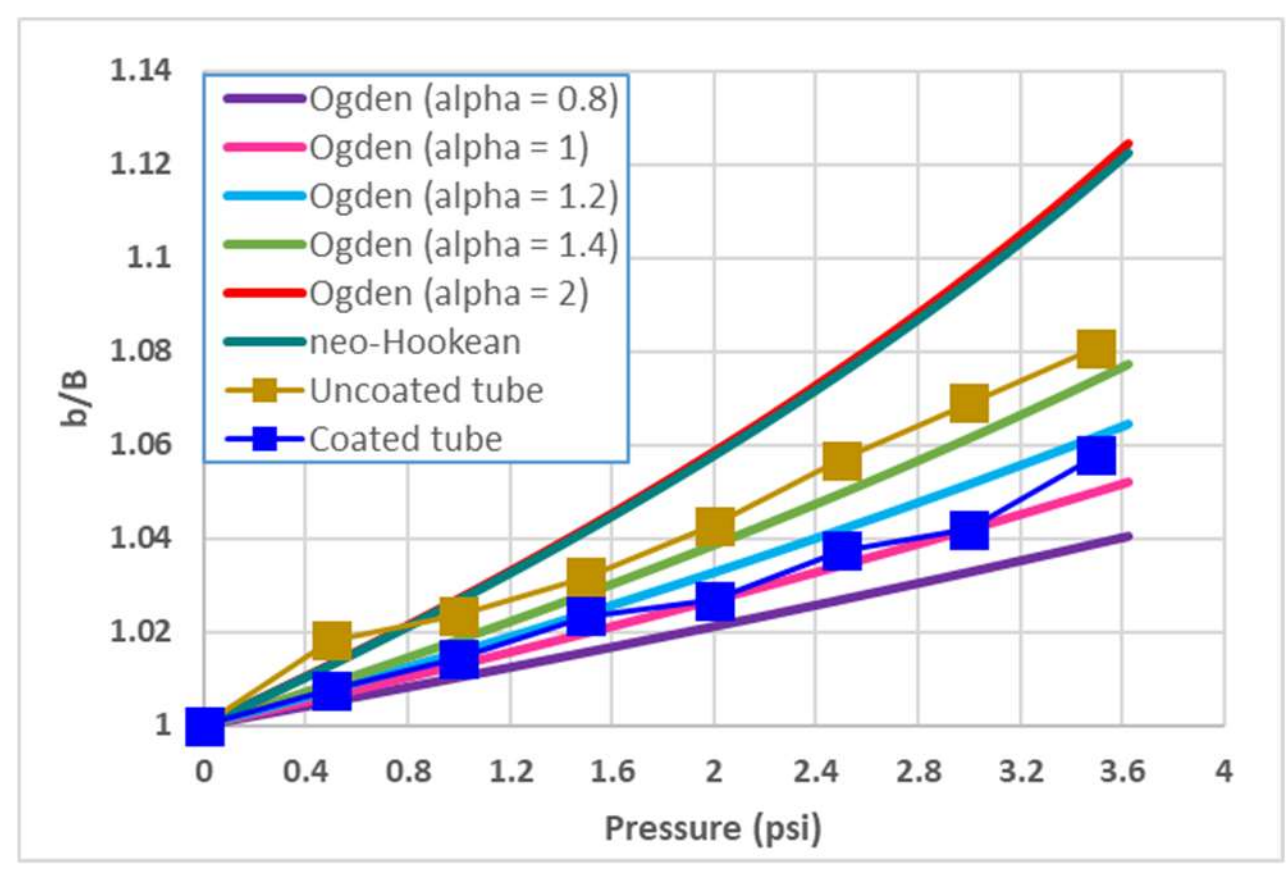

*Figure 4.7. Radial expansion ratio measured on uncoated and coated latex tubes as well as predicted using an Ogden material model (α = 0.8 to 2)*

Use of an Ogden model requires the determination of the material parameter α, and Figure 4.7 shows expansion ratio curves for values of α from 0.8 to 2 (note that with α = 2 the Ogden model reduces to the neo-Hookean model and those two curves lie essentially on top of one another). Based on the results shown in Figure 4.7, an Ogden model with α = 1.2 and μ = 0.65 MPa was chosen as a representation of the material properties of a cotton fiber wound latex FREE.

## 4.6 Validation

### Elongation and Rotation

Figures 4.8 and 4.9 display curves characterizing the rotation and elongation of a latex elastomer FREE with fiber winding angles of 20°, 40°, and 70° as predicted by FEA with an Ogden material model (α = 1.2, μ = 0.65 MPa). Also plotted are experimentally determined data points for three FREEs with the same winding angles, inside diameter 9.52 mm (3/8”), wall thickness 0.8 mm (1/32”), and 130 mm length. Both of the results closely follow the same trend and indicate that the finite element model reasonably predicts motions of FREEs.

To acquire experimental data imaging methods was used for FREE's elongation and rotation measurements (see Chapter 5 for the experimental apparatus). Figure 4.10 depicts the orientation of an indicator attached to the bottom of a FREE (Γ = 40°). The rotation of the end cap was measured statically at ten distinct pressures by tracking the angle of the straight line. From left to right in Figure 4.10, the orientations of the FREE at pressures of 0, 5, 8.7 psi are shown.

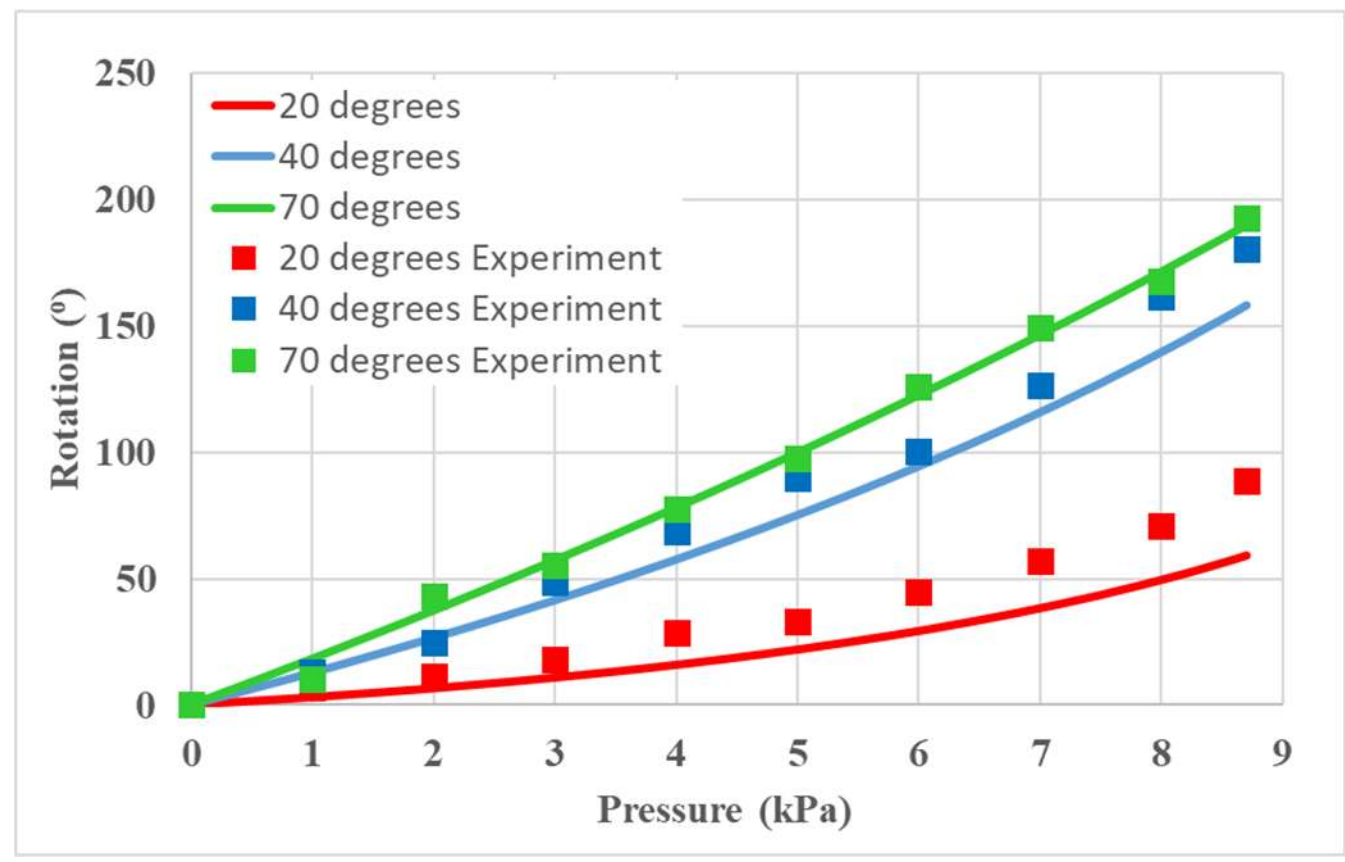

*Figure 4.8. FEA results obtained with an Ogden model (α = 1.2) showing rotation as a function of pressure for fiber angles of 20°, 40°, and 70° as well as corresponding experimental data points*

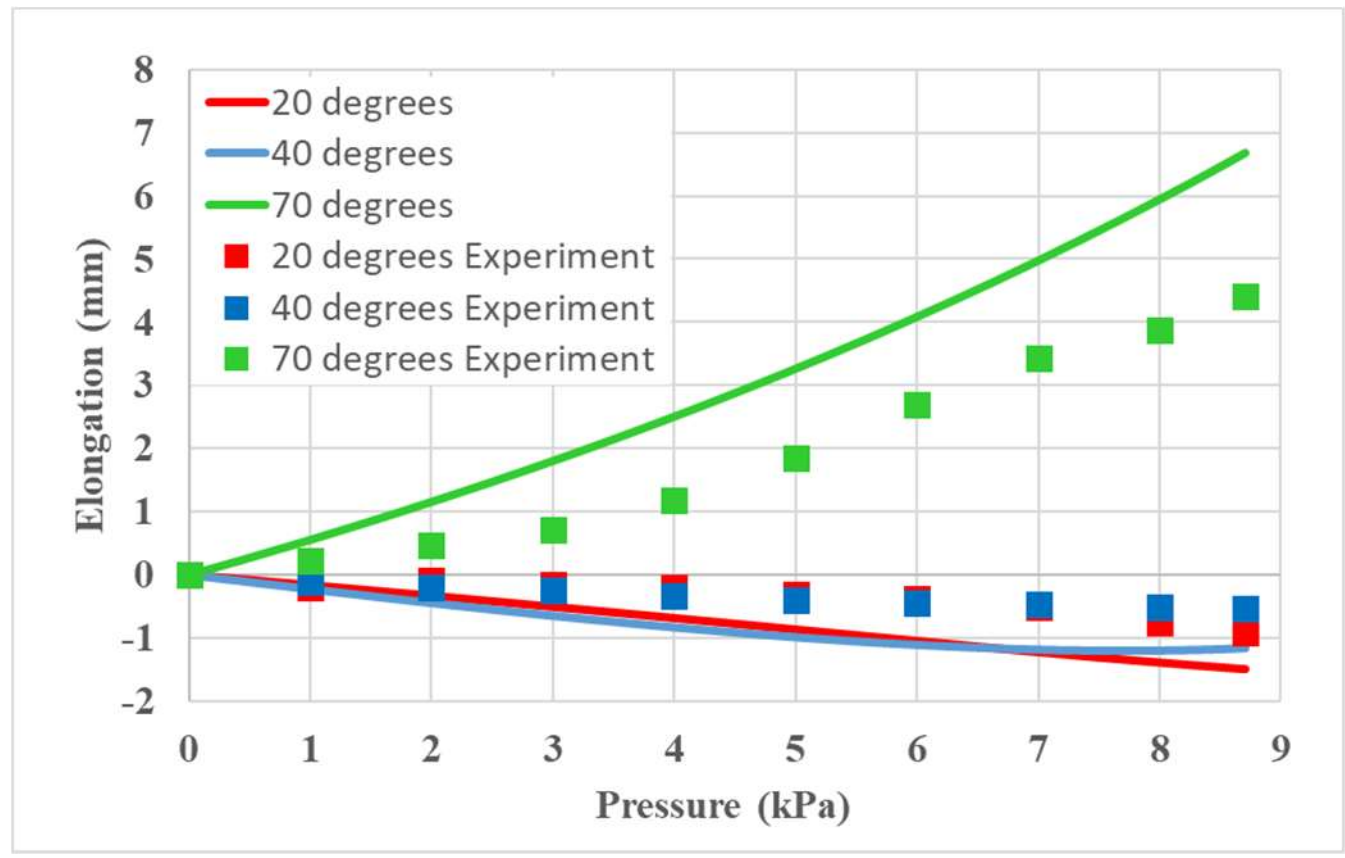

*Figure 4.9. FEA results obtained with an Ogden model (α = 1.2) showing elongation as a function of pressure for fiber angles of 20°, 40°, and 70° as well as corresponding experimental data points*

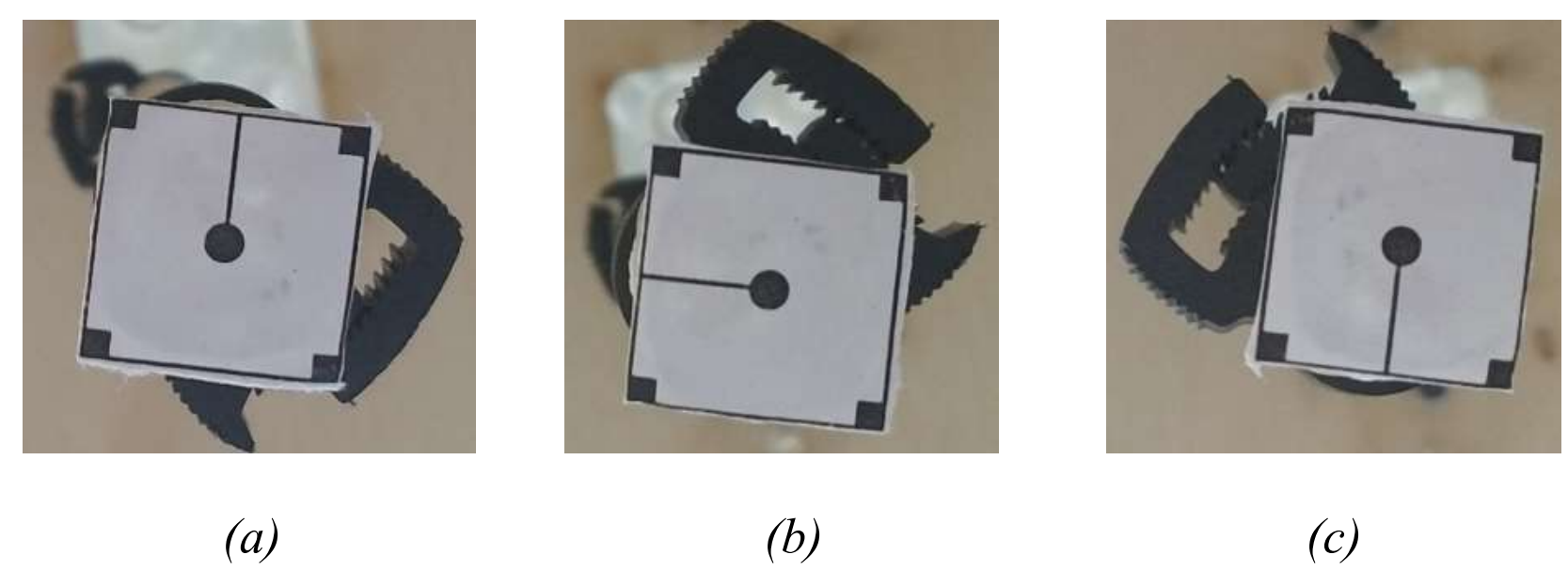

(a) (b) (c)

*Figure 4.10. Images of the FREE end cap at (a) 0 psi, (b) 5 psi, and (c) 8.7 psi*

Agreement between predicted and measured results is quite good, reinforcing confidence in the use of an Ogden model and in our general approach to modeling the behavior of the FREEs. Further evidence supporting the use of an Ogden model come from observations of the divergence of data between the neo-Hookean model and experimental results at higher pressures observed by Connolly *et al.* (2015) and displayed in Figure 4.11 [Fig. 2d in Connolly *et al.* (2015)]. The figure clearly shows stiffening (concavity) in the experimental expansion and rotation data at higher pressures, while the finite element results based on a neo-Hookean material model exhibit softening (convexity). Additionally, the model used in Connolly *et al*. (2015) required generating the desired internal pressure through thermal expansion to achieve convergence at large fiber angles (80$^{o}$ and up). Note that the winding angle convention used by Connolly *et al.* (2015) is the complementary angle of the convention used for FREEs in this thesis (see Chapter 2). This thermal expansion enforces a volume constraint on the FREE interior, making the analysis deformation driven in comparison to the load driven analysis used here. These convergence issues were not observed when using an Ogden model for the elastomer. This is likely due to the fact that the neo-Hookean model used in Connolly *et al*. (2015) cannot accurately reproduce the constitutive behavior of inflated elastomers at high strains (Holzapfel, 2002). The observed limitations in the use of a neo-Hookean model in describing the deformations of an unwound tube (see Figure 4.7) and the behavior observed in Connolly, *et al.* (2015) at higher pressures with a fiber wound tube (see Figure 4.11) suggest the advantage of an Ogden over a neo-Hookean material model at large pressures and deformations. Thus, an Ogden model is well suited for future modeling of FREEs.

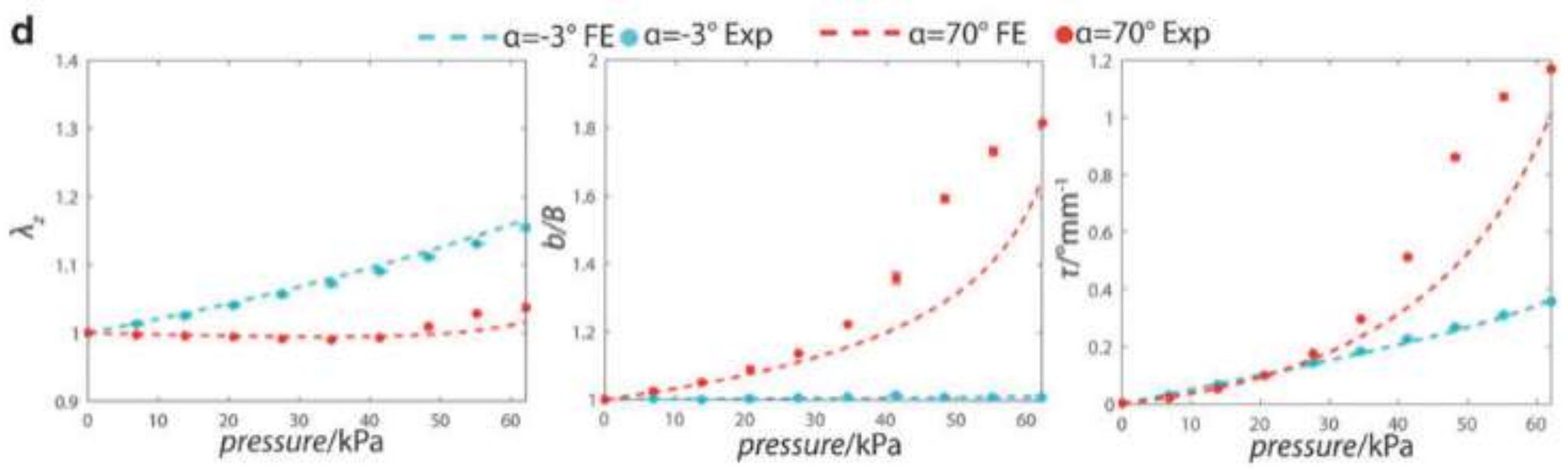

*Figure 4.11. Comparison between FEA and experiments for FREEs with fiber angles of -3° and 70° [Fig. 2d in Connolly et al. (2015)].*

### Force and Moment

Many of the modelling approaches [see examples in Sedal, Wineman, Gillespie, & Remy (2019)] used for soft robotic actuators have been created to facilitate the development of robotic systems such as soft manipulators, mobile robots, and rehabilitation robots. Predicting the load capabilities of soft actuators in various configuration is thus an important consideration. In particular, (Sedal *et al.*, 2019) has compared the prediction capabilities of three distinct FREE models: lumped-parameter, continuum, and neural network, in capturing the force and moment generation of FREEs. This section focuses on the flexibility and fidelity of a finite element model of a FREE in capturing "loading-deformation" characteristics. Sedal *et al.* (2019) tested eight FREE samples with 15°, 25°, 36°, 40°, 50°, 62°, 73°, and 76° winding angle using the testing apparatus shown in Figure 4.12 to measure applied force and torque for various combinations of elongation and twist angle. Each sample was tested for all possible combinations of initial elongation $\Delta l$ (mm) and twist angle $\Delta\varphi$ (°) over a range of internal pressures *P* (psi) in the following range:

$$\Delta l = \{-5, -4, \dots, -1, 0, 1, \dots, 4, 5\}\,, \tag{4.4}$$

$$\Delta\varphi = \{-120, -110, \dots, -20, -10, -1, 1, 10, 20, \dots, 110, 120\}\,, \tag{4.5}$$

$$P = \{0, 1, 2, 3, \dots, 14, 15\}\,. \tag{4.6}$$

For each proposed model (lumped-parameter, continuum, or neural network) in Sedal *et al.* (2019), parameter fitting was done for 80% of the experimentally measured datasets and then tested on the rest of the dataset. Finally, the predictive capability (root mean square model error) of each model was plotted in heat maps and the generality of them were compared [see Fig. 10 in Sedal *et al.*, (2019)].

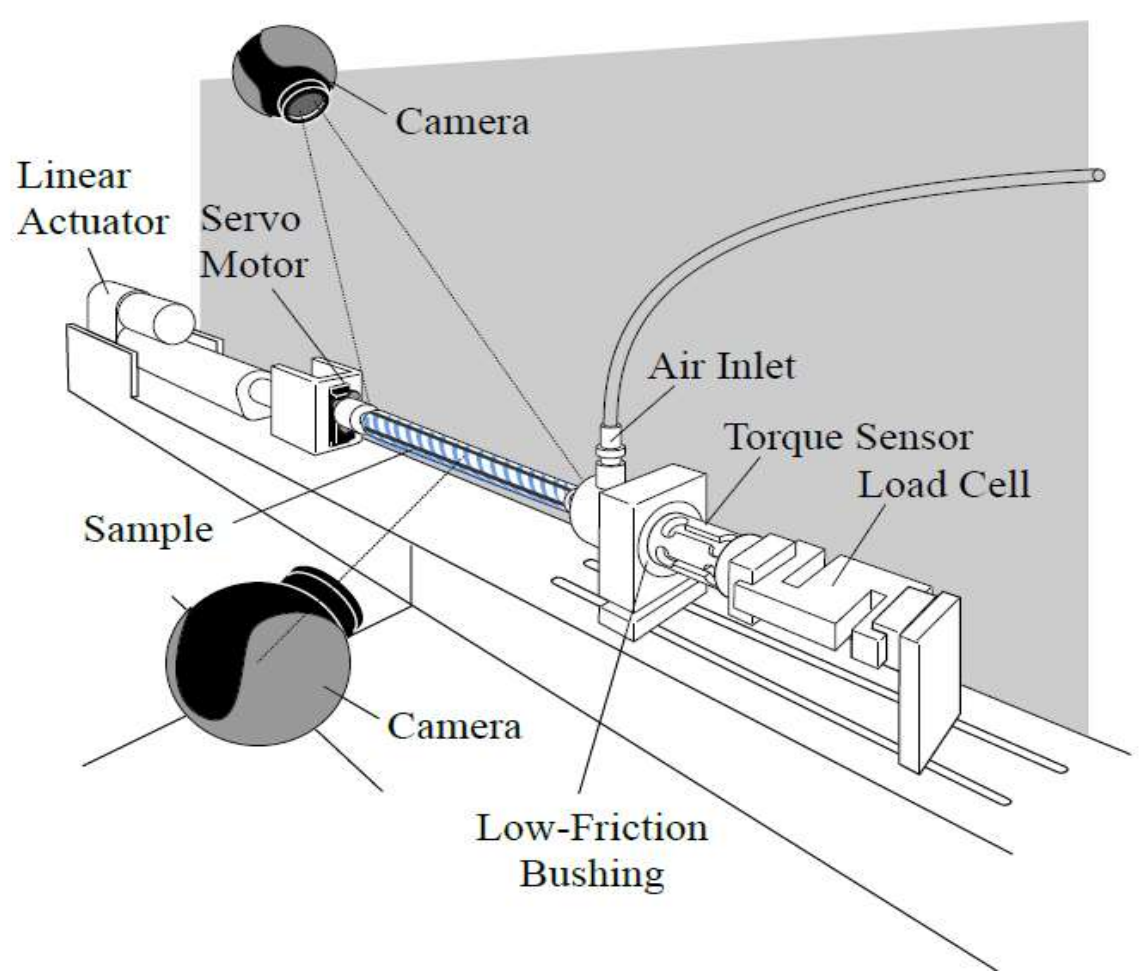

*Figure 4.12. Testing apparatus used by Sedal et al. (2019)*

In an effort to evaluate the effectiveness of using finite element analysis to accurately predict force and torque data, one of the tested FREEs (62° winding angle) was simulated as part of the analysis described within this thesis. The selected datasets for training and testing were limited to a smaller portion of the overall data collected by Sedal et al. (2019) due to convergence difficulties in the finite element model (described at the end of this section). For training the model, datasets with $\Delta l = 5.5$ mm and $\Delta\varphi = 40°$ were selected and the best combination of Ogden model parameters α and μ in Eq. (4.2) were tuned to yield the lowest error. Eq. (4.7) is used to calculate the model error $e$ as the normalized root-mean-square deviation (RMSD):

$$e = \sqrt{\frac{1}{n}\sum_{i=1}^{n}\left(\frac{F_{exp}^{i} - F_{sim}^{i}}{F_{max}}\right)^2 + \left(\frac{M_{exp}^{i} - M_{sim}^{i}}{M_{max}}\right)^2} \qquad (4.7)$$

where $F_{exp}$ and $M_{exp}$ are experimentally measured forces and moments, and $F_{sim}$ and $M_{sim}$ are force and moments calculated using FEA. $F_{max} = 43.9$ N and $M_{max} = 0.146$ Nm are the maximum measured values in the dataset of all cases. Figure 4.13 shows that μ dominants changes in the error and for this case $\mu = 0.4$ Mpa and $\alpha = 0.4$ yields the lowest total model error (4%). Note that the error is relatively insensitive to changes of α. Plots (b) and (c) in Figure 4.13 show the contribution of force and moment separately to the overall error. The red dots in Figure 4.13 are the distinct combinations of μ and α.

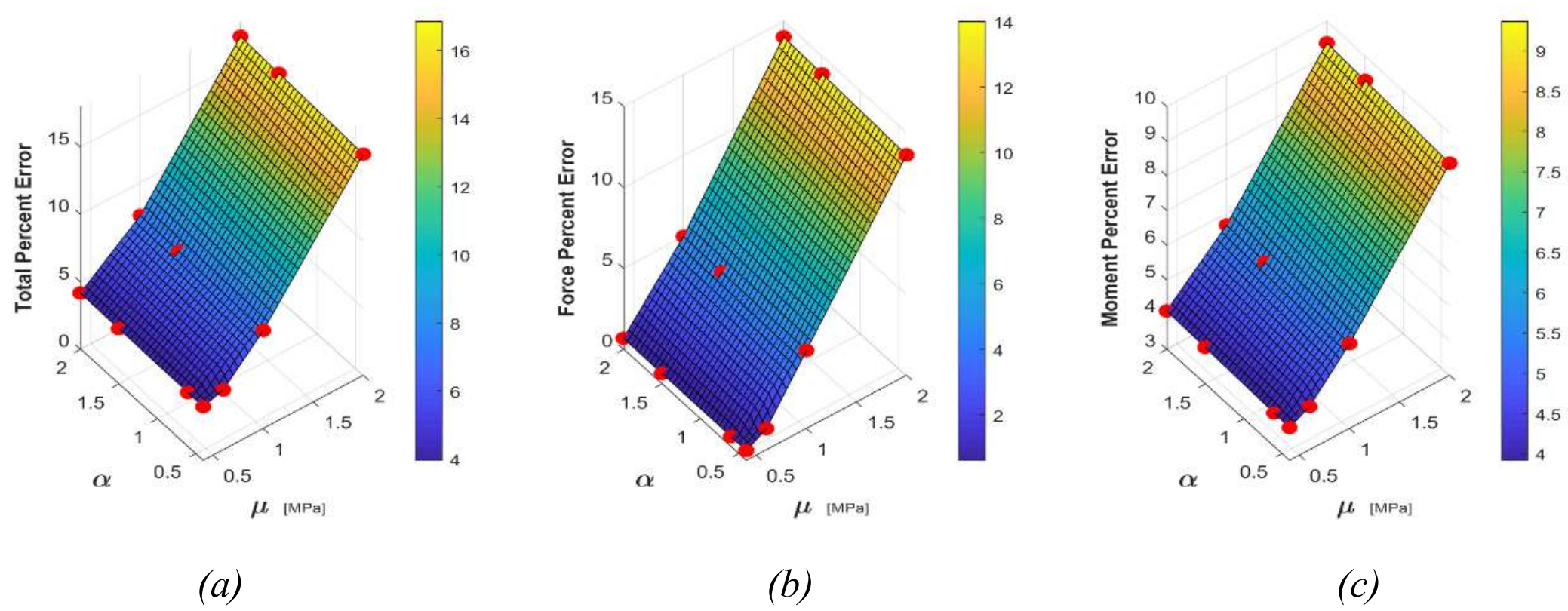

*(a)* *(b)* *(c)*

*Figure 4.13. Simulation error e as function of Ogden model parameters $\alpha$ and* μ

To evaluate the effectiveness of the determined Ogden parameters, simulations were run for datasets with angles of twist equal to $1°$, $20°, 60°$, $-1°$, $-20°$, $-40°$, $-60°$, and the corresponding errors are plotted in Figure 4.14. Error values are relatively high for negative twist angles, although the reason is not clear at this point since the model was evaluated only for one particular limited dataset. One possible explanation could be the difference between the directions of rotation of the pressured FREE and the twist angle. In other words, opposite signs between these two tests may cause large strains and correspondingly higher errors in the results. Overall, force errors make relatively smaller contributions than moment errors to the total errors.

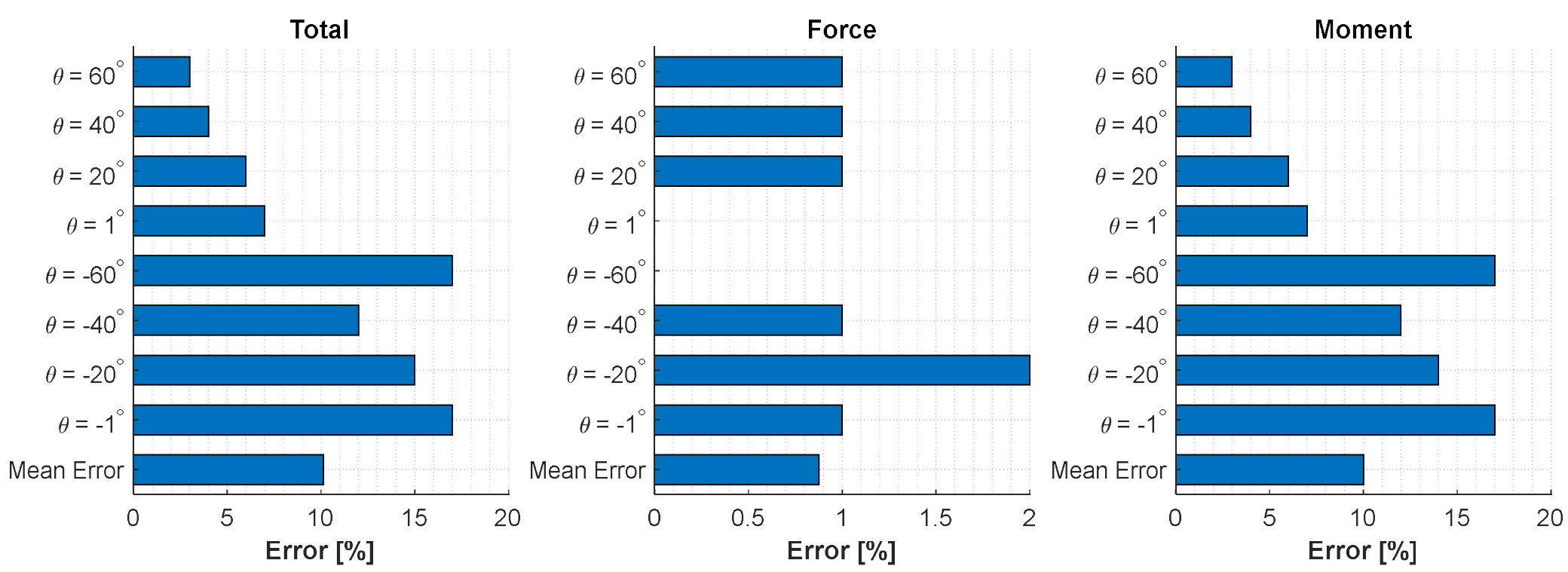

*Figure 4.14. Calculated total, force, and moment errors using the FEA model for winding angles* $1°, 20°, 60°, -1°, -20°, -40°, -60°$ *with Ogden parameters* μ $= 0.4$ *and* $\alpha = 0.4$

Although finite element analysis reasonably accurately predicts FREE applied force and moment, it lacks generality in predicting the performance of FREEs under arbitrary load and boundary conditions. For each possible test case [see Eqs. (4.4) and (4.5)] many parameter adjustments (solution increment or mesh size) were needed to achieve convergence. Additionally, a buckling without apparent pattern, as indicated in Figure 5 of Sedal *et al.* (2019), is another complicating issue for FREEs under external loads, although Sedal *et al.* (2019) found a tendency for buckling under axial compression and negative end-to-end rotation for some samples. Sample imperfection and non-uniform manufacturing have also been suggested reasons for buckling (Lee *et al*., 2016). Two cases of buckling of a 40° winding angle FREE simulated in Abaqus were observed. Figure 4.15 shows the buckled shape of a 40° FREE with a 60° twist angle and 5 mm of elongation. Figure 4.16 similarly shows the buckled shape with an axial compression of 5 mm and -20° twist angle. The buckling cases are similar to those noted by Sedal *et al*. (2019).

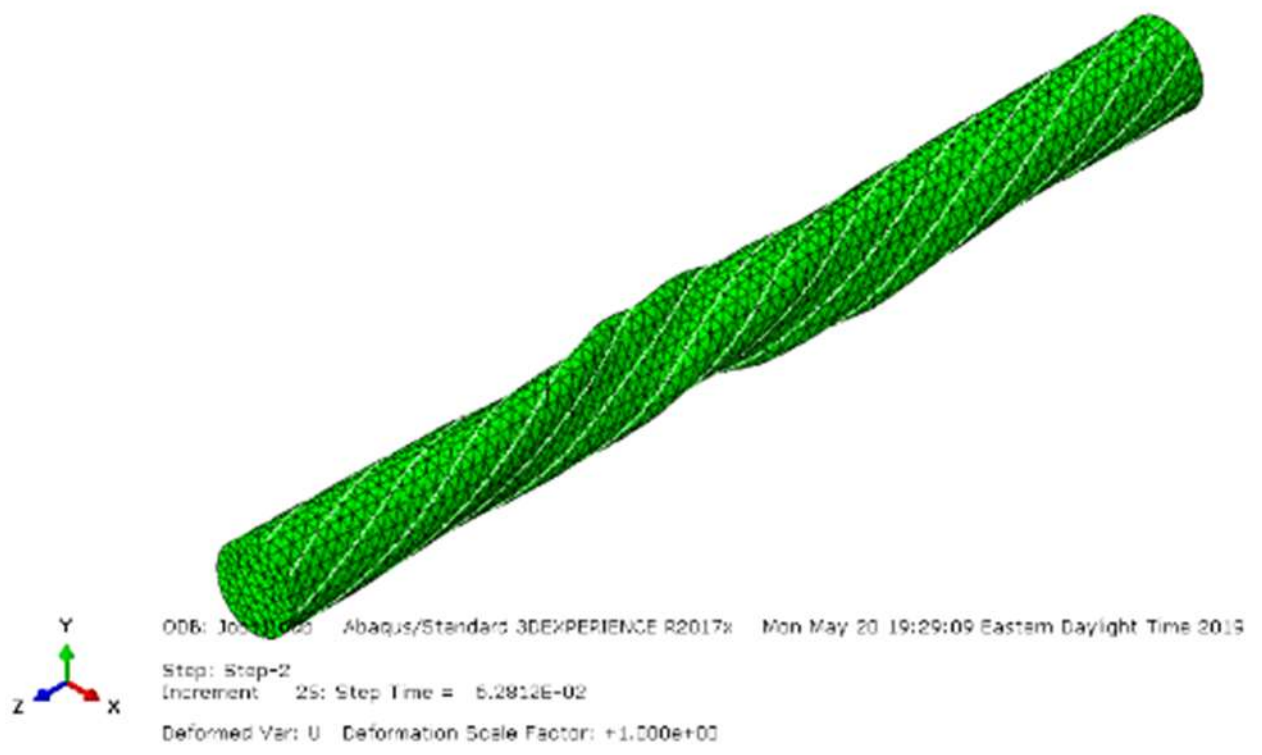

*Figure 4.15. Buckling of 40° FREE simulated in Abaqus with 60° twist angle and* $5\ mm$ *of elongation*

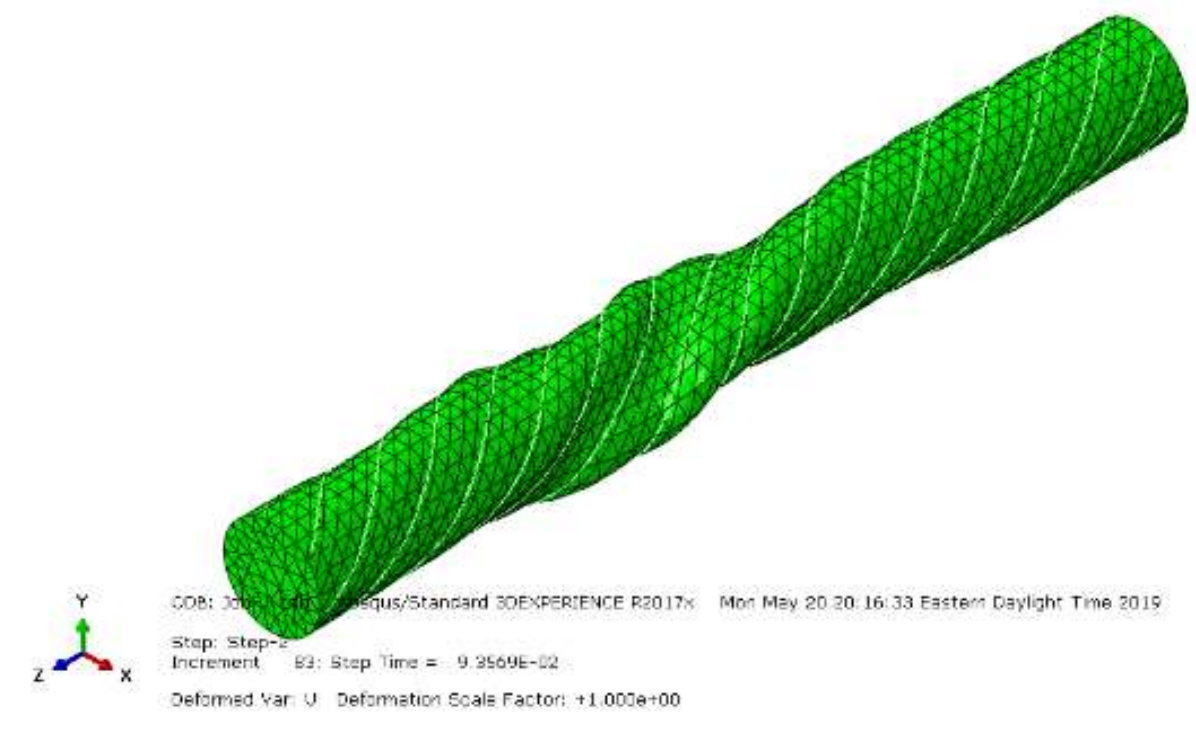

*Figure 4.16. Buckling of 40° FREE simulated in Abaqus with -20° twist angle and axial compression of* $5\ mm$

Because of the many challenges in analyzing each of the cases studied by Sedal *et al*. (2019), additional cases were not considered. A possible direction for future study is the use of Abaqus/Explicit to increase the robustness of the finite element model.

## 4.7 Parametric Studies

### Impact of Winding Angle

The theoretical winding angle at which a filament wound pressure vessel reverses direction between elongation and contraction is 54.7° (Roylance, 2001). The finite element analysis in this section provides a more refined modeling of FREE behavior that accounts for the significant impact of the elastomer and shows that a more accurate prediction of the transition between elongation and contraction is at approximately 45° to 50° for the system studied here, as shown in Figure 4.17. A fiber winding angle greater than 45° leads to contraction (at least at low pressures) and a fiber angle greater than about 45° leads to elongation of the FREE. The expansion behavior in Figure 4.18 shows an inverse relationship with fiber angle (greater expansion is observed with lower fiber angles), and maximum rotation (Figure 4.19) occurs at a fiber angle of approximately 30°. These results compare favorably to those in Connolly *et al.* (2015) with major differences being curve shape; extension, expansion, and rotation all appear to exhibit a higher degree of nonlinearity during pressurization in Connolly *et al.* (2015) compared to the results presented here. This is likely due to the difference in elastomer material properties [Connolly *et al.* (2015) analyze a silicon elastomer with Kevlar fibers], which has a large effect on behavior (see below subsection Impact of Material Properties). Nonetheless, the similarities in trends suggest that the following characteristics are similar for all FREE-like actuators: 1) a specific fiber angle at which extension transitions from negative to positive, 2) increasing expansion as fibers align with actuator length, and 3) a fiber angle at which rotation is maximum for a given pressure.

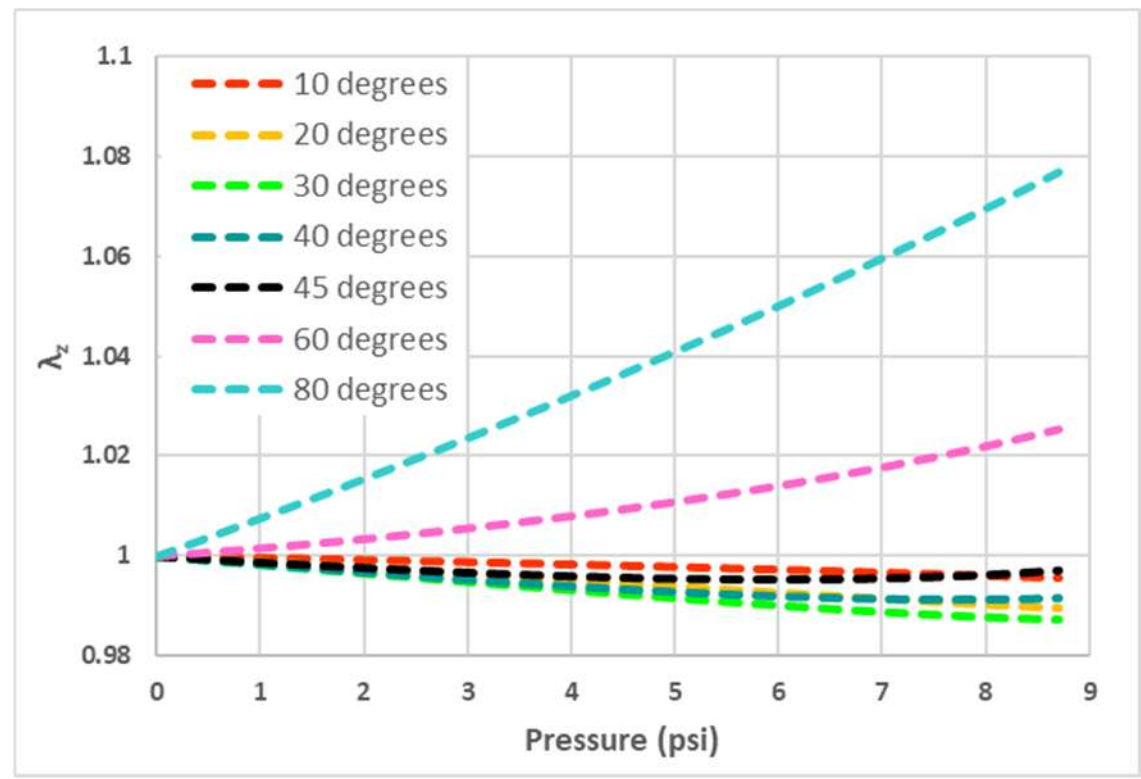

*Figure 4.17. FEA of FREEs for fiber angles over the range of 10° to 80° showing extension ($\lambda_z$) as a function of applied pressure*

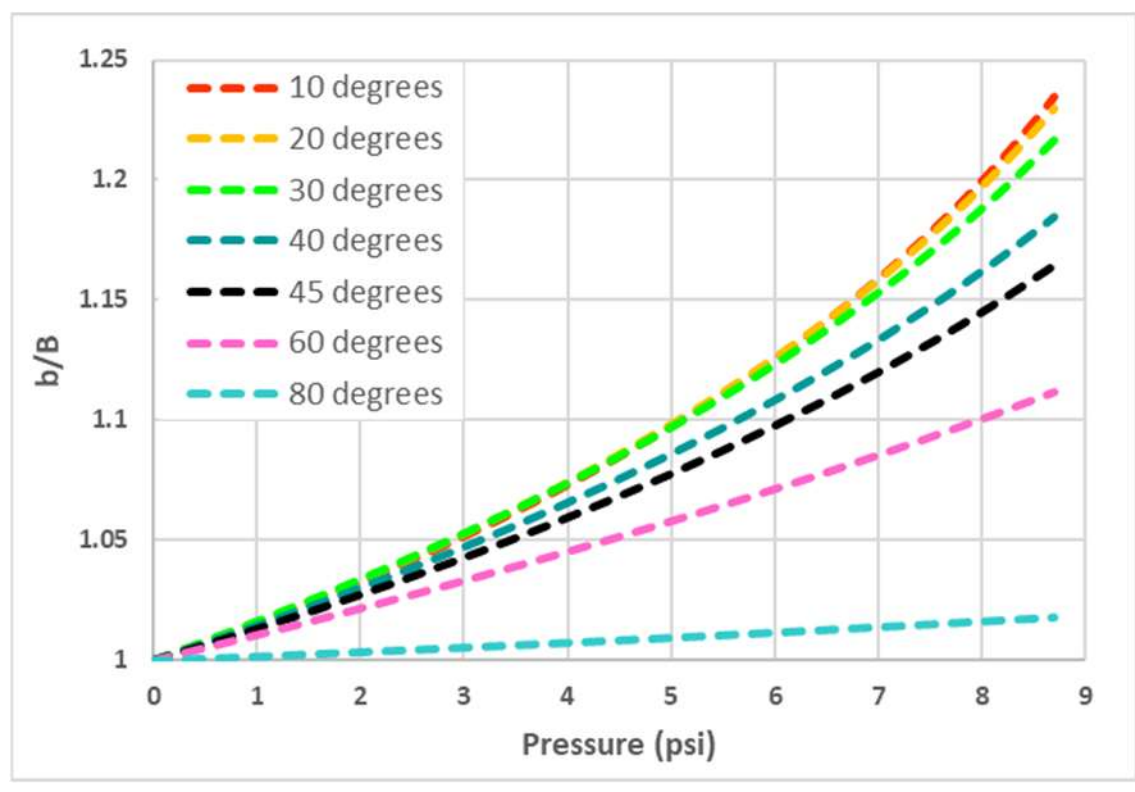

*Figure 4.18. FEA of FREEs for fiber angles over the range of 10° to 80° showing expansion (b/B) as a function of applied pressure*

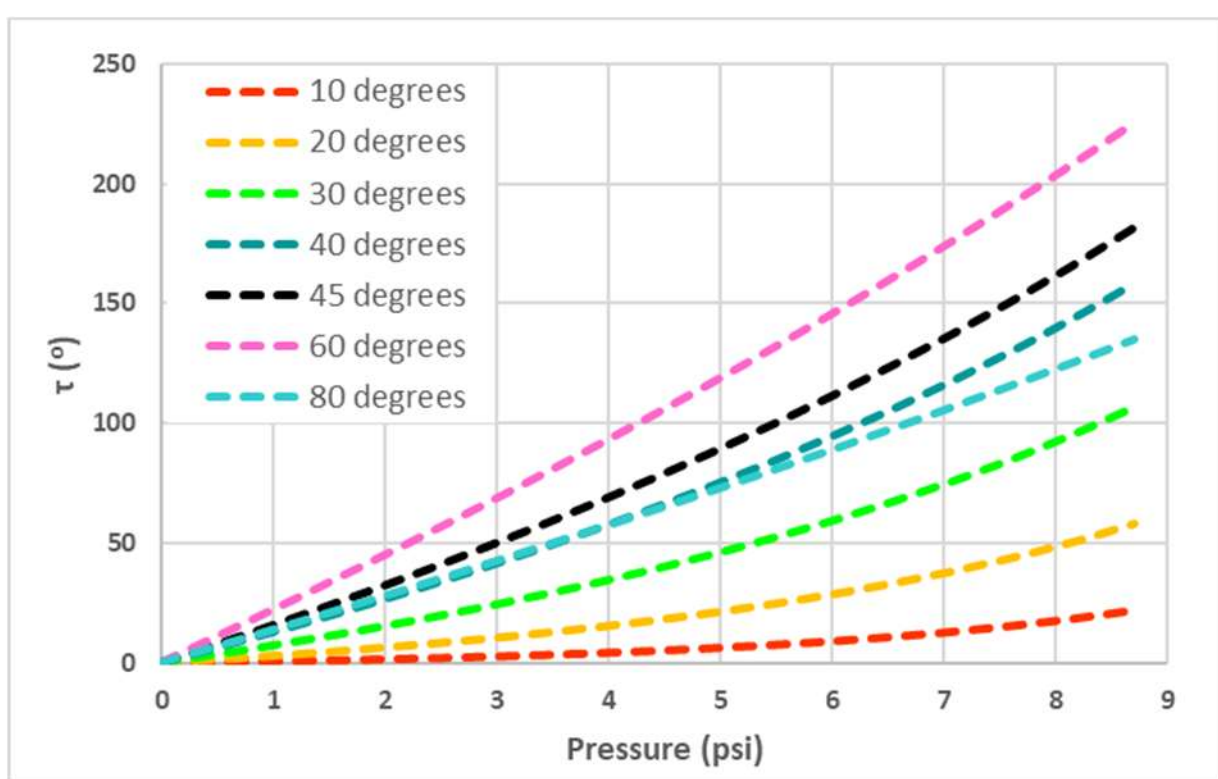

*Figure 4.19. FEA of FREEs for fiber angles over the range of 10° to 80° showing rotation per length (τ) as a function of applied pressure*

## Impact of Material Properties

To investigate the relative impact on response of the material properties of the fiber and elastomer, a parametric study was conducted using a FREE with a 20° fiber winding angle as a baseline. This winding angle yields significant extension, expansion, and rotation as compared to other winding angles, which may lead one of the three measures of deformation to approach zero. Moreover, this winding angle is particularly well suited for FREE-driven robotic devices in which significant extension and rotation are beneficial. The parameter study consisted of independently doubling and halving the elastomer and fiber stiffness, and the results obtained are show in Figure 4.20. Results show that changing the stiffness of the fibers has little effect on deformation, while elastomer stiffness greatly affects all deformations: extension, expansion, and rotation. Because the fiber deformation is so small relative to the elastomer, the exact fiber stiffness appears to have little effect on overall behavior so long as the fibers are significantly stiffer than the elastomer. Contrastingly, due to elastomer strains in excess of 25%, significant changes in elastomer stiffness can greatly affect deformation. The implications of these results when designing and manufacturing FREEs is that great attention must be given to accurately measuring, modeling, and understanding elastomer properties. A failure to do so may result in significant differences between desired performance of a FREE and observed behavior.

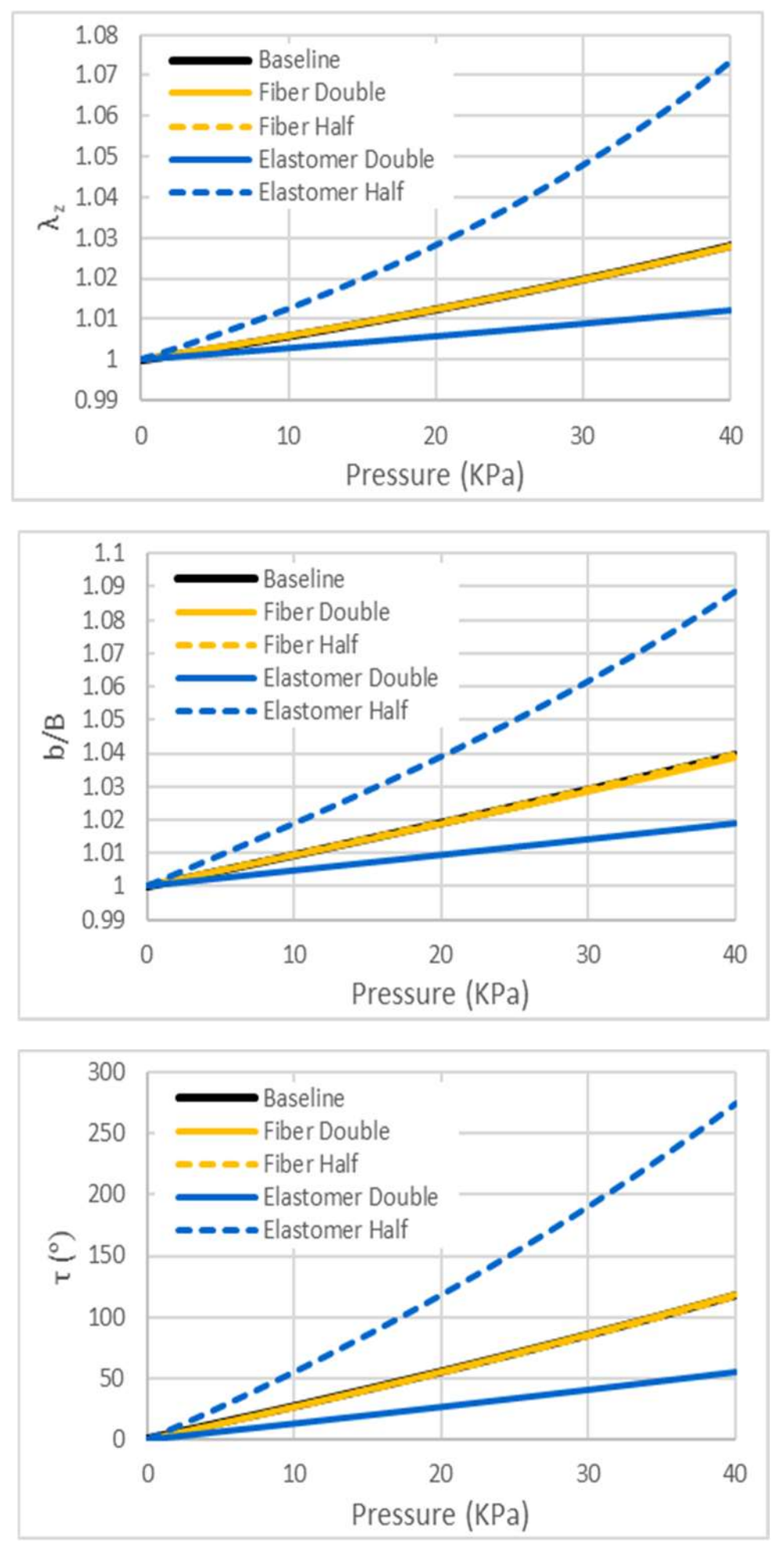

*Figure 4.20. FEA of FREEs with doubling and halving of elastomer and fiber material properties for a winding angle of 70° showing extension ratio ($\lambda_z$), expansion ratio (b/B), and rotation/length ($\tau$) as a function of pressure*

### Impact of Number of Fibers

As observed in the parametric study of material stiffness for both fibers and elastomer, it is understandable that fiber stiffness does not contribute significantly to the response of the system. However, having fibers as an important element of this composite (FREE) cannot be neglected. To explore the role of fibers more deeply, another parametric study was conducted varying the number of fibers in a FREE. Finding the optimal number of fibers for each winding angle needed to attain the desired performance is significant for design and manufacturing purposes. In this case study, winding angles of 10°, 30°, and 45° are selected and extension, expansion, and rotation of the FREE are compared within a range of number of fibers. Results in Figures 4.21, 4.22, and 4.23 show that at least four fibers are required for each the studied FREEs to fully benefit from the fiber reinforcement. Thus, changing the number of fibers above than a certain number has no contribution to motion (rotation, elongation, and expansion); however, it could still be beneficial for elastomer and fiber bonding in a FREE. The overall significance of this analysis suggests that in a single fiber family FREE, the number of fiber strings can be decreased to facilitate manufacturing, especially for smaller winding angles such as 20°. At the same time, fewer than a minimum number of fibers may not capture the role of fibers in the composite. Since fiber spacing becomes crucial in manufacturing, further studies are encouraged to better understanding the trade-offs between practical and theoretical considerations.

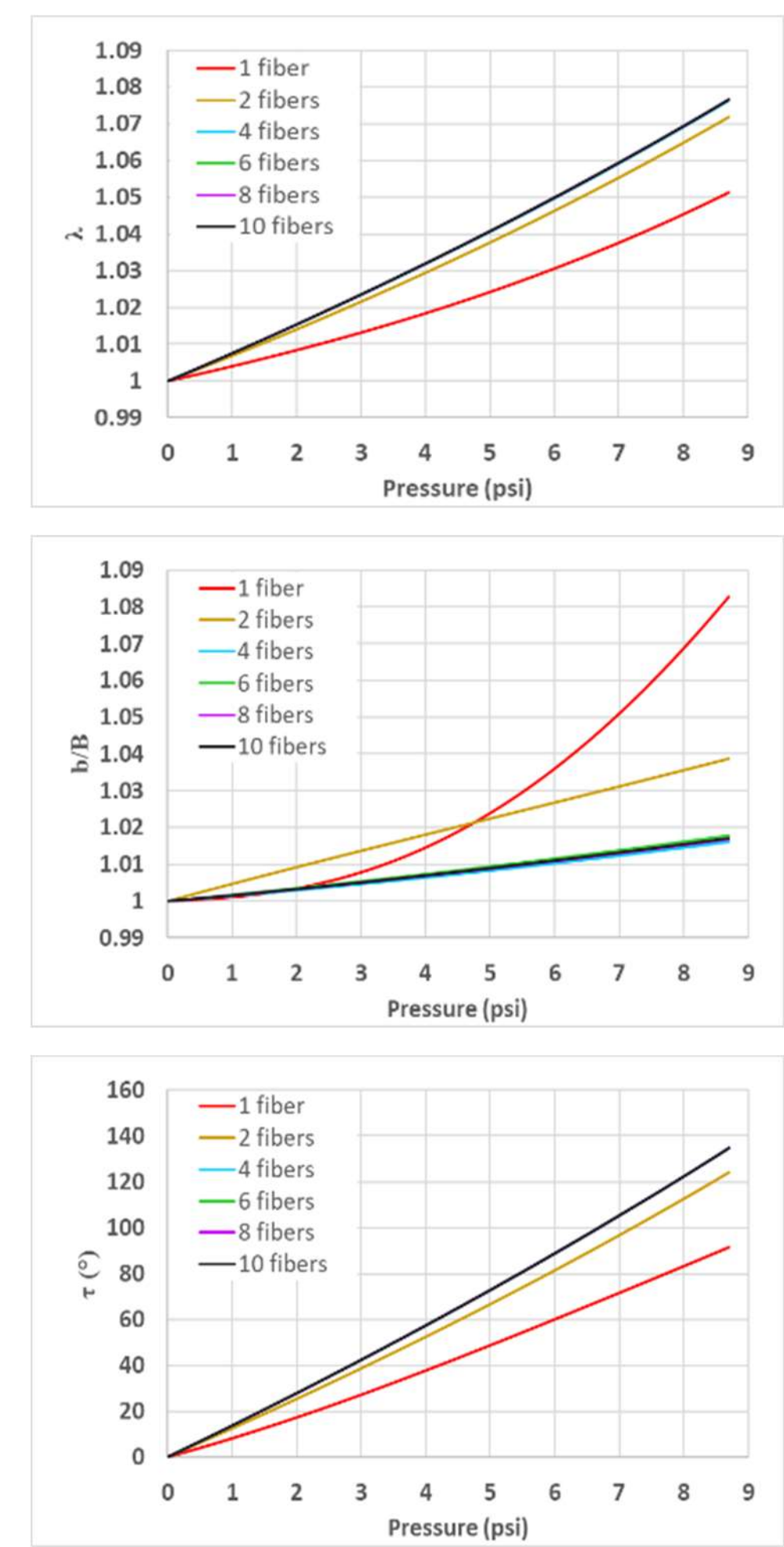

*Figure 4.21. FEA of FREEs with changing numbers of fibers for a winding angle of 10° showing extension ratio (λ), expansion ratio (b/B), and rotation (τ) as a function of pressure*

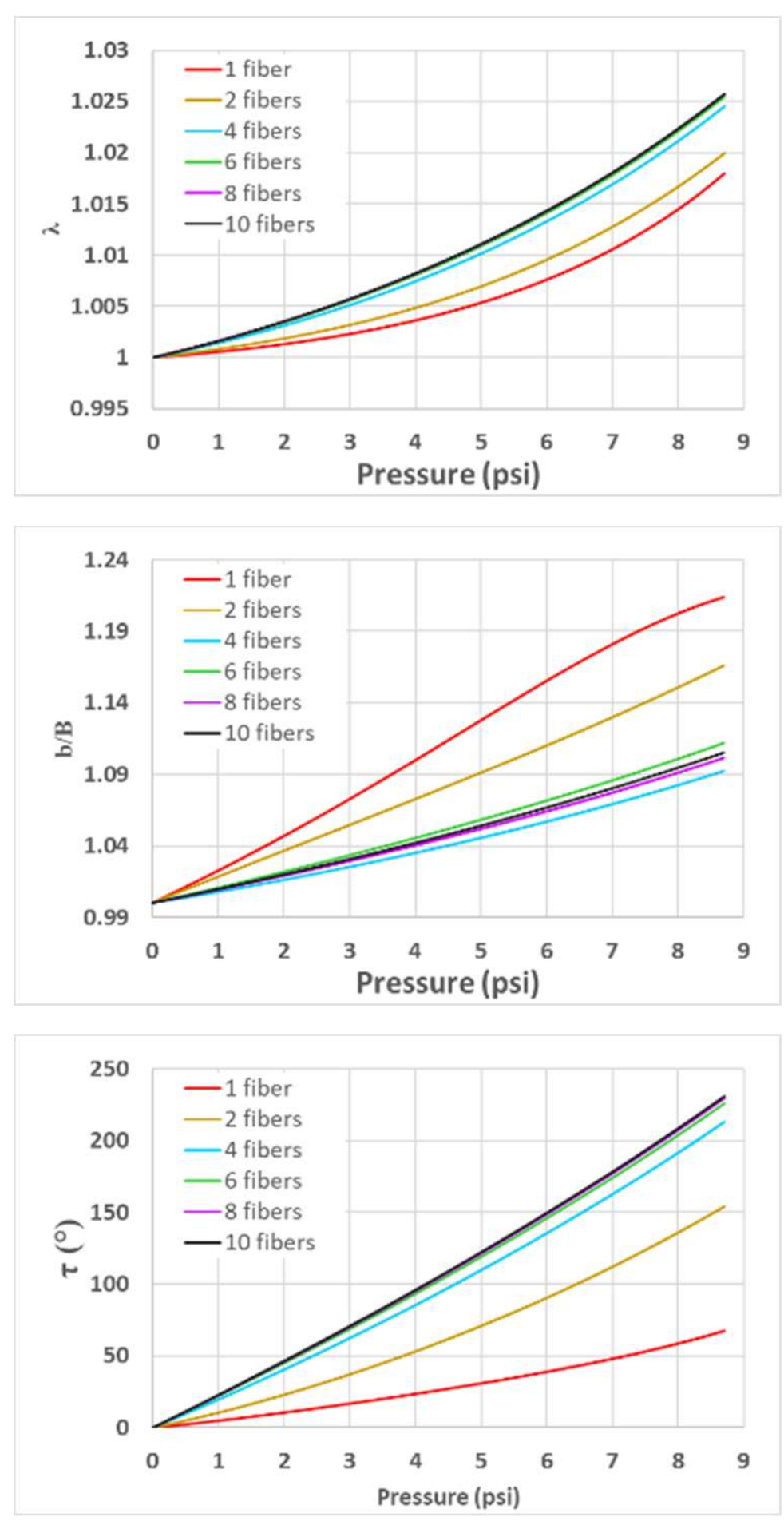

*Figure 4.22. FEA of FREEs with changing numbers of fibers for a winding angle of 30° showing extension ratio (λ), expansion ratio (b/B), and rotation/length (τ) as a function of pressure.*

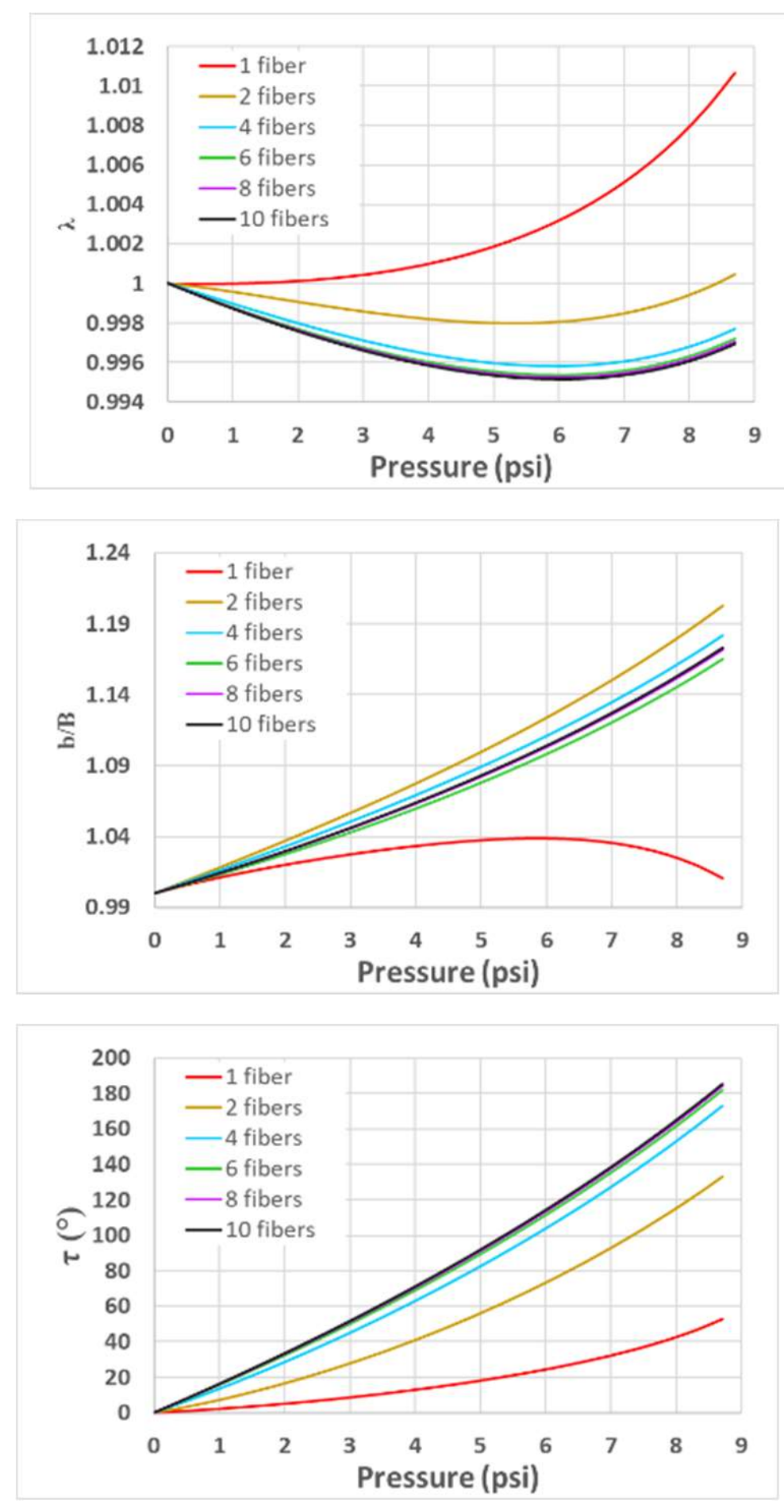

*Figure 4.23. FEA of FREEs with changing numbers of fibers for a winding angle of 45° showing extension ratio (λ), expansion ratio (b/B), and rotation/length (τ) as a function of pressure.*

## 4.8 FEA of Multiple FREEs

### Module Arrangement

A single FREE is not sufficient to create a soft robotic arm that is expected to perform various motions in space. The results in Section 4.7 suggest that although individual FREEs can be fabricated to produce unique displacement characteristics, an arrangement of four FREEs in a module has the potential to produce novel motions and forces (Bruder *et al.*, 2018) and useful performance characteristics of a robotic arm. Bruder *et al.* (2018) have explored the potential of combining multiple FREEs in a parallel configuration to obtain a "multi-dimensional soft actuation." There are various geometric arrangements for every distinct combination of FREEs in a module and thus experimentally investing all the possible combinations to develop a soft robotic arm is a tedious task. In this section, a finite element model a of square-module of four FREEs (Figure 4.24) has been created for three winding angles 30°, 40°, and 60°. This group of winding angles was selected because one is higher (60°) and the other two have lower (30° and 40°) than the theoretical critical winding angle (54.7°). The critical angle is the theoretical winding angle at which a fiber wound tube (pressure vessel) reverses direction between elongation and contraction (Roylance, 2011). The elongation of a pressure vessel wound with fibers at 54.7° is theoretically equal to zero (in reality it may not be the case due to variations in manufacturing). This phenomenon can also be verified by using the lumped-parameter model developed in Chapter 2. According to the Eq. (2.17), at static conditions without external loads, the elongation of a FREE goes to zero when the winding angle $\Gamma = 54.7°$.

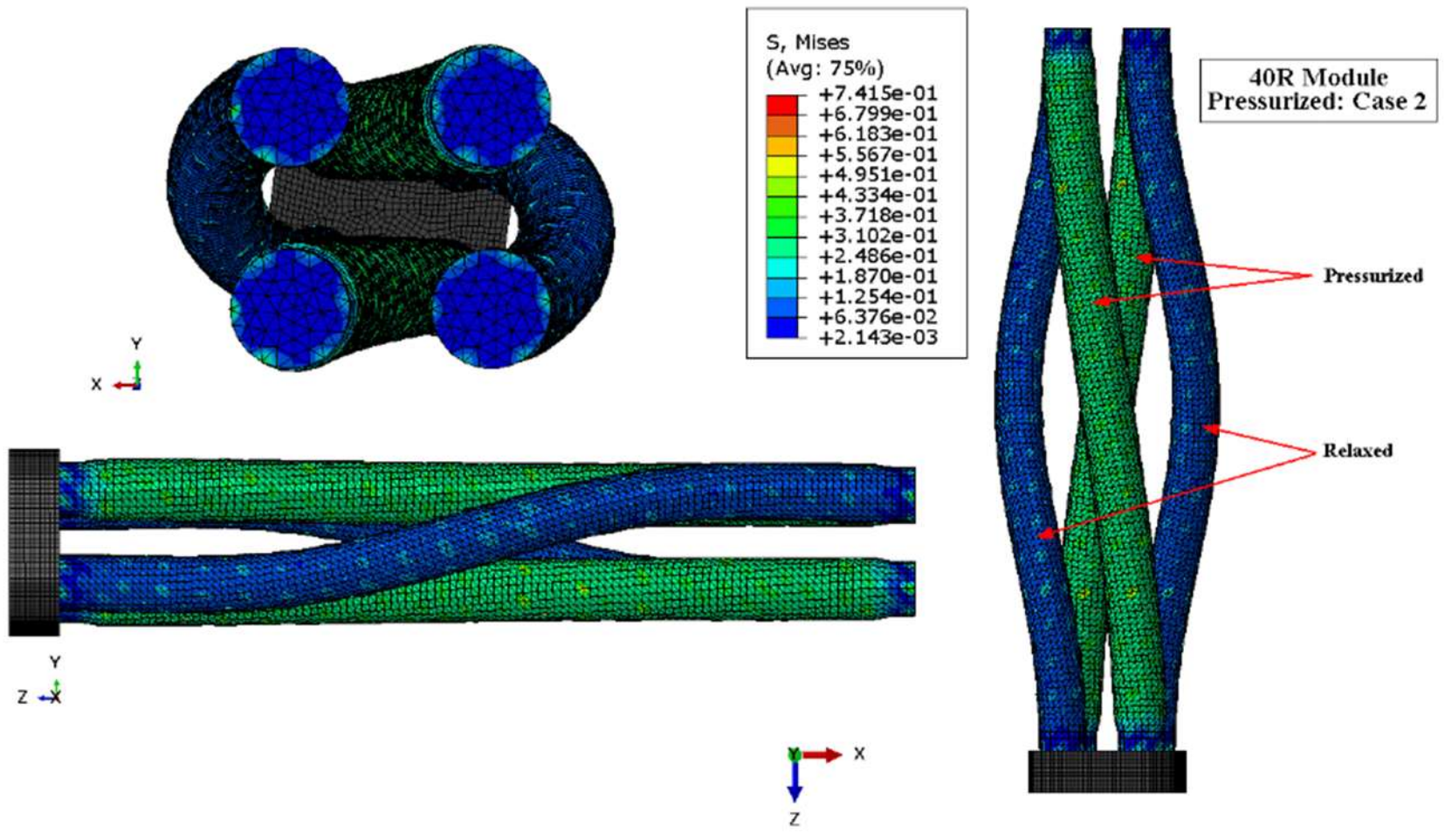

*Figure 4.24. Finite element model of a module consisting of four clockwise 40° FREEs in Abaqus (see Figures 4.25 and 4.26 for module conventions). Two FREEs diagonally across from one another are pressurized. The relaxed FREEs show relatively lower bending stiffness.*

Additionally, Bruder *et al.* (2018) found that fiber winding direction is important in the design of a module. The fibers can be wound in either a clockwise (R) or counterclockwise (L) direction (see Figure 4.25) which basically defines the direction of rotation. In the reminder of this section the significance of having the same or different winding directions in a module will be discussed along with simulation results. As a convention, “R” and “L” refer to a module with all clockwise or all counterclockwise winding angles respectively. Combining one pair of each winding direction in a module (two L and two R) is denoted as “RL” in which each pair of the same winding direction are diagonally across from one another.

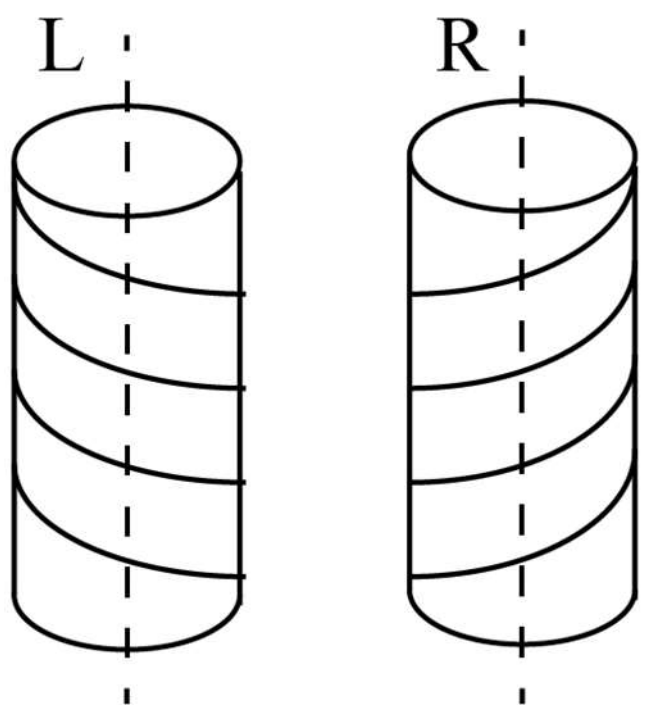

*Figure 4.25. FREE with clockwise and counterclockwise wound fibers*

For the purpose of this thesis, finite element models of four modules 30° (LR), 60° (LR), 40° (LR), and 40° (R) were created and tested over a range of pressures. Additionally, for each module, five different cases have been selected to explore the relationship between module pressurization and motion. The convention of numbered cases is shown in Figure 4.26, which displays which one of the FREEs numbered (1, 2, 3, and 4) is pressurized for each case.

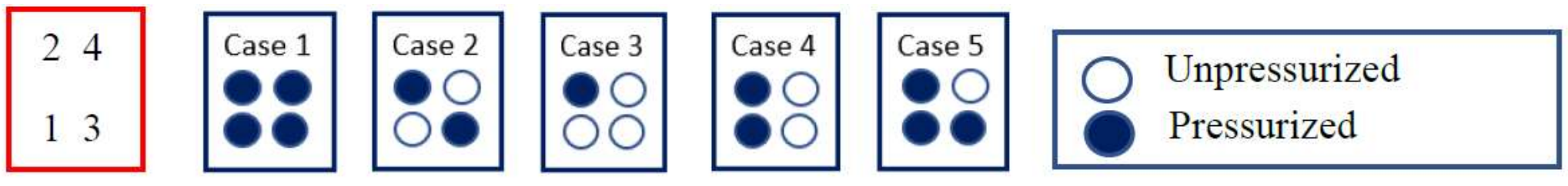

*Figure 4.26. The convention used for the combination of pressured FREEs in a module*

In all of the test cases, FREEs are modeled with 175 mm length, 9.52 mm inner diameter, and 0.8 mm thickness, and wound with six fibers. Table 4.1 to Table 4.4 indicate the results of simulations attempted by Brielle Cenci (2019) for each module using Abaqus software. Due to various reasons such as buckling, element distortion, and large deformations at higher pressures, the model did not converge in all cases. The green cells correspond to the successfully converged cases, red cells show cases that did not converge (some of them converged to within 98% of the full solution but nonetheless recorded as an unsuccessful test), and white cells represent cases that were not attempted. Note that the model was run for lower pressures (same for all FREEs 1, 2, 3, and 4) or by avoiding zero pressure in unpressured FREEs in cases 2, 3, 4, and 5. For example, in Table 4.1, the 30° (LR) model didn't converge for case 2 when pressurized to 10 psi and thus 7.25 psi of pressure was attempted to achieve convergence. Similarly in Table 4.4, 2 psi of pressure was applied in "unpressurized" FREEs in the 60° (LR) module (see Figure 4.26) to avoid the convergence issue for case 2.

| Pressure (psi) | 0 - 7.25 | 0 - 10 | 0 - 15 |
|---|---|---|---|
| case 1 | | | |
| case 2 | | | |
| case 3 | | | |
| case 4 | | | |
| case 5 | | | |

*Table 4.1. Convergence results for the 30° (LR) model*

| Pressure (psi) | 0 - 7.25 | 0 - 10 | 0 - 15 |
|---|---|---|---|
| case 1 | | | |
| case 2 | | | |
| case 3 | | | |
| case 4 | | | |
| case 5 | | | |

*Table 4.2. Convergence results for the 40° (LR) model*

| Pressure (psi) | 0 - 7.25 | 0 - 3.63 | 0 - 7.25[1] |
|---|---|---|---|
| case 1 | | | |
| case 2 | | | |
| case 3 | | | |
| case 4 | | | |
| case 5 | | | |

*Table 4.3. Convergence results for the 40° (R) model*

| Pressure (psi) | 0 - 7.25 | 0 - 3.63 | 0 - 7.25[1] | 0 - 10 | 0 - 15 |
|---|---|---|---|---|---|
| case 1 | | | | | |
| case 2 | | | | | |
| case 3 | | | | | |
| case 4 | | | | | |
| case 5 | | | | | |

*Table 4.4. Convergence results for the 60° (LR) model*

Results achieved by Cenci (2019) give valuable insight into the capabilities of this model and as well as the overall performance of a module with different configuration of FREEs. The FEA results suggest that case 1, case 2, and case 4 are the most fundamentally useful configurations in a module to create particular motions. Having all of the FREEs pressurized (case 1) in an LR-module produces pure elongation [Figure 4.27 (a)]. Case 2

[1] zero pressure avoided in unpressurized FREEs (cases 2 to 5 in Figure 4.26)

produces rotation without bending of an LR-module [Figure 4.27 (b)]. Pressurizing two adjacent FREEs (case 4) causes the module to bend in the opposite direction [Figure 4.27 (c)]. It is interesting to note at this point that cases 1 through 5 actual yield 15 unique motions. Case 1 yields one unique motion. Case 2 yields two by pressurizing opposite diagonals of FREEs. Case 3 yields four, case 4 four, and case 5 four, for a total of 15 unique motions.

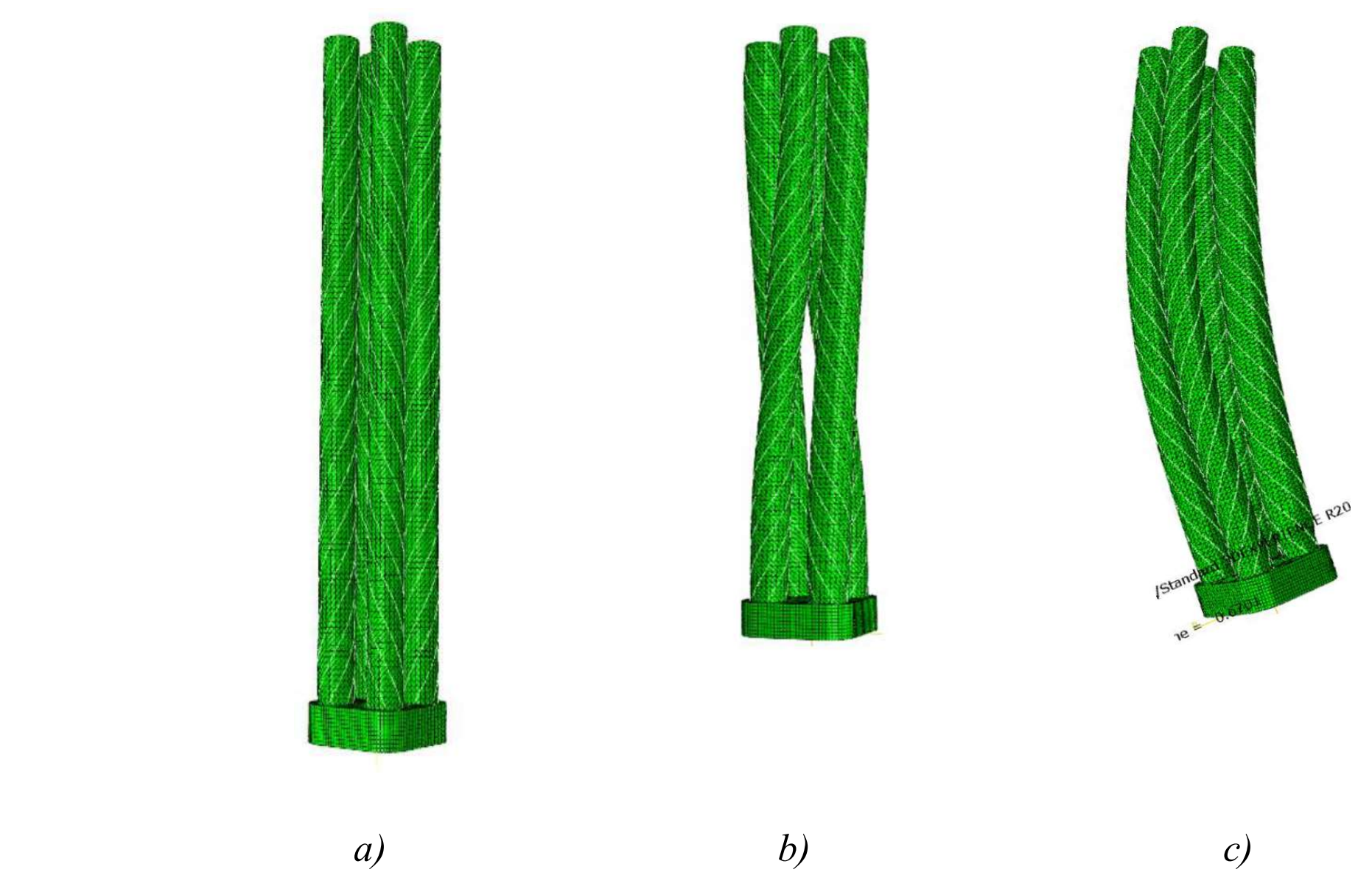

*a)* *b)* *c)*

*Figure 4.27. Finite element model of a 30° (RL) module showing deformation for a) case 1, b) case 2, and c) case 4*

The plots in Figure 4.28 to Figure 4.30 show rotation angle as a function of pressure for these RL modules as determined by the finite element model. Figure 4.28 shows a pressurized (LR) module, made of two pairs of FREEs with the opposite winding direction, that only generates contraction because rotations of the pressurized FREEs cancel out each other. Figure 4.29 shows the same type of module with two pressurized FREEs diagonally across from one another that only produces rotation since both pressurized FREEs rotate in the same direction. Figure 4.30 illustrates the bending motion of the same module; this time with two pressurized FREEs next to one another that their rotations cancel out each other and contribute to bending. Note that the direction of rotation or bending depends on the selection of the pressurized pair of FREEs in the module. Additionally, note that the motion in each case is not entirely one type (because the chosen node for extracting the results is not located at the center of the end cap) as the results show. Lastly, the results for the R-module are distinctly different than these for LR-module since the winding angles in all of the four FREEs are the same and they do not cancel out each other's rotation.

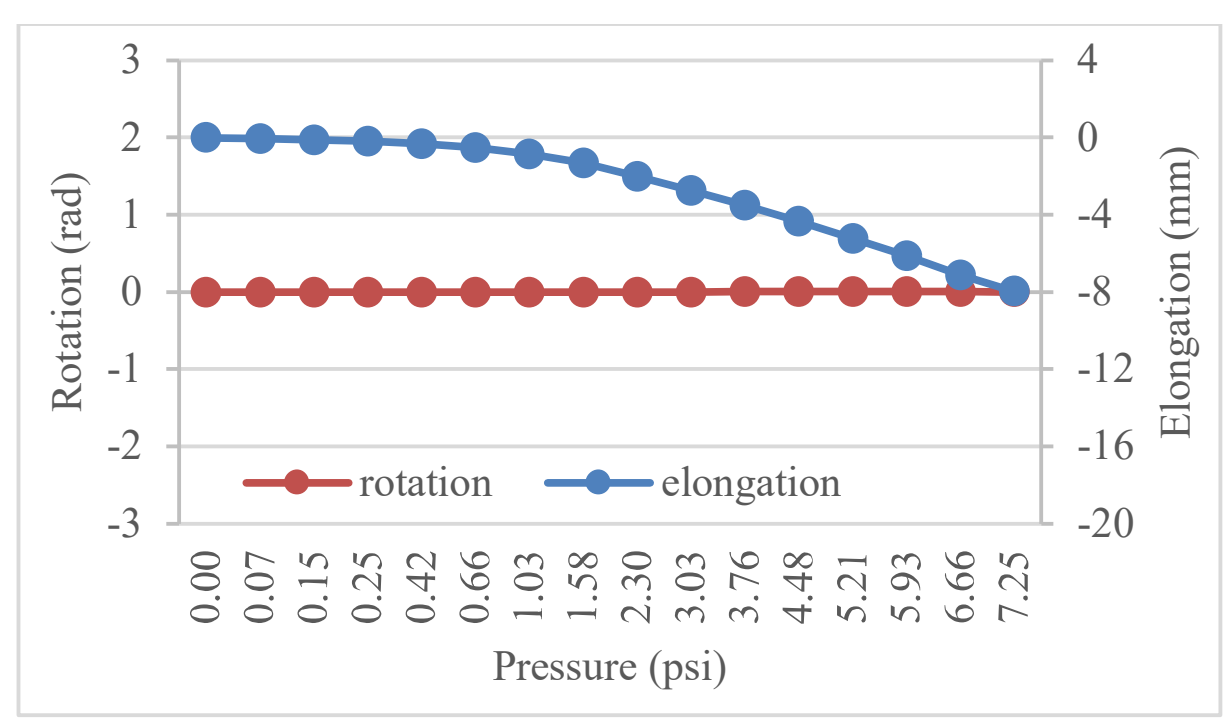

*Figure 4.28. Motion of 30° (LR) module with case 1*

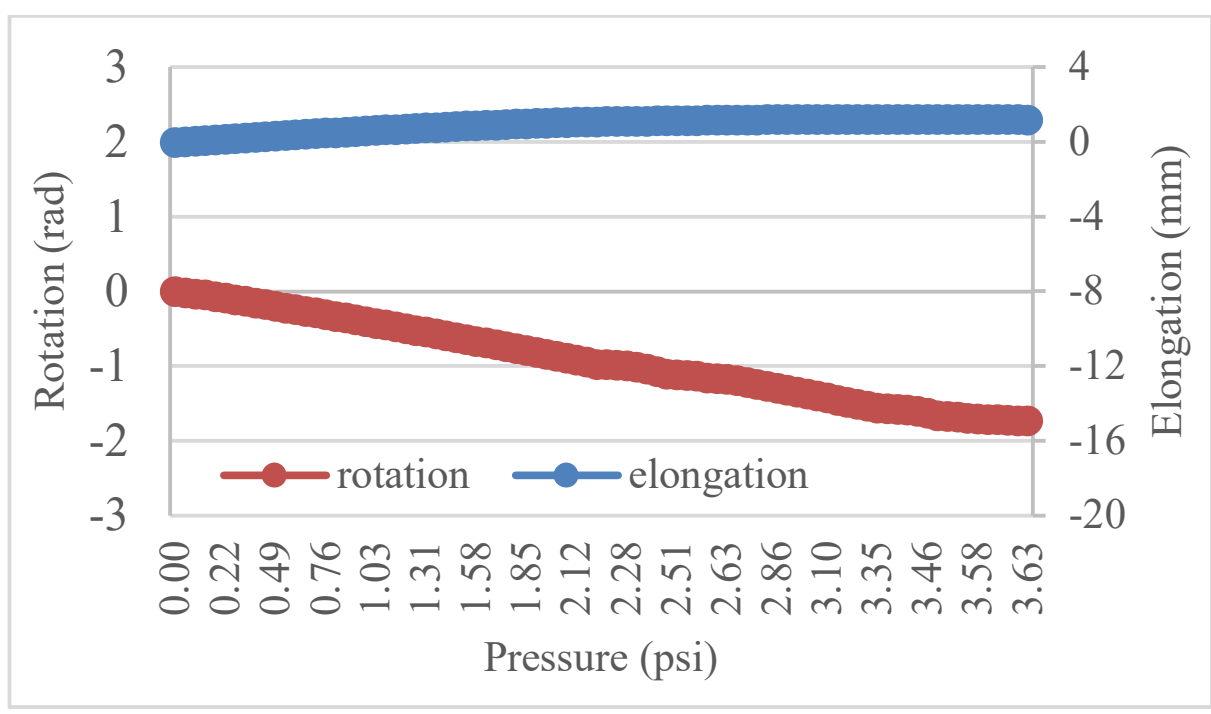

*Figure 4.29. Motion of 60° (LR) module with case 2*

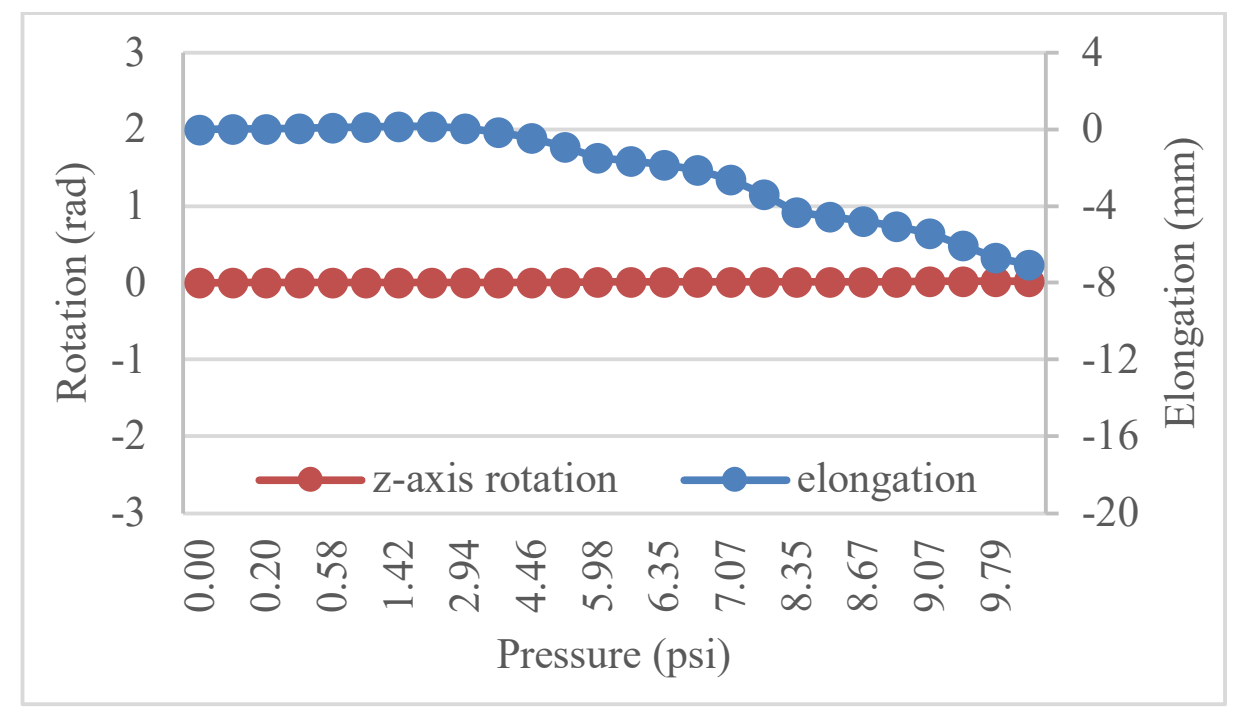

*Figure 4.30. Motion of 40° (LR) module with case 4*

## Workspace Study

The workspace of a manipulator is the reachable volume of space (including all the points in the space) that can be attained by the end effector. In general, kinematic design and geometrical arrangement considerations affect shape and type of the workspace. This matter is directly associated with the scale of tasks that can be performed by a robotic arm. According to (Craig, 1989), the workspace is categorized by two definitions: "Dexterous" and "Reachable" workspaces. The first one is referring to points in the workspaces that a manipulator can reach with all the possible orientations. The later definition is the volume of space that the arm can reach with at least one orientation.

In the case of a single FREE, the workspace is simply a one-dimensional line. As mentioned earlier, combining multiple FREEs in particular geometrical arrangements creates a workspace volume. Recalling the initial motivation of this project, developing a soft robotic arm made of FREEs, demands identifying the workspace of each module. To explore the reachable workspace (with any orientation) of a module, the 60° (LR) FEA model was run for the five pressurization cases (see Figure 4.26) and the position of the end effector was mapped in space. Figure 4.31 shows that cases 1 and 2 only generate axial motion. On other hand, cases 3, 4, and 5 produce bending motions which creates a concave shaped workspace. Note that each case of pressurization has its own unique orientation within the workspace and only reachable locations are studied in this analysis. Further studies need to be done to explore the existence or nonexistence of a kinematic solution in the workspace. At this point, observations from the FEA result suggest the locations of the boundaries of the workspace and provide insight into the kinematic capabilities of the 60° (LR) module. Similar results can be produced for other module configurations and the desired kinematic design selected to build a soft manipulator for a given application.

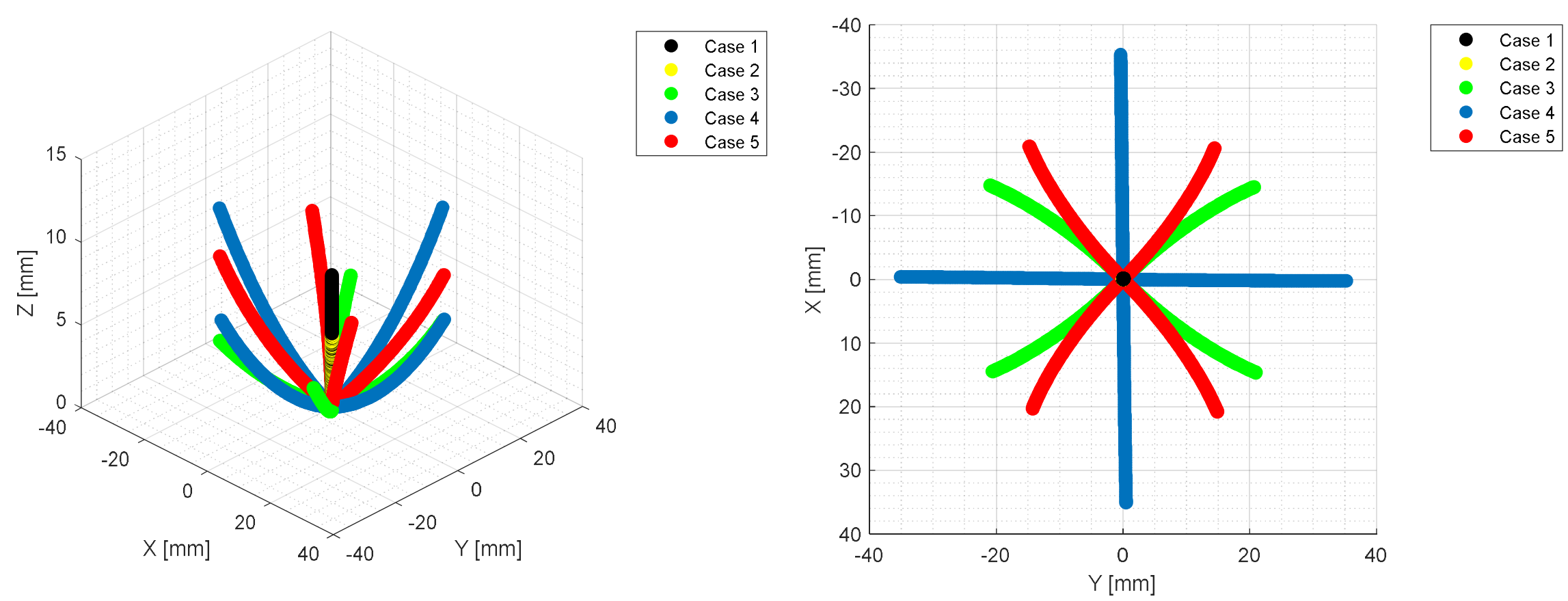

*Figure 4.31. Reachable points in space generated by pressurization cases in a 60° (LR) module*

To continue the workspace study, combinations of pressurization cases were used to investigate the locations of more reachable points between the lines shown in Figure 4.31, and linear interpolation was employed to find rough estimates for the overall workspace boundaries. This was done to obtain the overall shape of the module workspace, and additional surfaces between paths shown in Figure 4.31 have not been investigated. Figure 4.32 depicts this rough estimate of workspace boundaries as a function of pressure. Considering pressure as the input to the system is necessary to build a control system. The yellow points shown in the figure depict the locations that the end effector reaches at a pressure of 7.25 psi. This indicates that to reach a particular point in space, multiple pressurization cases can be used; however, some of them may be able to do it with lower pressures.

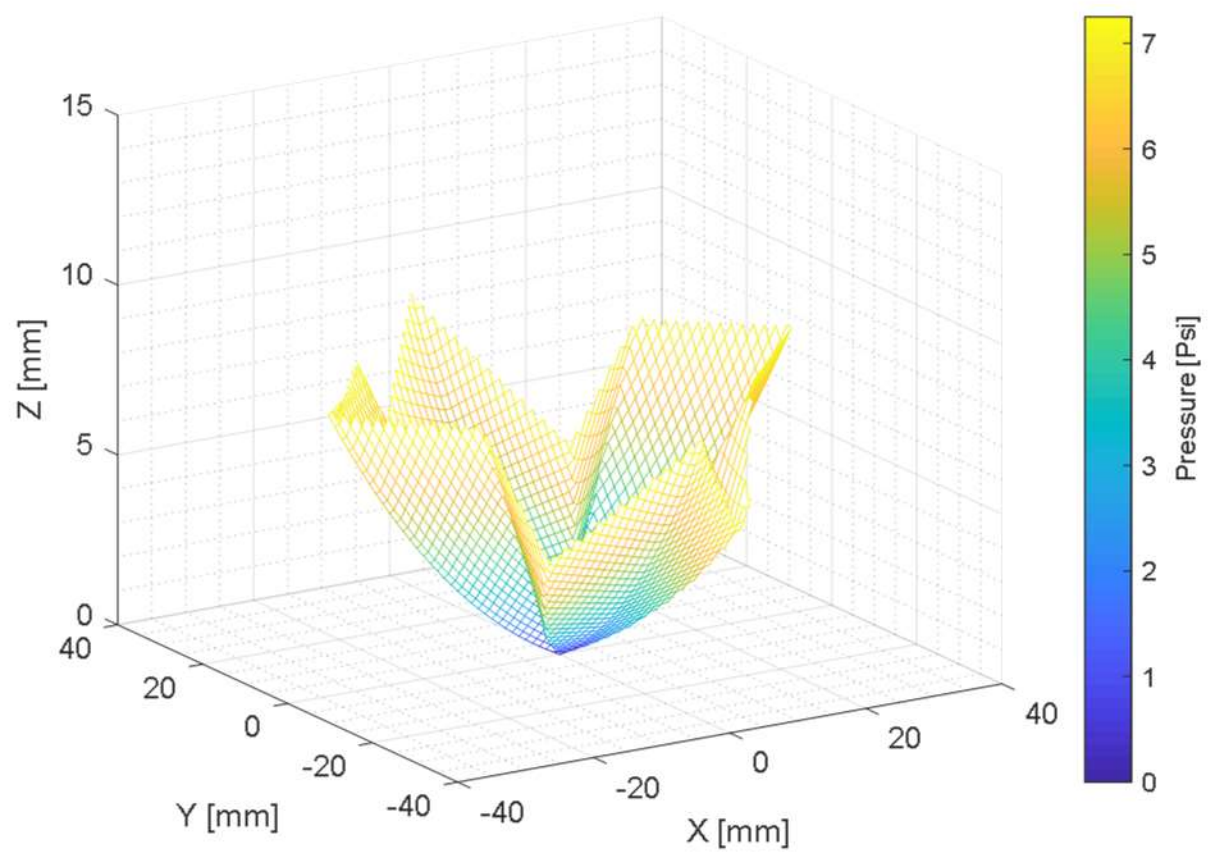

*Figure 4.32. Rough estimate of workspace of a 60° (LR) module*

One of the common issues that have been observed with FREEs is buckling under various conditions. For an example, Figure 4.33 is a snapshot from the finite element model of a 60° (LR) module that is pressurized to 7.25 psi that illustrates the buckling behavior. As described in Section 4.6, it is difficult to find a buckling pattern for all types of FREEs, although it is obvious that high torsions, bending, and contractions, particularly at lower pressures, usually cause buckling. In the cases of pressurized four-FREE module, buckling behavior was mostly observed for cases 1, 2, and 4, where large deformations exist. Hence, reaching all points within the workspace (see Figure 4.32) may not be feasible due to buckling of special cases.

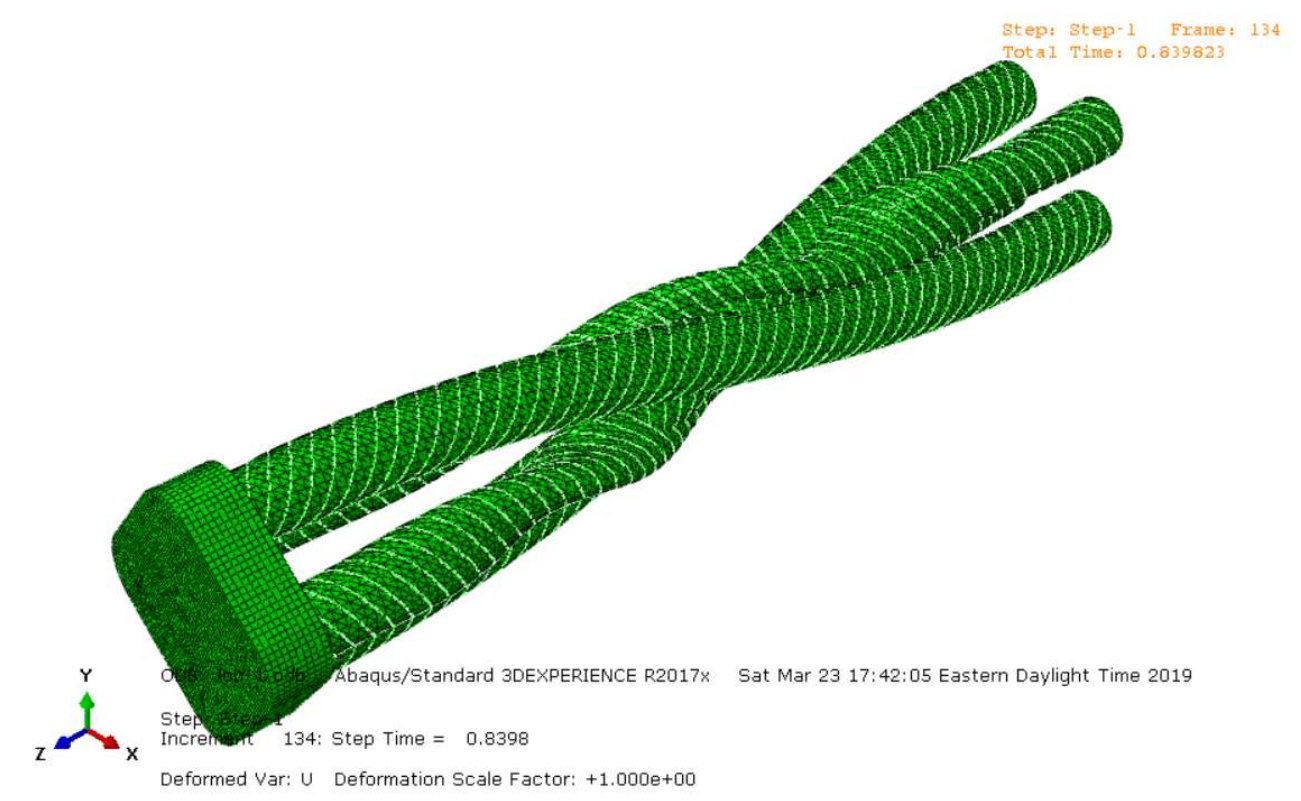

*Figure 4.33. Buckling of the 60° (LR) module pressured to 7.25 psi*

In general, there are potential reasons that FEA simulation of a nonlinear system such FREE cannot meet convergence: material instability (model parameters were fitted for a limited range of strains), increment size, and plastic effects of elastomers, which they were already considered for the FEA model of FREEs. On the other hand, convergence issues can exist in the presence of singularities. For example, for force-driven finite element analysis of a nonlinear system, there is not a unique solution (displacement). However, there is one solution (force) for a displacement-driven analysis and the simulation yield a better convergence as stated in Section 4.6. In the finite element analysis of a module, the force boundary conditions at the end cap generate multiple displacement solutions for the nonlinear system of equations, which can cause the lack of convergence in the simulation at large deformations. Similar convergence issues were observed in a force-driven simulation of a single FREE in Section 4.6. Nonetheless, the finite element simulation of modules that did not converge provided useful information on various buckling cases which were also observed experimentally.

## 4.9 Summary

The results of the finite element model presented in this chapter provide new insights into the behavior of single and multiple FREEs and demonstrate the effectiveness of using this tool for studying FREE-like actuators. The Ogden hyperelastic constitutive model for the elastomer prevents significant softening at high strains, and truss elements (only supporting axial loads) used to model fibers help to capture the buckling and bending behavior of FREEs. Parametric analyses show that the behavior of FREEs is highly sensitive to elastomer material properties but relatively insensitive to fiber stiffness. Additionally, once the fiber stiffness exceeds a certain level, the fiber essentially acts as an inextensible material relative to the elastomer. Finally, the finite element analysis of a module is a useful tool in finding the reachable points that lie within the workspace, aiding the designer in exploring various designs before manufacturing. As future directions for improving the finite element model, modifications should be considered for the Ogden constitutive model, the effects of wall thickness, and the element types of the elastomer since it effectively controls the behavior of FREEs.

# 5

# Experimental Apparatus

In order to validate the modeling approaches and simulation results presented in the preceding chapters, a set of experimental measurements and observations needed to be performed. This chapter describes the apparatus used to run various experiments and presents how the components of the system work together. The physical setup used for this project is similar to the version used at the RAM Lab at the University of Michigan (Robotics and Motion Laboratory at the University of Michigan, n.d.). It should be noted that major components of this system such as pressure regulators, digital-to-analogue converters, and filters are borrowed from the University of Michigan. This system is capable of running tests for various combinations of FREEs (up to four at a time) and determining spatial motions and orientations. Figure 5.1 shows a CAD model created in SolidWorks to optimize the arrangements of the major components.

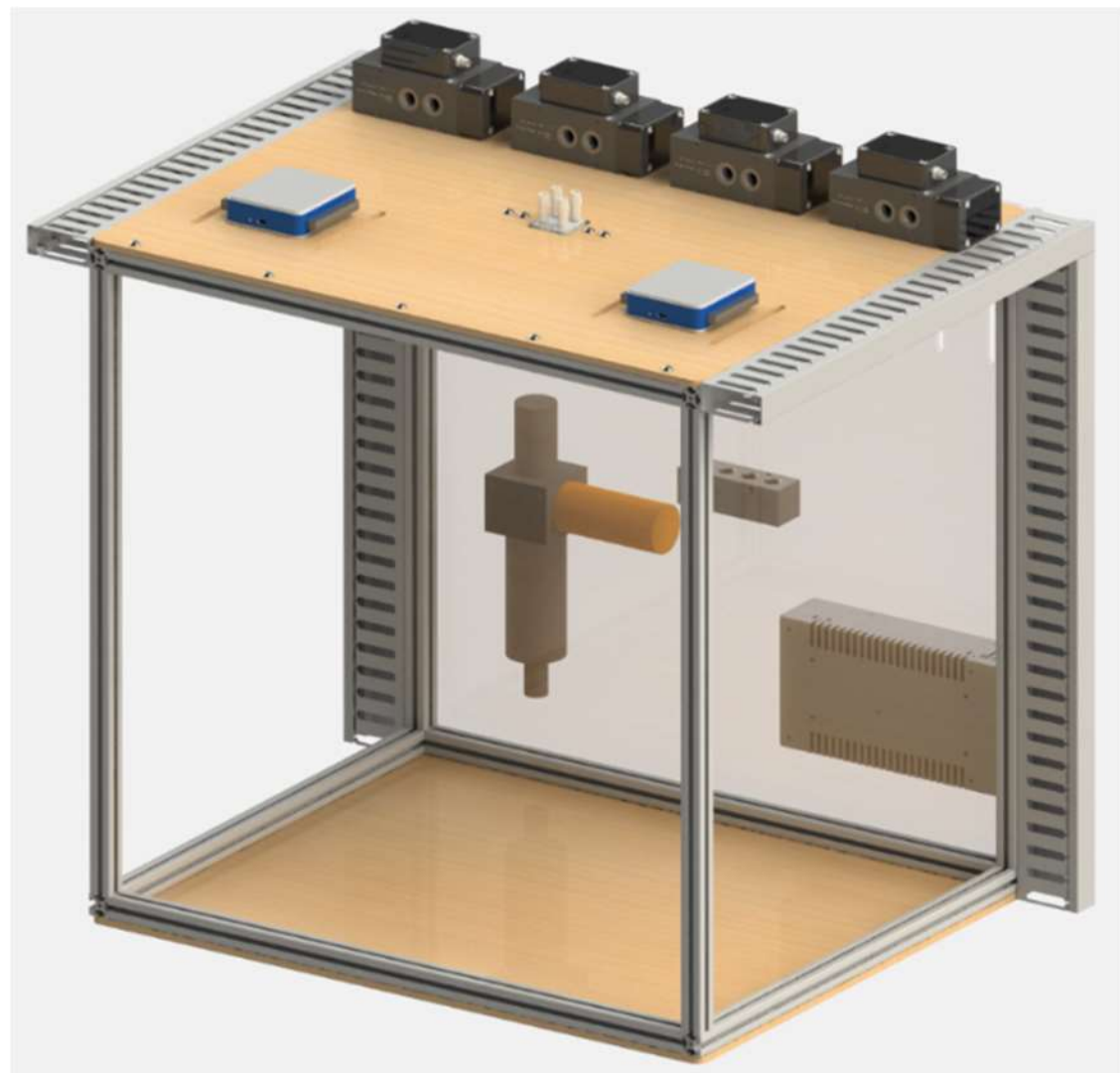

*Figure 5.1. CAD model of the experimental setup in SolidWorks*

## 5.1 Components

This section describes the developed apparatus (Figure 5.2) based on the CAD model in Figure 5.1 used to obtain the experimental data. Table 5.1 lists the major components with their application and designated number shown in Figure 5.4. Additionally, Figure 5.3 diagrammatically shows that the connection between those components. This system is capable of supplying air to four FREEs in a module. Due to the sensitivity of the pressure regulators, air filters have been mounted between the wall air supply and the inlet of the regulators. The laboratory compressed air supply is connected to two filters to remove particles, dust, oil, and moisture before entering the manifold inlet. The manifold divides the air-flow into four outlets. Each of them is filtered through an inline miniature filter and then connected to pneumatic regulators by ¼" tubing. The power supply (AC-DC) Model 360-12 is used to switch on or off the electrical supply for all of the devices. One data acquisition device (NI-USB DAQ 6001) is connected to each pair of pressure regulators to transmit signals independently, and communication with a supervisory computer is simplified by using a USB 2.0 10 port HUB.

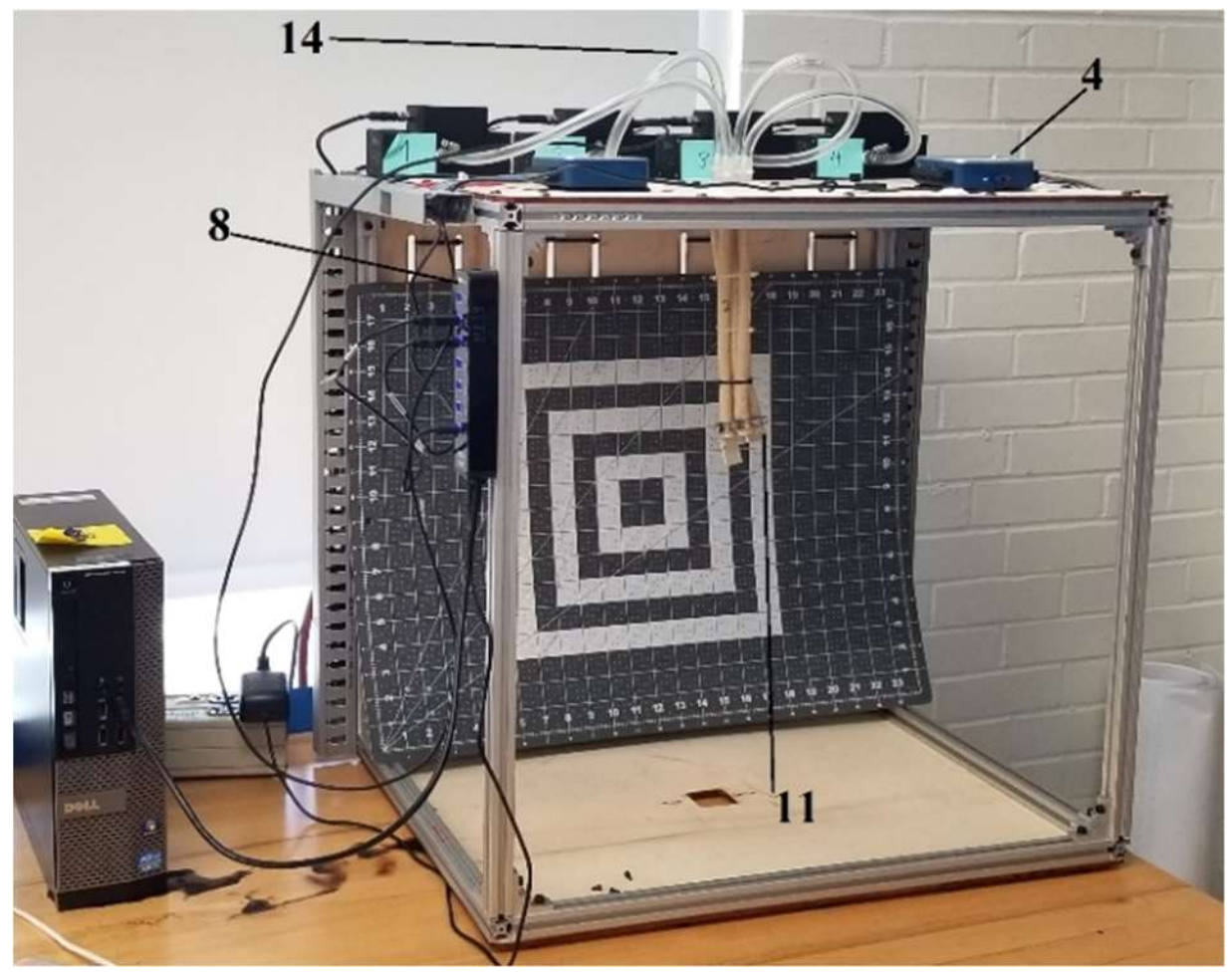

*Figure 5.2. Front view of the experimental setup*

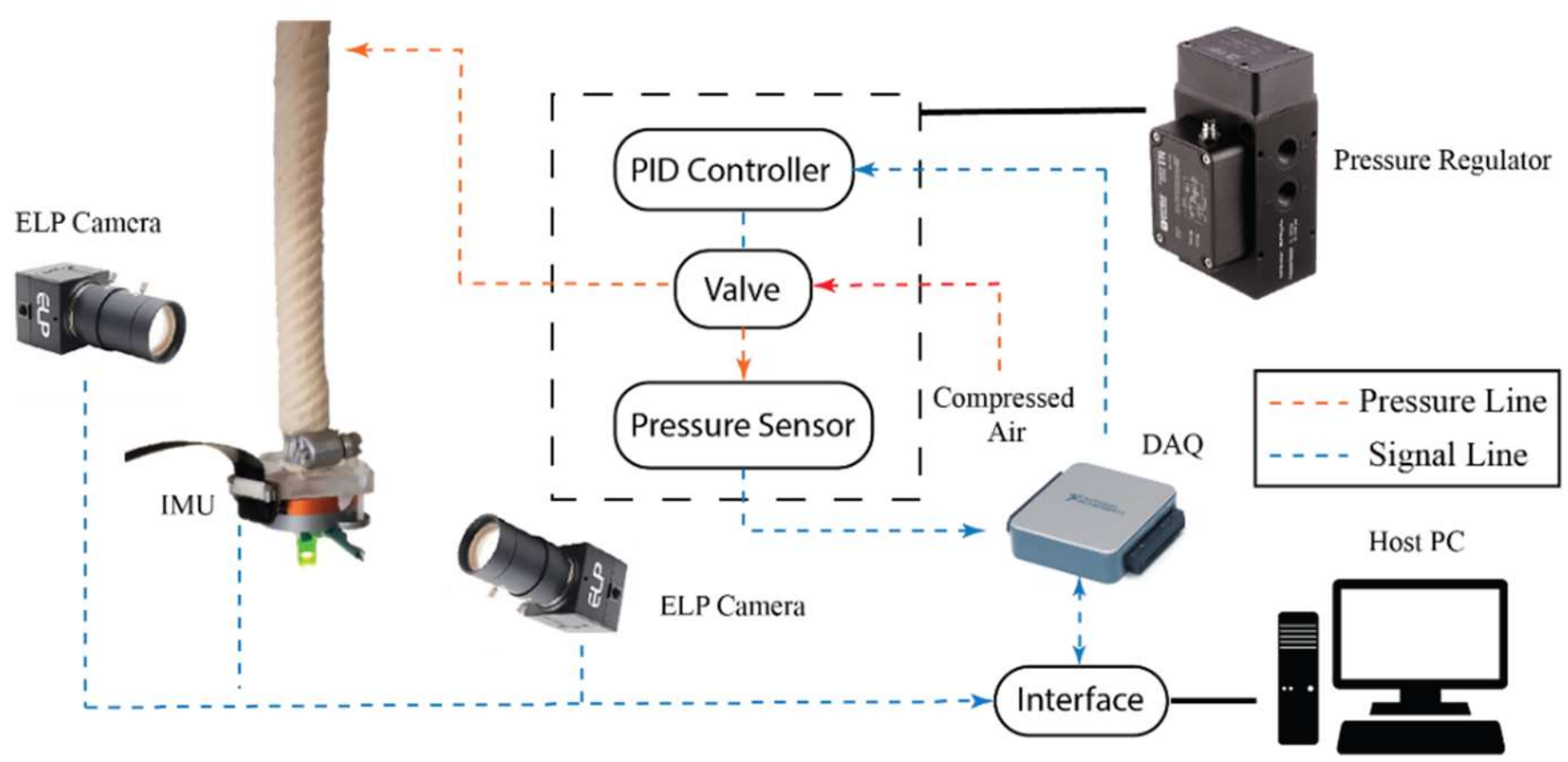

*Figure 5.3. Diagrammatic representation of the components of the experimental setup*

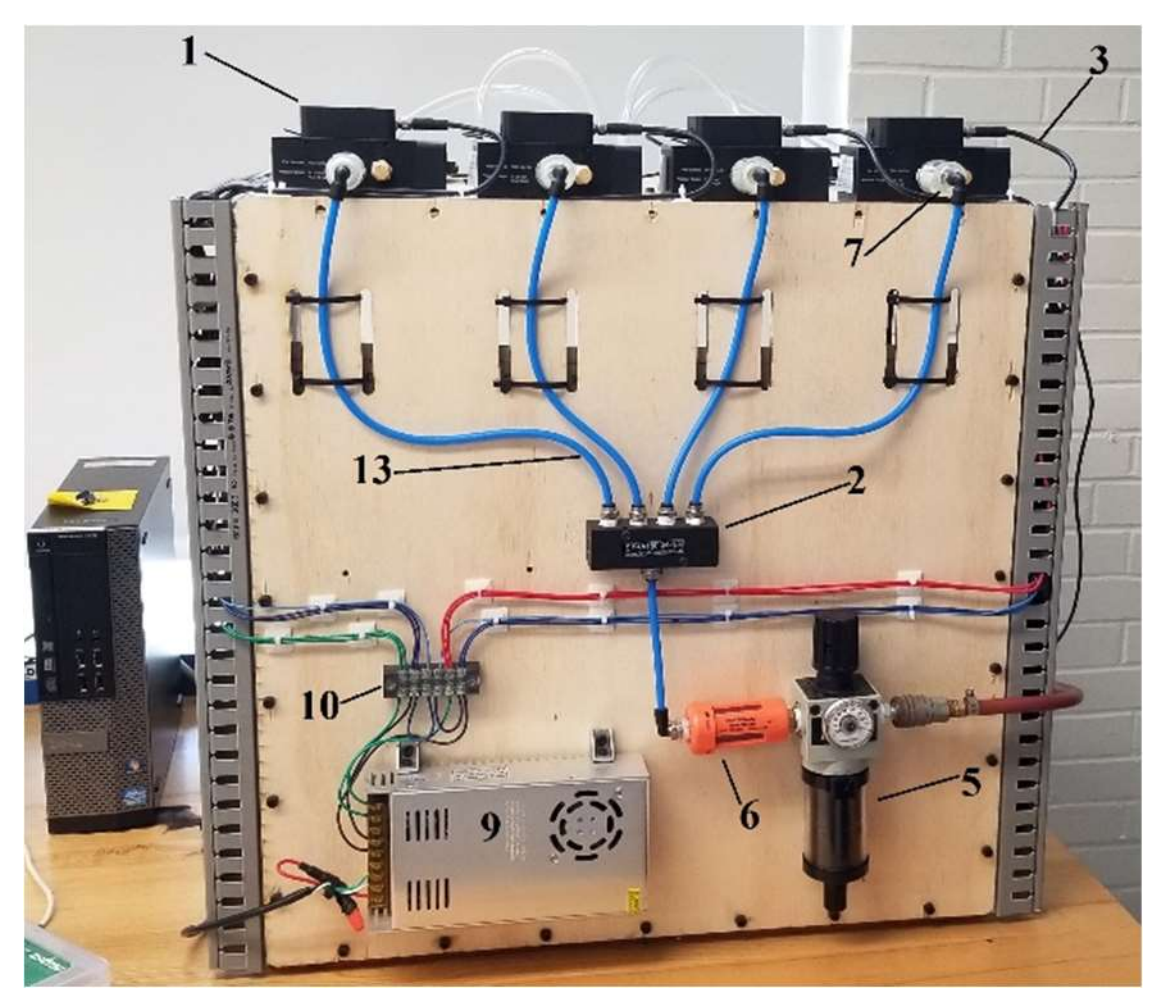

*Figure 5.4. Back view of the experimental setup*

| Number | Component | Quantity | Description |
|---|---|---|---|
| 1 | TR-025-g10-s | 4 | Electro-pneumatic pressure regulator |
| 2 | Header Manifold | 1 | Air pressure distributor |
| 3 | 4-pin M8 cable | 4 | Signal and power for TR regulators |
| 4 | NI-USB DAQ 6001 | 2 | Data logging/sending commands |
| 5 | A-APS-FRG air treatment unit | 1 | Air filtering/regulate/gauge |
| 6 | DD1008-2 Mini Desiccant Filter | 1 | Air dryer |
| 7 | 02FA10A PARKER WATTS | 4 | Coalescing air filter |
| 8 | USB 2.0 10 port HUB | 1 | Bus power connections |
| 9 | Model 360-12 switching power supply | 1 | AC to DC convertor |
| 10 | Screw-Type Terminal Block | 1 | Wire connections |
| 11 | SLA 3D printed end caps | 2 | FREE holder/cap |
| 12 | SparkFun 9DoF Razor IMU | 1 | Inertial measurement unit |
| 13 | 1/4” tube | | Air connections |
| 14 | 3/8” tube | | Air connections |

*Table 5.1. List of components used in the experimental setup*

### TR Pressure Regulator

The TR-025-g10-s (Figure 5.5) is an electro-pneumatic pressure regulator that converts a voltage or current input command into precise proportional air pressure as an output. It benefits from a technology called direct-acting voice-coil to deliver smooth, accurate air pressure control. This regulator can output 0-145 psi (0-10 bar) air pressure by scaling the input (voltage or current). The TR regulator has a Proportional-Integral-Derivative (PID) control system for adjusting the response depending on the connected physical system. The ranges for the command inputs are 0-10 VDC and 2-20 mA for voltage and current commands respectively. It also has an internal pressure sensor to measure the actual output pressure as represented by a 0-10 VDC signal. A 4-pin m8 cable makes the connection for power, command, and feedback output signals. There is a digital signal port to connect the device to a computer with a USB cable. All of the parameters can be adjusted through the TR configuration interface software (see Section 5.2).

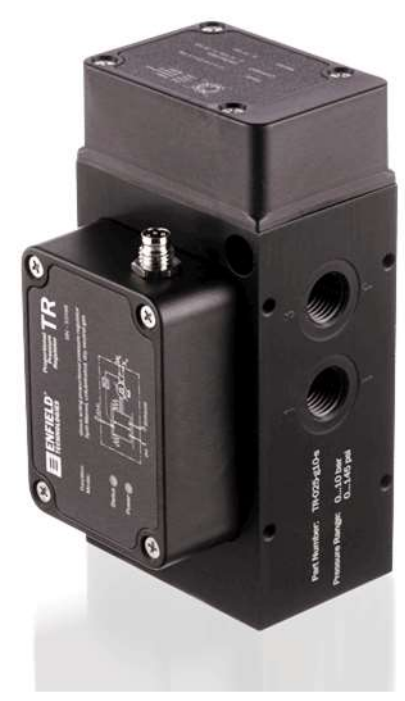

*Figure 5.5. Enfield TR-025-g10-s electronic pressure regulator*

**NI-USB DAQ 6001**

A National Instrument USB data acquisition (DAQ) device enables commands to be sent to the pneumatic pressure regulators. The NI-USB-6001 (Figure 5.6) has eight analog input (AI) channels, two analog output channels (AO), 13 digital input/output (DIO) channels, and a 32-bit counter. Each one of these devices can output analog signals to one pair of pneumatic valves. The DAQs are connected with USB cables to a computer and work as the communication device between pressure regulators and a LabVIEW user interface (see Section 5.3).

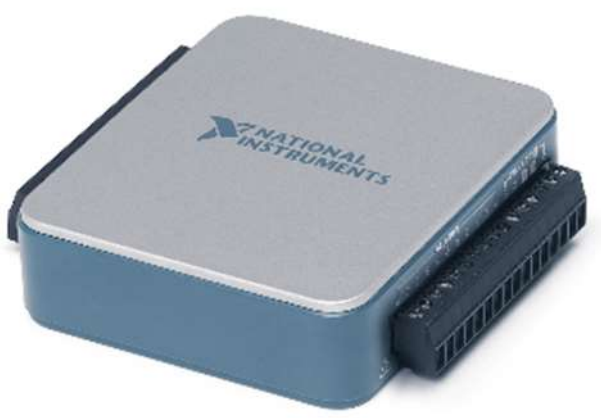

*Figure 5.6. National Instrument-USB DAQ 6001*

**Inertial Measurement Unit (IMU)**

In this project, several different models of IMU (Inertial Measurement Unit) were used to measure rotation of FREE (see Figure 5.7). All models include 3-axis accelerometers, gyroscopes, and magnetometers. Depending on the required communication protocol, each one was used in a particular experiment (see Table 5.2).

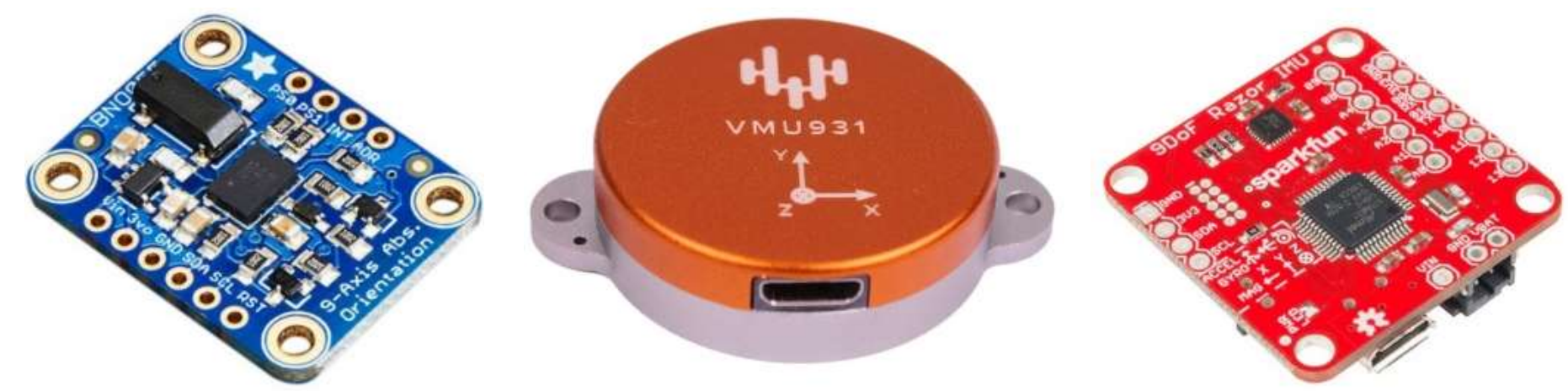

*Figure 5.7. Inertial Measurement Unit models from left Adafruit BNO055, VMU931, and SparkFun*

| Model | Communication | Software | Application |
| --- | --- | --- | --- |
| Adafruit BNO055 | I2C | Arduino | PID control |
| VMU931 | USB | VMU reader | Static Measurements |
| SparkFun 9 DOF Razor | USB | LabView | PID control |

*Table 5.2. Description of Inertial Measurement Units used for experiment*

Only the Euler rotation angle data of this multi-purpose IMU is being used in the current experimental measurements. Figure 5.8 shows a SparkFun IMU attached to the end of a FREE to measure rotation. The repeatability of the received data from the IMU is sufficient to capture the rotation and even small rotational vibrations of the system (see Section 5.3). However, a 2° change of rotation angle per second (drift) was observed which can cause problems in stationary measurements. This inherent error in IMU measurements can be reduced (Esser, Dawes, Collett, & Howells, 2009) to limit drift over of time but it was not the area of interest in this project. Another comment is that the (micro-USB) cable attached to the IMU exerts an additional load at the end of the FREE, and thus has an impact on the output data, although it was not found to be problem for closed-loop control and simply required higher pneumatic pressures to reach a given set point.

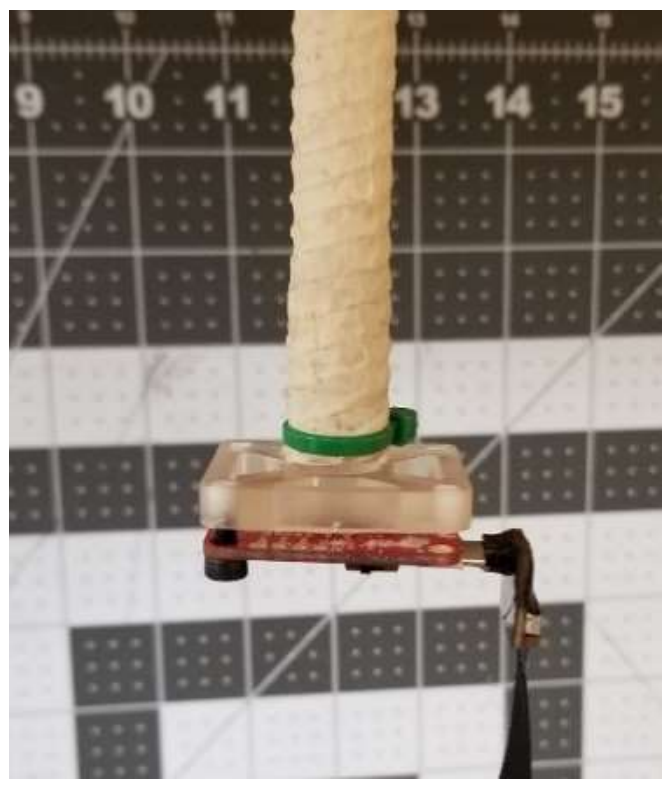

*Figure 5.8. SparkFun IMU attached to FREE*

## 5.2 System Calibration

The Enfield TR pressure regulators need to be tuned to the physical system, so the control gains were adjusted for each regulator with a closed output port (Figure 5.9). Gain adjustments were made by following steps similar to these in Section 3.2 to evaluate their influence on the response over time. In addition to the PID gains, there are other settings provided by the manufacturer in the Enfield software interface (Figure 5.10) to compensate for other variations in the system and produce a smoother response. Each of these are described below.

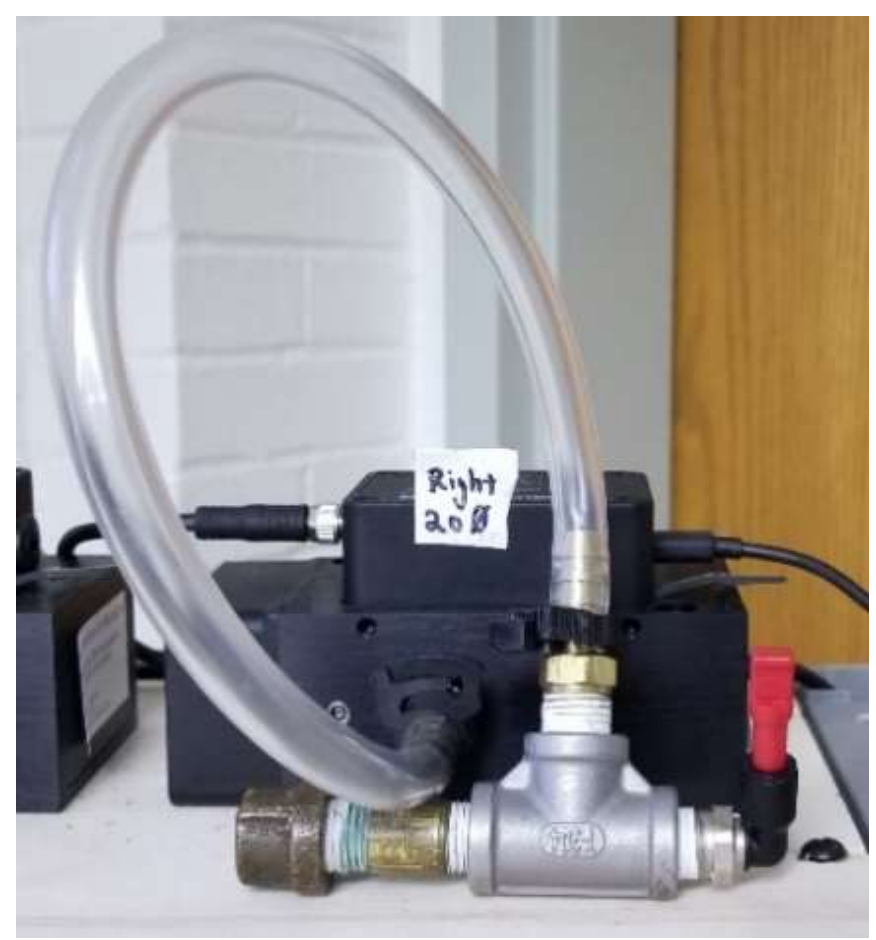

*Figure 5.9. TR pressure regulator with closed outlet*

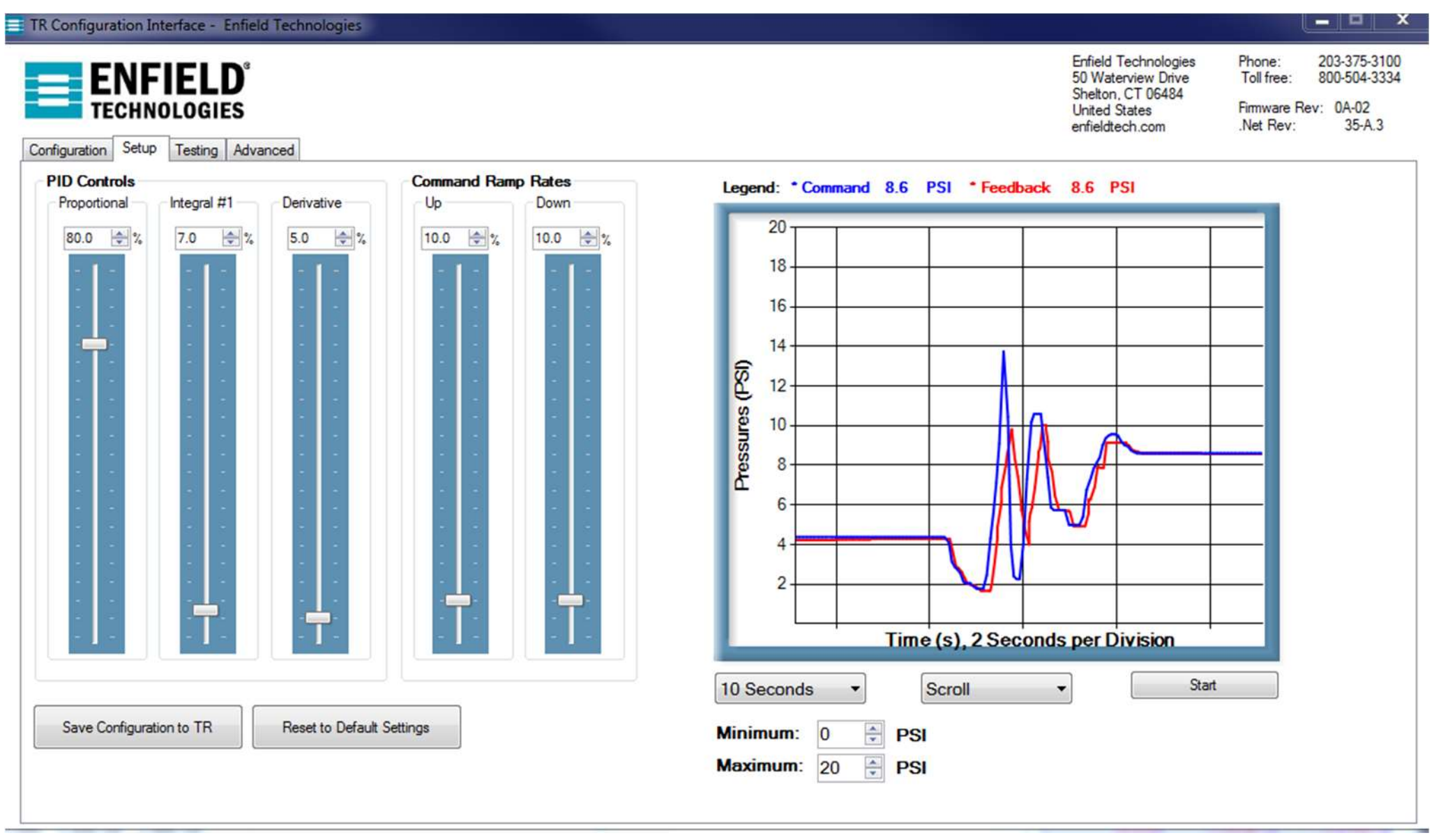

*Figure 5.10. Screenshot of TR user interface environment, PID gain adjustment tab*

**Proportional Gain**

This gain impacts the accuracy and the rate at which the pressure reaches a desired set point. Increasing the proportional gain generally improves accuracy and ability to follow the command pressure. An excessive amount of proportional gain increases overshoot.

**Integral Gain #1**

The integral gain is used to reduce steady-state error between the commanded pressure and the regulator's output. It is also useful to compensate for leakage from the controlled volume. Increasing the integral gain can make the system unstable and oscillatory.

**Derivative Gain**

The derivative gain influences overshoot and can produce an overdamped response with a longer response time.

**Command Ramp Rates**

The Ramp Up and Ramp Down settings determine the rate at which the command pressure can increase or decrease the controlled pressure per second. In other words, a lower percentage value produces a steeper slope in the response. This can produce a faster response but values below 10% may cause overshoot.

**Integral Window**

Basically, there are two different integral gains provided in this pneumatic pressure regulator. The integral window determines the threshold at which one of the integral gains is used. If the difference between the command pressure and the actual pressure is more than the integral window, integral gain #2 is used to control the outlet. Otherwise, integral gain #1 is used.

**Integral Gain #2**

As mentioned in the description of the integral window, this gain is the integral gain of the controller if the difference between the command and actual pressure exceeds the integral window.

**Filter**

This setting adjusts the response rate of the control algorithm depending on the volume downstream of the outlet port (pressure sensor region). The default value is set by the manufacturer at 1500 for a "1 liter volume at the end of 1 meter of 3/8" tubing" (Enfield Technologies, 2018).

**PID Tuning TR Pressure Regulator**

Based on the description of the proportional gain, integral #1 gain, and derivative gain, three values (two extremes and one mid-range) were used to tune the regulators. Figure 5.11 to Figure 5.15 show plots of system response to gain variations, as the pressure goes from zero to 5 psi. For each gain variation, value of the gain increases from left to right. Note that the gains are presented as a percentage on the TR user interface environment (Figure 5.10).

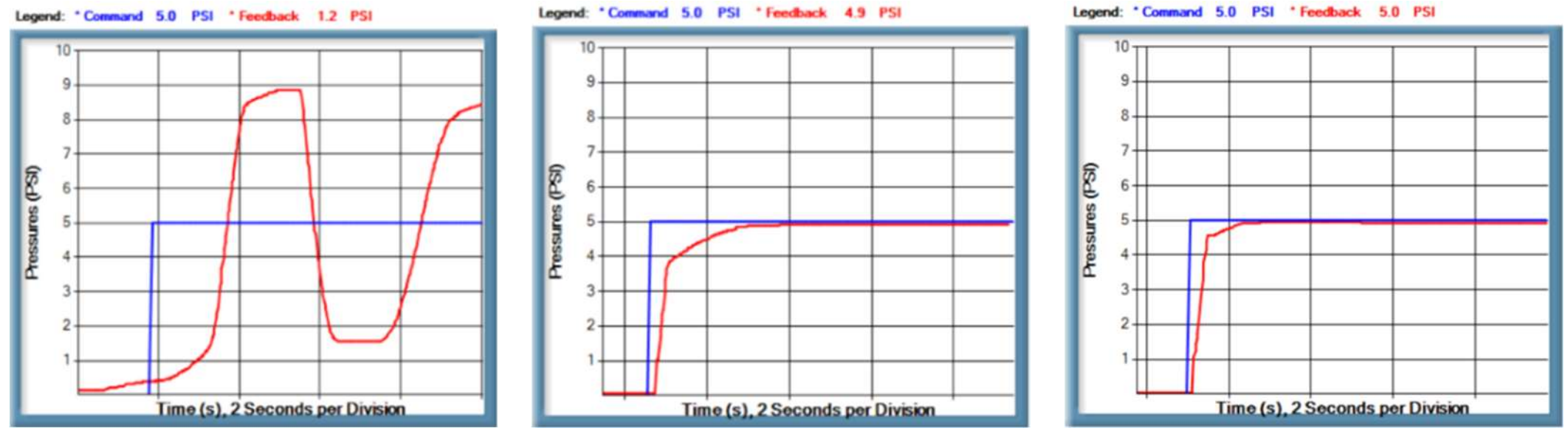

*Figure 5.11. System response to the Proportional gains of 0% (left), 40% (middle), and 80% (right), Integral #1 gain=7%, Derivative gain=5%, Ramp Up/Down rate=10%*

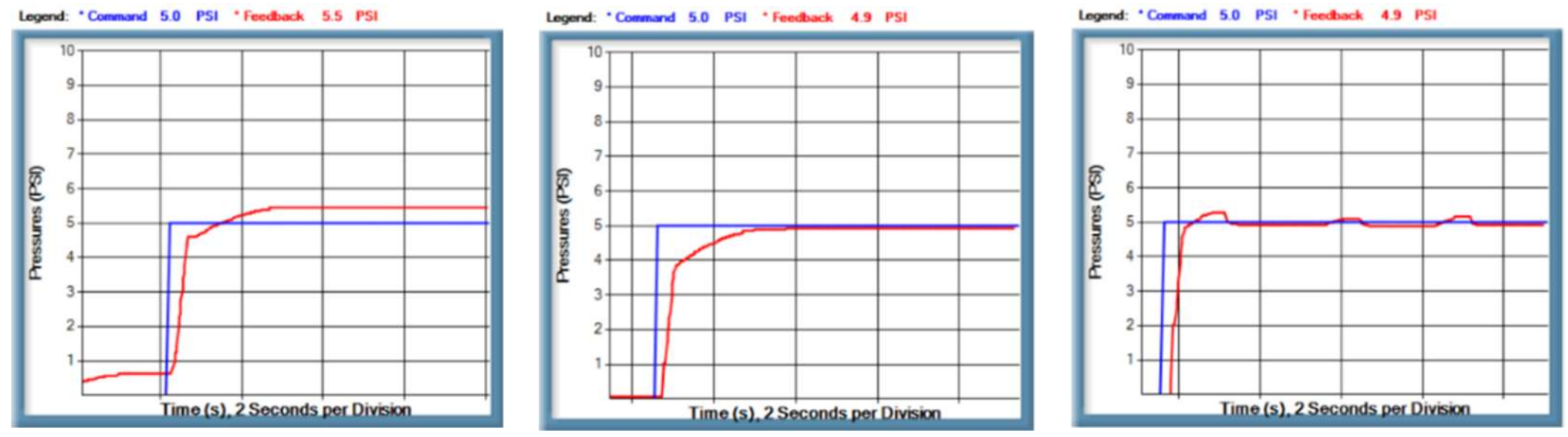

*Figure 5.12. System response to the Integral #1 gains of 0% (left), 7% (middle), and 15% (right), Proportional gain=80%, Derivative gain=5%, Ramp Up/Down rate=10%*

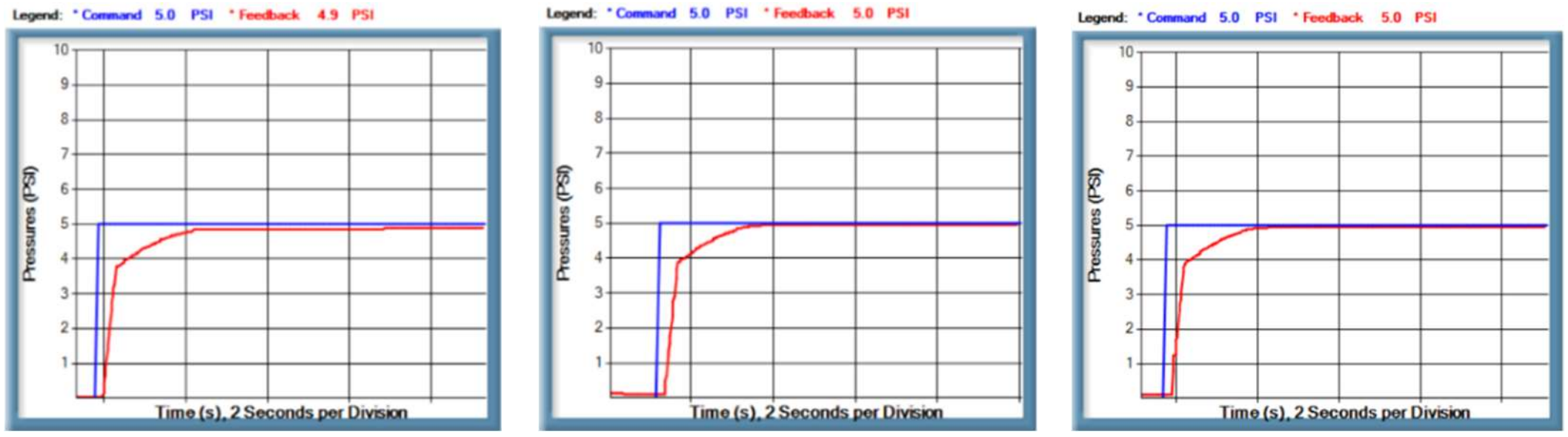

*Figure 5.13. System response to the Derivative gains of 0% (left), 3% (middle), and 5% (right), Proportional gain=80%, Integral #1 gain=7%, Ramp Up/Down rate=10%*

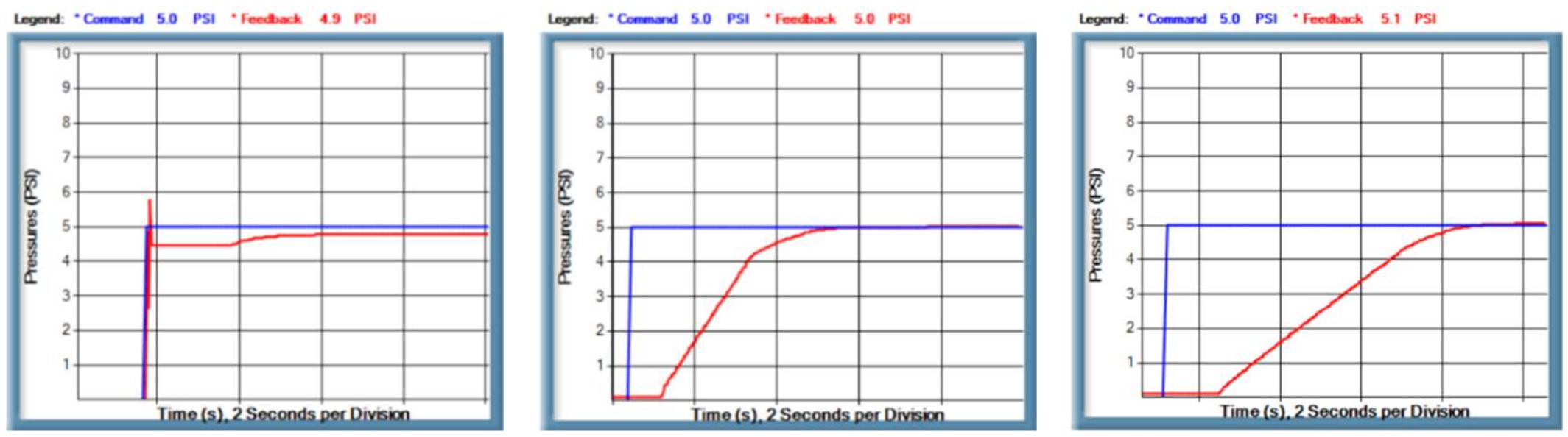

*Figure 5.14. System response to the Ramp Up rate of 0% (left), 5% (middle), and 10% (right), Proportional gain=80%, Integral #1 gain=7%, Derivative gain=5%, Ramp Down rate=10%*

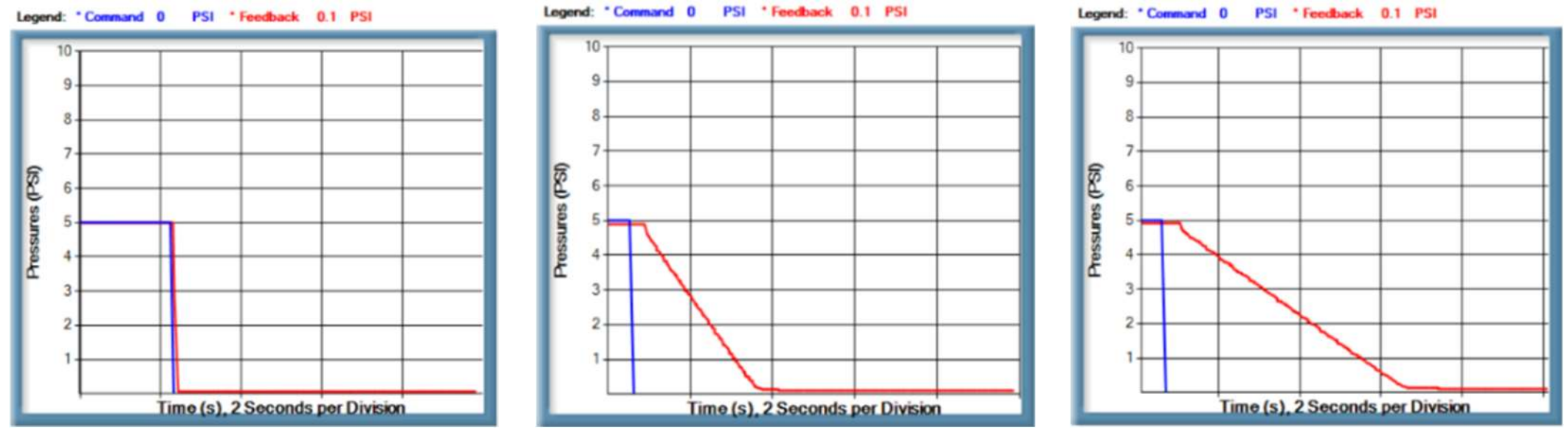

*Figure 5.15. System response to the Ramp Down rate of 0% (left), 5% (middle), and 10% (right), Proportional gain=80%, Integral #1 gain=7%, Derivative gain=5%, Ramp Up rate=10%*

The measured pressure of the Enfield TR regulator (red line in Figure 5.11 to Figure 5.15) suggests that *Proportional gain* = 80%, *Integral #1 gain* = 7%, *Derivative gain* = 5%, *Ramp Up/Down rate* = 10%, *Integral Window* = 2, *Integral #2 gain* = 5%, and *Filter* = 1500 yield the desired response. Variations in parameters such as the filter, integral window, and integral #2 had negligible impact and were not extensively explored and left at the manufacturer default settings because the integral #1 was sufficient to produce an acceptable response.

## 5.3 LabVIEW Interface

### Closed-loop Control of Rotation of a Single FREE

A simple VI was created in LabVIEW to explore the practicability of PID control of a FREE with pressure as the control variable. Further, the simulation results in 3 were compared with the actual response of a FREE to PID control of rotation. As shown in Figure 5.16, the user has access to the PID gains and rotation set-point selections. Figure 5.16 shows the front panel of the VI that includes three graphs, *PID Control, Command Pressure,* and *Feedback Pressure.* The orientation of the end cap (3D blue block) can also be seen above the commanded pressure.

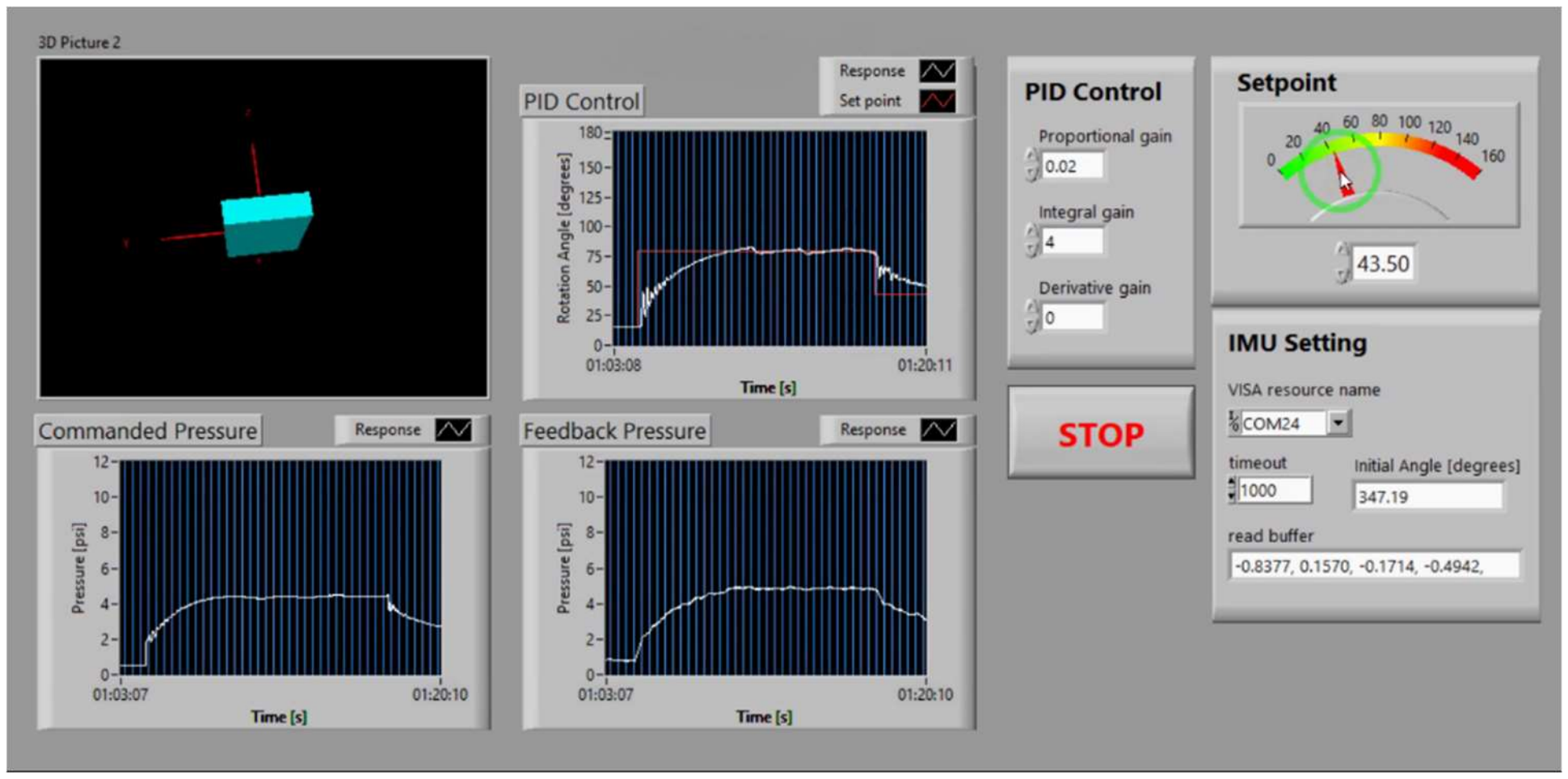

*Figure 5.16. Front panel of the LabVIEW VI for PID control*

The block diagram of this VI is similar to the module VI (see Figure 5.21) with minor changes. The *PID* VI implements a PID controller using pressure as the control variable. The controller maintains the required pressure until the operator commands zero pressure by clicking on the *STOP* button. Figure 5.17 depicts the PID control section of the VI with its variables and constants in the main block diagram.

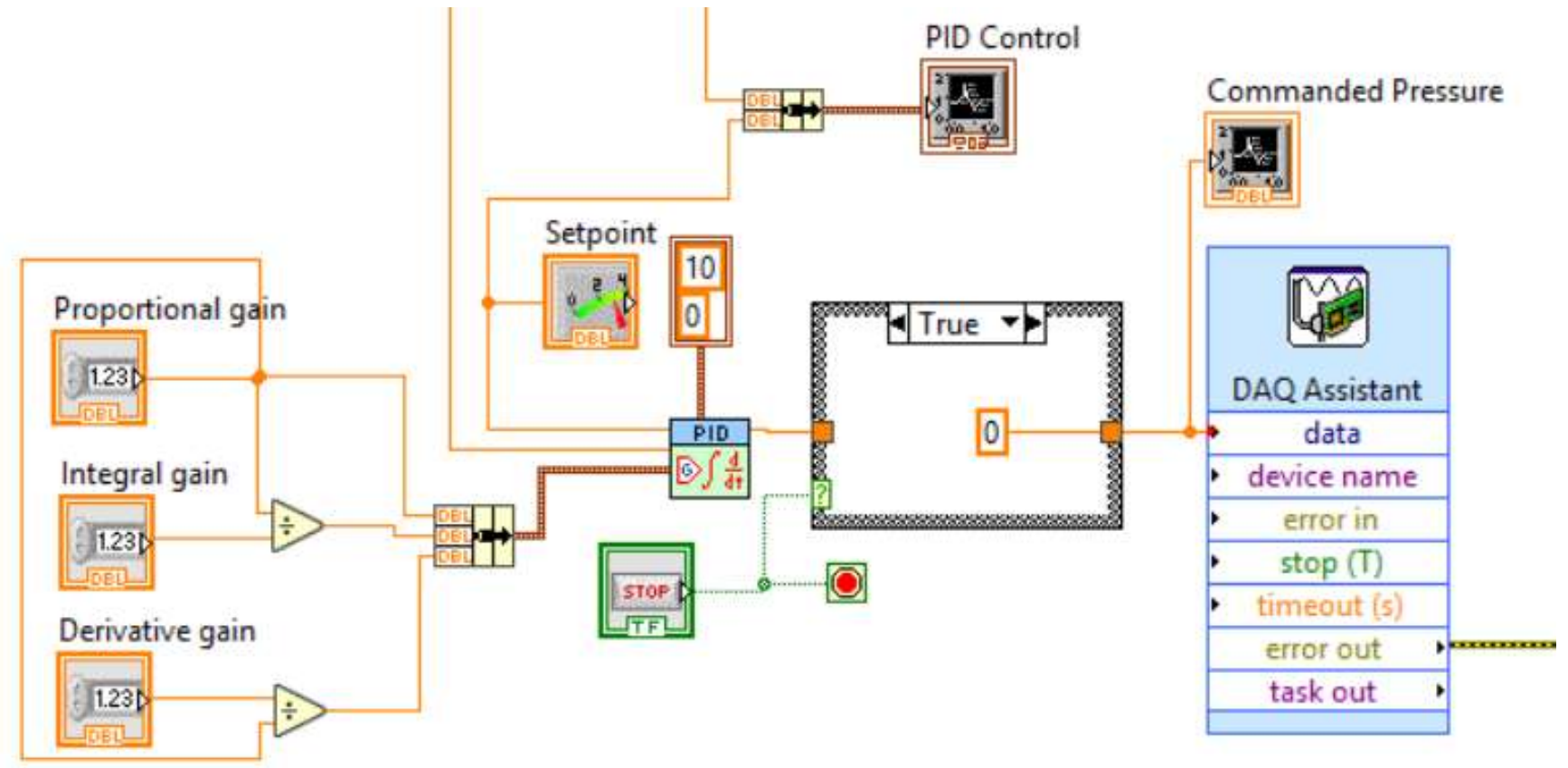

*Figure 5.17. PID control block in LabVIEW*

The pressure measured within the regulator is communicated through a *DAQ Assistant* function to the VI and filtered by a smoothing filter (see Figure 5.18) to provide a feedback signal with less noise. The smoothing filer is a *Rectangular Moving* average filter with half-width of 12. This number was found to be sufficient by observing the response of the control system.

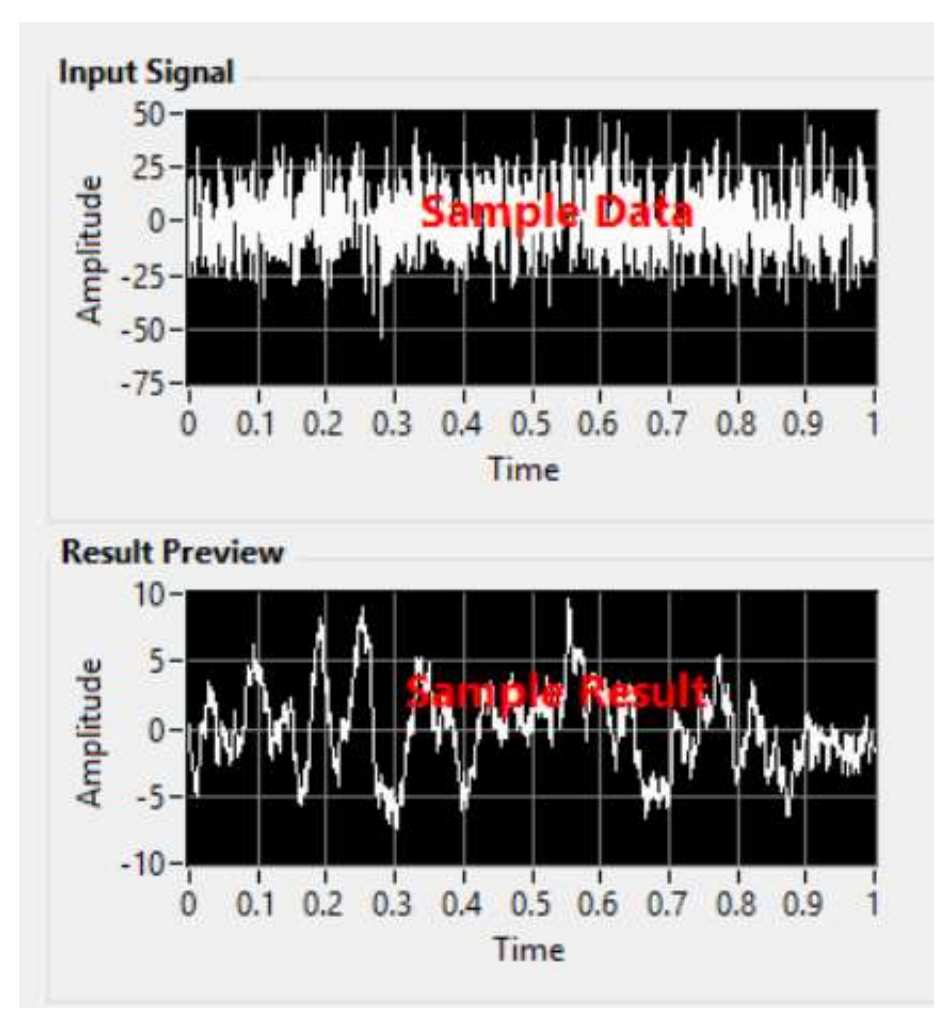

*Figure 5.18. Smoothing filtering, signal processing toolkit LabVIEW*

A quaternion representation of rotation is one of the possible outputs of the SparkFun IMU. The axis and angle of rotation calculated from the quaternion determines the orientation of the 3D object (end cap) displayed using the *3D Picture Control* toolkit in LabVIEW.

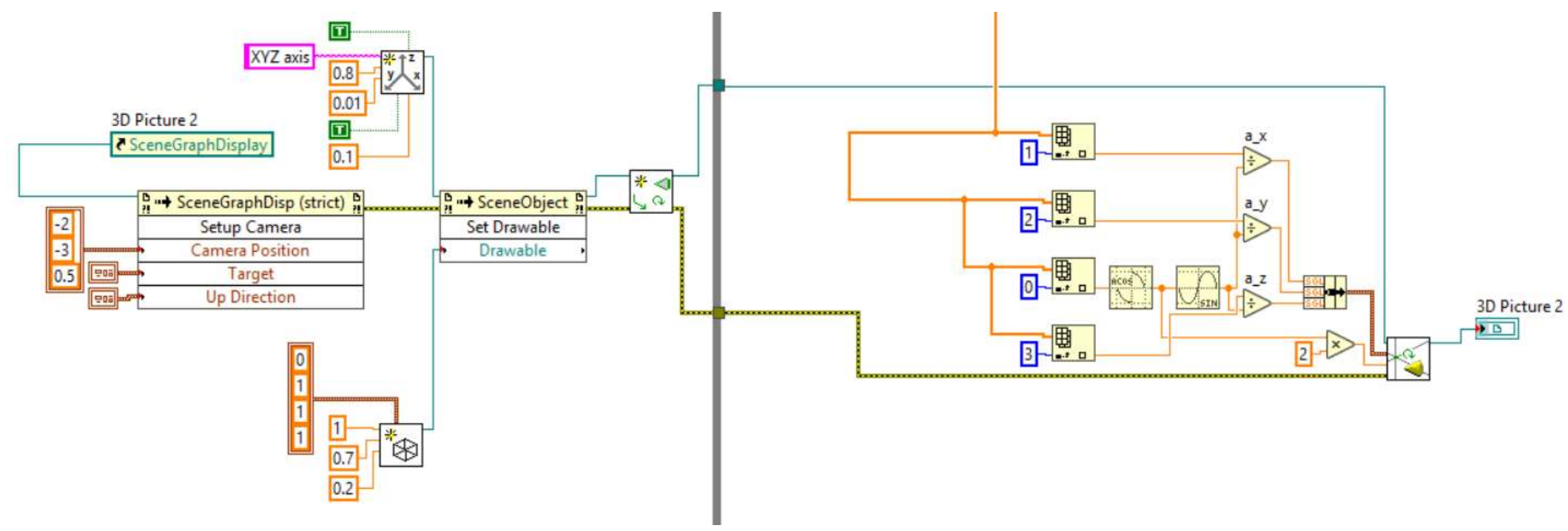

*Figure 5.19. 3D picture control blocks in LabVIEW using quaternion received from IMU*

## Open-Loop Control of a Module

The Enfield TR Configuration Interface software provides versatile control features to perform pneumatic calibration and actuation. However, controlling a group of regulators simultaneously is not possible since the software connects to only one regulator at a time. Hence, a LabVIEW interface was created to adjust the pressures in a module of FREEs. LabVIEW enables the operator to incorporate a variety of devices and sensors and monitor live data on the displayed front panel. For this system, the two USB DAQ 6001 are used to communicate signals (command/output) between the computer (PC) and the sensor and actuators. Figure 5.20 shows the front panel of the LabVIEW interface developed to actuate (open-loop) a module of FREEs.

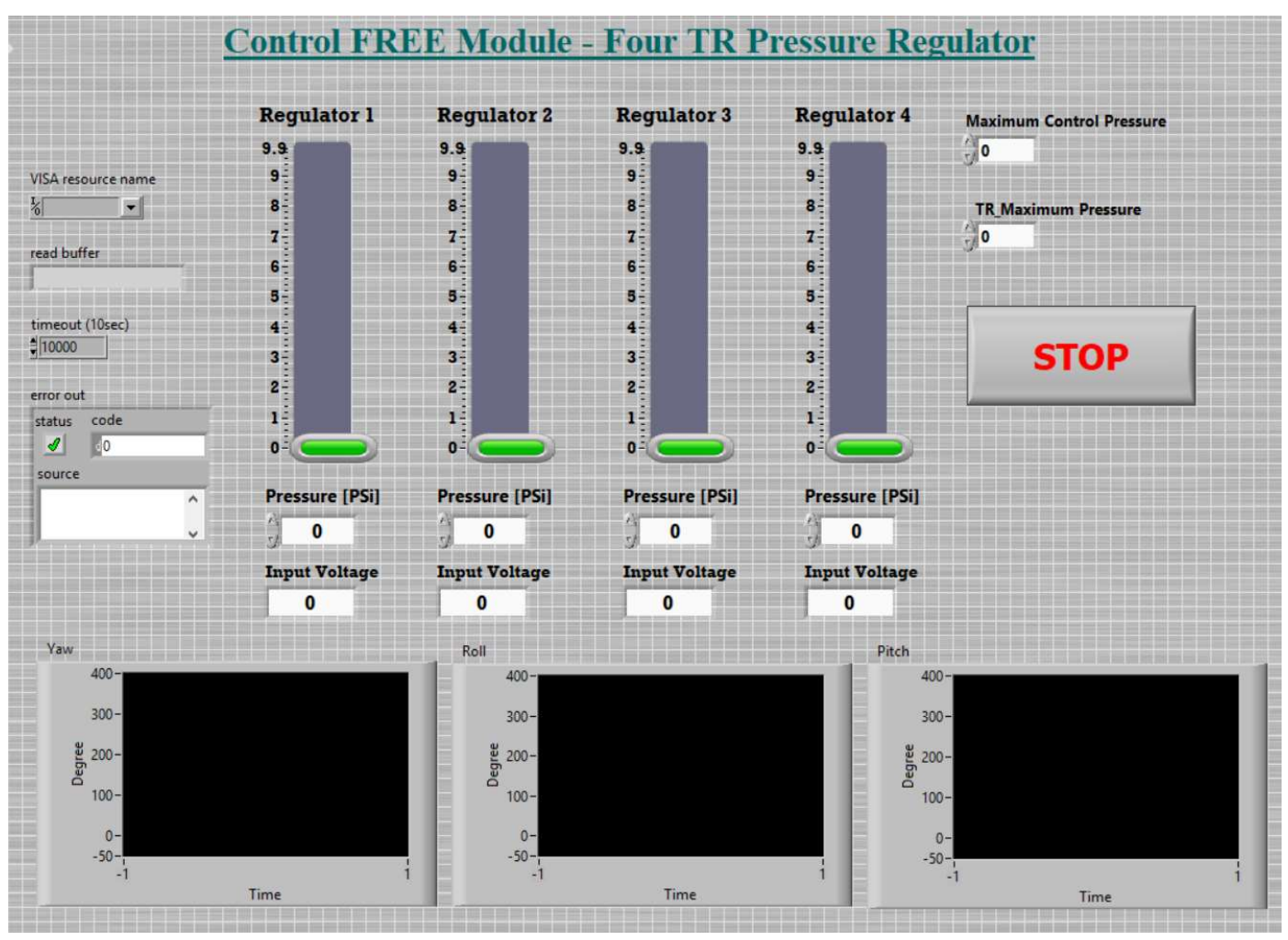

*Figure 5.20. LabVIEW front panel for module open-loop control*

Within this interface, each FREE (connected to a pressure regulator) can be pressurized individually up to the specified maximum pressures. The *"TR Maximum Pressure"* is the value set within the pressure regulator software and *"Maximum Control Pressure"* is the value set in LabVIEW for the range of sliders on the front panel. Output displayed for the system includes, graphs of *Roll*, *Pitch*, and *Yaw*, which give the orientation of the end cap in space obtained from the SparkFun IMU. Figure 5.21 shows a block diagram of the LabVIEW code which can be modified for any particular application through the National Instruments toolkits. Two *DAQ Assistant* functions were added to the VI (virtual instrument) to set up data logging and triggering. This VI has been configured to communicate with the USB DAQ 6001 to take user inputs (voltages) and receive the feedback signals from the regulators. Additionally, the VISA Configure Serial Port VI initializes the serial port connection to the SparkFun IMU. The String to Array function puts the measured angles in an array for plotting with respect to time.

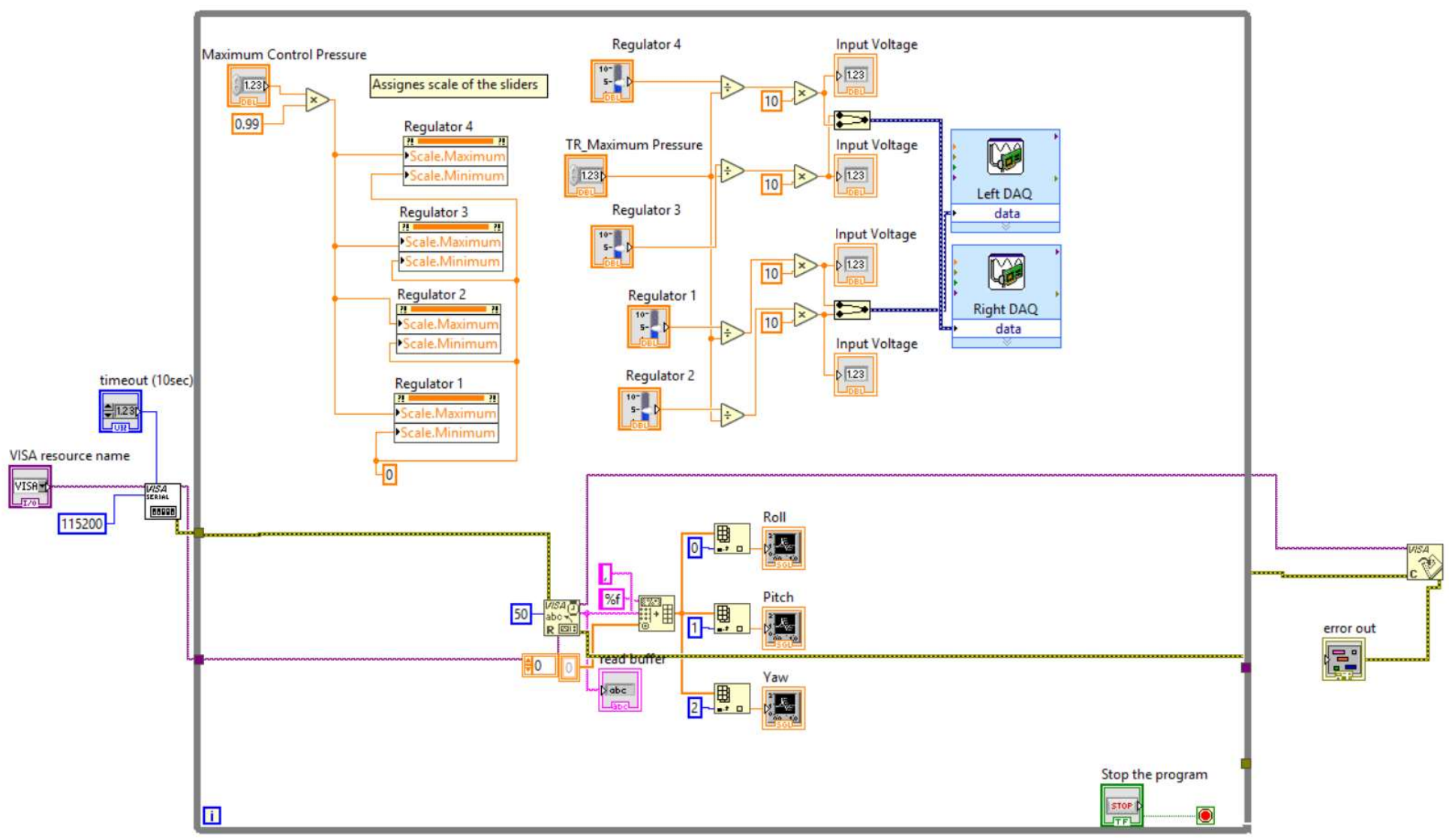

*Figure 5.21. LabVIEW block diagram for module open-loop control*

## 5.4 Summary

The apparatus used to run experiments on FREEs enabled the validation of the modeling approaches and simulation results presented in the previous chapters. The TR pressure regulator provided smooth and accurate air pressure control; however, only 10% of its output capacity (145 psi) is used to control the FREEs. Therefore, the use of pressure regulators with a lower output range and better pressure resolution is recommended for future experiments. Multiple Inertial Measurement Units (IMUs) were used to track the rotation angle of FREEs but more prone to drift during the experiments. Among the models, the VMU931 operated with minimum drift over a short period of time. Overall, image tracking was found to be most useful for gathering data and should be used for future research. Finally, the LabVIEW interface was recognized as a suitable design-and-development platform due to its powerful visual programming language and graphical presentation.

# 6
# Further Model Evaluation

This chapter provide additional data collected from actual FREEs with winding angles of 40°, 50°, 60°, and 70° and compare them to the results calculated with the lumped parameter model of Chapter 2 and the finite element model of Chapter 4, thus further demonstrating the capability of the proposed simulation models to predict the behavior of FREEs.

## 6.1 Experimental Measurements

In this section, the experimental results obtained with single FREEs, along with a description of tested samples, are presented. The data of 40° FREEs are separately analyzed to relate variations in experimental results and geometry of samples. Experimental methods for obtaining stiffnesses and damping constants of the 40° FREE are provided. The gathered experimental data is used in the next section to validate the model predictions.

**Rotation and Elongation**

Fourteen FREE samples (five 40°; three of each 50°, 60°, and 70° winding angles) were created from cotton threads (Red Heart Yarn, 2019) and latex tubing with 9.52 mm inner diameter, and 0.8 thickness (Kent Elastomer, 2019). After carefully applying a thin layer of latex coating, FREEs with wall thicknesses ranging from 1.4 mm to 2.2 mm resulted. Table 6.1 shows the thickness of each sample and the number of fibers used for each winding angle. All samples were cut to 11 cm lengths to minimize variations in the experiment. Each FREE was fit into the experimental setup described in the previous Chapter and displacements were measured at pressures in the range of 0-10 psi. Note that the rotation angle and the extension were each measured two times by ramping up the pressure and then back down. Figure 6.1 to Figure 6.4 show plots of averaged rotation and extension of FREE samples under the same conditions.

| Winding Angle | # of Fibers | Wall Thickness (mm) | | | | | | |
|---|---|---|---|---|---|---|---|---|
| | | Sample 1 | Sample 2 | Sample 3 | Sample 4 | Sample 5 | Average | Range |
| 40° | 6 | 1.55 | 1.82 | 1.86 | 2 | 2 | 1.85 | 0.45 |
| 50° | 6 | 1.8 | 1.4 | 1.4 | - | - | 1.56 | 0.4 |
| 60° | 5 | 2 | 2.2 | 1.95 | - | - | 2.05 | 0.25 |
| 70° | 5 | 1.88 | 1.76 | 1.86 | - | - | 1.83 | 0.12 |

*Table 6.1. Description of FREE samples used for experiments*

The experimental results suggest that the difference between rotation angles increases at higher pressures. Elongation direction changes between the 40° and 50° winding angles, close to the theoretical critical winding angle of 54.7° (see Section 4.8).

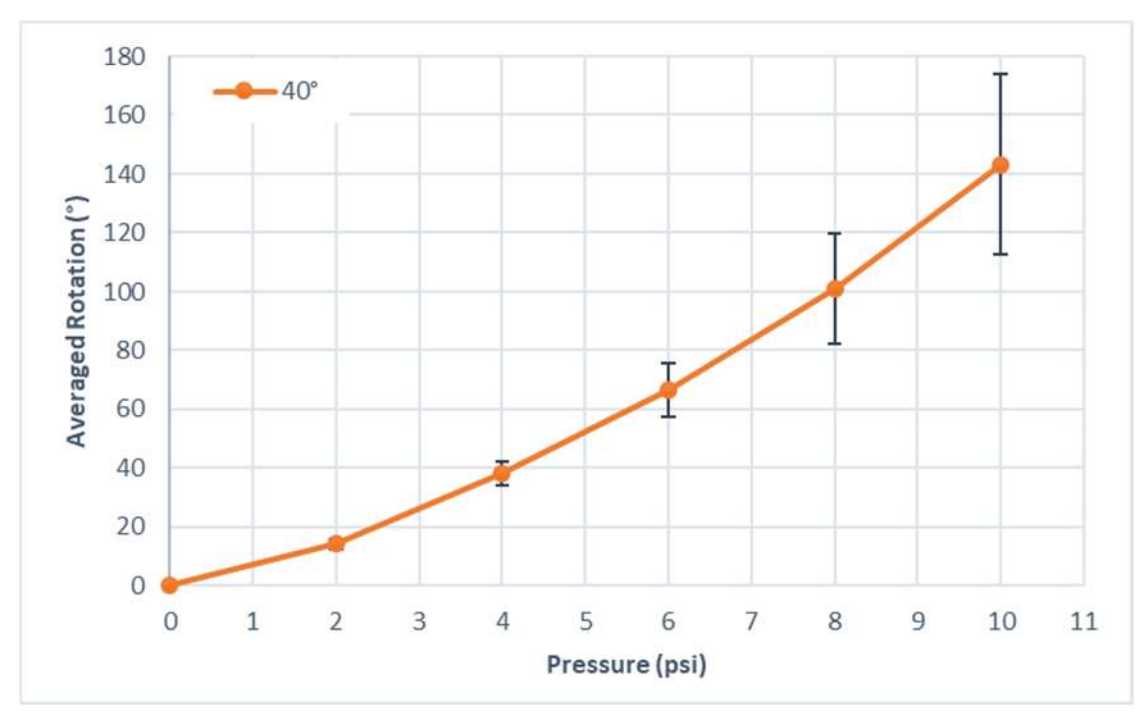

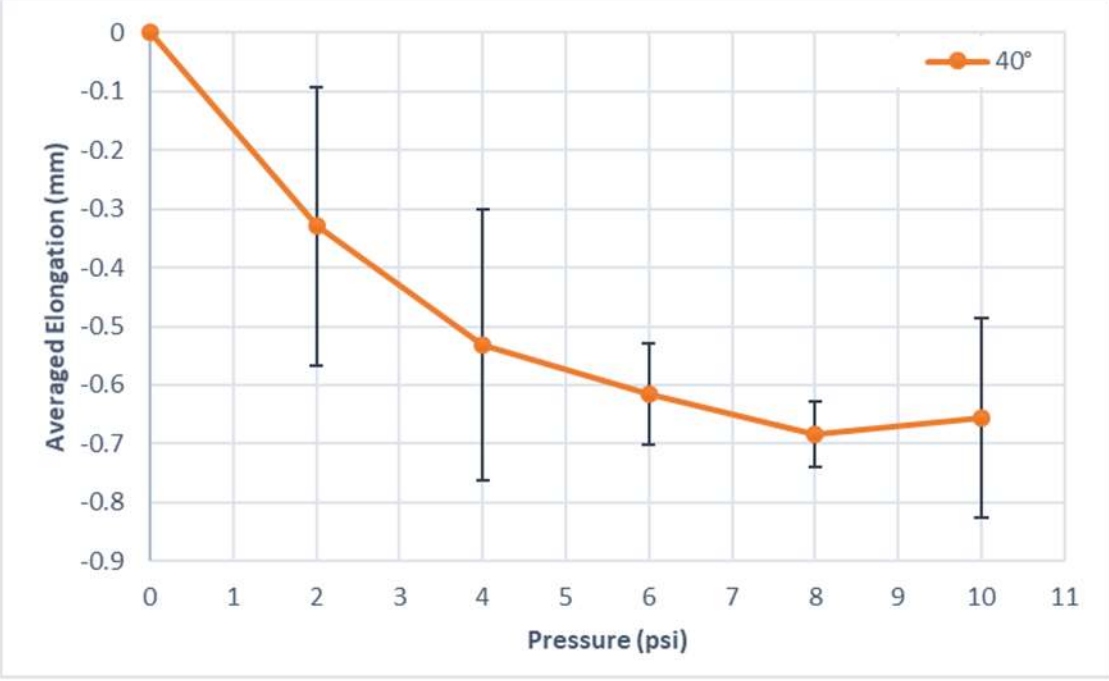

*Figure 6.1. Experimental results of FREEs for winding angle of 40° (bars represent standard deviation)*

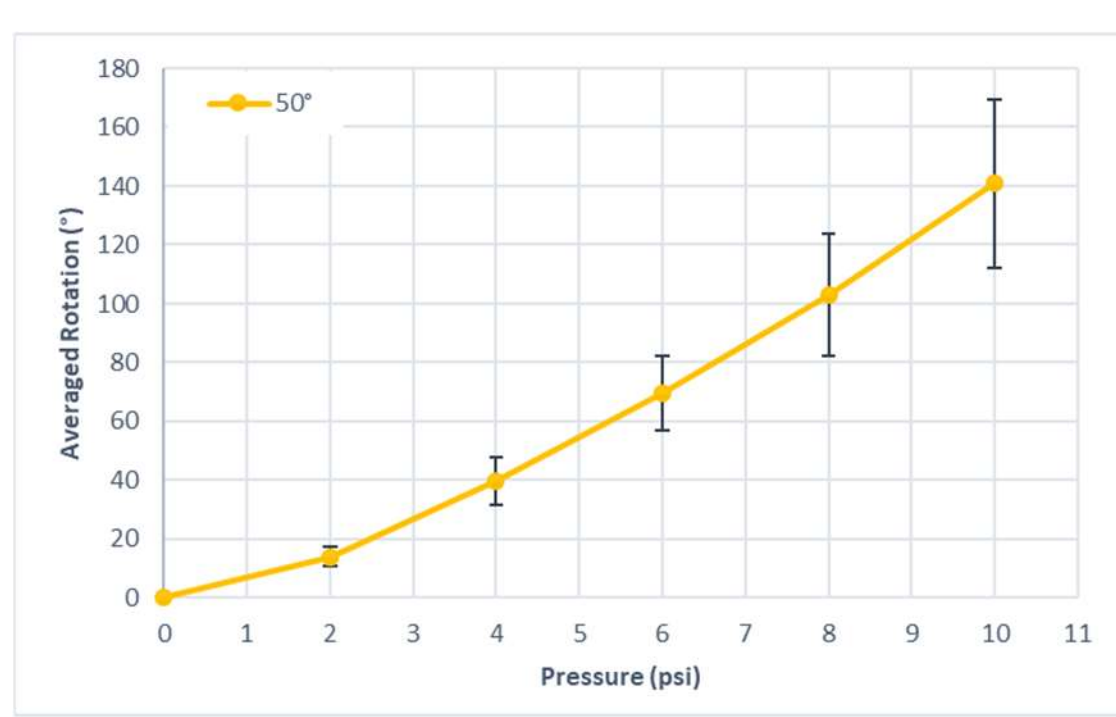

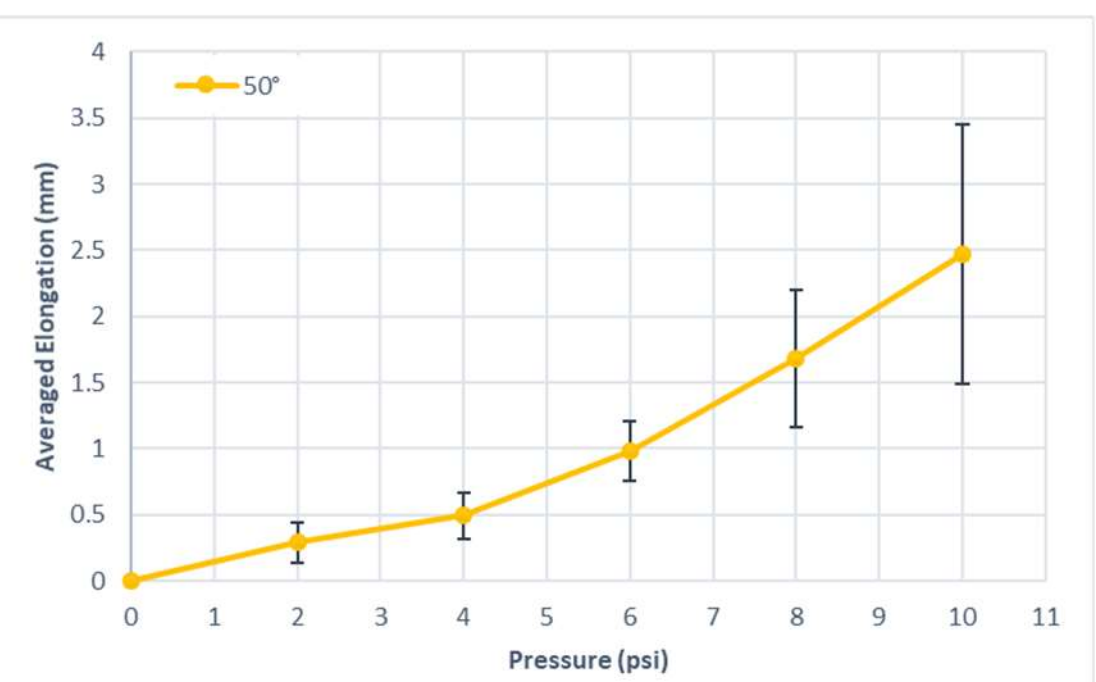

*Figure 6.2. Experimental results of FREEs for winding angle of 50° (bars represent standard deviation)*

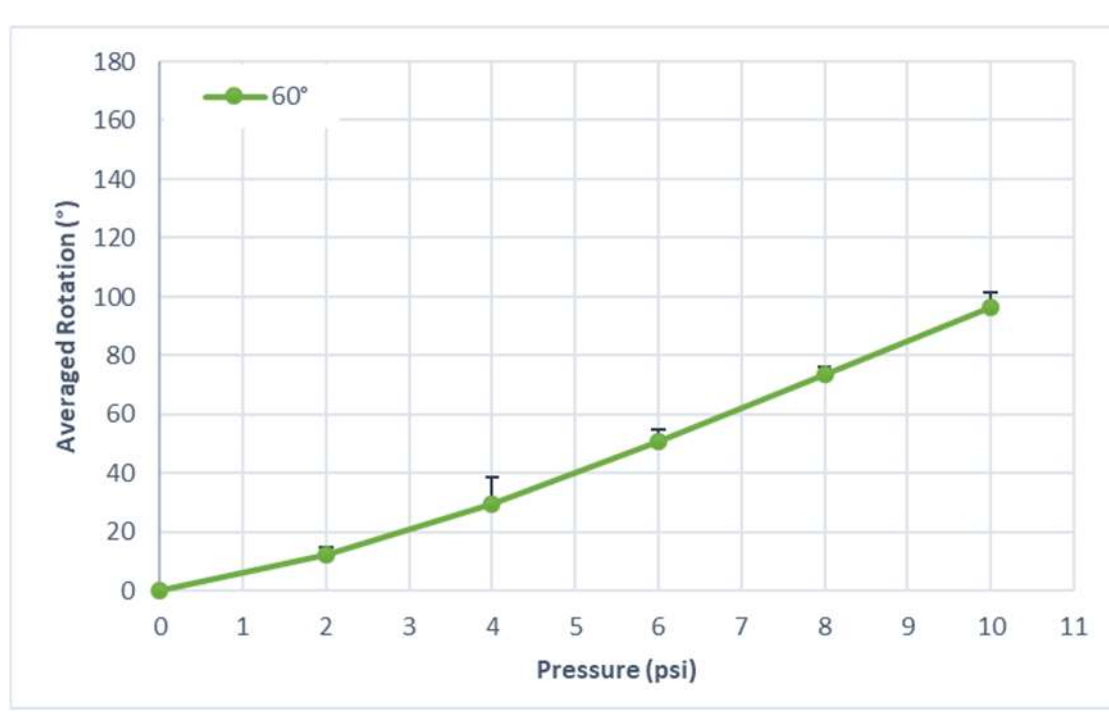

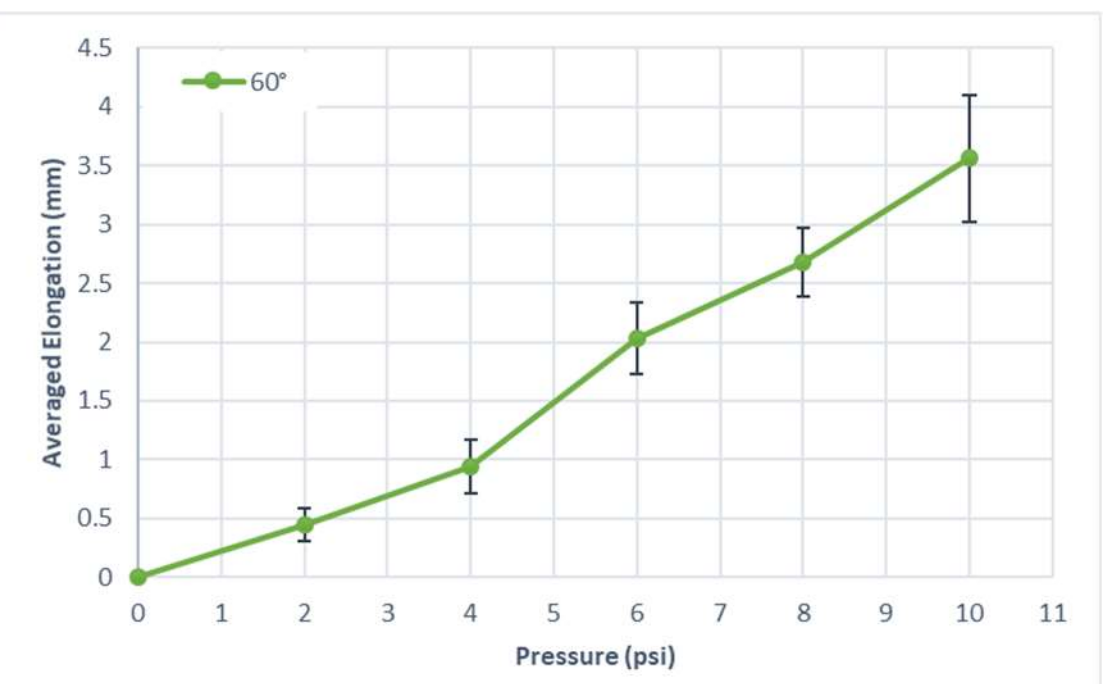

*Figure 6.3. Experimental results of FREEs for winding angle of 60° (bars represent standard deviation)*

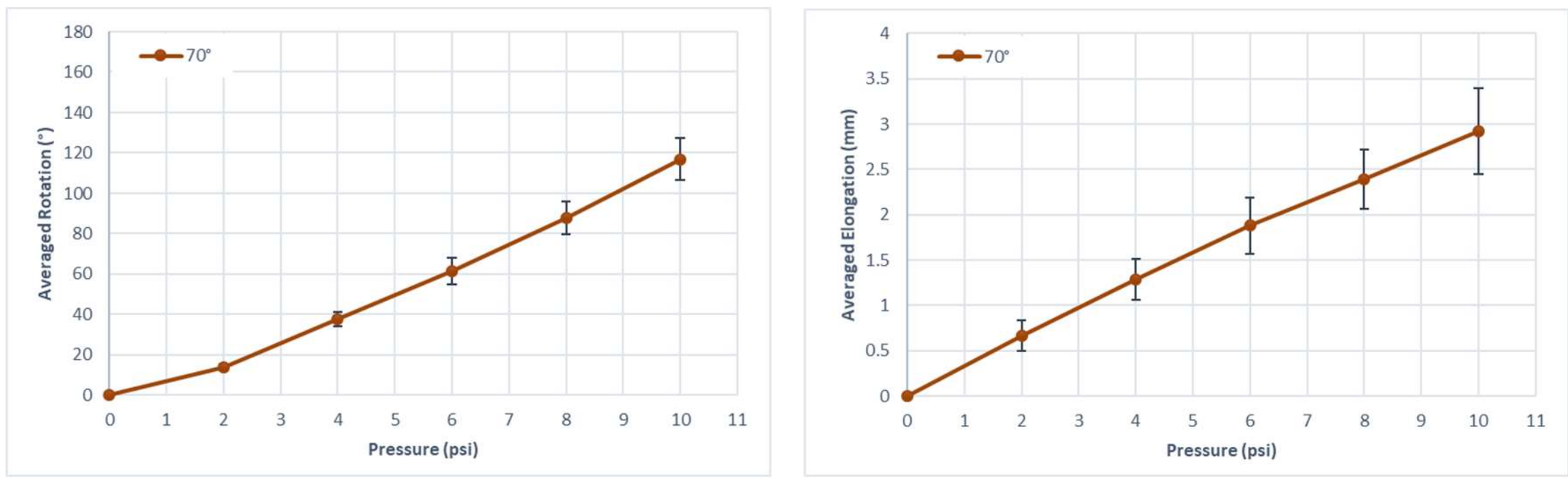

*Figure 6.4. Experimental results of FREEs for winding angle of 70° (bars represent standard deviation)*

The plots of Figure 6.1 to Figure 6.4 do not follow the same trend as observed in finite element analysis in Chapter 4. For example, according to the FEA results, the 60° FREE produces the largest rotation at 10 psi, relative to the other winding angles, however the experimental data in Figure 6.1 shows the smallest rotation for this winding angle. The similar comparisons indicate that there is a mismatch between the FEA model results and the experimental data.

Since there are variations, up to about 0.8 mm, in the wall thickness between all samples, the experimental measurements were scrutinized in more detail. The deformations of five samples of 40° FREE are individually plotted (Figure 6.5) as a function of pressure to clarify the impact of wall thickness.

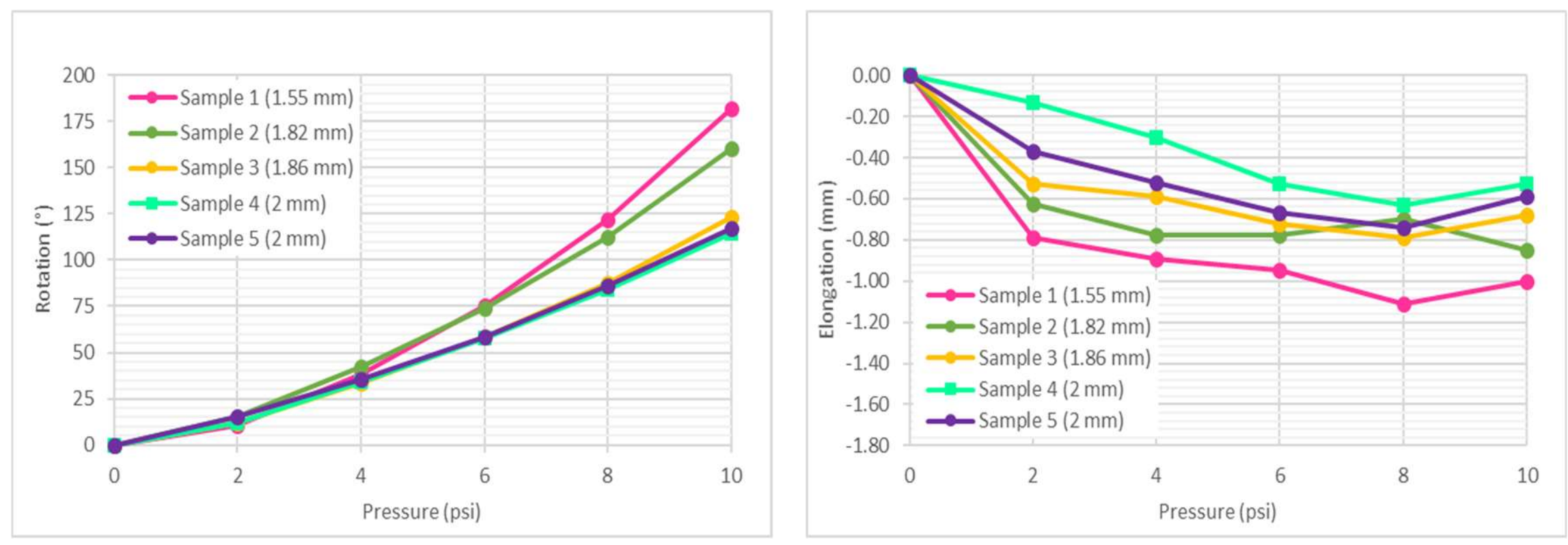

*Figure 6.5. Experimental results of 40° FREEs samples*

Observations from the experiments suggest that the 0.45 mm difference in the wall thickness of sample 1 and sample 5, causes the rotation and elongation to be different by about 67° and 0.4 mm (at 10 psi pressure), respectively. Variations in wall thickness thus largely alter the deformations, which reflect the significant importance of the manufacturing process. If fabrication was consistent for all FREEs tested, then the trends of plots (as a function of winding angle) in Figure 6.1 would be different.

## FREE Stiffnesses

To determine the axial and rotational behavior of a FREE using the lumped-parameter model, the elastomer's extensional and torsional stiffnesses and damping constants are required. The stiffness can be determined when the unpressurized FREE is in equilibrium. From the data recorded experimentally for the extension $s$ and rotation $\varphi$ under various external loadings (axial and torsional), assuming a linear spring relationship for the elastomer, the best estimate of these stiffnesses is obtained by performing a linear regression between $F_e$ ($M_e$) and $s$ ($\varphi$). These two simple tests were performed for all these FREE samples in Table 6.1 to statically measure the extension and twist as a function of applied force and moment, respectively. Figure 6.6 shows the apparatus used to measure the static displacements of the 40° FREE samples as a function of applied load. The elongation was measured each time weight was added using a vision system to track an LED attached to the end of the FREE. Similar steps were taken to measure the rotation of the FREE by using an IMU.

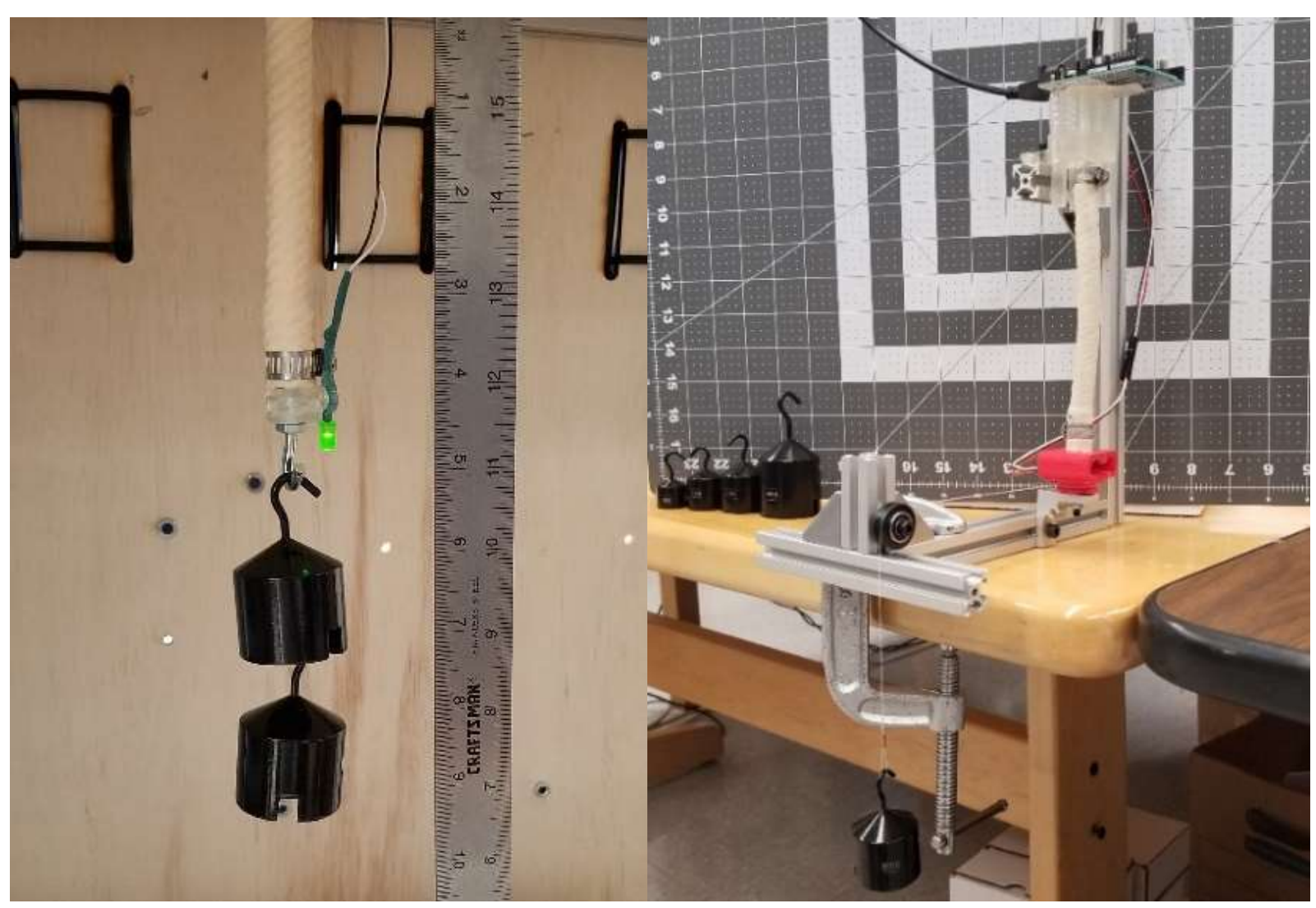

*Figure 6.6. Axial and torsional loading of a FREE*

To obtain the stiffness of a 40° FREE, for example, displacements of three samples (1, 2, and 3) were averaged (plotted in Figure 6.7) and a linear regression was used to produce a value of the extensional stiffness $k_e = 600\ \frac{N}{m}$ and a value of the torsional stiffness $k_t = 0.018\ \frac{Nm}{rad}$.

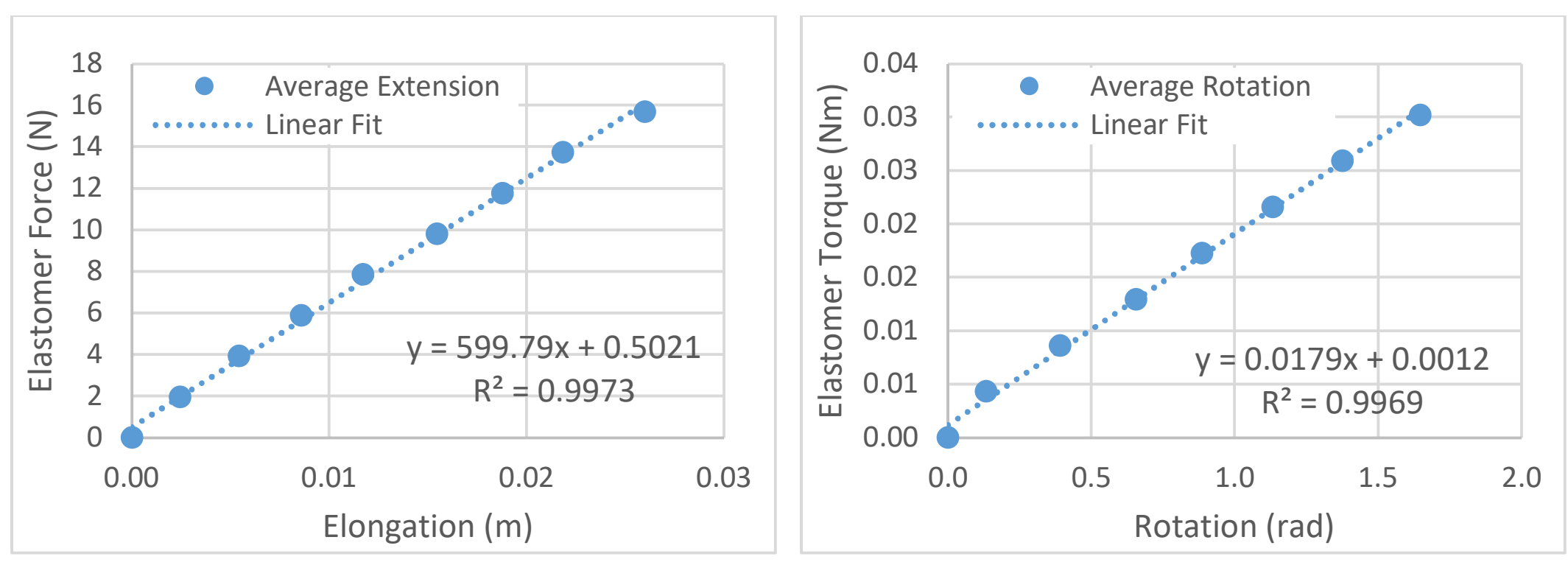

*Figure 6.7. Elastomer force vs. elongation, elastomer torque vs. rotation of 40° FREE*

These constants then used in Eqs. (2.17) and (2.18) to determine deformations of the 40° FREE at various pressures based on the lumped-parameter model. The calculated rotation and elongation were significantly different from the experimental measurements of the previous section. It was concluded that the stiffnesses of unpressurized FREEs do not accurately represent the real physical system since the internal pressure directly affects those values. As an alternative, the averaged experimental measurements of rotation and elongation of five 40° FREE samples were used in Eqs. (2.17) and (2.18) to calculate the extensional stiffness $k_e$ and torsional stiffness $k_t$ at corresponding pressures. The averaged experimental data include five data sets of rotation and elongation at pressures of 2, 4, 6, 8, and 10 psi. Figure 6.8 presents a bar graph of calculated stiffness values for each dataset using the lumped-parameter model. Since there are variations in stiffnesses at each pressure, the averaged values of $k_e$ and $k_t$ were selected as representative stiffnesses of the 40° FREE. The same plot for winding angles of 50°, 60°, and 70° can be found in Appendix C.

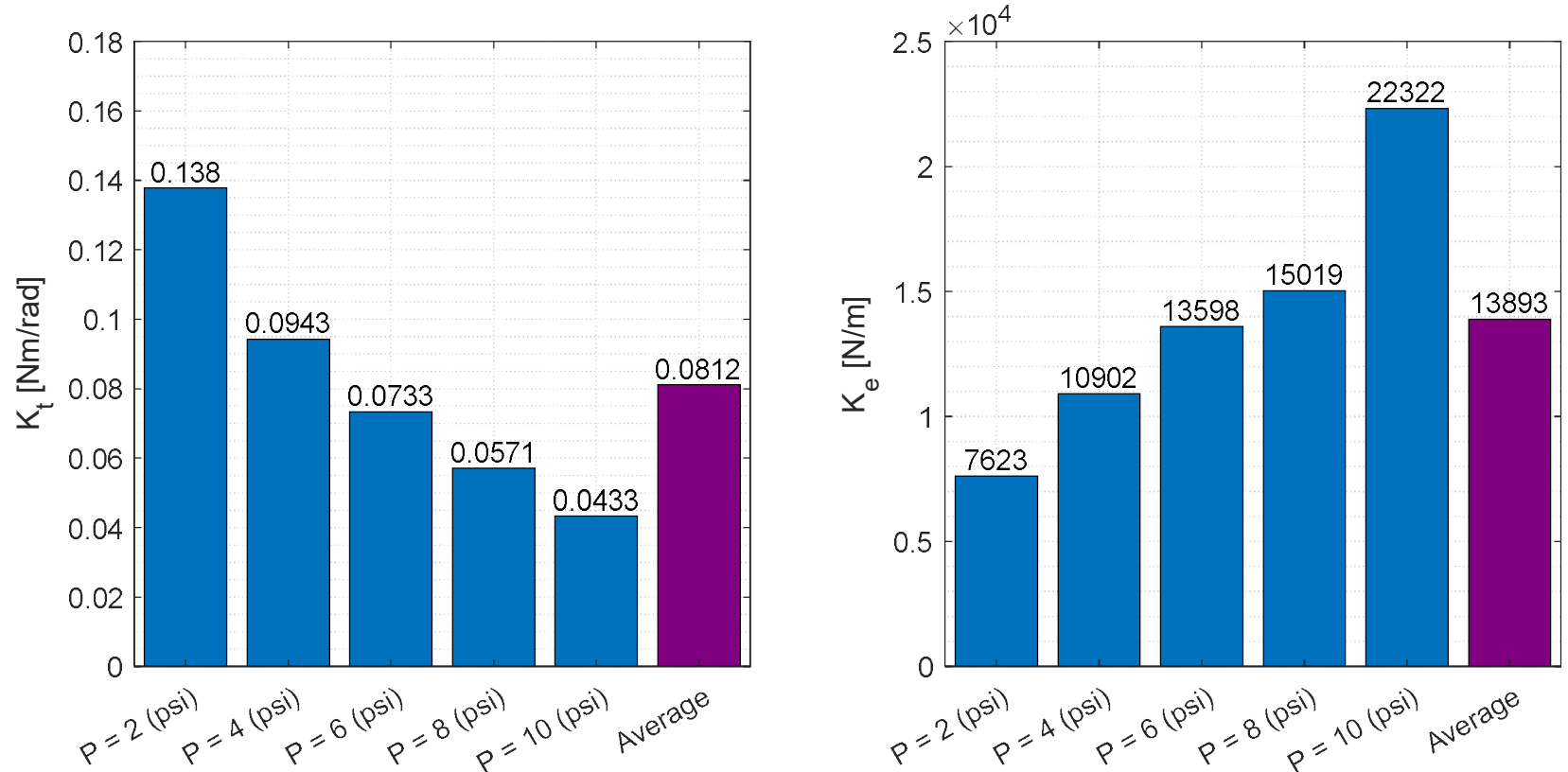

*Figure 6.8. Extensional stiffness $k_e$ and torsional stiffness $k_t$ of the 40° FREE at various pressures*

### FREE Damping Constants

Using the lumped-parameter model also requires the damping constants of the FREE to be determined. This subsection analyzes the experimental data of rotational and axial vibration (Figure 6.9) of a 40° FREE to obtain those constants. Bailis (2019) outlines the approach as follows: 1) Calculate logarithmic decrements and damping ratios using the vibration data, 2) Compute damped and undamped natural frequencies, 3) Insert the damping ratios and undamped natural frequencies into the standard form of a vibratory system (Hutton, 1981), and 4) Express Eqs. (2.17) and (2.18) in the form of the equations of step 3.

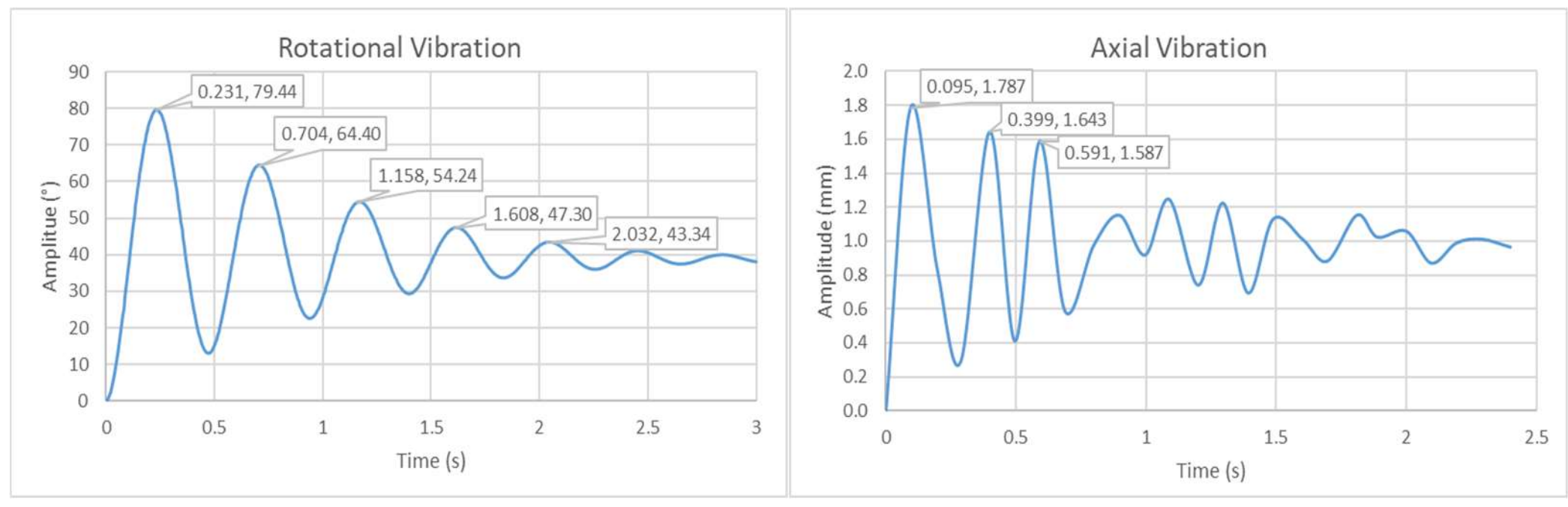

*Figure 6.9. Free axial and rotational vibration of a 40° FREE*

By using the above steps, the damping constants of the 40° FREE are:

$$c_e = 0.34 \frac{N - s}{m} \tag{6.1}$$

$$c_t = 0.0000397 \frac{Nm - s}{rad} \tag{6.2}$$

## 6.2 Model Validation

### Static Displacements

This section compares static results obtained with the lumped-parameter model and the finite element model with the collected experimental data of Section 6.1 and discusses the capabilities of both models. To accurately simulate the actual FREE samples, the stiffness and damping constants obtained in the previous section were used in the lumped-parameter model [i.e., in Eqs. (2.17) and (2.18)] and finite element models of 40°, 50°, 60°, and 70°, all with the same geometry. Note that the Ogden model parameters of Section 4.5 were used to produce the FEA results. Given these conditions, displacement of these samples in the three cases of the FEA model, lumped-parameter model, and experiment were plotted for each winding angle in Figure 6.10 to Figure 6.13.

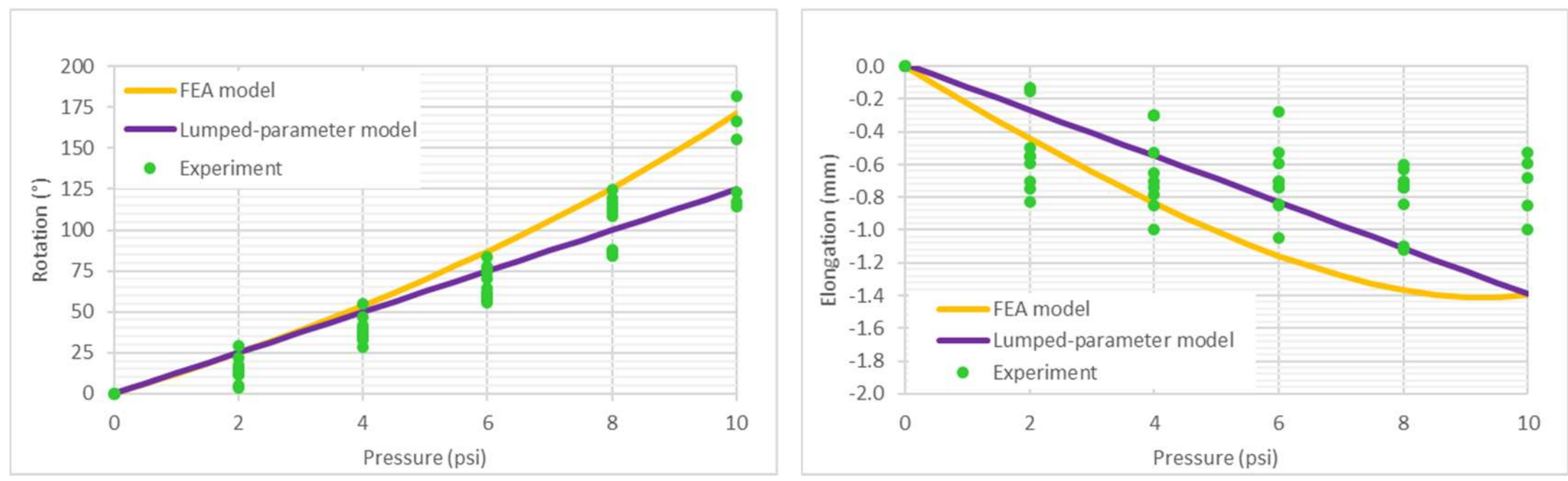

*Figure 6.10. Model prediction for rotation and elongation of the 40° FREE as well as corresponding experimental data*

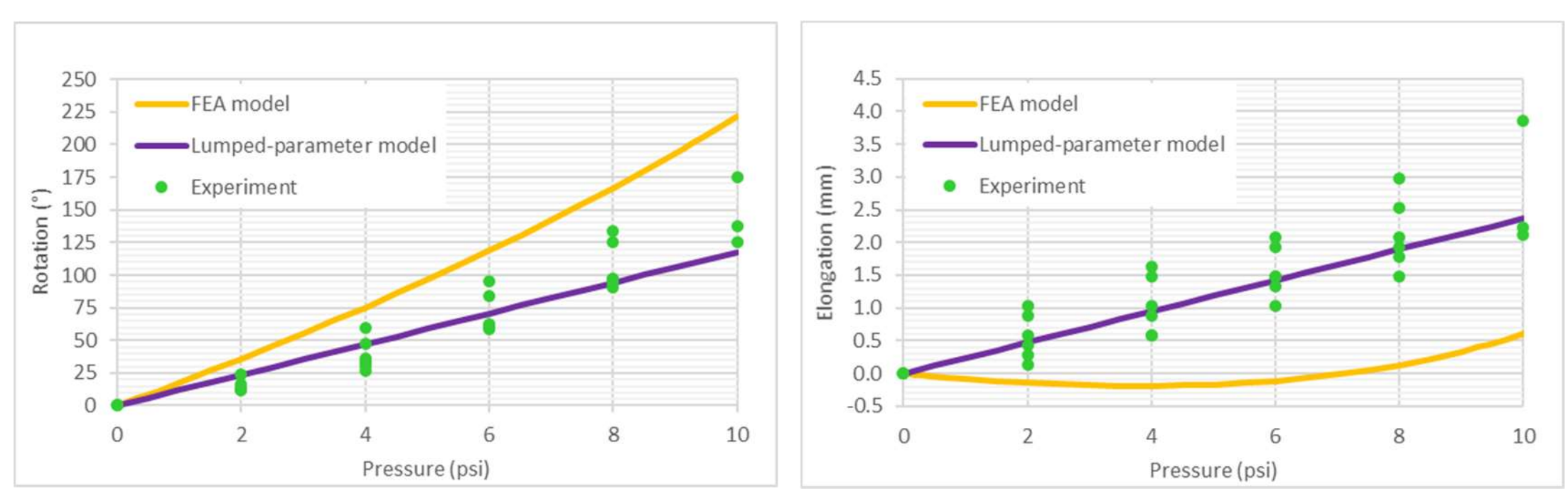

*Figure 6.11. Model prediction for rotation and elongation of the 50° FREE as well as corresponding experimental data*

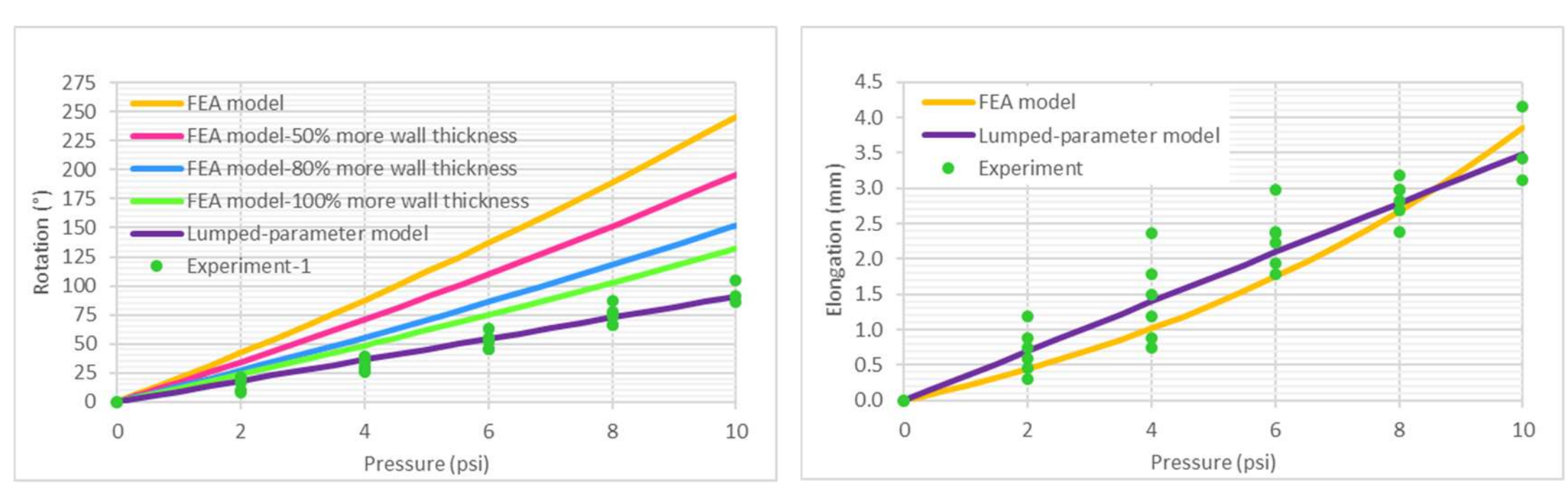

*Figure 6.12. Model prediction for rotation and elongation of the 60° FREE as well as corresponding experimental data*

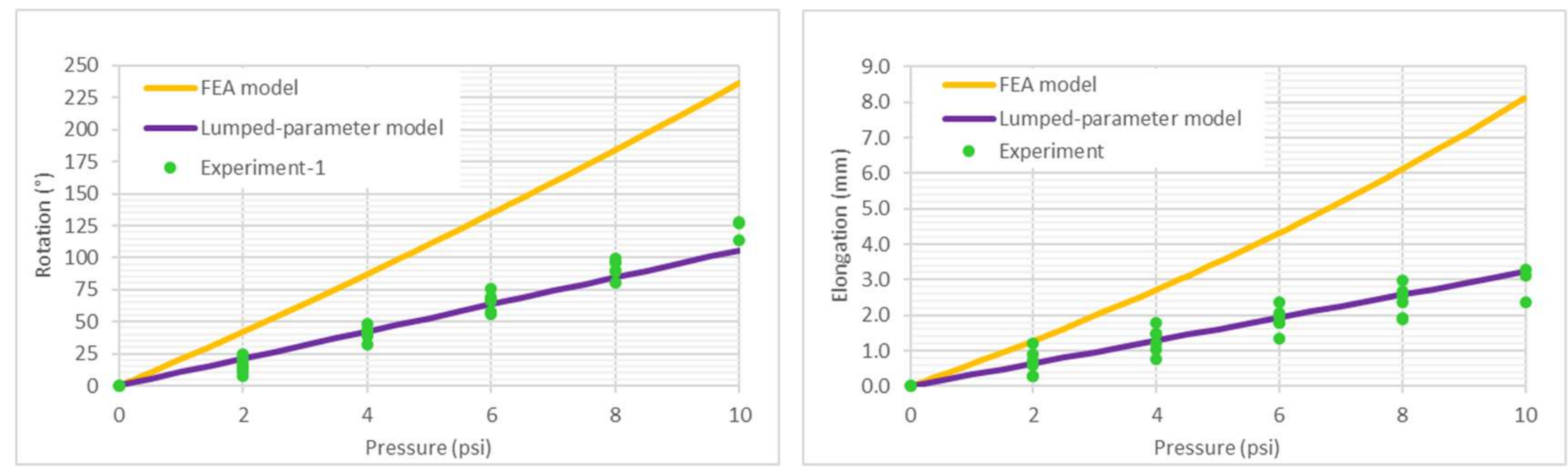

*Figure 6.13. Model prediction for rotation and elongation of the 70° FREE as well as corresponding experimental data*

All cases show that the lumped-parameter model closely follows the experimental data since stiffness and damping constants are derived from the same data. Additionally, the finite element model and experimental results are nonlinear, while the lumped-parameter model results are linear. To compare experimental and computational results, Eq. (6.3) is used to calculate model error as the normalized root-mean-square deviation (RMSD):

$$error = \sqrt{\frac{1}{n}\sum_{i=1}^{n}\left(\frac{X_{exp}^{i} - X_{sim}^{i}}{X_{max}}\right)^{2}} \tag{6.3}$$

where $X_{exp}$ is experimentally measured values, and $X_{sim}$ is values calculated using each model to predict rotation/elongation. $X_{max}$ is the maximum experimentally measured value of rotation/elongation in the dataset of each winding angle. Figure 6.14 shows the RMSD error for both models as compared to the experimental data of Section 6.1. Errors for the lumped-parameter model are below 10% for all cases except in predicting the elongation of 40° FREEs, which is likely due to a lack of adequate ability to experimentally measure small elongations of FREEs. Overall, the finite element model lead to greater error in predicting the experimental data, and the reasons are highlighted in the discussion of results below.

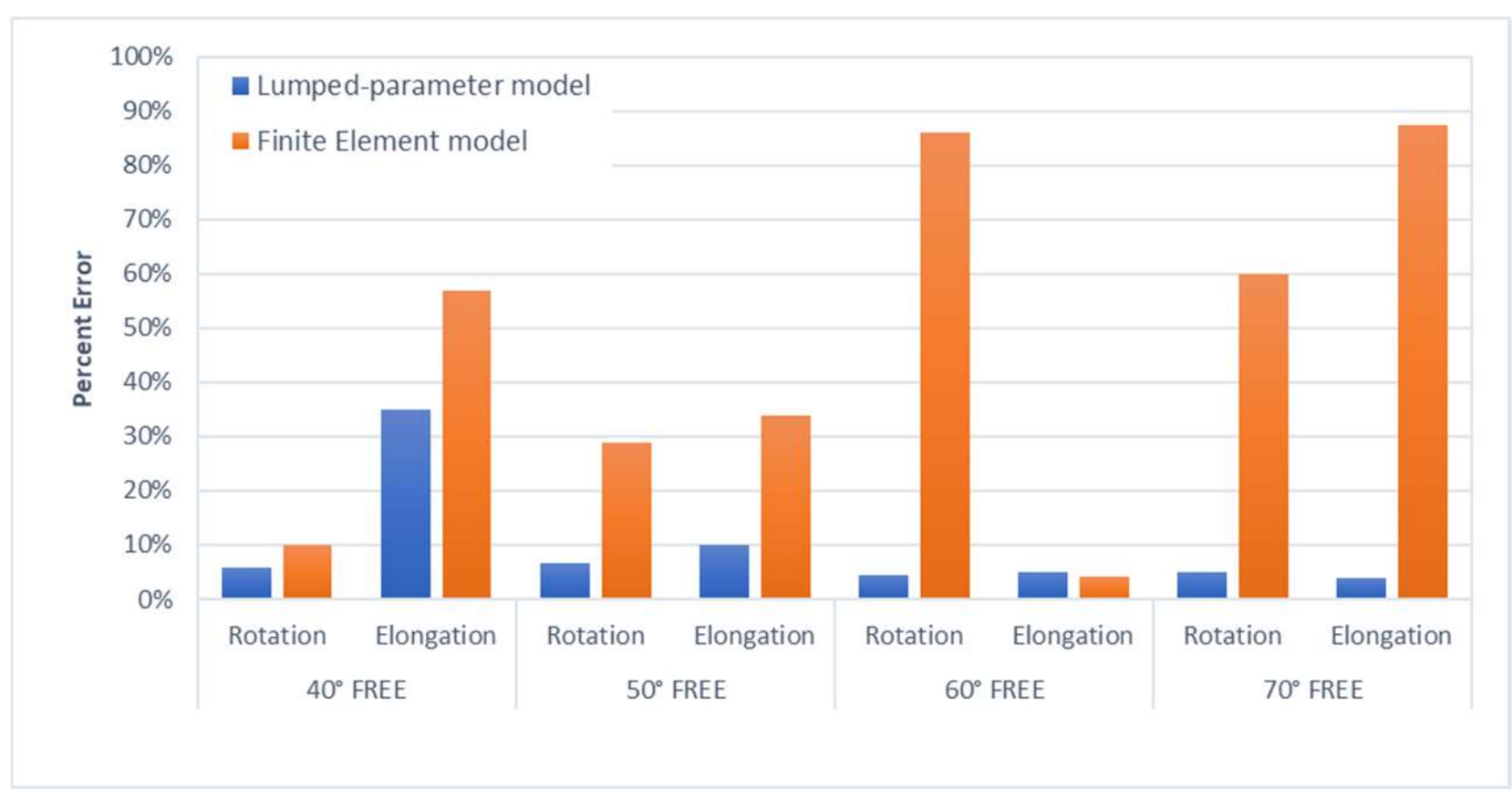

*Figure 6.14. RMSD error for the lumped-parameter and finite element models as compared to static experimental measurements of rotation and elongation for 40°, 50°, 60°, and 70° FREEs*

Note that Figure 6.10 through Figure 6.13 show that the FEA results diverge from the experimental data as the winding angle increases, especially at higher pressures. Remembering the method used to find the Ogden model parameters in Section 4.5 (identifying the best combination of α and μ to produce FEA result close to the data collected from a simple tube expansion experiment) suggests that these parameters need to be adjusted for the new samples based on the average thickness of the FREE coating. In other words, the thickness of coating plays a major role in determining the deformations.

Examining the trends in finite element and experimental data is helpful to gain further insights. Figures 6.1 through 6.4 showed that the 60° and 70° FREEs have less rotation than the 40° and 50° FREEs, which is different from the trend shown in finite element results in Figure 6.10 to Figure 6.13 and also in Section 4.7 (the 60° FREE produces the highest rotation angle at 10 psi). This mismatch between trends can be clarified by considering the variations in wall thickness displayed in Table 6.1 and Figure 6.15. The finite element model is intentionally simplified and does not account for the coating of latex on the tube. To understand the impact of this simplification, consider Figure 6.12 and the rotation curves shown for a 60° FREE with various wall thicknesses determined with the FEA model. The latex tube used to model the FREEs in the FEA model has a wall thickness of 0.8 mm. Increasing this thickness to 1.6 mm (100% more) in the FEA model produces results closer to the experiment data obtained with FREEs with a 2.05 mm average wall thickness (Table 6.1). This suggests that the difference between experimental measurements and simulation results are due to in consistent geometry and that both simulation models are useful as long as they are tuned with a consistent set of experiments.

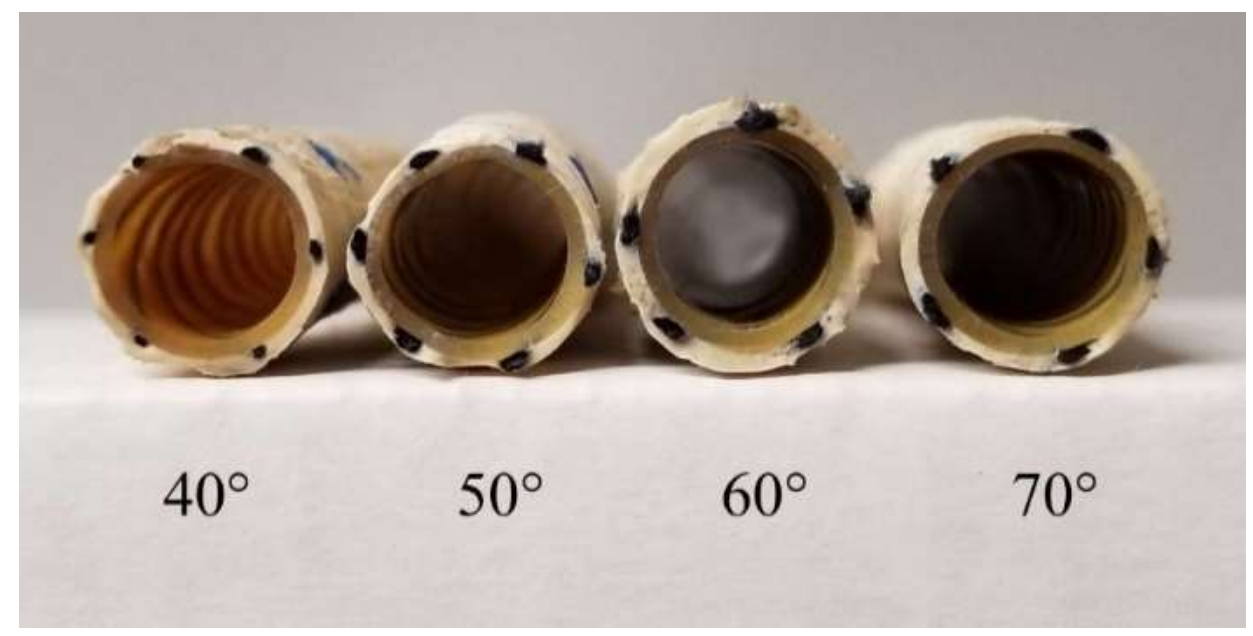

*Figure 6.15. FREEs used for experiments, cross-sectional point of view*

**PID Control**

Another application of the lumped-parameter model is to study the dynamic behavior of FREEs when part of a closed-loop control system. As described in Chapter 3, PID control of pressure enables the FREE to rotate smoothly to a desired angle of rotation. Finding the proper gain values for the PID controller is discussed in Section 3.2. Here, the controllability of the 40° FREE ($L = 12$ cm, $R = 0.7$ cm, $F_l = F_{gravity}$, $M_l = 0$, $k_e = 16478\ \frac{\mathrm{N}}{\mathrm{m}}$, $k_t = 0.0862\ \frac{\mathrm{N}}{\mathrm{rad}}$, $c_e = 0.34\ \frac{\mathrm{N.s}}{\mathrm{m}}$, and $c_t = 0.0000397\ \frac{\mathrm{Nm.s}}{\mathrm{rad}}$) is experimentally studied and compared to the corresponding simulation (Figure 6.16). See Appendix A for the MATLAB simulation function and Appendix C for plots of elongation and pressure.

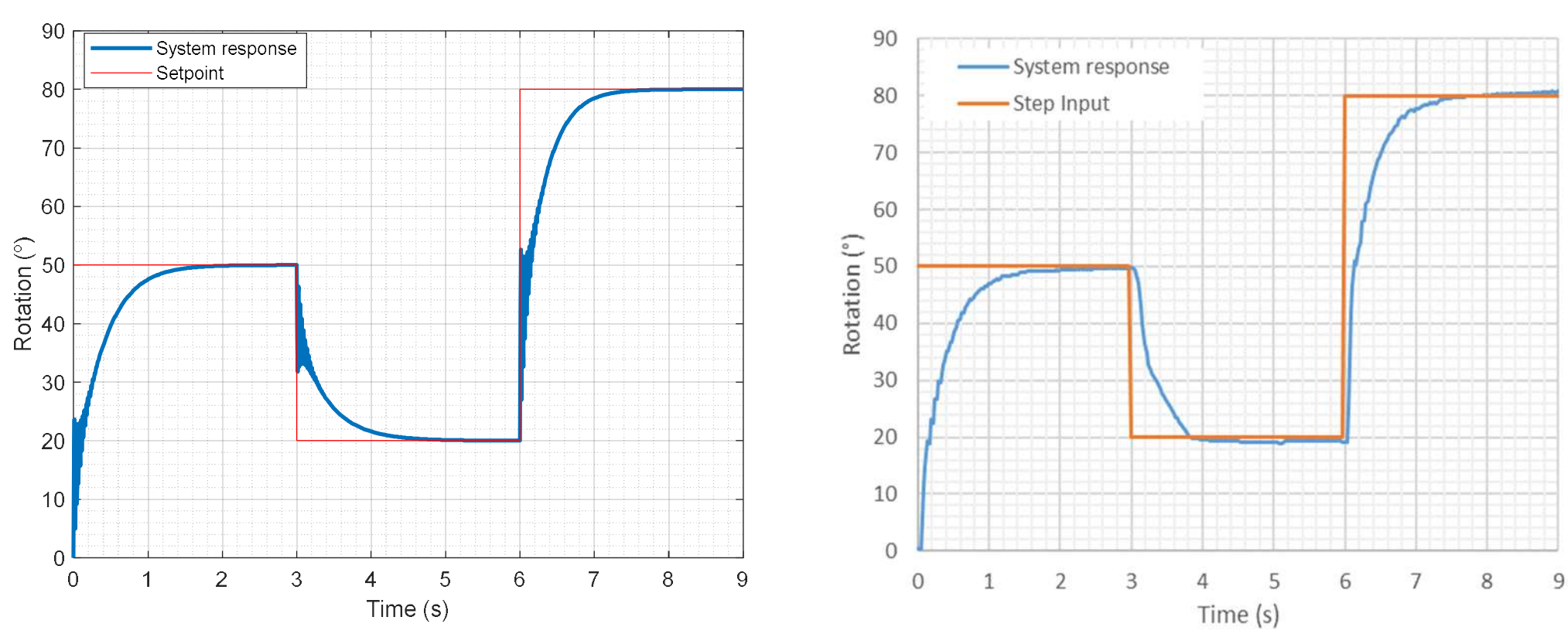

*Figure 6.16. Simulated response (left) versus actual response (right) of 40° FREE to various command angles ($K_p$ =10342 $\frac{Pa}{rad}$, $K_i$=94803 $\frac{Pa}{rad.s}$, $K_d$=0 $\frac{Pa.s}{rad}$)*

Note that the same stiffness and damping constants as obtained in Section 6.1 were used in the PID simulation. Additionally, the experimental results were collected by the method employed by Baumgart (2017): Arduino code (see Appendix C) and the pressure transducer shown Figure 4.2 in Baumgart (2017) were used to produce the PID control experimental data with a baud rate of 9600. Comparing the actual and simulation results suggest that the lumped-parameter model fairly predicts the motion of the FREE with the PID controller. The RMSD (root-mean-square deviation) error between simulation and experimental is 4% over the total duration of the maneuver. There are slight differences in the shape of curves in both plots of Figure 6.16, especially at the beginning of each step, due to 1) the sampling rate in the simulation being considerably higher than in data sets collected by experiment and 2) the physical characteristics of the pressure regulator not being included in simulation.

### Trajectory Planning

Controlling the angle of rotation following a customized trajectory is discussed in Section 3.3. This subsection compares the trajectory following behavior of the 40° FREE ($L =$ 12 cm, $R = 0.7$ cm, $F_l = F_{gravity}$, $M_l = 0$, $k_e = 16478\ \frac{\text{N}}{\text{m}}$, $k_t = 0.0862\ \frac{\text{N}}{\text{rad}}$, $c_e =$ $0.34\ \frac{\text{N.s}}{\text{m}}$ , and $c_t = 0.0000397\ \frac{\text{Nm.s}}{\text{rad}}$) in both simulation and experiment with the PID controller (Figure 6.17). See Appendix A for the MATLAB simulation function and Appendix C for plots of elongation and pressure.

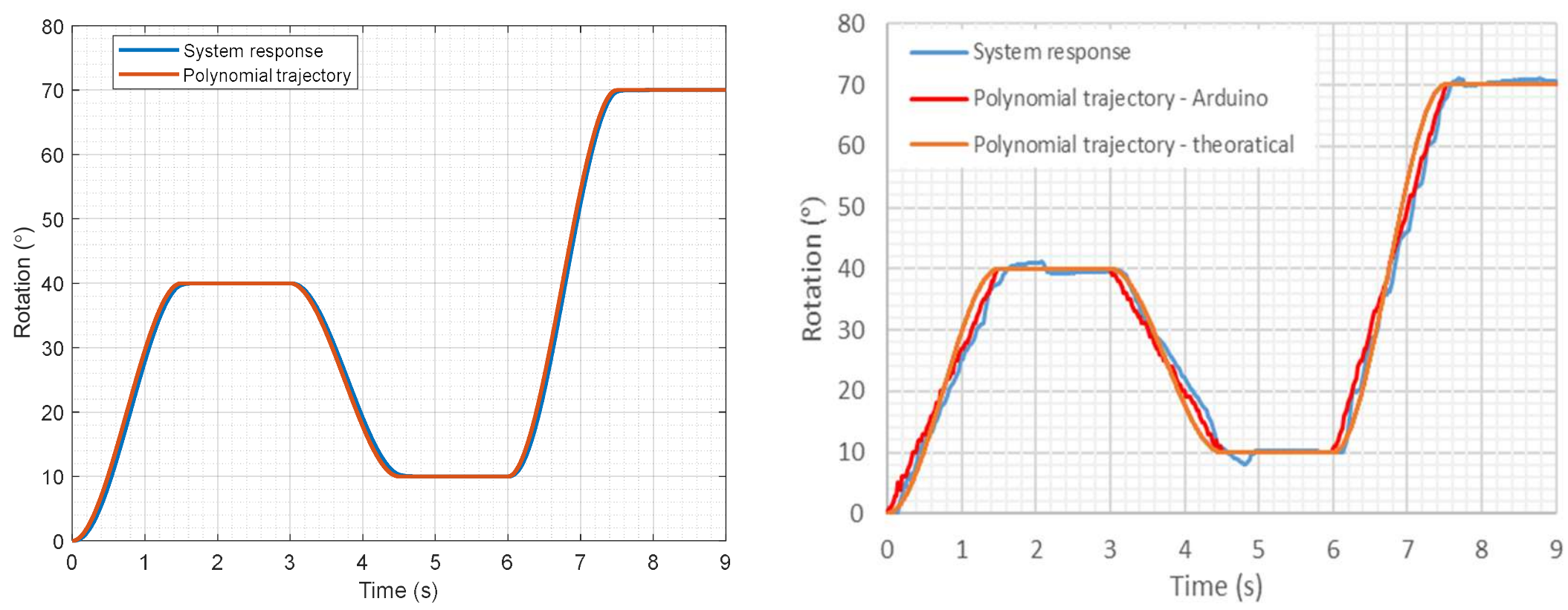

*Figure 6.17. Simulated response (left) versus actual response (right) of 40° FREE to trajectory following command angles ($K_p$ =17237 $\frac{Pa}{rad}$, $K_i$=603290 $\frac{Pa}{rad.s}$ , $K_d$=0 $\frac{Pa.s}{rad}$)*

Note that the same stiffness and damping constants as obtained in Section 6.1 were used in the simulation, and the same apparatus as in the PID control experiment was used for the experiment. Identical to the PID control response, the trajectory following experiment yields results matching with the simulation, although the relatively low speed of the Arduino control loop created a trajectory in the experiment slightly different from the theoretical one. This produced a more oscillatory response to the commanded path. The RMSD (root-mean-square deviation) error between simulation and experimental results is 3% over the total duration of the maneuver. One important insight obtained from the trajectory following simulation, besides its importance to actual applications, is the realization that high frequency oscillations observed in the PID simulation (Figure 6.16) are likely due to rapid changes in the desired angle of rotation. In the trajectory following maneuvers, the desired angle varies gradually. Additionally, the high frequency oscillations exhibits at the beginning of each step in Figure 6.16 can be eliminated by introducing derivative feedback. Figure 6.18 shows the PID response of the 40° FREE with a derivative gain of $K_d = 1723.7\ \frac{\text{Pa.s}}{\text{rad}}$ that eliminated the oscillatory behavior.

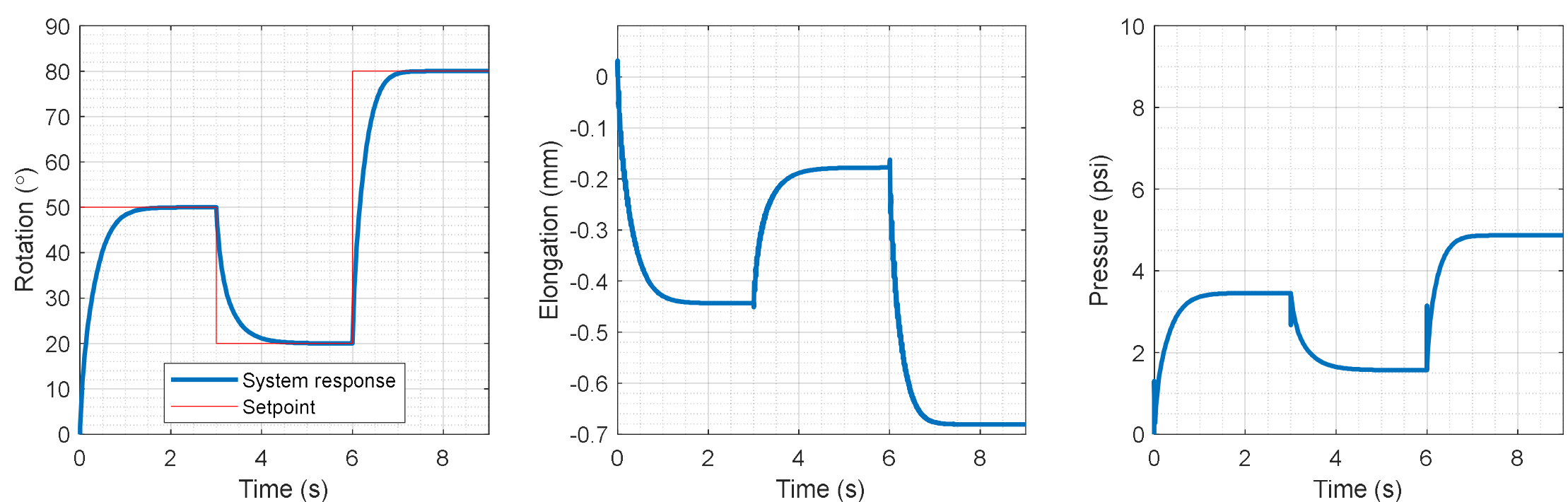

*Figure 6.18. Simulated response of 40° FREE to various command angles ($K_p$ =17237 $\frac{Pa}{rad}$, $K_i$=603290 $\frac{Pa}{rad.s}$, $K_d$ =1723.7 $\frac{Pa.s}{rad}$)*

Trajectory following maneuvers are beneficial in avoiding vibrations caused by the rapid changes in internal pressure. Experimental observation of the behavior of FREE module indicates substantial vibrations during motions. Hence, trajectory following could be used to control the rotation of a module by controlling the internal pressure of all four FREEs with one controller to create a smooth motion.

## 6.3 Summary

The comparisons in this chapter indicate that the lumped-parameter model predicts the behavior of FREEs better than the finite element model since the stiffness and damping coefficients used in simulation were derived directly from experimental data. Depending on the application and consistency between actual FREEs, one of the models can be preferred over the other in different cases. For example, if the actual system consists of FREEs with varying wall-thickness, the lumped-parameter is suitable to predict the average behavior. On the other hand, the finite element model captures the nonlinearity of the system better if the wall-thickness variation is minimized in manufacturing and the Ogden model parameters are adjusted carefully. Otherwise, the wall thickness needs to be accurately measured and set in the finite element model for each FREE. Additionally, the finite element model is found to be more convenient for analyzing the behavior of modules, since the lumped-parameter model would be excessively complicated. In conclusion, both models are complementary tools to characterize the behavior of FREEs and to utilize either model effectively, careful attention should be paid to the manufacturing process.

Investigating useful methods to create uniform wall-thickness, moldable elastomers, and the use of a precise fiber winding mechanism are potential future directions for the fabrication process.

# 7

# Discussion and Conclusions

In this thesis, a special type of soft robotic actuators, Fiber Reinforced Elastomeric Enclosures (FREEs) were modeled and experimentally analyzed in an attempt to understand the practicability of this actuator for use in a soft robotic arm. First, a lumped-parameter model was developed to simulate the dynamic and static behavior of a single FREE based on a consideration of the pressure and internal/external loads of the FREE. The presented simple model is capable of predicting the motion of FREEs to within a good approximation even when the complexities of the system are neglected. This analysis suggests that the winding angle and radius of a FREE do not change significantly after pressurization and that a linear model, using a constant winding angle and radius, can sufficiently predict the motion of FREEs at low pressures. However, creating a similar mathematical model for a module of FREEs was found to be tedious due to the complexity of establishing a relationship between reaction moments and forces. The dynamics of the lumped-parameter model were studied with a PID controller to determine the response of the FREE when used to study the control of the orientation of the end-effector. A defined step input and a trajectory following function for the rotation angle were implemented experimentally and theoretically with a PID controller. Both methods showed that the FREE successfully reaches the desired rotation by simply controlling the internal pressure. As a future research direction, trajectory following could be investigated for controlling the end effector position and rotation of a module along a path, which is required for a robotic arm.

The Finite Element Method was used to predict the effect of various parameters and arrangements of FREEs. The material properties of components of a FREE were determined experimentally from stress-strain relationships to formulate material models. A single FREE was modeled using two different nonlinear material models, neo-Hookean and Ogden. The results showed that an Ogden constitutive model has greater robustness and accuracy than a neo-Hookean model, especially at higher pressures, because the neo-Hookean model is a simple material characterization based on linear approximations of the strain invariants. The Ogden model was examined and validated by replicating experimental trials of loading and displacement. Parametric studies of the FREE pointed to the major role of the elastomer material properties in determining the characteristics of the FREE. They also showed that since fibers are essentially inextensible relative to the elastomer, the number of fibers or their stiffness after reaching a certain level do not change the FREE's behavior. Further explorations of the finite element model suggested that a FREE with an 90° winding angle has the largest elongation, and a 60° winding angle enables a FREE to rotate more than other angles geometries. Additionally, finite element analysis was able to demonstrate physical limitations (buckling) of FREEs under various boundary and loading conditions. All of these findings efficiently yield a comprehensive understanding of FREEs' behavior and facilitate their use in a parallel configuration as a module. The finite element model was able to determine the workspace of multiple FREEs in a module, which is an essential building block in creating a soft robotic arm. The finite element model had limitation when attempting to model large loads and displacements, although some of the limitations associated with buckling issues were observed in experimental testing as well. The use of an Ogden model to represent hyperelastic behavior requires parameter calibration for each particular case of study, and thus developing a more sophisticated hyperelastic material law, capturing the viscoelastic effects of the nonlinear elastomer, could be a future direction of research.

Considering the basic design specifications of a soft robotic arm, it is necessary to determine the workspace, the payload capacity, and the controllability of each degree of freedom in space. This classification of requirements can effectively identify the future directions of research on FREEs to develop a soft robotic arm. The workspace of a module has been explored using the finite element model; however, the workspace of multiple connected modules still needs to be studied. The outcomes of the current research on motion of a module, consisting of the most deformable FREEs, show that it has insufficient bending, contraction, and extension capabilities. This feature is because a pressurized single FREE generates small length change and zero bending. Hence, employing other types of pneumatic actuators combined with FREEs will be practical to create a more flexible arm.

Regarding the payload capacity, a holistic study of FREEs' force and torque generation needs to be conducted to determine their relationship to the displacements, and their contribution to the module's payload. Employing the understanding obtained from the workspace and payload studies of a module assists the designer in the future work to choose a suitable control system for modules to perform real-world tasks. On the other hand, establishing a cheap and efficient method for localizing the end-effector in space is essential for the control system and it could be one of the essential research directions in the future.

As another future improvement that would affect all of the studies, the manufacturing process needs to be improved to efficiently create uniform FREEs. Inconsistencies in the geometries of actual FREEs significantly affected experimental results and model comparisons, so it is highly crucial for future research.

In summary, the models developed in this project effectively predict the behavior of FREEs. However, each model cannot completely represent all behaviors, but they can nonetheless assist the design process in a number of many ways and potentially be used for future studies of FREEs and similar soft actuators.

# Appendix A

## Dynamic Simulation

```
%% Program freesolver.m: M-file to simulate the dynamic response of a
FREE
% Created: October 2018 by Soheil Habibian
% Latest Revision: March 28, 2019 by Soheil Habibian

% Remove all variables from the workspace
clear;
% Close all open figure windows
% close all;

%% Establish global variables
global Gamma L R m_l I_l dphi M_l F_l ke kt ce ct kp kd ki Pmax
%% Constants (based on values in Clayton's report)
Gamma = (40)*pi/180;    % fiber angle of relaxed FREE [rad]
L = 0.12;                % length of relaxed FREE [m]
R = 0.007;               % radius of relaxed FREE [m]
m_l = 0.028;             % mass of end cap [kg]
I_l = m_l*R^2;           % mass moment of inertia of end cap [kg.m^2]
dphi = (-20)*pi/180;     % desired rotation [rad]
Pmax = 10;       % Max pressure [psi]
%% External load and torque
M_l = 0;                 % external torque
F_l = m_l*9.81;          % external force
%% Stiffness and damping factors of FREE
ke = 10110.1;            % FREE's stiffness [N/m]
kt = 0.18557;            % FREE's torsional stiffness [Nm/rad]
ce = 5;                  % FREE's axial stiffness [N-s/m]
ct = 0.005;              % FREE's torsional damping constant [Nm-s/rad]
%% PID control gains
kp = 32000; kd = 0 ; ki = 1200000;      % Control gains

% figure('NumberTitle','off',...
%     'Position',[100 100 1000 500])
%% Perform Simlation
tend = 0.2;
tspan = [0 tend];
opts = odeset('RelTol',1e-6,'AbsTol',1e-6);
[t,y] = ode45('freefunction',tspan,[0 0 0 0 0],opts);

%% Create first figure of results
% figure('NumberTitle','off',...
%     'Position',[50 150 1000 500])

% Plot rotation
subplot(2,3,[4 5])
plot(t,y(:,2)*180/pi,'LineWidth',1.2);
hold on
line([0,t(end)],[dphi*180/pi,dphi*180/pi],'Color','red');
xlabel('time[s]');
```

```
ylabel('\theta^{\circ}');
% axis([0 t(end) 0 dphi*180/pi+10])
% legend('geometric rel.','constant','setpoint')
grid on; grid minor

% Plot elongation
subplot(2,3,2)
plot(t,y(:,1)*1000,'LineWidth',0.8);
xlabel('time[s]');
ylabel('elongation [mm]');
grid on; grid minor
hold on
% axis([0 0.35 -20 10])

% Calculate pressure for plotting based on controller (P must be > 0)
N = length(t);
Pplot = zeros(1,N);
for i=1:N
    Pplot(i) = -(kp*(dphi-y(i,2)) - kd*y(i,4) + ki*(dphi*t(i) - y(i,5)));
% PID control of Pressure
if Pplot(i) < 0
    Pplot(i) = 0;
elseif Pplot(i) > Pmax*6894.76
    Pplot(i) = Pmax*6894.76;
end
end

% Plot pressure
subplot(2,3,1)
plot(t,Pplot/6894.76,'LineWidth',0.8);
xlabel('time[s]');
ylabel('pressure [psi]')
grid on; grid minor;
hold on
% axis([0 0.35 0 10])

% Calculate fiber angle and tube radius for plotting
gammaplot = zeros(1,N);
rplot = zeros(1,N);
for i=1:N
    gammaplot(i) = acos((L+y(i,1))*cos(Gamma)/L);                        %
deformed fiber angle
    rplot(i) = L*tan(gammaplot(i))/((L*tan(Gamma)/R) + y(i,2));          %
deformed radius
end

% Plot FREE's fiber angle
subplot(2,3,3)
plot(t,gammaplot*180/pi,'LineWidth',0.8);
xlabel('time[s]');
ylabel('\gamma^{\circ}');
grid on; grid minor
hold on
% axis([0 0.35 0 45])

% Plot FREE's radius
subplot(2,3,6)
```

```
plot(t,rplot*1000,'LineWidth',0.8);
xlabel('time[s]');
ylabel('radius [mm]');
grid on; grid minor
hold on
% axis([0 0.35 -20 0])
```

## Function of Differential Equations of Motion

```
function ydot = freefunction(t,y)
% Function freefunction(t,y): function used by freesolver.m in simulating
% the dynamic response of a FREE
% Created: October 2018 by Soheil Habibian
% Latest Revision: November 17, 2018 by Keith W. Buffinton

% Establish global variables
global Gamma L R m_l I_l dphi M_l F_l ke kt ce ct kp kd ki Pmax

%% variables S(elongation), phi(rotation), derivatives, and integral
% y(1) = s
% y(2) = phi
% y(3) = sdot
% y(4) = phidot
% y(5) = phiint

ydot(1)= y(3);
ydot(2)= y(4);
ydot(5)= y(2);
%% Calculate pressure based on controller (P must be non-negative)
P = -(kp*(dphi-y(2)) - kd*y(4) + ki*(dphi*t-y(5)));   % PID control of
Pressure
if P < 0
    P = 0;
elseif P > Pmax*6894.76
    P = Pmax*6894.76;
end
%% Geometrical relationships
gamma = acos((L+y(1))*cos(Gamma)/L);        % deformed fiber angle
r = L*tan(gamma)/(L*tan(Gamma)/R + y(2));   % deformed radius
%% Elastomer's force and torque
F_e = -ke*y(1) - ce*y(3);                   % elastomer force (stiffness)
M_e = -kt*y(2) - ct*y(4);                   % elastomer torque (torsional
stiffness)
%% Differential equations
ydot(3) = (P*pi*r^2*(1-2*cot(gamma)^2) + F_l + F_e)/m_l;
ydot(4) = (-2*P*pi*r^3*cot(gamma) + M_l + M_e)/I_l;
ydot=ydot';
end
```

## PID Control

```
function PID
%% Function PID control: Simulates PID control of the FREE
```

```
% Created: June 2019 by Soheil Habibian
% Latest Revision: October 4, 2019
% Remove all variables from the workspace
clc;
% clear open figure windows
clf;
%% Establish global variables
global Gamma L R m_l I_l M_l F_l ke kt ce ct kp kd ki Pmax t dphi ddphi
dphi_int
%% Constants (based on values in Clayton's report)
Gamma = (-40)*pi/180;    % fiber angle of relaxed FREE [rad]
L = 0.12;                % length of relaxed FREE [m]
R = 0.007;               % radius of relaxed FREE [m]
m_l = 0.028;             % mass of end cap [kg]
I_l = m_l*R^2;           % mass moment of inertia of end cap [kg.m^2]
Pmax = 10;       % Max pressure [psi]
%% External load and torque
M_l = 0;                 % external torque
F_l = m_l*9.81;          % external force
%% Stiffness and damping factors of FREE
ke = 16478;            % FREE's stiffness [N/m]
kt = 0.0862;            % FREE's torsional stiffness [Nm/rad]
ce = 0.34;                  % FREE's axial stiffness [N-s/m]
ct = 0.0000397;              % FREE's torsional damping constant [Nm-
s/rad]
%% PID control gains
kp = (0.6)*17236.9;       % porpotional gain [Pa/rad]
ki = 5.5*17236.9;        % porpotional gain [Pa/rad-s]
kd = 0;                 % derivative gain [Pa-s/rad]
%% Trajectory planning parameters
phi_f1 = 50*pi/180;     % desired rotation [rad]
phi_f2 = 20*pi/180;     % desired rotation [rad]
phi_f3 = 80*pi/180;     % desired rotation [rad]
%% Perform Simlation
inc = 9;                 % length of the whole simulation
tspan = [0 inc];
opts = odeset('RelTol',1e-6,'AbsTol',1e-6);
[t,y] = ode45(@freefunction,tspan,[0 0 0 0 0],opts);
%% Create first figure of results
% figure('NumberTitle','off',...
%     'Position',[100 150 900 400])

% Calculate pressure for plotting based on controller (P must be > 0)
N = length(t);
Pplot = zeros(1,N);
dphi = zeros(1,N);
ddphi = zeros(1,N);
dphi_int = zeros(1,N);

for ii=1:N
    if (0 <= t(ii)) && (t(ii) < inc/3)                    % 1st step
starts here
        dphi(ii) = phi_f1;
        ddphi(ii) = 0;
        dphi_int(ii) = phi_f1*(t(ii));
```

```
    elseif (inc/3 <= t(ii)) && (t(ii) < 2*inc/3)          % 2nd step
starts here

         dphi(ii) = phi_f2;
         ddphi(ii) = 0;
         dphi_int(ii) = phi_f2*(t(ii)-inc/3)...
             + phi_f1*(inc/3);

    else                                        % 3rd step starts here

        dphi(ii) = phi_f3;
        ddphi(ii) = 0;
        dphi_int(ii) = phi_f3*(t(ii)-2*inc/3)...
            + phi_f2*(inc/3)...
            + phi_f1*(inc/3);

    end

   Pplot(ii) = kp*(dphi(ii)-y(ii,2)) + kd*(ddphi(ii)-y(ii,4))+
ki*(dphi_int(ii)-y(ii,5));  % PID control of Pressure
if Pplot(ii) < 0
    Pplot(ii) = 0;
elseif Pplot(ii) > Pmax*6894.76
    Pplot(ii) = Pmax*6894.76;
end

end

% Plot rotation response and trajectory
subplot(1,3,1)
plot(t,y(:,2)*180/pi,'LineWidth',2);
hold on

line([0,inc/3],[phi_f1*180/pi,phi_f1*180/pi],'Color','red');
line([inc/3,inc/3],[phi_f1*180/pi,phi_f2*180/pi],'Color','red');
line([inc/3,2*inc/3],[phi_f2*180/pi,phi_f2*180/pi],'Color','red');
line([2*inc/3,2*inc/3],[phi_f2*180/pi,phi_f3*180/pi],'Color','red');
line([2*inc/3,inc],[phi_f3*180/pi,phi_f3*180/pi],'Color','red');

xlabel('Time (s)');
ylabel('Rotation (\circ)');
legend('System response','Setpoint','Location', 'Best')
grid on; grid minor
axis([0 9 0 90]);

% Plot elongation response and trajectory
subplot(1,3,2)
plot(t,y(:,1)*1000,'LineWidth',2);
xlabel('Time (s)');
ylabel('Elongation (mm)');
grid on; grid minor
axis([0 9 -0.7 0.1]);

% Plot pressure
subplot(1,3,3)
plot(t,Pplot/6894.76,'LineWidth',2);
xlabel('Time (s)');
```

```
ylabel('Pressure (psi)')
grid on; grid minor;
hold on
axis([0 9 0 10])
```

## Root-Locus

```
% This script calculate/plot root-loci from the equation of motions of
free
% Y(s)= rotation of FREE in laplace domain
% Y(d)= desired twist angle of FREE in laplace form
% Last modified by Soheil Habibian, March 28, 2019

%% constant parameters
Gamma = (-20)*pi/180;    % fiber angle of relaxed FREE [rad] (considered
constant)
R = 0.007;               % radius of relaxed FREE [m] (considered
constant)
c2 = 2*pi*R^3*cot(Gamma);
m_l = 0.028;             % mass of end cap [kg]
I_l = m_l*R^2;           % mass moment of inertia of end cap [kg.m^2]
ct = 0.005;              % FREE's torsional damping constant [Nm-s/rad]
kt = 0.18557;            % FREE's torsional stiffness [Nm/rad]

%% PID control gains
kp = 32000; kd = 0 ; ki = 1200000;

%% Matlab's function for plotting root-loci - kp variation
% Y(s)/Y(d) = (-c2(ki+kp*s)/(s*A))/(1+kp(-c2*s/A))
% where A = I_l*s^3 + ct*s^2 + kt*s -c2*ki
num=[-c2 0]; den = [I_l ct kt -c2*ki];
figure('Name','Kp variation','Position',[50 50 700
500]);rlocus(num,den)

%% Matlab's function for plotting root-loci - ki variation
% Y(s)/Y(d) = (-c2(ki+kp*s)/(s*A))/(1+ki(-c2/A))
% where A = I_l*s^3 + ct*s^2 + (kt-c2*kp)*s
num=[-c2]; den = [I_l ct kt-c2*kp 0];
figure('Name','Ki variation','Position',[50 50 700
500]);rlocus(num,den)
%
% % Matlab's function for plotting root-loci - kd variation
% Y(s)/Y(d) = (-c2(ki+kp*s)/(s*A))/(1+ki(c2*s^2/A))
% where A = I_l*s^3 + ct*s^2 + (kt-c2*kp)*s -c2*ki
num=[-c2 0 0]; den = [I_l ct kt-c2*kp -c2*ki];
figure('Name','Ki variation','Position',[50 50 700
500]);rlocus(num,den)
```

## Trajectory Following

```
function ydot = freefunction(t,y)
%%  freefunction(t,y): Simulating the dynamic response of a FREE
% Created: October 2018 by Soheil Habibian
```

```matlab
% Latest Revision: November 17, 2018 by Keith W. Buffinton
%% variables S(elongation), phi(rotation), derivatives, and integral
% y(1) = s
% y(2) = phi
% y(3) = sdot
% y(4) = phidot
% y(5) = phiint

ydot(1)= y(3);
ydot(2)= y(4);
ydot(5)= y(2);
%% Calculate pressure based on controller (P must be non-negative)
if (0 <= t) && (t < inc/3)                         % 1st step starts here
    dphi = phi_f1;
    ddphi = 0;
    dphi_int = phi_f1*(t);

elseif (inc/3 <= t) && (t < 2*inc/3)               % 2nd step starts here

     dphi = phi_f2;
     ddphi = 0;
     dphi_int = phi_f2*(t-inc/3)...
         + phi_f1*(inc/3);

else           % 3rd step starts here

    dphi = phi_f3;
    ddphi= 0;
    dphi_int = phi_f3*(t-2*inc/3)...
         + phi_f2*(inc/3)...
         + phi_f1*(inc/3);

end

P = kp*(dphi-y(2)) + kd*(ddphi-y(4))+ ki*(dphi_int-y(5));  % PID
control of Pressure
if P < 0
    P = 0;
elseif P > Pmax*6894.76
    P = Pmax*6894.76;
end

%% Geometrical relationships
gamma = acos((L+y(1))*cos(Gamma)/L);        % deformed fiber angle
r = L*tan(gamma)/(L*tan(Gamma)/R + y(2));   % deformed radius

%% Elastomer's force and torque
F_e = -ke*y(1) - ce*y(3);                   % elastomer force
(stiffness)
M_e = -kt*y(2) - ct*y(4);                   % elastomer torque
(torsional stiffness)

%% Differential equations
ydot(3) = (P*pi*r^2*(1-2*cot(gamma)^2) + F_l + F_e)/m_l;
ydot(4) = (-2*P*pi*r^3*cot(gamma) + M_l + M_e)/I_l;
ydot=ydot';
end
```

```
end
function trajectory
%% Function trajectory: Simulates the trajectory following of the FREE
% Created: June 2019 by Soheil Habibian
% Latest Revision: October 3, 2019
% Remove all variables from the workspace
clc;
% clear open figure windows
clf;
%% Establish global variables
global Gamma L R m_l I_l M_l F_l ke kt ce ct kp kd ki Pmax t dphi ddphi
phiplot dphi_int
%% Constants (based on values in Clayton's report)
Gamma = (-40)*pi/180;    % fiber angle of relaxed FREE [rad]
L = 0.12;                % length of relaxed FREE [m]
R = 0.007;               % radius of relaxed FREE [m]
m_l = 0.028;             % mass of end cap [kg]
I_l = m_l*R^2;           % mass moment of inertia of end cap [kg.m^2]
Pmax = 10;       % Max pressure [psi]
%% External load and torque
M_l = 0;                 % external torque
F_l = m_l*9.81;          % external force
%% Stiffness and damping factors of FREE
ke = 16478;           % FREE's stiffness [N/m]
kt = 0.0862;           % FREE's torsional stiffness [Nm/rad]
ce = 0.34;                 % FREE's axial stiffness [N-s/m]
ct = 0.0000397;              % FREE's torsional damping constant [Nm-
s/rad]
%% PID control gains
kp = (1)*17236.9;      % porpotional gain [Pa/rad]
ki = 35*17236.9;       % porpotional gain [Pa/rad-s]
kd = 0;                % derivative gain [Pa-s/rad]
%% Trajectory planning parameters
phi_i = 0;       % initial angle of rotation, first step [rad]
phi_f1 = 40*pi/180;     % desired rotation [rad]
phi_f2 = 10*pi/180;     % desired rotation [rad]
phi_f3 = 70*pi/180;     % desired rotation [rad]
%% Perform Simlation
inc = 9;                 % length of the whole simulation
t_d = 1.5;                 % desired time to reach the set point angle
of rotation [s]
tspan = [0 inc];
opts = odeset('RelTol',1e-6,'AbsTol',1e-6);
[t,y] = ode45(@freefunction,tspan,[0 0 0 0 0],opts);
%% Create first figure of results
% figure('NumberTitle','off',...
%     'Position',[100 150 900 400])

% Calculate pressure for plotting based on controller (P must be > 0)
N = length(t);
Pplot = zeros(1,N);
dphi = zeros(1,N);
ddphi = zeros(1,N);
dphi_int = zeros(1,N);

for ii=1:N
```

```
    if (t(ii) <= t_d)                                        % 1st step
starts here

        dphi(ii) = phi_i + 3*(phi_f1 - phi_i)*(t(ii)^2)/(t_d^2) -
2*(phi_f1 - phi_i)*(t(ii)^3)/(t_d^3);
        ddphi(ii) = 6*(phi_f1 - phi_i)*t(ii)/(t_d^2) - 6*(phi_f1 -
phi_i)*(t(ii)^2)/(t_d^2);
        dphi_int(ii) = phi_i*t(ii) + (phi_f1 - phi_i)*(t(ii)^3)/(t_d^2)
- (phi_f1 - phi_i)*(t(ii)^4)/((t_d^3)*2);

    elseif (t_d <= t(ii)) && (t(ii) < inc/3)

        dphi(ii) = phi_f1;
        ddphi(ii) = 0;
        dphi_int(ii) = phi_f1*(t(ii)-t_d)...
            + phi_i*t_d + (phi_f1 - phi_i)*(t_d^3)/(t_d^2) - (phi_f1 -
phi_i)*(t_d^4)/((t_d^3)*2);

    elseif (inc/3 <= t(ii)) && (t(ii) < inc/3+t_d)           % 2nd step
starts here

        dphi(ii) = phi_f1 + ((t(ii)-inc/3)^2)*3*(phi_f2 -
phi_f1)/(t_d^2) - ((t(ii)-inc/3)^3)*2*(phi_f2 - phi_f1)/(t_d^3);
        ddphi(ii) = (t(ii)-inc/3)*6*(phi_f2 - phi_f1)/(t_d^2) -
((t(ii)-inc/3)^2)*6*(phi_f2 - phi_f1)/(t_d^3);
        dphi_int(ii) = phi_f1*(t(ii)-inc/3) + (phi_f2 - phi_f1)*(t(ii)-
inc/3)^3/(t_d^2) - (phi_f2 - phi_f1)*(t(ii)-inc/3)^4/((t_d^3)*2)...
            + phi_f1*(inc/3-t_d)...
            + phi_i*t_d + (phi_f1 - phi_i)*(t_d^3)/(t_d^2) - (phi_f1 -
phi_i)*(t_d^4)/((t_d^3)*2);

    elseif (inc/3+t_d <= t(ii)) && (t(ii) < 2*inc/3)

         dphi(ii) = phi_f2;
         ddphi(ii) = 0;
         dphi_int(ii) = phi_f2*(t(ii)-inc/3-t_d)...
            + phi_f1*(t_d) + (phi_f2 - phi_f1)*(t_d)^3/(t_d^2) -
(phi_f2 - phi_f1)*(t_d)^4/((t_d^3)*2)...
            + phi_f1*(inc/3-t_d)...
            + phi_i*t_d + (phi_f1 - phi_i)*(t_d^3)/(t_d^2) - (phi_f1 -
phi_i)*(t_d^4)/((t_d^3)*2);

    elseif (2*inc/3 <= t(ii)) && (t(ii) < 2*inc/3+t_d)         % 3rd
step starts here

        dphi(ii) = phi_f2 + ((t(ii)-2*inc/3)^2)*3*(phi_f3 -
phi_f2)/(t_d^2) - ((t(ii)-2*inc/3)^3)*2*(phi_f3 - phi_f2)/(t_d^3);
        ddphi(ii) = (t(ii)-2*inc/3)*6*(phi_f3 - phi_f2)/(t_d^2) -
((t(ii)-2*inc/3)^2)*6*(phi_f3 - phi_f2)/(t_d^3);
        dphi_int(ii) = phi_f2*(t(ii)-2*inc/3) + (phi_f3 -
phi_f2)*(t(ii)-2*inc/3)^3/(t_d^2) - (phi_f3 - phi_f2)*(t(ii)-
2*inc/3)^4/((t_d^3)*2)...
            + phi_f2*(inc/3-t_d)...
            + phi_f1*(t_d) + (phi_f2 - phi_f1)*(t_d)^3/(t_d^2) -
(phi_f2 - phi_f1)*(t_d)^4/((t_d^3)*2)...
            + phi_f1*(inc/3-t_d)...
```

```
            + phi_i*t_d + (phi_f1 - phi_i)*(t_d^3)/(t_d^2) - (phi_f1 -
phi_i)*(t_d^4)/((t_d^3)*2);

    else

        dphi(ii) = phi_f3;
        ddphi(ii) = 0;
        dphi_int(ii) = phi_f3*(t(ii)-2*inc/3-t_d)...
            + phi_f2*(t_d) + (phi_f3 - phi_f2)*(t_d)^3/(t_d^2) -
(phi_f3 - phi_f2)*(t_d)^4/((t_d^3)*2)...
            + phi_f2*(inc/3-t_d)...
            + phi_f1*(t_d) + (phi_f2 - phi_f1)*(t_d)^3/(t_d^2) -
(phi_f2 - phi_f1)*(t_d)^4/((t_d^3)*2)...
            + phi_f1*(inc/3-t_d)...
            + phi_i*t_d + (phi_f1 - phi_i)*(t_d^3)/(t_d^2) - (phi_f1 -
phi_i)*(t_d^4)/((t_d^3)*2);

    end

   Pplot(ii) = kp*(dphi(ii)-y(ii,2)) + kd*(ddphi(ii)-y(ii,4))+
ki*(dphi_int(ii)-y(ii,5));  % PID control of Pressure
if Pplot(ii) < 0
    Pplot(ii) = 0;
elseif Pplot(ii) > Pmax*6894.76
    Pplot(ii) = Pmax*6894.76;
end

end

% Calculate trajectory for plotting based on controller (P must be > 0)
phiplot = zeros(1,N);
for jj=1:N
    if (t(jj) <= t_d)
        phiplot(jj) = phi_i + 3*(phi_f1 - phi_i)*(t(jj)^2)/(t_d^2) -
2*(phi_f1 - phi_i)*(t(jj)^3)/(t_d^3);
    elseif (t_d <= t(jj)) && (t(jj) < inc/3)
        phiplot(jj) = phi_f1;
    elseif (inc/3 <= t(jj)) && (t(jj) < inc/3+t_d)
        phiplot(jj) = phi_f1 + ((t(jj)-inc/3)^2)*3*(phi_f2 -
phi_f1)/(t_d^2) - ((t(jj)-inc/3)^3)*2*(phi_f2 - phi_f1)/(t_d^3);
    elseif (inc/3+t_d <= t(jj)) && (t(jj) < 2*inc/3)
        phiplot(jj) = phi_f2;
    elseif (2*inc/3 <= t(jj)) && (t(jj) < 2*inc/3+t_d)
        phiplot(jj) = phi_f2 + ((t(jj)-2*inc/3)^2)*3*(phi_f3 -
phi_f2)/(t_d^2) - ((t(jj)-2*inc/3)^3)*2*(phi_f3 - phi_f2)/(t_d^3);
    else
        phiplot(jj) = phi_f3;
    end
end

% Plot rotation response and trajectory
subplot(1,3,1)
plot(t,y(:,2)*180/pi,'LineWidth',2);
hold on
plot(t,phiplot*180/pi,'LineWidth',2);
xlabel('Time (s)');
ylabel('Rotation (\circ)');
```

```
legend('System response','Polynomial trajectory','Location', 'Best')
grid on; grid minor

% Plot elongation response and trajectory
subplot(1,3,2)
plot(t,y(:,1)*1000,'LineWidth',2);
xlabel('Time (s)');
ylabel('Elongation (mm)');
grid on; grid minor

% Plot pressure
subplot(1,3,3)
plot(t,Pplot/6894.76,'LineWidth',2);
xlabel('Time (s)');
ylabel('Pressure (psi)')
grid on; grid minor;
hold on
axis([0 9 0 10])

function ydot = freefunction(t,y)
%%  freefunction(t,y): Simulating the dynamic response of a FREE
% Created: October 2018 by Soheil Habibian
% Latest Revision: November 17, 2018 by Keith W. Buffinton
%% variables S(elongation), phi(rotation), derivatives, and integral
% y(1) = s
% y(2) = phi
% y(3) = sdot
% y(4) = phidot
% y(5) = phiint

ydot(1)= y(3);
ydot(2)= y(4);
ydot(5)= y(2);
%% Calculate pressure based on controller (P must be non-negative)
if (t <= t_d)                                   % 1st step starts here

    dphi = phi_i + 3*(phi_f1 - phi_i)*(t^2)/(t_d^2) - 2*(phi_f1 -
phi_i)*(t^3)/(t_d^3);
    ddphi = 6*(phi_f1 - phi_i)*t/(t_d^2) - 6*(phi_f1 -
phi_i)*(t^2)/(t_d^2);
    dphi_int = phi_i*t + (phi_f1 - phi_i)*(t^3)/(t_d^2) - (phi_f1 -
phi_i)*(t^4)/((t_d^3)*2);

elseif (t_d <= t) && (t < inc/3)

    dphi = phi_f1;
    ddphi = 0;
    dphi_int = phi_f1*(t-t_d)...
        + phi_i*t_d + (phi_f1 - phi_i)*(t_d^3)/(t_d^2) - (phi_f1 -
phi_i)*(t_d^4)/((t_d^3)*2);

elseif (inc/3 <= t) && (t < inc/3+t_d)          % 2nd step starts here

    dphi = phi_f1 + ((t-inc/3)^2)*3*(phi_f2 - phi_f1)/(t_d^2) - ((t-
inc/3)^3)*2*(phi_f2 - phi_f1)/(t_d^3);
    ddphi = (t-inc/3)*6*(phi_f2 - phi_f1)/(t_d^2) - ((t-
inc/3)^2)*6*(phi_f2 - phi_f1)/(t_d^3);
```

```
    dphi_int = phi_f1*(t-inc/3) + (phi_f2 - phi_f1)*(t-inc/3)^3/(t_d^2)
- (phi_f2 - phi_f1)*(t-inc/3)^4/((t_d^3)*2)...
        + phi_f1*(inc/3-t_d)...
        + phi_i*t_d + (phi_f1 - phi_i)*(t_d^3)/(t_d^2) - (phi_f1 -
phi_i)*(t_d^4)/((t_d^3)*2);

elseif (inc/3+t_d <= t) && (t < 2*inc/3)

     dphi = phi_f2;
     ddphi = 0;
     dphi_int = phi_f2*(t-inc/3-t_d)...
        + phi_f1*(t_d) + (phi_f2 - phi_f1)*(t_d)^3/(t_d^2) - (phi_f2 -
phi_f1)*(t_d)^4/((t_d^3)*2)...
        + phi_f1*(inc/3-t_d)...
        + phi_i*t_d + (phi_f1 - phi_i)*(t_d^3)/(t_d^2) - (phi_f1 -
phi_i)*(t_d^4)/((t_d^3)*2);

elseif (2*inc/3 <= t) && (t < 2*inc/3+t_d)          % 3rd step starts
here

    dphi = phi_f2 + ((t-2*inc/3)^2)*3*(phi_f3 - phi_f2)/(t_d^2) - ((t-
2*inc/3)^3)*2*(phi_f3 - phi_f2)/(t_d^3);
    ddphi = (t-2*inc/3)*6*(phi_f3 - phi_f2)/(t_d^2) - ((t-
2*inc/3)^2)*6*(phi_f3 - phi_f2)/(t_d^3);
    dphi_int = phi_f2*(t-2*inc/3) + (phi_f3 - phi_f2)*(t-
2*inc/3)^3/(t_d^2) - (phi_f3 - phi_f2)*(t-2*inc/3)^4/((t_d^3)*2)...
        + phi_f2*(inc/3-t_d)...
        + phi_f1*(t_d) + (phi_f2 - phi_f1)*(t_d)^3/(t_d^2) - (phi_f2 -
phi_f1)*(t_d)^4/((t_d^3)*2)...
        + phi_f1*(inc/3-t_d)...
        + phi_i*t_d + (phi_f1 - phi_i)*(t_d^3)/(t_d^2) - (phi_f1 -
phi_i)*(t_d^4)/((t_d^3)*2);

else

    dphi = phi_f3;
    ddphi= 0;
    dphi_int = phi_f3*(t-2*inc/3-t_d)...
        + phi_f2*(t_d) + (phi_f3 - phi_f2)*(t_d)^3/(t_d^2) - (phi_f3 -
phi_f2)*(t_d)^4/((t_d^3)*2)...
        + phi_f2*(inc/3-t_d)...
        + phi_f1*(t_d) + (phi_f2 - phi_f1)*(t_d)^3/(t_d^2) - (phi_f2 -
phi_f1)*(t_d)^4/((t_d^3)*2)...
        + phi_f1*(inc/3-t_d)...
        + phi_i*t_d + (phi_f1 - phi_i)*(t_d^3)/(t_d^2) - (phi_f1 -
phi_i)*(t_d^4)/((t_d^3)*2);

end

% dphi = a0 + a2*t^2 + a3*t^3;
% ddphi = 2*a2*t + 3*a3*t^2;
% dphi_int = a0*t + a2*t^3/3 + a3*t^4/4;

P = kp*(dphi-y(2)) + kd*(ddphi-y(4))+ ki*(dphi_int-y(5));  % PID
control of Pressure
if P < 0
    P = 0;
```

```
elseif P > Pmax*6894.76
    P = Pmax*6894.76;
end

%% Geometrical relationships
gamma = acos((L+y(1))*cos(Gamma)/L);       % deformed fiber angle
r = L*tan(gamma)/(L*tan(Gamma)/R + y(2));  % deformed radius

%% Elastomer's force and torque
F_e = -ke*y(1) - ce*y(3);                  % elastomer force
(stiffness)
M_e = -kt*y(2) - ct*y(4);                  % elastomer torque
(torsional stiffness)

%% Differential equations
ydot(3) = (P*pi*r^2*(1-2*cot(gamma)^2) + F_l + F_e)/m_l;
ydot(4) = (-2*P*pi*r^3*cot(gamma) + M_l + M_e)/I_l;
ydot=ydot';
end
end
```

## Finite Element Analysis

```
clc; clear;

%% Program plot_disps.m: M-file to calculate displacements of a FREE
from
% the results generated from FEA model in Abaqus
% Created: March 2018 by Soheil Habibian

%% Parameters
L = 140; % length

%% reading the Excel file
filename = '40.xlsx';

pp = xlsread(filename,'A:A');
p = pp*60;

% base cordinates of the bottom node
x = xlsread(filename,'I2:I2');
y = xlsread(filename,'J2:J2');
z = xlsread(filename,'K2:K2');

% base cordinates of the side node
xx = xlsread(filename,'M2:M2');
yy = xlsread(filename,'N2:N2');
zz = xlsread(filename,'O2:O2');

% disp. the bottom node
u1 = xlsread(filename,'B:B');
u2 = xlsread(filename,'D:D');
u3 = xlsread(filename,'F:F');

% disp. the side node
```

```
uu1 = xlsread(filename,'C:C');
uu2 = xlsread(filename,'E:E');
uu3 = xlsread(filename,'G:G');

% NEW cordinates of the bottom node
x1 = x + u1;
y1 = y + u2;

% NEW cordinates of the side node
xx1 = xx + uu1;
yy1 = yy + uu2;
vvv = [x1 y1];

% calculating rotation
theta1 = atan2d(y,x);
s = size(u1);
theta2 = zeros(s(1),1);
for i = 1:s(1)
theta2(i,1) = atan2d(y1(i),x1(i));
if theta2(i,1) > 0
    theta2(i,1) = theta2(i,1)-360;
end
end

% calculating expansion
B = 2*sqrt(xx^2 + yy^2);
b = zeros(s(1),1);
for j = 1:s(1)
b(j,1) = 2*sqrt(xx1(j)^2 + yy1(j)^2);
end

extn = (L+u3)/L;
rotn = (theta1 - theta2);
for i = 2:s(1)
if rotn(i,1) < rotn(i-1,1)
    rotn(i,1) = rotn(i,1)+360;
end
end
% rotn(170,1) = rotn(170,1)+360;
% rotn(171,1) = rotn(171,1)+360;
% rotn(172,1) = rotn(172,1)+360;

expn = b/B;
% for j = 78:172
% if expn(j,1) < expn(j-1,1)
%     expn(j,1) = -expn(j,1)+(2*expn(j-1,1));
% end
% end

soheil= [p extn expn rotn];
%% Plotting
subplot(1,3,1)
plot(p,extn,'m','LineWidth',1.2)
% title('\alpha = 40^{\circ}')
xlabel('Pressure (Kpa)')
ylabel('\lambda_z')
```

```
% axis([0 60 1 1.32]);
grid on

subplot(1,3,2)
plot(p,expn,'m','LineWidth',1.2)
% title('\alpha = 40^{\circ}')
xlabel('Pressure (Kpa)')
ylabel('b/B')
% axis([0 60 1 1.25]);
grid on

subplot(1,3,3)
plot (p,rotn,'m','LineWidth',1.2)
% title('\alpha = 40^{\circ}')
xlabel('Pressure (Kpa)')
ylabel('\tau^{\circ}')
% axis([0 60 0 450]);
grid on
```

# Appendix B

## PID Control Arduino

```
// PID Control of FREE
// Created: Summer 2017 by Clayton Baumgart
// Latest Revision: October 4, 2019 by Soheil Habibian
// Control rotation of a single FREE
// Arduino output routes to pressure transducer attached to FREE

#include <Wire.h>
#include <Adafruit_Sensor.h>
#include <Adafruit_BNO055.h>
#include <utility/imumaths.h>

// create sensor object
Adafruit_BNO055 bno = Adafruit_BNO055(55);

int out = 9; //output pin - attach to pressure transducer input

//*******************************************
double omega_d = 0.0; //desired angle value (deg)
//******************************************

//** controller gains *******************************
double kp = 0.6;
double ki=5.5;
double kd=0;
//***************************************************

double outV = 0; //output voltage
unsigned long lastTime;
unsigned long time;
double angle;
double postn;
double initial_angle;
double error, errSum, errD, lastError;
double timeChange;
int timestp=3000;
//*************************************************
void setup()
{
 Serial.begin(9600);
 // throw error if connection is not found with sensor
 if (!bno.begin())
 {
  Serial.println("Not connected.");
 }
 pinMode(out, OUTPUT);
 delay(800);
 imu:: Vector<3> euler = bno.getVector(Adafruit_BNO055::VECTOR_EULER);
 initial_angle=euler.x();
}
```

```
double getAngle()
{
  imu:: Vector<3> euler = bno.getVector(Adafruit_BNO055::VECTOR_EULER);
  angle=euler.x();
  postn= angle-initial_angle;

  if(postn<0)
  {
    postn=-postn;
  }

  return postn;
}

double controlAction()
{
  unsigned long now = millis();
  timeChange = (double)(now-lastTime);
  if (now>timestp){
    omega_d=50;
  if (2*timestp<now)
    omega_d=20;
  if (3*timestp<now)
    omega_d=80;
 }
  error=omega_d-postn;

  error=map(error,0.0,180.0,0.0,255.0);
  errSum += error*(timeChange/1000);

  //fix error overshoots
  if (errSum>255)
  {
    errSum=255;
  }
  if (errSum<0)
  {
    errSum=0;
  }

  //define derivative error term
  errD = (error-lastError)/(timeChange/1000);

  if (error != 0)
  {
    outV = kp*error + ki*errSum + kd*errD;
    if (outV>255)
    {
      outV=255;
    }
  }

  lastError = error;
  lastTime = now;
```

```
  if (outV<0)
  {
   outV=0;
  }

  return outV;
}

void loop()
{
  postn=getAngle();
  outV=controlAction();
  analogWrite(out, outV);
  time = millis();

Serial.print(postn);
Serial.print("  ,  ");
//Serial.print(angle);
//Serial.print("  ,  ");
Serial.print(omega_d);
Serial.print("  ,  ");
Serial.println(time);}
```

## Trajectory Following

```
// Trajectory Following
// Created: October 4, 2019 by Soheil Habibian
// Control rotation of a single FREE
// Arduino output routes to pressure transducer attached to FREE

#include <Wire.h>
#include <Adafruit_Sensor.h>
#include <Adafruit_BNO055.h>
#include <utility/imumaths.h>

// create sensor object
Adafruit_BNO055 bno = Adafruit_BNO055(55);

int out = 9; //output pin - attach to pressure transducer input

//*******************************************
double omega_d = 0.0; //desired angle value (deg)
//******************************************

//** controller gains *******************************
double kp = 0.6;
double ki=25;
double kd=0;
//***************************************************

double outV = 0; //output voltage
unsigned long lastTime;
unsigned long time;
```

```
double angle;
double postn;
double initial_angle;
double error, errSum, errD, lastError;
double timeChange;
int t_d= 1500;
int phi1 = 40;
int  phi2 = 10;
int phi3 = 70;
//**********************************************
void setup()
{
  Serial.begin(9600);
  // throw error if connection is not found with sensor
  if (!bno.begin())
  {
    Serial.println("Not connected.");
  }
  pinMode(out, OUTPUT);
  delay(800);
  imu:: Vector<3> euler = bno.getVector(Adafruit_BNO055::VECTOR_EULER);
  initial_angle=euler.x();
}

double getAngle()
{
  imu:: Vector<3> euler = bno.getVector(Adafruit_BNO055::VECTOR_EULER);
  angle=euler.x();
  postn= angle-initial_angle;

  if(postn<0)
  {
    postn=-postn;
  }

  return postn;
}

double controlAction()
{
  unsigned long now = millis();
  timeChange = (double)(now-lastTime);

if (now > (2*t_d) && now <= 3*t_d)
  {
    omega_d = (3*phi1*((now-2*t_d)^2)/(t_d^2))-(2*phi1*((now-2*t_d)^3)/(t_d^3));
  }
  else if (now > (3*t_d) && now <= 4*t_d){
    omega_d =phi1;
  }
  else if (now > (4*t_d) && now <= 5*t_d)
  {
    omega_d = phi1-((3*(phi1-phi2)*((now-4*t_d)^2)/(t_d^2))-(2*(phi1-phi2)*((now-4*t_d)^3)/(t_d^3)));
  }
  else if (now > (5*t_d) && now <= 6*t_d)
  {
```

```
    omega_d =phi2;
  }
  else if (now > (6*t_d) && now <= 7*t_d)
  {
    omega_d = phi2+((3*(phi3-phi2)*((now-6*t_d)^2)/(t_d^2))-(2*(phi3-phi2)*((now-6*t_d)^3)/(t_d^3)));
  }
  else if (now > (7*t_d) && now <= 8*t_d)
  {
    omega_d =phi3;
  }
  else{
    omega_d=0;
  }

  error=omega_d-postn;
//  if (error<0)
//  {
//    error=0;
//  }

  error=map(error,0.0,180.0,0.0,255.0);
  errSum += error*(timeChange/1000);

  //fix error overshoots
  if (errSum>255)
  {
    errSum=255;
  }
  if (errSum<0)
  {
    errSum=0;
  }

  //define derivative error term
  errD = (error-lastError)/(timeChange/1000);

  if (error != 0)
  {
    outV = kp*error + ki*errSum + kd*errD;
    if (outV>255)
    {
      outV=255;
    }
  }

  lastError = error;
  lastTime = now;

  if (outV<0)
  {
    outV=0;
  }

  return outV;
}
```

```
void loop()
{
  postn=getAngle();
  outV=controlAction();
  analogWrite(out, outV);
  time = millis();

Serial.print(postn);
Serial.print("  ,  ");
Serial.println(omega_d);
//Serial.print(omega_d);
//Serial.print("  ,  ");
//Serial.println(time);

}
```

# Appendix C

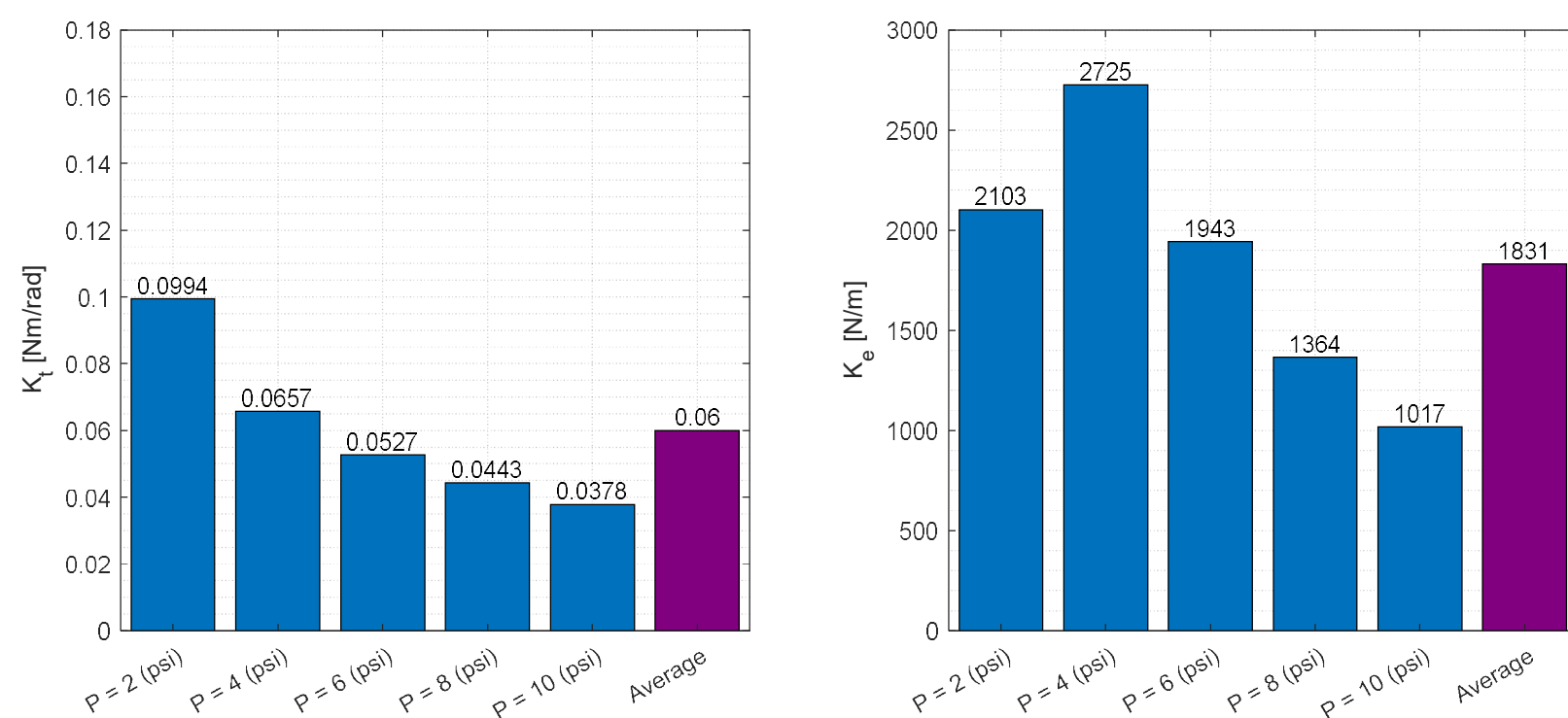

*Extensional stiffness $k_e$ and torsional stiffness $k_t$ of the 50° FREE at various pressures*

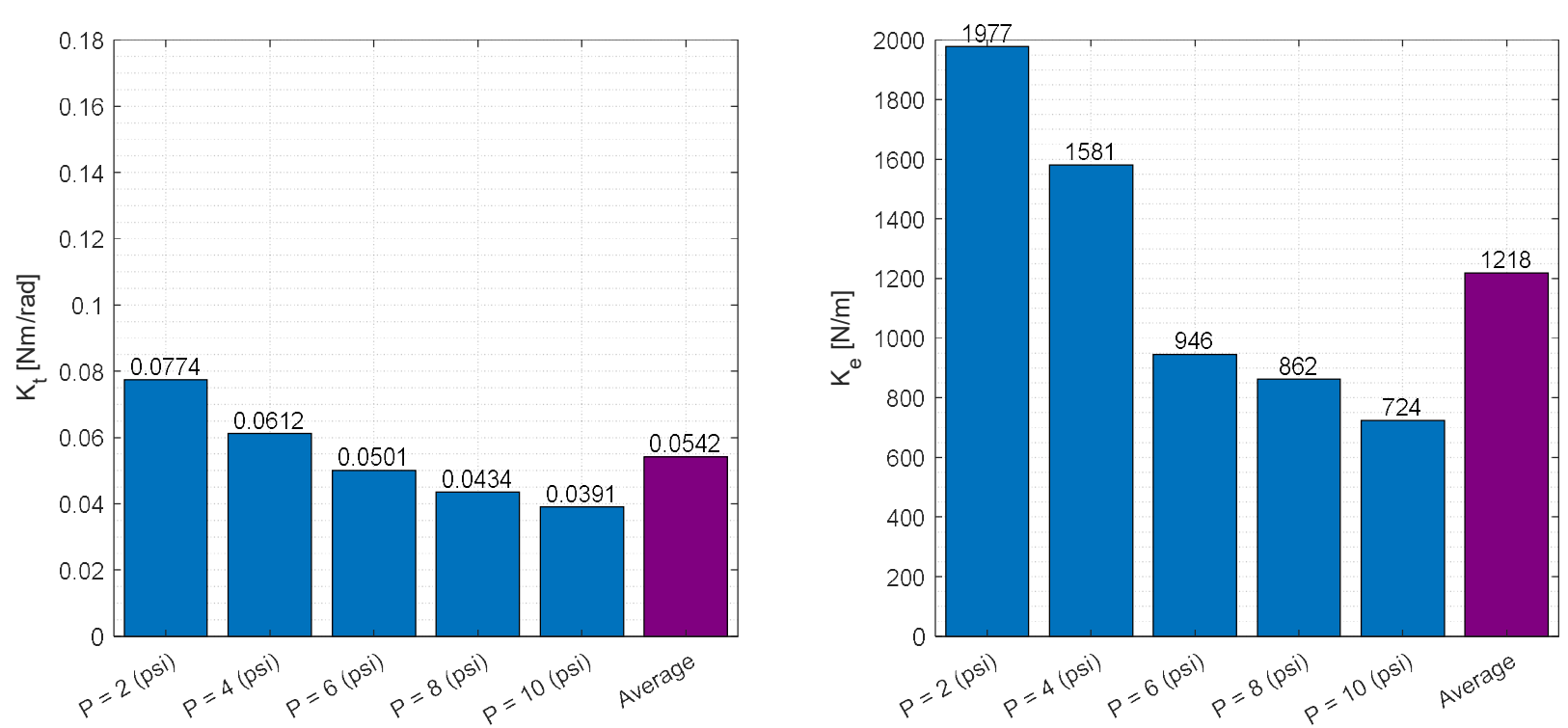

*Extensional stiffness $k_e$ and torsional stiffness $k_t$ of the 60° FREE at various pressures*

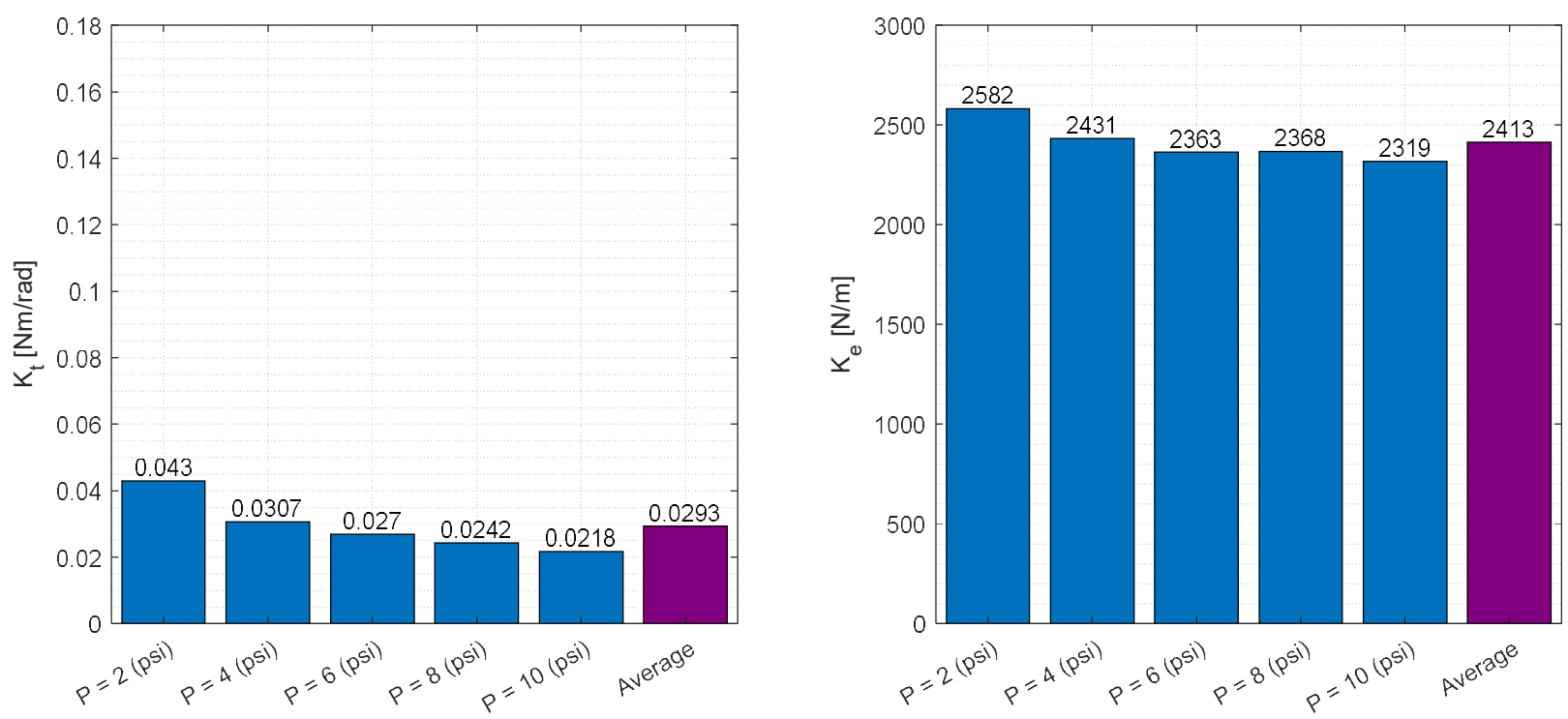

*Extensional stiffness $k_e$ and torsional stiffness $k_t$ of the 70° FREE at various pressures*

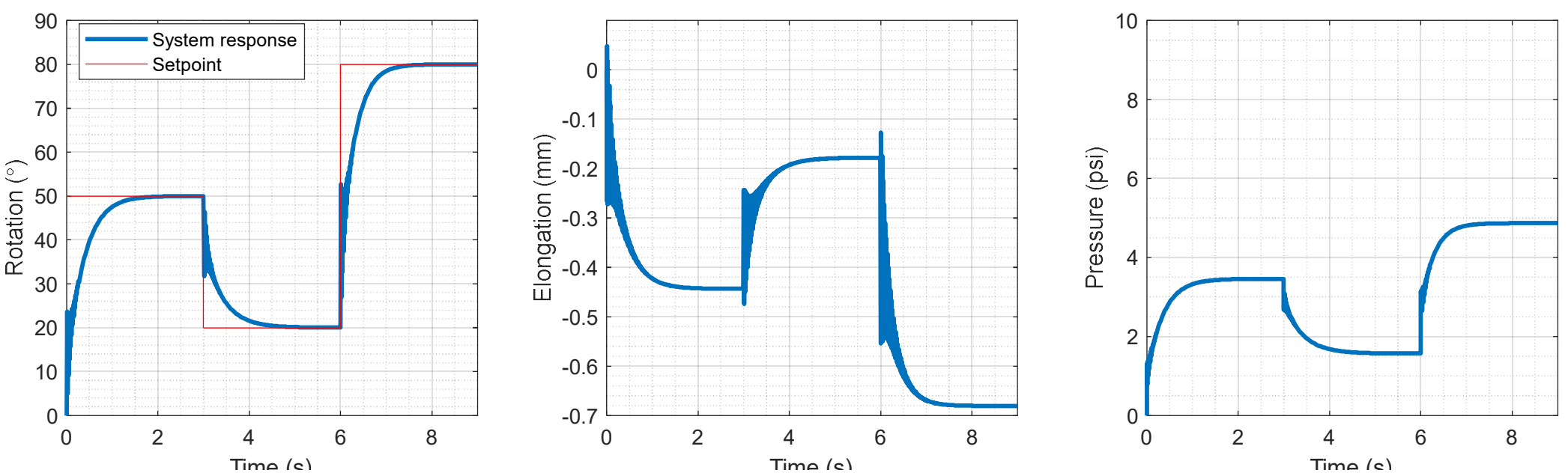

*Simulated response of 40° FREE to various command angles ($K_p$ =10342 $\frac{Pa}{rad}$ , $K_i$=94803 $\frac{Pa}{rad.s}$)*

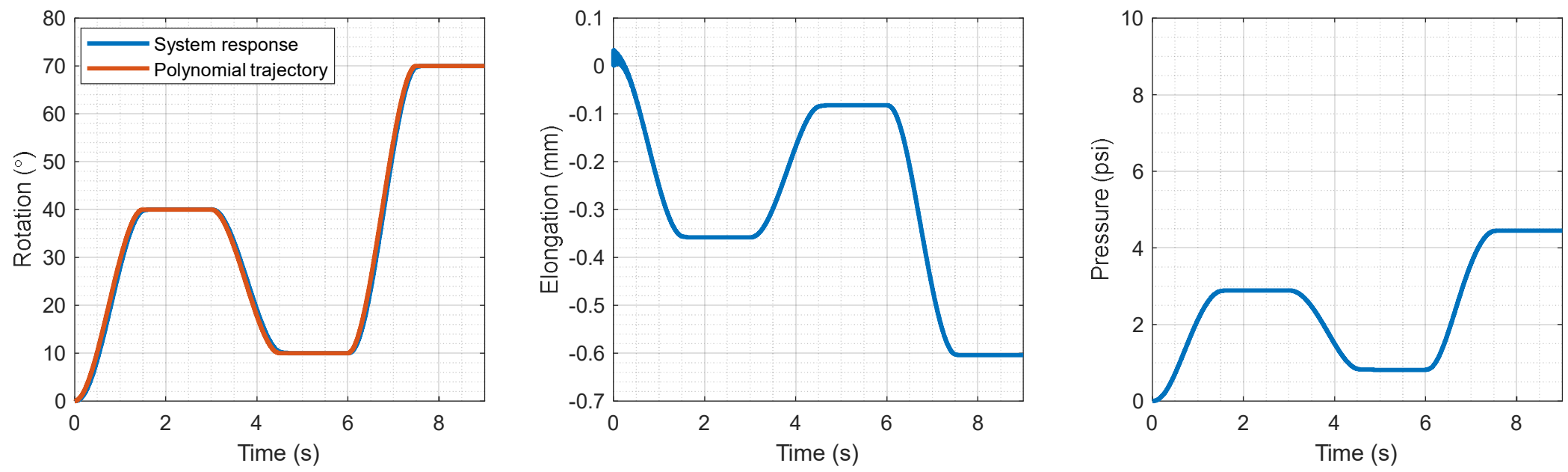

*Simulated response of 40° FREE to trajectory following command angles (Kp =17237 $\frac{Pa}{rad}$ , Ki=603290 $\frac{Pa}{rad.s}$)*